\documentclass[twoside,11pt,preprint]{article}

\usepackage[T1]{fontenc}
\usepackage{lmodern}
\usepackage{blindtext}

\usepackage{amsthm} 
\usepackage{jmlr2e}

\usepackage{lastpage}
\jmlrheading{23}{2022}{1-\pageref{LastPage}}{1/21; Revised 5/22}{9/22}{21-0000}{Victor Boone}

\ShortHeadings{Asymptotically optimal regret in communicating MDPs}{Boone}
\firstpageno{1}

\usepackage{subfiles}

\usepackage{url}            
\usepackage{soul}           

\hypersetup
{
	colorlinks=true,
	linkcolor=MidnightBlue,
	filecolor=magenta,
	urlcolor=FireBrick,
	citecolor=MidnightBlue,
}

\usepackage{booktabs}       
\usepackage{multicol}

\usepackage[svgnames,dvipsnames]{xcolor}         
\usepackage{graphicx} 
\usepackage{tikz}
\usetikzlibrary{positioning}
\usetikzlibrary{backgrounds}
\usetikzlibrary{shapes.geometric} 
\graphicspath{{figures/}}

\tikzset{
    state/.style={
        draw,
        circle,
        fill=none,
    },
    action/.style={
        draw,
        circle,
        color=white,
        fill=black,
        inner sep=0pt,
        minimum size=4mm,
        anchor=center,
    },
    reward/.style={
        font={\scriptsize}
    },
    transition/.style={
        ->,
        >=stealth,
    },
}

\colorlet{MyRed}{Crimson!60!DarkRed}
\colorlet{MyBlue}{DodgerBlue!75!black}
\colorlet{MyGreen}{DarkGreen}
\colorlet{MyViolet}{DarkMagenta}

\colorlet{MyLightBlue}{DodgerBlue!20}
\colorlet{MyLightGreen}{MyGreen!20}

\colorlet{PrimalColor}{MyBlue}
\colorlet{PrimalFill}{MyLightBlue}
\colorlet{DualColor}{MyRed}

\colorlet{AlertColor}{MyRed}	
\colorlet{BadColor}{MyRed}	
\colorlet{GoodColor}{MyGreen}	
\colorlet{LinkColor}{MediumBlue}	
\colorlet{MacroColor}{MyViolet}
\colorlet{RevColor}{MediumBlue}	

\usepackage[most]{tcolorbox}

\usepackage[inline,shortlabels]{enumitem}

\newenvironment{enum}{
    \begin{enumerate}[{\upshape (1)}, itemsep=-.33em]
}{
    \end{enumerate}
}

\newenvironment{ronum}{
    \begin{enumerate}[{\upshape (I)}, itemsep=-.33em]
}{
    \end{enumerate}
}

\usepackage{mdpmath}
\usepackage[sort&compress,capitalize,nameinlink,noabbrev]{cleveref}

\newtheorem{theorem}{Theorem}
\newtheorem{corollary}[theorem]{Corollary}
\newtheorem{lemma}[theorem]{Lemma}
\newtheorem{proposition}[theorem]{Proposition}

\newtheorem{definition}{Definition}
\newtheorem{assumption}{Assumption}

\theoremstyle{definition}

\newenvironment{subproof}{
    \def\proofname{Proof}
    \def\proofqed{$\square$}
    \begin{proof}
}{
    \end{proof}
}

\usepackage{algorithm,algorithmic}

\usepackage{xargs}

\usepackage[showdeletions]{color-edits}	
\newcommand{\draft}[1]{#1}	

\newcommand{\strong}[1]{\textbf{#1}} 
\newcommand{\hideme}[1]{} 

\usepackage{textgreek}

\allowdisplaybreaks 
\begin{document}

    \def\ECOE{\texttt{ECoE}}    
    \def\qed{\QED}
	
    \title{Asymptotically optimal regret in communicating Markov decision processes}

    \author{%
        \name Victor Boone 
        \email victor.boone@inria.fr 
        \\
        \addr Univ.~Grenoble Alpes, Inria, CNRS, Grenoble INP, LIG, 38000 Grenoble, France.
    }
    \editor{My editor}
	\maketitle
	
	\begin{abstract}%
        In this paper, we present a learning algorithm that achieves asymptotically optimal regret for Markov decision processes in average reward under a communicating assumption. 
        That is, given a communicating Markov decision process $\model$, our algorithm has regret $\regretlb(\model) \log(T) + \oh(\log(T))$ where $T$ is the number of learning steps and $\regretlb(\model)$ is the best possible constant. 
        This algorithm works by explicitly tracking the constant $\regretlb(\model)$ to learn optimally, then balances the trade-off between exploration (playing sub-optimally to gain information), co-exploration (playing optimally to gain information) and exploitation (playing optimally to score maximally). 
        We further show that the function $\regretlb(\model)$ is discontinuous, which is a consequence challenge for our approach.
        To that end, we describe a regularization mechanism to estimate $\regretlb(\model)$ with arbitrary precision from empirical data.
        \\[1em]
    \end{abstract}

    \section{Introduction}

    There is a Markov decision process, and there is a learning algorithm doing its best at picking actions to gather as much reward as possible. 
    The algorithm is learning, in that it is unaware of the structure of the Markov decision process; The algorithm learns forever and there is no reset mechanism whatsoever, in that the time horizon is infinite and that rewards are not discounted. 
    As such, the so-described system is a Markov decision process in average reward of which the reward and transition functions are unknown to the algorithm.
    In this paper, we measure the learning performance by the \strong{regret}, that compares the amount of reward collectible by an ominous algorithm that knows the Markov decision process in advance and plays optimal actions, to the learning algorithm that has to learn everything. 
    Following \cite{auer_near_optimal_2009} and all the subsequent works on regret minimization in average reward Markov decision processes, we define the regret after $T$ learning steps as
    \begin{equation}
    \label{equation_regret_introduction}
        T \optgain - \sum_{t=1}^{T} \Reward_t
    \end{equation}
    where $\optgain$ is the optimal gain of the unknown Markov decision process \cite{puterman_markov_2014} and $\Reward_t$ is the reward earned by the learning algorithm at time $t$.
    The smaller the regret is, the better the algorithm performs. 

    In this work, we are interested in \strong{asymptotic} regret guarantees in the \strong{model dependent} setting and in \strong{expectation}.
    That is, the unknown Markov decision process $\model$ is fixed and we want to design a learning algorithm $\learner$ such that the expectation of \eqref{equation_regret_introduction} under $\model$ and $\learner$, denoted $\Reg(T; \model, \learner)$, is as small as possible when $T \to \infty$. 
    This very problem goes back to the bandit literature, especially to the seminal paper of \cite{lai_asymptotically_1985}.
    Despite an immense literature in the surroundings of bandit problems, where the problem is mostly solved---see the book of \cite{lattimore_bandit_2020} for instance, gaping holes are left in the more general world of Markov decision processes. 
    Existing works either stick to a strong ergodic assumption \cite{agrawal_asymptotically_1988,burnetas_optimal_1997,graves1997asymptotically,auer_logarithmic_2006,pesquerel_imed_rl_2022}, or provide partial solutions \cite{tranos_regret_2021}, or have no proof of asymptotic optimality \cite{auer_near_optimal_2009,filippi_optimism_2010,saber_logarithmic_2024}. 
    It was only recently \cite{boone_regret_2025} that a general lower bound for the asymptotic regret of ``reasonable'' algorithms (see \Cref{section_lower_bound} and \Cref{definition_consistency}) has been described under the weaker \strong{communicating} assumption (\Cref{assumption_communicating}): For every reasonable learning algorithm, we have
    \begin{equation}
    \label{equation_bound_introduction}
        \Reg(T; \model, \learner) 
        \ge
        \regretlb(\model) \log(T)
        + \oh(\log(T))
        \qquad
        \text{as~$T \to \infty$}
    \end{equation}
    where $\regretlb(\model) \in \RR_+$ is a constant that describes the learning difficulty of $\model$ and that we recall in \Cref{theorem_lower_bound}. 

    In this paper, we show that the bound of \eqref{equation_bound_introduction} is tight: We describe an algorithm $\learner$ such that $\Reg(T; \model, \learner) = \regretlb(\model) \log(T) + \oh(\log(T))$ for all communicating Markov decision process $\model$ simulatenously, therefore solving the regret minimization problem at first order under the communicating assumption. 

    \section{Preliminaries}
    
In this section, we provide some background on average reward Markov decision processes and the associated regret minimization problem.

\paragraph{General notations.}
Given a set $\mathcal{X}$, we denote $\probabilities(\mathcal{X})$ the set of probability distributions over $\mathcal{X}$. 
We write $p \ll q$ if $p \in \probabilities(\mathcal{X})$ is absolutely continuous with respect to $q \in \probabilities(\mathcal{X})$ and $p \sim q$ if $p \ll q$ and $q \ll p$. 
Given two finite sets $\mathcal{X}$ and $\mathcal{Y}$, a transition kernel from $\mathcal{X}$ to $\mathcal{Y}$, or \strong{kernel}, is a map $k : \mathcal{X} \to \probabilities(\mathcal{Y})$ and we write $k(y|x)$ the probability that $Y = y$ under $Y \sim k(x)$.
Given $k : \mathcal{X} \to \probabilities(\mathcal{Y})$ a transition kernel, we write $\norm{k}_\infty := \sup_{x \in \mathcal{X}} \norm{k(x)}_1$ its infinity norm. 
We write $\KL(p||q) := \integral \log(\frac{\dd p}{\dd q})~\dd p$ the Kullback-Leibler divergence from distribution $q$ to distribution $p$. 
Given a vector $u \in \RR_+^d$, we write $\dmin(u) := \min \braces*{u_i: u_i > 0}$ its definite minimum. 
The span of a vector $u \in \RR^d$ is denoted $\vecspan{u} := \max(u) - \min(u)$. 

\subsection{Markov decision processes in average reward}

A \strong{Markov decision process} (or \strong{model}) is a tuple $\model \equiv (\states, \actions, \kernel, \rewardd)$ where $\states$ is the state space and $\actions \equiv \product_{\state \in \states} \actions(\state)$ is the action space, together forming the pair space $\pairs := \product_{\state \in \states} \braces{\state} \times \actions(\state)$; $\kernel : \pairs \to \probabilities(\states)$ is the transition kernel and $\rewardd : \pairs \to \probabilities([0, 1])$ is the reward function. 
Its mean reward function is $\reward : \pairs \to [0, 1]$ and is given by $\reward(\pair) := \integral x~\dd \rewardd(\pair)(x)$.
We write $\State_t, \Action_t, \Reward_t$ the $t$-th step state, action and reward observed when interacting with $\model$, and we further denote $\Pair_t := (\State_t, \Action_t)$ the played state-action pair at time $t$. 
By definition,
\begin{equation*}
    \State_{t+1} \sim \kernel(\Pair_t)
    \quad\text{and}\quad
    \Reward_t \sim \rewardd(\Pair_t)
\end{equation*}
are sampled independently of each other and independently of the previous history of play, given by $\History_t := (\State_1, \Action_1, \Reward_1, \ldots, \State_t)$, making the system memoryless.
The space of histories is denoted $\histories := \bigcup_{t \ge 1} ((\pairs \times [0, 1])^{t-1} \times \states)$. 
Given $\pairs' \subseteq \pairs$ a subset of pairs, we write $\states(\pairs') := \braces{\state \in \states: \braces{\state} \times \actions (\state) \cap \pairs' \ne \emptyset}$ the projection of $\pairs'$ onto $\states$. 
All throughout the paper, we will assume that $\model$ has a finite state-action space, see \Cref{assumption_finite} below.

\begin{assumption}
\label{assumption_finite}
    The pair space $\pairs$ is fixed and finite. 
\end{assumption}

\subsubsection{Randomized policies, their gain, bias \& gap functions}
\label{section_policies}

While new states and rewards are generated by the Markov decision process, actions are chosen by an exterior mechanism, and the simplest way to choose actions are \strong{policies}.
A (stationary) randomized policy is a kernel $\policy : \states \to \probabilities(\actions)$ from states to probabilities over legal actions, and it is said deterministic if of the form $\policy : \states \to \actions$.
We write $\randomizedpolicies$ and $\policies$ the respective sets of randomized and deterministic policies. 

Once a policy $\policy \in \randomizedpolicies$, a model $\model$ and an initial state $\state_0 \in \states$ are fixed, the distribution of $\History_t$ is completely determined for $t \ge 1$ and we write $\Pr^{\model, \policy}_{\state_0}(-)$ and $\EE^{\model, \learner}_{\state_0}[-]$ the probability and expectation operators over the induced probability space on $\histories$. 
Note that a fixed policy defines a Markov reward process on $\states$, and in particular a Markov chain. 
We say that a policy is \strong{ergodic}, \strong{recurrent} or \strong{unichain} if the associated Markov chain is respectively ergodic (irreducible and aperiodic), irreducible or if it has a unique stationary probability distribution (unique recurrent component), see \cite[§1]{levin_markov_2017} for more details on the terminology.

Given a policy $\policy \in \randomizedpolicies$, its \strong{gain} $\gainof{\policy}$ is the amount of reward that the policy collects in average and in the long run; Its \strong{bias} $\biasof{\policy}$ is the amount of reward that it collects in addition to the gain, visually:
\begin{equation}
\label{equation_policy_gain_and_bias}
\begin{aligned}
    \gainof{\policy}(\state; \model)
    := 
    \lim_{T \to \infty} 
    \EE_{\state}^{\model, \policy} \brackets*{
        \frac 1T \sum_{t=1}^{T} \Reward_t
    }
    ~~\text{and}~~
    \biasof{\policy}(\state; \model)
    :=
    \Clim_{T \to \infty} 
    \EE_{\state}^{\model, \policy} \brackets*{
        \sum_{t=1}^{T} \parens*{
            \Reward_t
            - \gainof{\policy}(\State_t; \model) 
        }
    }
\end{aligned}
\end{equation}
where $\Clim$ is the Cesàro-limit. 
The \strong{gap function} of a policy $\policy \in \randomizedpolicies$ is $\gapsof{\policy}(\state, \action; \model) := \gainof{\policy}(\state; \model) + \biasof{\policy}(\state; \model) - \reward(\state, \action) - \kernel(\state, \action) \biasof{\policy}(\model)$.
When too heavy to carry, the dependence on $\model$ may be dropped in notations to lighten up everything. 
Following standard Markov decision process theory, all these objects are well-defined, see \cite[§9]{puterman_markov_2014} for e.g.

\subsubsection{Communicating Markov decision processes and diameter}
\label{section_communicating}

A central assumption that is adopted in the following pages is the communicativity of the underlying model (\Cref{assumption_communicating}), meaning that one can travel from any state to any other in finite time under the right policy, i.e., that the diameter is finite. 
Equivalently, it can be shown that the diameter is finite if, and only if every policy $\policy \in \randomizedpolicies$ of full support, i.e., every policy such that $\policy(\action|\state) > 0$ for all $(\state, \action) \in \pairs$, is recurrent. 
The space of fully supported randomized policies, or \strong{fully randomized policies}, is denoted $\fullyrandomizedpolicies$. 

\begin{assumption}
\label{assumption_communicating}
    The model $\model$ is communicating, i.e., it has finite diameter:
    \begin{equation}
    \label{equation_diameter}
        \diameter(\model)
        :=
        \max_{\state \ne \state'}
        \min_{\policy \in \policies}
        \EE_{\state}^{\model, \policy} \brackets*{
            \inf \braces*{
                t \ge 1 : \State_t = \state'
            }
        }
        < 
        \infty
        .
    \end{equation}
\end{assumption}

\Cref{assumption_communicating} is fairly standard in the literature on regret minimization for average reward Markov decision processes, and it can be found in \cite{auer_near_optimal_2009,filippi_optimism_2010,tossou_near_optimal_2019,bourel_tightening_2020,fruit_improved_2020,zhang_regret_2019,boone_regret_2023,saber_logarithmic_2024,boone_logarithmic_2025} for example. 
This assumption is much weaker than the well-known ergodic assumption, found in \cite{agrawal_asymptotically_1988,burnetas_optimal_1997} for e.g..
Indeed, if a Markov decision process is ergodic, every state is recurrent under every policy, so that the diameter (\Cref{assumption_communicating}) is automatically finite.
The converse is wrong: All the examples of Markov decision processes that we provide in this work (see \Cref{figure_example_coexploration,figure_regret_discontinuity,figure_example_leveling,figure_example_discontinuous_gaps,figure_regret_discontinuity}) are communicating but not ergodic. 

\subsubsection{Optimal policies, gain, bias and the Bellman gaps}
\label{section_optimal_policies}

Given a model $\model$, a policy $\optpolicy \in \randomizedpolicies$ is said \strong{gain optimal}, written $\optpolicy \in \optpolicies(\model)$, if it achieves maximal gain over all possible policies, i.e.,
\begin{equation}
    \gainof{\optpolicy}(\state; \model)
    =
    \optgain(\state; \model)
    :=
    \max_{\policy \in \randomizedpolicies}
    \gainof{\policy}(\state; \model)
\end{equation}
regardless of the initial state $\state \in \states$. 
The vector $\optgain(\model) \in \RR^\states$ is the optimal gain vector. 
A policy $\optpolicy$ is said \strong{bias optimal} if it maximizes the bias among gain optimal policies, i.e., if $\optpolicy \in \optpolicies(\model)$ and $\biasof{\optpolicy}(\state; \model) = \optbias(\state; \model) := \max_{\policy \in \optpolicies(\model)} \biasof{\policy}(\state; \model)$ for all initial state $\state \in \states$.
The vector $\optbias(\model) \in \RR^\states$ is the optimal bias vector. 
According to standard theory \cite[§9]{puterman_markov_2014}, deterministic gain and bias optimal policies exist. 
Lastly, the \strong{Bellman gaps} are the gaps of bias optimal policies, given by
\begin{equation}
    \ogaps(\state, \action; \model)
    :=
    \optgain(\state; \model) + \optbias(\state; \model)
    - \reward(\state, \action) - \kernel(\state, \action) \optbias(\model)
    .
\end{equation}
Under the communicating assumption (\Cref{assumption_communicating}), it is well-known that the optimal gain does not depend on the initial state with $\vecspan{\optgain(\model)} = 0$ and that Bellman gaps are non-negative. 

\subsubsection{Optimal and weakly optimal state-action pairs}
\label{section_pairs}

We borrow the classification of pairs from \cite{boone_regret_2025}.

\begin{definition}[Optimal, weakly optimal and sub-optimal pairs]
\label{definition_classification_pairs}
    Let $\model \equiv (\pairs, \kernel, \rewardd)$ be a communicating model (\Cref{assumption_communicating}). 
    A pair $\pair \in \pairs$ is said \strong{weakly optimal} if $\ogaps(\pair) = 0$, written $\pair \in \weakoptimalpairs(\model)$;
    A pair $\pair \equiv (\state, \action) \in \pairs$ is said \strong{optimal} if it is recurrent under some gain optimal policy, i.e., there exists $\policy \in \optpolicies(\model)$ such that $\Pr_{\state}^{\model, \policy}(\forall n, \exists m \ge n: \Pair_m = \pair) = 1$, and in that case we write $\pair \in \optpairs(\model)$. 
\end{definition}

When a model $\model$ is communicating, it is known that $\optpairs(\model) \subseteq \weakoptimalpairs(\model)$ with equality if, and only if there exists a recurrent gain optimal policy \cite{boone_regret_2025}. 
In spirit, optimal pairs are all the pairs that are played infinitely often by gain optimal policies, while weakly optimal pairs that are not optimal correspond to their transient counterparts. 
Weakly optimal pairs are pairs that can be played without increasing the regret, see \Cref{definition_regret} and the discussion on \eqref{equation_regret_idea}.
In light of this property, a pair $\pair \notin \weakoptimalpairs(\model)$ is said \strong{sub-optimal}.

\subsection{Learning algorithms, regret minimization and lower bounds}
\label{section_algorithms}

Formally, a \strong{learning algorithm} is a kernel $\learner : \histories \to \probabilities(\actions)$, mapping histories to probabilities over actions. 
Similarly to policies, the choice of a learning algorithm $\learner$, a Markov decision process $\model$ and an initial state $\state_0 \in \states$ completely determines the distribution of the history $\History_t$ for all $t \ge 1$; We write $\Pr_{\state_0}^{\model, \learner}(-)$ and $\EE_{\state_0}^{\model, \learner}[-]$ the probability and expectation operators over the induced probability space on $\histories$. 
The \strong{visit count} at time $t \ge 1$ of a pair $\pair \in \pairs$ is the number of time this pair has been visited prior to time $t$, denoted $\visits_t (\pair) := \sum_{i=1}^{t-1} \indicator{\Pair_i = \pair}$. 

\subsubsection{The expected regret}
\label{section_regret}

In this paper, we measure the learning performance of learning algorithms with the regret.
\begin{definition}[Regret, \cite{auer_near_optimal_2009}]
\label{definition_regret}
    The \strong{expected regret} of a learning algorithm $\learner$ in the model $\model$ at horizon $T \ge 1$ and from the initial state $\state_0 \in \states$ is given by
    \begin{equation*}
        \Reg(T; \model, \learner, \state_0)
        :=
        \EE_{\state_0}^{\model, \learner} \brackets*{
            T \optgain(\state_0; \model)
            - \sum_{t=1}^{T-1} \Reward_t
        }
        .
    \end{equation*}
\end{definition}
Under the communicating assumption (\Cref{assumption_communicating}), it appears that the initial state $\state_0 \in \states$ is not important conceptually, and we will often willingly forget it in equations. 

The idea behind the definition of the regret \Cref{definition_regret} is that by writing $\EE[\Reward_t|\State_t,\Action_t] = \optgain(\State_t) + \optbias(\State_t) - \optbias(\State_{t+1}) - \ogaps(\State_t, \Action_t)$ and summing for $t \ge 1$, we have
\begin{equation}
\label{equation_regret_idea}
    \EE_{\state_0}^{\model, \learner} \brackets*{
        \sum_{t=1}^{T}
        \Reward_t
    }
    =
    T \optgain(\state_0) 
    + \optbias(\state_0)
    - \EE_{\state_0}^{\model, \learner} \brackets*{
        \optbias(\State_{T+1})
    }
    - \EE_{\state_0}^{\model, \learner} \brackets*{
        \sum_{t=1}^{T}
        \ogaps(\Pair_t)
    }
    .
\end{equation}
In \eqref{equation_regret_idea}, the quantity $\sum_{t=1}^T \ogaps(\Pair_t)$ is non-negative by non-negativity of Bellman gaps. 
It follows that for every learning algorithm $\learner$, we have $\EE^{\model, \learner}[\sum_{t=1}^T \Reward_t] \le T \optgain + \vecspan{\optbias}$. 
Similarly, we obtain that $\EE^{\model, \policy}[\sum_{t=1}^T \Reward_t] \ge T \optgain - \vecspan{\optbias}$ for every bias optimal policy $\policy$. 
So, $T \optgain$ is a good comparison proxy for the performance of the optimal policy---hence the definition of the regret. 
Note that \eqref{equation_regret_idea} also shows that the expected regret is the sum of the Bellman gaps up to a bounded error, with $\Reg(T; \model, \learner, \state_0) = \EE_{\state_0}^{\model, \learner}[\sum_{t=1}^T \ogaps(\Pair_t)] + \OH(1)$, meaning that the regret only grows when sub-optimal pairs are played. 

\subsubsection{Consistent algorithms and lower bound of \texorpdfstring{\cite{boone_regret_2025}}{(Boone, 2025)}}
\label{section_lower_bound}

Every regret lower bound requires a form of assumption on the class of learning algorithms that we consider.
The point is to discard algorithms that overspecialize their performance to a happy few Markov decision processes. 
Following a large panel of the literature, we restrict the focus to \strong{consistent} algorithms (see \Cref{definition_consistency}), also known as strong consistency or uniformly fast convergence, see \cite{lattimore_bandit_2020} for e.g.

\begin{definition}[Consistency]
\label{definition_consistency}
    A learning algorithm $\learner$ is said \strong{consistent} on a space of Markov decision processes $\models$ if, for all $\eta > 0$, all $\model \in \models$ and $\state_0 \in \states$, $\Reg(T; \model, \learner, \state_0) = \oh(T^\eta)$.
\end{definition}

Consistency is the assumption under which \cite{boone_regret_2025} state their regret lower bound $\Reg(T; \model, \learner) \ge \regretlb(\model) \log(T) + \oh(\log(T))$ that we recall in \Cref{theorem_lower_bound}.

The constant $\regretlb(\model) \in [0, \infty]$ is the solution of an optimization problem that is about gathering a sufficient amount of information for a minimal cost. 
That optimization problem goes over $\imeasure$ within the polytope of invariant measures $\imeasures(\model)$ (\Cref{definition_invariant_measure}); The objective function is $\sum_{\pair \in \pairs} \imeasure(\pair) \ogaps(\pair)$, that is directly related to the regret; And $\imeasure \in \imeasures(\model)$ must further satisfy an information constraint, consisting in that $\sum_{\pair \in \pairs} \imeasure(\pair) \KL(\model(\pair)||\model^\dagger(\pair)) \ge 1$ for every model $\model^\dagger \in \confusing(\model)$.
In that information constraint, $\confusing(\model)$ is the confusing set (\Cref{definition_confusing_set}), which is in the collection of models that coincide with $\model$ on optimal pairs $\optpairs(\model)$ and such that $\model$ and $\model^\dagger$ do not share any gain optimal policy. 
That information constraint can be interpreted as a rejection principle: Enough information must be gathered in order to ``reject'' every confusing model. 

\begin{definition}[Invariant measures]
\label{definition_invariant_measure}
    An \strong{invariant measure} of a Markov decision process $\model$ is a vector $\imeasure \in \RR_+^\pairs$ such that, for all $\state \in \states$, we have
    $
        \sum_{\action \in \actions(\state)}
        \imeasure(\state, \action)
        =
        \sum_{\pair \in \pairs}
        \imeasure(\pair)
        \kernel(\state|\pair) 
        .
    $
\end{definition}

\begin{definition}[Confusing set, \cite{boone_regret_2025}]
\label{definition_confusing_set}
    Fix $\models$ a space of Markov decision process.
    The \strong{confusing set} of a communicating model $\model \in \models$ is 
    \begin{equation*}
        \confusing(\model) 
        := 
        \braces*{
            \model^\dagger \in \models 
            : 
            \model^\dagger \gg \model, 
            \model^\dagger = \model \mathrm{~on~} \optpairs(\model) 
            \mathrm{~and~} \optpolicies(\model^\dagger) \cap \optpolicies(\model) = \emptyset
        }
    \end{equation*}
    where $\model^\dagger \gg \model$ means that $\rewardd^\dagger(\pair) \gg \rewardd(\pair)$ and $\kernel^\dagger(\pair) \gg \kernel(\pair)$ for all $\pair \in \pairs$, or equivalently, that $\support(\rewardd^\dagger(\pair)) \supseteq \support(\rewardd(\pair))$ and $\support(\kernel^\dagger(\pair)) \supseteq \support(\kernel(\pair))$ for all $\pair \in \pairs$. 
\end{definition}

With \Cref{definition_invariant_measure} and \Cref{definition_confusing_set} in hand, we can state the lower bound of \cite{boone_regret_2025}.

\begin{theorem}[\cite{boone_regret_2025}]
\label{theorem_lower_bound}
    Fix $\models$ a space of Markov decision process. 
    For every learning algorithm $\learner$ that is consistent on $\models$, for all $\model \in \models$ that is communicating (\Cref{assumption_communicating}) and $\state_0 \in \states$, we have $\Reg(T; \model, \learner, \state_0) \ge \regretlb(\model) \log(T) + \oh(\log(T))$ where $\regretlb(\model) \in [0, \infty]$ is given by
    \begin{equation}
    \label{equation_definition_regretlb}
        \inf \braces*{
            \sum_{\pair \in \pairs}
            \imeasure(\pair)
            \ogaps(\pair)
            :
            \imeasure \in \imeasures(\model)
            \mathrm{~and~}
            \inf_{\model^\dagger \in \confusing(\model)}
            \braces*{
                \sum_{\pair \in \pairs}
                \imeasure(\pair)
                \KL(\model(\pair)||\model^\dagger(\pair))
            }
            \ge
            1
        }
    \end{equation}
    where $\KL(\model(\pair)||\model^\dagger(\pair)) := \KL(\rewardd(\pair)||\rewardd^\dagger(\pair)) + \KL(\kernel(\pair)||\kernel^\dagger(\pair))$.
\end{theorem}

Any measure $\imeasure \in \imeasures(\model)$ achieving $\regretlb(\model)$ is called an \strong{optimal exploration measure}.\footnote{
    We show that such measures exist in \Cref{appendix_regularized}.
}

In the remaining of this paper, we show that the lower bound described in \Cref{theorem_lower_bound} is tight.
We provide a consistent learning algorithm $\learner$ such that $\Reg(T; \model, \learner) \le \regretlb(\model) \log(T) + \oh(\log(T))$ under mild condition on the ambient space $\models$.
As a consequence, the constant $\regretlb(\model)$ given above is the right measure of the learning difficulty of the communicating model $\model$ in the asymptotic regime $T \to \infty$.

    \section{A framework of asymptotically optimal algorithms: \ECOE}
    \label{section_framework_ecoe}

To begin our paper, we present the framework of \ECOE{}, standing for \texttt{E}xploration-\texttt{Co}exploration-\texttt{E}xploitation. 
That framework of algorithms is inspired from the lower bound of \Cref{theorem_lower_bound} and from intermediate results of \cite{boone_regret_2025}.
We will start by describing the framework from a high level perspective, then provide an instance $\learner$ of \ECOE{} in \Cref{section_final_ecoe} that we prove to be asymptotically optimal.
More specifically, we prove that our algorithm $\learner$ satisfies $\Reg(T; \model, \learner, \state_0) \le \regretlb(\model) \log(T) + \oh(\log(T))$ for all communicating model $\model$ and initial state $\state_0 \in \states$. 

The \ECOE{} framework combines three main components:

\begin{enum}
    \item 
        \strong{Exploration}, that is about playing sub-optimal pairs (\Cref{section_pairs}) to obtain information. 
        It turns out the cost of exploration is of order $\log(T)$ and accounts for the dominant part of the regret lower bound.
        Accordingly, $\regretlb(\model)$ and the associated optimization problem \eqref{equation_definition_regretlb} describe the optimal way to explore the environment $\model$.
        In \Cref{section_exploration}, we explain how to obtain an optimal exploration policy from \eqref{equation_definition_regretlb}.
    
   \item
        \strong{Co-Exploration}, that makes sure that \emph{all} optimal pairs (\Cref{definition_classification_pairs}) are visited over-logarithmically. 
        The cost of co-exploration is $\oh(\log(T))$, that is negligible in front of the cost of exploration. 
        For general communicating Markov decision processes, co-exploration brings new challenges that are absent from the ergodic setting and multi-armed bandits. 
        They are discussed in more details in \Cref{section_coexploration}.

    \item 
        \strong{Exploitation}, that is about scoring maximally when there is no apparent need to collect information. 
        In \Cref{section_exploitation}, we explain how to choose the exploitation policy and how to determine whether information is lacking or not. 
\end{enum}

\subsection{Retrieving a good exploration policy by estimating \texorpdfstring{$\regretlb(\model)$}{K(M)}}
\label{section_exploration}

The regret lower bound of \Cref{theorem_lower_bound} is only about gathering information outside of $\optpairs(\model)$; Indeed, as the information constraint ``$\inf_{\model^\dagger \in \confusing(\model)} \sum_{\pair \in \pairs} \imeasure(\pair) \KL(\model(\pair)||\model^\dagger(\pair)) \ge 1$'' only involves models $\model^\dagger$ that coincide with $\model$ on $\optpairs(\model)$ by definition, the bound of \Cref{theorem_lower_bound} tells nothing about the behavior of consistent learning algorithms outside of $\optpairs(\model)$. 
So, essentially, \Cref{theorem_lower_bound} claims that the dominant part of the regret is scaling logarithmically and is due to exploration exclusively.  

In fact, it also says \emph{how} to explore optimally.

Indeed, \Cref{equation_definition_regretlb} can be rewritten in terms of an optimization problem over policies.
The policy obtained by solving it can be used to explore optimally; Such a policy is called an optimal exploration policy. 
The precise alternative form for $\regretlb(\model)$ is found in \Cref{proposition_regretlb_policies}, where we show that the computation of the lower bound is about finding a policy $\policy$ that optimizes the ratio of the gain-loss $\optgain(\model) - \gainof{\policy}(\model)$ and of the amount of gathered information $\inf_{\model^\dagger \in \confusing(\model)} \sum_{\pair \in \pairs} \imeasureof{\policy}(\pair) \KL(\model(\pair)||\model^\dagger(\pair))$ under that policy.

\begin{proposition}[Policy-wise formulation of $\regretlb(\model)$]
\label{proposition_regretlb_policies}
    Fix $\models$ a space of Markov decision processes.
    Let $\model \in \models$ be a communicating model. 
    For $\policy \in \fullyrandomizedpolicies$ a fully randomized policy (\Cref{section_communicating}), denote by $\imeasureof{\policy}$ its unique probability invariant measure in $\model$ (i.e., the unique probability invariant measure of the Markov chain that $\policy$ induces on $\pairs$, see \Cref{proposition_measures_policies_correspondence} (1)). 
    We have
    \begin{equation}
    \label{equation_regretlb_2}
        \regretlb(\model)
        =
        \inf_{
            \policy \in \fullyrandomizedpolicies
        }
        \braces*{
            \frac{
                \optgain(\model) - 
                \gainof{\policy}(\model)
            }{
                \inf_{\model^\dagger \in \confusing(\model)}
                \sum_{\pair \in \pairs}
                \imeasureof{\policy}(\pair)
                \KL(\model(\pair)||\model^\dagger(\pair))
            }
        }
    \end{equation}
    with the convention $x / 0 = + \infty$ for $x \in \RR_+$. 
\end{proposition}

Note that for $\policy \in \fullyrandomizedpolicies$, we have $\imeasureof{\policy}(\state, \action) = \lim_{T \to \infty} \frac 1T \EE_{\state}^{\model, \policy} [\sum_{t=1}^T \indicator{\Pair_t = (\state, \action)}]$. 

The infimum in \eqref{equation_regretlb_2} is not a minimum in general, meaning that optimal exploration is not guaranteed to be reached by fully uniform policies.
This concern is minor, because the learning algorithm will only have access to a noisy version of $\model$, so getting an approximation of $\regretlb(\model)$ is fine anyway---provided that the approximation can be arbitrarily precise.
The take-away is that
\Cref{equation_regretlb_2} encourages a natural approach: Compute a fully randomized {exploration policy} that optimizes the cost/information gain trade-off well enough and use it to explore efficiently.
So, provided that we can solve \eqref{equation_regretlb_2}, the problem of optimal exploration is dealt with.

\bigskip
\def\proofname{Proof~of~\Cref{proposition_regretlb_policies}}
\begin{proof}
    If $\regretlb(\model) = + \infty$, there is nothing to prove.
    Assume $\regretlb(\model) < \infty$. 

    Denote $\ivalue(\imeasure; \model) := \inf_{\model^\dagger \in \confusing(\model)} \sum_{\pair \in \pairs} \imeasure(\pair) \KL(\model(\pair)||\model^\dagger(\pair))$ the amount of information gathered by a measure $\imeasure \in \RR_+$. 
    Note that every element $\imeasure \in \imeasures(\model)$ is of the form $\lambda \imeasure'$ with $\lambda \in \RR_+$ and $\imeasure' \in \imeasures(\model) \cap \probabilities(\pairs)$. 
    So, from \eqref{equation_definition_regretlb}, we have
    \begin{equation*}
        \regretlb(\model)
        =
        \inf_{\imeasure \in \imeasures(\model) \cap \probabilities(\pairs)}
        \braces*{
            \frac{
                \sum_{\pair \in \pairs} 
                \imeasure(\pair) \ogaps(\pair)
            }{
                \ivalue(\imeasure; \model)
            }
        }
    \end{equation*}
    with the convention $x / 0 = + \infty$ for $x \in \RR_+$.
    Now, we replace $\imeasures(\model) \cap \probabilities(\pairs)$ by $\mathcal{I}_{>0}(\model) := \braces{\imeasure \in \imeasures(\model): \imeasure \in \probabilities(\pairs) \mathrm{~and~} \support(\imeasure) = \pairs}$ the set of fully supported probability invariant measures of $\imeasure$. 
    It is obvious that
    $
        \regretlb(\model)
        \le
        \inf_{\imeasure \in \mathcal{I}_{>0}(\model)}
        \braces{
            \sum_{\pair \in \pairs} 
            \imeasure(\pair) \ogaps(\pair)
            /\ivalue(\imeasure; \model)
        }
    $
    and we focus on the reverse inequality, that is obtained by approximating probability invariant measures by fully supported ones using \Cref{theorem_invariant_measures_approximate_with_covering}.
    Given $\epsilon > 0$, let $\imeasure_\epsilon \in \imeasures(\model) \cap \probabilities(\pairs)$ be such that $\sum_{\pair \in \pairs} \imeasure_\epsilon (\pair) \ogaps(\pair) / \ivalue(\imeasure_\epsilon; \model) \le \regretlb(\model) + \epsilon$. 
    By \Cref{theorem_invariant_measures_approximate_with_covering}, for all $\epsilon' > 0$, there exists $\imeasure_{\epsilon, \epsilon'} \in \mathcal{I}_{>0}(\model)$ such that $\norm{\imeasure_{\epsilon, \epsilon'} - \imeasure_{\epsilon}}_\infty \le \epsilon'$ and $\imeasure_{\epsilon, \epsilon'} \ge (1 - \epsilon') \imeasure_\epsilon$.
    So, 
    \begin{align*}
        \inf_{\imeasure \in \mathcal{I}_{>0}(\model)}
        \braces*{
            \frac{
                \sum_{\pair \in \pairs} 
                \imeasure(\pair) \ogaps(\pair)
            }{
                \ivalue(\imeasure; \model)
            }
        }
        & \le
        \inf_{\epsilon > 0}
        \inf_{\epsilon' > 0}
        \braces*{
            \frac{
                \sum_{\pair \in \pairs} 
                \imeasure_{\epsilon, \epsilon'}(\pair) \ogaps(\pair)
            }{
                \inf_{\model^\dagger \in \confusing(\model)}
                \imeasure_{\epsilon, \epsilon'}(\pair)
                \KL(\model(\pair)||\model^\dagger(\pair))
            }
        }
        \\
        & \overset{(\dagger)}\le
        \inf_{\epsilon > 0}
        \inf_{\epsilon' > 0}
        \braces*{
            \frac{
                \sum_{\pair \in \pairs} 
                \imeasure_{\epsilon}(\pair) \ogaps(\pair)
                + \epsilon' \norm{\ogaps}_\infty
            }{
                (1 - \epsilon')
                \inf_{\model^\dagger \in \confusing(\model)}
                \imeasure_{\epsilon}(\pair)
                \KL(\model(\pair)||\model^\dagger(\pair))
            }
        }
        \\
        & \overset{(\ddagger)}\le
        \inf_{\epsilon > 0}
        \inf_{\epsilon' > 0}
        \braces*{
            \frac 1{1-\epsilon'} \parens*{
                \regretlb(\model) + \epsilon
                + \frac{\epsilon' \norm{\ogaps}_\infty}{\ivalue(\imeasure_\epsilon; \model)}
            }
        }
        \\
        & \overset{(\S)}=
        \inf_{\epsilon > 0}
        \braces*{
            \regretlb(\model) + \epsilon
        }
        = \regretlb(\model)
    \end{align*}
    where
    $(\dagger)$ follows by definition of $\imeasure_{\epsilon, \epsilon'}$;
    $(\ddagger)$ follows by definition on $\imeasure_{\epsilon}$; and
    $(\S)$ follows from $\ivalue(\imeasure_\epsilon; \model) > 0$, that holds because $\sum_{\pair \in \pairs} \imeasure_\epsilon (\pair) \ogaps(\pair) / \ivalue(\imeasure_\epsilon; \model) < \infty$.

    We conclude using \Cref{proposition_measures_policies_correspondence} to rewrite the optimization problem over $\fullyrandomizedpolicies$:
    \begin{align*}
        \inf_{\imeasure \in \mathcal{I}_{>0}(\model)}
        \braces*{
            \frac{
                \sum_{\pair \in \pairs} 
                \imeasure(\pair) \ogaps(\pair)
            }{
                \ivalue(\imeasure; \model)
            }
        }
        & \overset{(\dagger)}=
        \inf_{\policy \in \fullyrandomizedpolicies}
        \braces*{
            \frac{
                \sum_{\pair \in \pairs} 
                \imeasureof{\policy}(\pair) \ogaps(\pair)
            }{
                \ivalue(\imeasureof{\policy}; \model)
            }
        }
        \\
        & \overset{(\ddagger)}=
        \inf_{\policy \in \fullyrandomizedpolicies}
        \braces*{
            \frac{
                \lim \EE_{\state_0}^{\model, \policy} \brackets*{
                    \frac 1T 
                    \sum_{\pair \in \pairs}
                    \sum_{t=1}^{T}
                    \indicator{\Pair_t = \pair}
                    \ogaps(\pair)
                }
            }{
                \ivalue(\imeasureof{\policy}; \model)
            }
        }
        \\
        & \overset{(\S)}=
        \inf_{\policy \in \fullyrandomizedpolicies}
        \braces*{
            \frac{
                \lim \EE_{\state_0}^{\model, \policy} \brackets*{
                    \frac 1T 
                    \sum_{t=1}^{T}
                    \parens*{
                        \optgain
                        + \optbias(\State_t) - \optbias(\State_{t+1})
                        - \Reward_t
                    }
                }
            }{
                \ivalue(\imeasureof{\policy}; \model)
            }
        }
        \\
        & \overset{(\$)}=
        \inf_{\policy \in \fullyrandomizedpolicies}
        \braces*{
            \frac{
                \optgain(\state_0) - \gainof{\policy}(\state_0)
            }{
                \ivalue(\imeasureof{\policy}; \model)
            }
        }
        \overset{(\#)}=
        \inf_{\policy \in \fullyrandomizedpolicies}
        \braces*{
            \frac{
                \optgain - \gainof{\policy}
            }{
                \ivalue(\imeasureof{\policy}; \model)
            }
        }
    \end{align*}
    where
    $(\dagger)$ follows from \Cref{proposition_measures_policies_correspondence}~(2);
    $(\ddagger)$ follows from \Cref{proposition_measures_policies_correspondence}~(1) and introduces $\state_0 \in \states$ arbitrary;
    $(\S)$ expands the gaps as $\ogaps(\Pair_t) = \optgain(\State_t) + \optbias(\State_t) - \reward(\Pair_t) - \kernel(\Pair_t) \optbias$;
    $(\$)$ follows by definition of the gain of $\policy$; and 
    $(\#)$ uses that $\state_0 \in \states$ is arbitrary. 
\end{proof}
\def\proofname{Proof}

\subsection{Co-exploration: Traveling between components of optimal pairs}
\label{section_coexploration}

By construction of $\regretlb(\model)$, exploration is not meant to gather information on optimal pairs. 
\cite{boone_regret_2025} show that optimal pairs are treated differently than others, in that they are visited over-logarithmically.
They show that, for all optimal pair $\pair \in \optpairs(\model)$, all consistent learning algorithm $\learner$ and provided that $\models$ is rich enough, we have
\begin{equation}
\label{equation_coexploration_requirement}
    \liminf_{T \to \infty}
    \frac{
        \EE_{\state_0}^{\model, \learner} \brackets*{
            \visits_T (\pair)
        }
    }{\log(T)}
    =
    + \infty.
\end{equation}
To guarantee that \eqref{equation_coexploration_requirement} is satisfied, \ECOE{} is careful at its choice for the exploitation policy, for how long it iterates it, and when it iterates it, for the consequence of \eqref{equation_coexploration_requirement} is that during play, $\optpairs(\model)$ must be covered uniformly fast.
That motivates the ideas thereafter.
\begin{ronum}
    \item 
        The exploitation policy $\policy^+$ must be of maximal support among $\optpairs(\model)$. 
        Ideally, we want to pick $\policy^+(-|\state)$ uniform on $\braces{\action \in \actions(\state): (\state, \action) \in \optpairs(\model)}$.
    \item 
        The exploitation policy is iterated up to \strong{regeneration}, i.e., we wait for a return to the initial state it was first iterated from. 
        By doing so, we guarantee that all the states of the current recurrent classes are visited with positive probability.
\end{ronum}

\subsubsection{An issue: The number of travels}

\begin{figure}[ht]
    \centering
    \begin{tikzpicture}
        \node[state] (1) at (0, 0) {$1$};
        \node[state] (2) at (3, 0) {$2$};
        \draw[transition] (1) to[bend left] node[midway, above] {$0$} (2);
        \draw[transition] (2) to[bend left] node[midway, below] {$0$} (1);
        \draw[transition] (1) to[loop] node[midway, above] {$0.5+\theta$} (1);
        \draw[transition] (2) to[loop] node[midway, above] {$0.5$} (2);
    \end{tikzpicture}
    \caption{
    \label{figure_example_coexploration}
        A class of Markov decision processes with deterministic transitions parameterized by $\theta$ where co-exploration is troublesome.
        Arrows are choices of actions that deterministically lead to the pointed state and labels are rewards. 
    }
\end{figure}
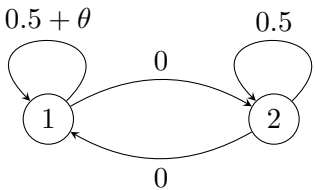

One difficulty arise---one that does not exist for ergodic Markov decision processes: The set of optimal pairs $\optpairs(\model)$ may be split into several components that may only be accessible by playing sub-optimal pairs (see \Cref{definition_component} for a formal definition of components). 
One such example is provided in \Cref{figure_example_coexploration}.

Consider the class of Markov decision processes $\models := \braces{\model_\theta : \abs{\theta} \le \frac 12}$ where $\model_\theta$ is the model with Bernoulli rewards described \Cref{figure_example_coexploration}.
Note that the policy alternating between the states $1$ and $2$ is never gain optimal for $\model_\theta \in \models$.
Now, consider $\model_0 \in \models$ and let $\learner^*$ be an asymptotically optimal algorithm for $\models$, i.e., assume that $\learner^*$ is such that $\Reg(T; \model_\theta, \learner^*) \le \regretlb(\model_\theta) \log(T) + \oh(\log(T))$ for all $\theta \in \Theta$. 
In $\model_0$, optimal pairs consists in the two loops $\optpairs(\model_0) = \braces{(1, 1), (2, 2)}$ and are separated by sub-optimal transitions of null reward. 
In the first hand, the algorithm $\learner^*$ must alternate infinitely often between the states $1$ and $2$ because \eqref{equation_coexploration_requirement} is satisfied.
On the second hand, we see that the confusing set of $\model_0$ relatively to $\models$ is empty, so that $\regretlb(\model_0) = 0$.
Accordingly, $\Reg(T; \model_0, \learner^*) = \oh(\log(T))$.
So,
\begin{equation*}
    \EE^{\model_0, \learner^*} \brackets*{
        \visits_T (1, 2) + \visits_T (2, 1)
    }
    =
    \oh (\log(T))
    .
\end{equation*}
Interestingly, $\model_0$ is the same example as in \cite[Figure~2]{ortner_online_2010} that is used to motivate the use of episodes in \texttt{UCYCLE} \cite{ortner_online_2010}, \texttt{UCRL2} \cite{auer_near_optimal_2009} and all the algorithms that followed. 
These algorithms make $\Omega(\log(T))$ episodes, and because of it, we see that on $\model_0$ above, they switch $\Omega(\log(T))$ times between the loops on the states $1$ and $2$.
Therefore, the regret of these algorithms is $\Omega(\log(T))$ on $\model_0$ and they are not asymptotically optimal---provided that \Cref{theorem_lower_bound} is tight, which is yet to be shown. 

\begin{definition}[Component of pairs]
\label{definition_component}
    Let $\model \equiv (\pairs, \kernel, \rewardd)$ be a model and let $\pairs' \subseteq \pairs$ be a subset of pairs.
    A communicating component of $\pairs'$, or just \strong{component}, is a maximal set $\pairs'' \subseteq \pairs'$ such that the restrained model $\model'' := (\pairs'', \kernel|_{\pairs''}, \rewardd|_{\pairs''})$ is well-defined\footnote{By \strong{well-defined}, we mean that by writing $\pairs'' \equiv \product_{\state'' \in \states''} \braces{\state''} \times \actions''(\state'')$, we have $\support \kernel(\pair'') \subseteq \states''$ for all $\pair'' \in \pairs''$ and $\actions''(\state'') \ne \emptyset$ for all $\state'' \in \states''$.} and communicating. 
\end{definition}

\subsubsection{Reducing the number of travels with the square trick}

To reduce the number of travels between the several possibles {components} (\Cref{definition_component}) of $\optpairs(\model)$, \ECOE{} uses a variant of the doubling trick (see \cite{auer_near_optimal_2009}) that we call the \strong{square trick} \eqref{equation_square_trick}.

\begin{ronum}
    \setcounter{enumi}{2}
    \item 
        Before exploiting, we check if it is worth traveling to another component of $\optpairs(\model)$:
        We decompose $\optpairs(\model)$ into components $\pairs_1, \ldots, \pairs_c$, and denote $\pairs_0 \in \braces*{\pairs_1, \ldots, \pairs_c}$ the component the algorithm is currently on.
        If 
        \begin{equation}
        \tag{\texttt{ST}}
        \label{equation_square_trick}
            \min_{1 \le i \le c}
            \min_{\pair \in \pairs_i} 
            \visits(\pair) 
            < 
            \sqrt{
                \min_{\pair \in \pairs_0}
                \visits(\pair)
            },
        \end{equation}
        then, we travel to the least visit component of $\optpairs(\model)$; Otherwise, we exploit on the current component $\pairs_0$ as described in points (1) and (2) upstream. 
\end{ronum}

Note the ressemblance of \eqref{equation_square_trick} with the forced exploration rule of the D-Tracking rule of \cite[§3.1]{garivier_optimal_2016} in the setting of best-arm identification. 
Similarly to the D-Tracking rule, the square root of the square trick \eqref{equation_square_trick} is not really important, and $\sqrt{\min_{\pair \in \pairs_0} \visits(\pair)}$ in \eqref{equation_square_trick} could be replaced by $(\min_{\pair \in \pairs_0} \visits(\pair))^\alpha$ for arbitrary $\alpha \in (0, 1)$. 

\subsection{The exploration test: Testing if information is lacking or not}
\label{section_exploitation}

The last component that we need to discuss is how the algorithm should decide whether information is lacking or not. 
By \cite[Corollary~7]{boone_regret_2025}, provided that $\learner$ is a consistent learning algorithm, for all $\model^\dagger \in \confusing(\model)$, we have
\begin{equation}
\label{equation_information_idea}
    \EE^{\model, \learner} \brackets*{
        \sum_{\pair \in \pairs}
        \visits_T(\pair)
        \KL(\model(\pair)||\model^\dagger(\pair))
    }
    \ge
    \log(T) + \oh(\log(T))
    .
\end{equation}
\Cref{equation_information_idea} translates the idea that consistent learning algorithms must gather enough information in order to ``reject'' confusing model.
The term $\EE^{\model, \learner} [\sum_{\pair \in \pairs} \visits_T(\pair) \KL(\model(\pair)||\model^\dagger (\pair)]$ is an expected log-likelihood ratio.
One recognizes in $\sum_{\pair \in \pairs} \visits_T (\pair) \KL(\model(\pair)||\model^\dagger(\pair))$ the dominant term in the likelihood that $\model^\dagger$ is to generate empirical data resembling $\model(\pair)$ after $\visits_T (\pair)$ samples at the pair $\pair \in \pairs$ for every $\pair \in \pairs$, as stated by Sanov's theorem.
Overall, a good proxy for the above is to state that $\learner$ lacks information under history $\History_T$ if
\begin{equation}
\label{equation_exploration_idea}
    \inf_{\model^\dagger \in \confusing(\model)}
    \braces*{
        \sum_{\pair \in \pairs}
        \visits_{T}(\pair) 
        \KL(\model(\pair)||\model^\dagger(\pair))
    }
    < (1 + f(\History_T)) \log(T)
\end{equation}
where $f : \histories \to \RR_+$ is some well-chosen vanishing function that accounts for second order errors in Sanov's theorem. 
Therefore, we will require \ECOE{} to explore ``often enough'' whenever information is lacking in the sense of \eqref{equation_exploration_idea}.

\subsection{The algorithmic scheme of \ECOE{}}

Combining the ideas of \Cref{section_exploration,section_coexploration,section_exploitation}, we end up with an algorithmic scheme that we name \ECOE{} and summarize in \Cref{algorithm_ecoe_informal}.

\begin{algorithm}[h]
\begin{algorithmic}[1]
    \FOR{phases $k=1,2,\ldots$}
        \STATE Estimate optimal pairs $\optpairs(\model)$ out of empirical data;
        \STATE Compute the exploitation policy $\policy^+$ as uniform over $\optpairs(\model)$, see \Cref{section_exploitation}, (1);
        \STATE Compute the exploration policy $\policy^-$ by estimating $\regretlb(\model)$ as in \eqref{equation_regretlb_2};
        \IF{
            $\State_t$ is not recurrent under $\policy^+$ or lacking information in the sense of \eqref{equation_exploration_idea}
        }
            \STATE Explore with $\policy^-$ one step;
            \hfill \textit{$\triangleright$ Vanilla exploration}
        \ELSIF{
            \eqref{equation_square_trick} one component of $\optpairs(\model)$ is sub-visited 
        }
            \STATE Explore with $\policy^-$ one step;
            \hfill \textit{$\triangleright$ Exploration travel triggered by co-exploration}
        \ELSE 
            \STATE Exploit by iterating $\policy^+$ until regeneration, see \Cref{section_exploitation} (2);
            \hfill \textit{$\triangleright$ Exploitation}
        \ENDIF
    \ENDFOR
\end{algorithmic}
\caption{
\label{algorithm_ecoe_informal}
    The \ECOE{} framework, informal. 
}
\end{algorithm}

The algorithm \ECOE{} works into phases.
For every phase, \ECOE{} estimates the set of optimal pairs $\optpairs(\model)$, an exploitation policy $\policy^+$ and an exploration policy $\policy^-$.
If \ECOE{} is outside of the recurrent components of $\policy^+$, it explores.
Otherwise, \ECOE{} checks if it is lacking information, in which case it explores by iterating $\policy^-$ once.
Otherwise, \ECOE{} checks if information is unbalanced on $\optpairs(\model)$, in which case it tries to reach the least visited component using the exploration policy $\policy^-$---it would be better to use a specifically designed policy $\policy^\pm$ that travels within the environment more efficiently than $\policy^-$, and this is definitively something to take into consideration for any serious implementation of \ECOE{}, but the regret analysis will be easier this way. 
Otherwise, the current amount of information is considered to be enough everywhere, and \ECOE{} exploits by playing $\policy^+$ until regeneration.

    \section{Estimating the regret lower bound $\regretlb(\model)$ out of noise}
    \allowdisplaybreaks

There is a fundamental issue with the algorithmic scheme \ECOE{} described in \Cref{algorithm_ecoe_informal}.

\Cref{algorithm_ecoe_informal} aims for asymptotically optimal regret by approximating the regret lower bound frontally.
At the beginning of every phase, \ECOE{} estimates the optimal pairs $\optpairs(\model)$ that it later splits into communicating components, computes an exploitation policy and computes an exploration policy by estimating $\regretlb(\model)$. 
This is where the issue lies: Obviously, \ECOE{} does not have access to $\model$, so all these estimations must be done out of empirical data.

The naive approach is to estimate $\optpairs(\model)$ and $\regretlb(\model)$ via the maximum likelihood estimator of $\model$, i.e., as $\optpairs(\hat{\model})$ and $\regretlb(\hat{\model})$ respectively.
This approach fails completely. 
Indeed, the set of optimal pairs $\optpairs(\model)$ and the regret lower bound $\regretlb(\model)$ are both discontinuous functions of $\model$ and $\model \approx \hat{\model}$ does not guarantee that $\optpairs(\model) \approx \optpairs(\hat{\model})$ nor $\regretlb(\model) \approx \regretlb(\hat{\model})$. 
So, in order to estimate optimal pairs $\optpairs(\model)$ and the regret lower bound $\regretlb(\model)$ out of the empirical estimate $\hat{\model}$, we need smooth approximates of these two; This is the aim of this section.
We begin by describing the discontinuity issue in \Cref{section_discontinuity}, then define regularized versions of $\optpairs(\model)$ and $\regretlb(\model)$ in \Cref{section_leveling,section_definition_regularized}.
The well-behavior of our regularized lower bound is described in \Cref{section_properties_regularized}.

\subsection{Difficulties due to the discontinuity of the lower bound}
\label{section_discontinuity}

The discontinuity of the lower bound is due to the discontinuity of $\weakoptimalpairs(\model)$, the set of weakly optimal pairs (\Cref{definition_classification_pairs}), therefore producing discontinuities of optimal pairs $\optpairs(\model)$ and of the confusing set $\confusing(\model)$.
To be fair, such discontinuities are already present in the simpler settings of multi-armed bandits and ergodic Markov decision processes.
Yet, nobody ever complained about it.
As a matter of fact, algorithms such as \texttt{IMED} of \cite{honda_non_asymptotic_2015}, \texttt{KLUCB} of \cite{cappe_kullback_2013} or \texttt{TS} of \cite{thompson_likelihood_1933,agrawal_analysis_2012,kaufmann_thompson_2012} (for bandits) and \texttt{IMED-RL} \cite{pesquerel_imed_rl_2022} (for ergodic Markov decision processes) all are asymptotically optimal and none need to handle the discontinuity of $\regretlb(\model)$ like we do in this work. 
But because of the design of \ECOE{}, we have no choice. 
Indeed, \ECOE{} requires the computation of $\regretlb(\model)$ and $\optpairs(\model)$ for every of its three components.
For exploration purposes, \ECOE{} approximates a near-optimal exploration policy using \eqref{equation_regretlb_2} via the computation of $\regretlb(\model)$;
For co-exploration purposes, \ECOE{} explicitly needs $\optpairs(\model)$ to balance the visits counts on components of $\optpairs(\model)$ via the square trick \eqref{equation_square_trick};
For exploitation purposes, \ECOE{} explicitly needs $\optpairs(\model)$ because its exploitation policy is uniform on $\optpairs(\model)$. 

So, these discontinuities must be dealt with by regularizing $\regretlb(\model)$ and $\optpairs(\model)$. 
In \Cref{section_example_discontinuity}, we discuss an example of discontinuity to motivate our regularization solution.
Especially, we stress that it all boils down to the smoothing of $\optpairs(\model)$. 
The regularization principle is stated in \Cref{section_principle_regularization} and is extensively discussed in the next \Cref{section_leveling}.

\subsubsection{An example of discontinuity with deterministic  transtions}
\label{section_example_discontinuity}

\begin{figure}[h]
    \centering
    \begin{tikzpicture}
        \node at (0, 0) {$\model$};

        \node[draw,circle] (1) at (-1.5, 0) {$1$};
        \node[draw,circle] (2) at (+1.5, 0) {$2$};

        \node[draw,circle,fill=Crimson,color=Crimson,minimum size=1mm,label=center:{\color{white}$\boldsymbol{*}$}] (a) at (-1, 0.43) {};
        \node[draw,circle,fill=Crimson,color=Crimson,minimum size=1mm,label=center:{\color{white}$\boldsymbol{\dagger}$}] (b) at (1, -.43) {};
        \node[draw,circle,fill=Crimson,color=Crimson,minimum size=1mm,label=center:{\color{white}$\boldsymbol{\ddagger}$}] (c) at (1.5, 1) {};
        \node[draw,circle,fill=black,minimum size=1mm,label=center:{\color{white}$\boldsymbol{\S}$}] (d) at (-1.5, 1) {};

        \draw[->,>=stealth,color=Crimson] (1) to (a) to[bend left] node[midway, above] {\scriptsize $0.5$} (2);
        \draw[->,>=stealth,color=Crimson] (2) to (b) to[bend left] node[midway, below] {\scriptsize $0.5$} (1);
        \draw[->,>=stealth,color=Crimson] (2) to (c) to[out=0,in=45] node[midway, right] {\scriptsize $0.5$} (2);
        \draw[->,>=stealth] (1) to (d) to[out=180,in=135] node[midway, left] {\scriptsize $0.1$} (1);
    \end{tikzpicture}
    \hspace{3em}
    \begin{tikzpicture}
        \node at (0, 0) {$\model_\theta$};
        
        \node[draw,circle] (1) at (-1.5, 0) {$1$};
        \node[draw,circle] (2) at (+1.5, 0) {$2$};

        \node[draw,circle,fill=black,minimum size=1mm,label=center:{\color{white}$\boldsymbol{*}$}] (a) at (-1, 0.43) {};
        \node[draw,circle,fill=black,minimum size=1mm,label=center:{\color{white}$\boldsymbol{\dagger}$}] (b) at (1, -.43) {};
        \node[draw,circle,fill=Crimson,color=Crimson,minimum size=1mm,label=center:{\color{white}$\boldsymbol{\ddagger}$}] (c) at (1.5, 1) {};
        \node[draw,circle,fill=black,minimum size=1mm,label=center:{\color{white}$\boldsymbol{\S}$}] (d) at (-1.5, 1) {};

        \draw[->,>=stealth] (1) to (a) to[bend left] node[midway, above] {\scriptsize $0.5 - \theta$} (2);
        \draw[->,>=stealth] (2) to (b) to[bend left] node[midway, below] {\scriptsize $0.5 - \theta$} (1);
        \draw[->,>=stealth, color=Crimson] (2) to (c) to[out=0,in=45] node[midway, right] {\scriptsize $0.5$} (2);
        \draw[->,>=stealth] (1) to (d) to[out=180,in=135] node[midway, left] {\scriptsize $0.1$} (1);
    \end{tikzpicture}%
    \caption{
    \label{figure_regret_discontinuity}
        An example of discontinuity of the regret lower bound.
        The displayed transitions are deterministic and the labels represent the means of the attached Bernoulli rewards.
        Actions are distinguished by unique symbols for better readability of the contraction. 
        Optimal pairs are colored in \textcolor{Crimson}{\bf red}.
    }
\end{figure}
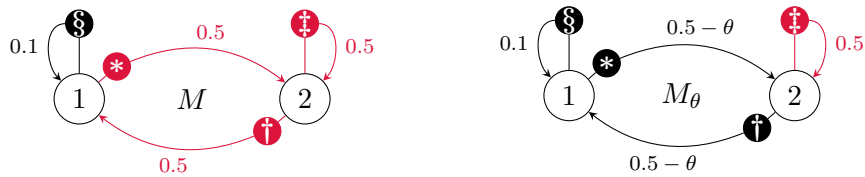

In this paragraph, we provide an explicit example with a discontinuity of $\optpairs(\model)$ and of $\regretlb(\model)$, see \Cref{figure_regret_discontinuity}. 

In \Cref{figure_regret_discontinuity}, we present two models $\model$ and $\model_\theta$ that are statistically indistinguishable when $\theta \to 0$. 
On $\model \equiv \model_0$, the optimal pairs are $\optpairs(\model) = \braces{(1, *), (2, \dagger), (2, \ddagger)}$ and are displayed in red.
On $\model_\theta$ for $\theta > 0$, the optimal pairs are $\optpairs(\model_\theta) = \braces{(2, \dagger)}$, also displayed in red.
So, we observe a discontinuity of $\optpairs$ at $\theta = 0$. 
Note that this discontinuity drastically changes the way the model can be navigated.
For $\model \equiv \model_0$, the state $1$ can be reached at any time with zero-cost but for $\model_\theta$, it can only be reached by playing sub-optimal pairs. 
In addition, this discontinuity of optimal pairs produces a discontinuity in confusing sets, since $\model_\theta \notin \confusing(\model)$ but $\model \in \confusing(\model_\theta)$. 
Taking the ambient set $\models := \braces{\model_\theta : \theta \in [-0.5, 0.5]}$ with Bernoulli rewards, we can derive the regret lower bounds in closed form:
\begin{equation*}
    \regretlb(\model)
    =
    \frac {4}{10~\kl(0.1, 0.5)}
    \quad\text{and}\quad
    \regretlb(\model_\theta)
    \underset{\theta \to 0^+}{\sim}
    \frac{\theta}{\kl(0.5 - \theta, 0.5)}
    \underset{\theta \to 0^+}{\sim}
    \frac 1{2\theta}
\end{equation*}
where $\kl(x, y) := x \log(\frac xy) + (1-x) \log(\frac{1-x}{1-y})$ is the Kullback-Leibler divergence between $\mathrm{Ber}(x)$ and $\mathrm{Ber}(y)$. 
So, $\regretlb$ is discontinuous at $\model \equiv \model_0$.
Because $\model$ and $\model_\theta$ are indistinguishable as $\theta \to 0$, it follows that when the underlying model is $\model$, no learner is capable of claiming that the current model isn't $\model_\theta$ for $\theta > 0$ with overwhelming confidence. 
Yet, in this example, when one is doubting about whether the model is $\model_\theta$ or $\model$, it means that the pairs $(1, *)$ and $(2, \dagger)$ are nearly optimal.
In that case, these pairs can arguably be used to navigate the model more easily while scoring near optimally. 
After all, if the true model is some $\model_\theta$ for $\theta > 0$ rather than $\model_0$, then the optimality of $(1, *)$ and $(2, \dagger)$ will be ruled out at some point. 

\subsubsection{A need for a notion of near-optimal pairs}
\label{section_principle_regularization}

From this discussion, we make the following observations.

\begin{ronum}
    \item If the optimality of a pair is uncertain, it should be considered optimal: Navigation is made easier and if the pair is actually sub-optimal, its optimality will be rejected eventually.
\end{ronum}

By following the above principle, we will solve the discontinuity problems due to the navigation constraints as well as those due to the discontinuity of the confusing set. 

\subsection{Retrieving optimal pairs under noise via the leveling transform}
\label{section_leveling}

To smooth the regret lower bound, the central technique is to level the Markov decision process to make all near-optimal pairs simultaneously optimal, so that $\optpairs(\model)$ can be retrieved from $\model' \approx \model$. 
The idea of the leveling operation (\Cref{definition_leveled_mdp}) is to increase the rewards at pairs that are nearly optimal by just the right amount. 
By ``nearly optimal'', we mean pairs with Bellman gaps below a given threshold $\epsilon > 0$. 

\begin{definition}[Leveling transform]
\label{definition_leveled_mdp}
    Let $\model \equiv (\pairs, \kernel, \reward)$ a Markov decision process and fix $\epsilon > 0$.
    The \strong{$\epsilon$-leveled} transform of $\model$ is the model $\flatmodel{\model}{\epsilon} \equiv (\pairs, \kernel, \flatmodel{\reward}{\epsilon})$ with reward function 
    \begin{equation*}
        \flatmodel{\reward}{\epsilon}(\pair) 
        :=
        \reward(\pair)
        + 
        \indicator{\ogaps(\pair; \model) < \epsilon}
        \ogaps(\pair; \model).
    \end{equation*}
\end{definition}

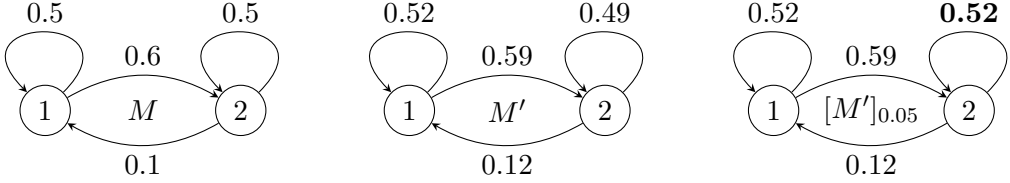
\begin{figure}[ht]
    \centering
    \begin{tikzpicture}
        \node at (0, 0) {$M$};
        \node[state] (1) at (-1.3, 0) {$1$};
        \node[state] (2) at ( 1.3, 0) {$2$};
        \draw[transition] (1) to[bend left] node[midway, above] {$0.6$} (2);
        \draw[transition] (2) to[bend left] node[midway, below] {$0.1$} (1);
        \draw[transition, loop] (1) to node[midway, above] {$0.5$} (1);
        \draw[transition, loop] (2) to node[midway, above] {$0.5$} (2);
    \end{tikzpicture}%
    \hspace{-1em}%
    \begin{tikzpicture}
        \node at (0, 0) {$M'$};
        \node[state] (1) at (-1.3, 0) {$1$};
        \node[state] (2) at ( 1.3, 0) {$2$};
        \draw[transition] (1) to[bend left] node[midway, above] {$0.59$} (2);
        \draw[transition] (2) to[bend left] node[midway, below] {$0.12$} (1);
        \draw[transition, loop] (1) to node[midway, above] {$0.52$} (1);
        \draw[transition, loop] (2) to node[midway, above] {$0.49$} (2);
    \end{tikzpicture}%
    \hspace{-1em}%
    \begin{tikzpicture}
        \node at (0, 0) {$\flatmodel{M'}{0.05}$};
        \node[state] (1) at (-1.3, 0) {$1$};
        \node[state] (2) at ( 1.3, 0) {$2$};
        \draw[transition] (1) to[bend left] node[midway, above] {$0.59$} (2);
        \draw[transition] (2) to[bend left] node[midway, below] {$0.12$} (1);
        \draw[transition, loop] (1) to node[midway, above] {$0.52$} (1);
        \draw[transition, loop] (2) to node[midway, above] {\bf 0.52} (2);
    \end{tikzpicture}
    \caption{
    \label{figure_example_leveling}
        An example of leveling transform.
        On the left, a Markov decision process $\model$ with two states and two actions per state.
        Transitions are deterministic and represented by arrows. 
        Labels are rewards. 
        The model $\model'$ is a noisy version of $\model$. 
    }
\end{figure}

An example of the leveling operation is provided in \Cref{figure_example_leveling}.
For the model $\model$ on \Cref{figure_example_leveling}, the two optimal pairs correspond to the two loops $1 \to 1$ and $2 \to 2$ with associated gain $\optgain = 0.5$. 
In the noisy copy $\model'$, only the left loop $1 \to 1$ is optimal with gain $\optgain = 0.52$
After leveling $\model'$ with threshold $\epsilon = 0.05$, the right loop $2 \to 2$ becomes optimal again and the optimal gain is $\optgain = 0.52$. 
So, $\model$ and $\flatmodel{\model}{\epsilon}$ have the same optimal pairs and gain optimal policies.
On this example, we observe that leveling with the right threshold recovers the set of optimal pairs and of gain optimal policies. 

Motivated by this example, we define the $\epsilon$-leveled optimal pairs as the optimal pairs of the $\epsilon$-leveled transform. 

\begin{definition}[Leveled optimal pairs]
\label{definition_near_optimal_pairs}
    Let $\model \equiv (\pairs, \kernel, \reward)$ be a Markov decision process and fix $\epsilon > 0$.
    The \strong{$\epsilon$-leveled optimal pairs} of $\model$ are $\flatme{\optpairs}{\epsilon}(\model) := \optpairs(\flatmodel{\model}{\epsilon})$. 
\end{definition}

In \Cref{proposition_continuity_near_optimality}, we show that this leveling operation works as intended. 
Provided that $\epsilon > 0$ is small enough, that $\model'$ is close enough to $\model$ relatively to a well-chosen metric (\Cref{definition_stared_distance}) and that $\epsilon$ is large enough relatively to the distance between $\model'$ and $\model$, then the optimal structure of the $\epsilon$-leveled transform of $\model'$ is the same as $\model$ at first order: $\flatmodel{\model'}{\epsilon}$ and $\model$ have same gain optimal optimal policies and the same optimal pairs. 

\begin{definition}[Distances between models]
\label{definition_stared_distance}
    For $\model \equiv (\pairs, \kernel, \rewardd)$ and $\model' \equiv (\pairs, \kernel', \rewardd')$ two Markov decision processes with the same pair space, we introduce the following distances:
    \begin{equation*}
    \begin{aligned}
        \norm{\model' - \model} 
        & := \max_{\pair \in \pairs} \braces*{
            \abs{\reward'(\pair) - \reward(\pair)} 
            + \norm{\kernel'(\pair) - \kernel(\pair)}_1
        };
        \quad \text{and}
        \\
        \snorm{\model' - \model} 
        & := \max_{\pair \in \pairs} \braces*{
            \abs{\reward'(\pair) - \reward(\pair)} 
            + \norm{\kernel'(\pair) - \kernel(\pair)}_1
            + \infty \cdot \indicator{\kernel'(\pair) \not\sim \kernel(\pair)}
        }.
    \end{aligned}
    \end{equation*}
    The operator $\snorm{-}$ will be called the \strong{support-aware norm}.
\end{definition}

In \Cref{proposition_continuity_near_optimality} downstream, we express the properties of the leveling transform relatively to two classical quantities of the Markov decision process. 
The first is the \strong{gain gap}, given by $\gaingap(\model) := \dmin \braces*{\norm{\optgain(\model) - \gainof{\policy}(\model)}_\infty : \policy \in \policies}$, that is the smallest gap between the gain of optimal policies and the gain of sub-optimal policies.
The second if the \strong{worst diameter} $\wdiameter(\model) := \max_{\policy \in \policies} \diameter(\policy; \model)$ where $\diameter(\policy; \model)$ is the policy diameter of $\policy$ (\Cref{definition_policy_diameter}), that is the largest diameter over all possible policies.

\begin{proposition}[Fundamental properties of leveling]
\label{proposition_continuity_near_optimality}
    For every communicating Markov decision process $\model \equiv (\pairs, \kerneld, \rewardd)$, there exists a quantity $C(\model) \in \RR_+$, that is a rational function of $\gaingap(\model)$ and $\worstdiameter(\model)$, such that for all $\epsilon > 0$ and $\model' \in \models$ satisfying
    \begin{equation}
        \epsilon + 2 C (\model) \snorm{\model' - \model} < \gaingap(\model)
        \quad\text{and}\quad
        C (\model)
        \snorm{\model' - \model}
        < \epsilon
        ,
    \end{equation}
    we have
    \begin{enum}
        \item $\flatmodel{\model'}{\epsilon}$ and $\model$ have the same gain optimal policies;
        \item $\optpairs(\flatmodel{\model'}{\epsilon}) = \optpairs(\model)$, i.e., $\flatme{\optpairs}{\epsilon}(\model') = \optpairs(\model)$; 
    \end{enum}
\end{proposition}

The proof of \Cref{proposition_continuity_near_optimality} is deferred to \Cref{appendix_near_optimality}. 

\subsection{The regularized lower bound \texorpdfstring{$\regretlb_\EPSILON (\model)$}{Ke(M)}}
\label{section_definition_regularized}

With the properties of the leveling transform in hand, we can describe the regularized lower bound $\regretlb_\epsilon(\model)$, see \Cref{definition_regularized_regret_lower_bound}. 
We combine three regularization mechanisms. 

\begin{ronum}
    \item 
        Approximating the confusing set by the $\epsilon$-leveled confusing set (\Cref{definition_near_confusing_set}) via the leveling transform: While the confusing set cannot be estimated under noise because of discontinuity issues, its leveled version is continuous.
    \item 
        Approximating invariant measures by $\epsilon$-uniform invariant measures (\Cref{definition_uniform_invariant_measure}): Optimizing over uniform invariant measures is encouraged by the formulation \eqref{equation_regretlb_2} of $\regretlb(\model)$ and is key to the smoothness properties of the regularized $\regretlb_\epsilon (\model)$.  
        Also, using fully supported exploration policies guarantees that during exploration, the set of pairs is uniformly covered, which is helpful in the regret analysis.
    \item 
        Adding a strongly convex regularizer to the objective function $\sum_{\pair \in \pairs} \imeasure(\pair) \ogaps(\pair)$ of \eqref{equation_definition_regretlb}: This makes the optimizer unique. 
        This one is rather useful than necessary.
\end{ronum}

Uniform invariant measures and leveled confusing sets are defined as follows.

\begin{definition}[Uniform invariant measure]
\label{definition_uniform_invariant_measure}
    For $\epsilon > 0$, an \strong{$\epsilon$-uniform invariant measure} of a model $\model$ is an element $\imeasure \in \imeasures(\model)$ such that $\imeasure(\state, \action) \ge \epsilon \sum_{\action' \in \actions(\state)} \imeasure(\state, \action')$ for all $\state \in \states$. 
    The set of $\epsilon$-uniform invariant measures is denoted $\imeasures(\epsilon; \model)$.
\end{definition}

\begin{definition}[Leveled confusing set]
\label{definition_near_confusing_set}
    Fix $\models$ a space of Markov decision processes and fix $\epsilon > 0$.
    The \strong{$\epsilon$-leveled confusing set} of a communicating model $\model \in \models$ is
    \begin{equation*}
        \confusing(\epsilon; \model)
        \equiv
        \confusing_\epsilon(\model)
        :=
        \braces*{
            \model^\dagger \in \models
            :
            {
                \model^\dagger \gg \model,
                \model^\dagger = \model \mathrm{~on~} \flatme{\optpairs}{\epsilon}(\model) 
                \atop
                \mathrm{~and~} \optpolicies(\model^\dagger) \cap \optpolicies(\model) = \emptyset
            }
        }
        .
    \end{equation*}
\end{definition}

It can be shown that uniform invariant measures of communicating Markov decision processes are fully supported, see \Cref{appendix_invariant_measures}.
Also, note that only difference between the leveled confusing set $\confusing_\epsilon(\model)$ and the vanilla version $\confusing(\model)$ is that elements of $\confusing_\epsilon(\model)$ are required to coincide with $\model$ on $\flatme{\optpairs}{\epsilon}(\model)$ rather than $\optpairs(\model)$.
As one can show that $\flatme{\optpairs}{\epsilon}(\model)$ satisfies $\flatme{\optpairs}{\epsilon}(\model) \supseteq \optpairs(\model)$, we have $\confusing_\epsilon(\model) \subseteq \confusing(\model)$ for all $\epsilon \ge 0$. 
Moreover, by \Cref{proposition_continuity_near_optimality}, for $\epsilon < \gaingap(\model)$, we have $\flatme{\optpairs}{\epsilon}(\model) = \optpairs(\model)$ so that $\confusing_\epsilon(\model) = \confusing(\model)$. 

We have everything in hand to give a complete description of the regularized regret lower bound. 
The regularized version (\Cref{definition_regularized_regret_lower_bound}) uses a \strong{regularization hyperparameter} $\EPSILON \equiv (\epsilonflat, \epsilonunif, \epsilonreg) \in \RR_+^3$ to tune the different regularization components (I--III). 
(I)~The first $\epsilonflat$ is the \strong{leveling regularizer} and tunes the leveling threshold used to estimate the confusing set;
(II)~The second $\epsilonunif$ is the \strong{uniformization regularizer} and tunes how uniform are the invariant measures that we are optimizing over; and
(III)~The third $\epsilonreg$ is the \strong{convexification regularizer} that makes the objective function $\epsilonreg$-strongly convex.

\begin{definition}[Regularized regret lower bound]
\label{definition_regularized_regret_lower_bound}
    Fix $\models$ a space of Markov decision processes and fix $\EPSILON \equiv (\epsilonflat, \epsilonunif, \epsilonreg) \in \RR_+^3$ a regularization hyperparameter. 
    The \strong{$\EPSILON$-regularized regret lower bound} is the solution $\regretlb_\EPSILON(\model) \in [0, \infty]$ of the optimization problem below:
    \begin{equation*}
        \inf
        \sum_{\pair \in \pairs}
        \imeasure(\pair)
        \ogaps(\pair)
        + \epsilonreg \norm{\imeasure}_2^2
        \quad \mathrm{s.t.} \quad
        \begin{cases}
            & \displaystyle
            \imeasure \in \imeasures(\epsilonunif; \model)
            , \quad \mathrm{and}
            \\
            & \displaystyle
            \inf_{\model^\dagger \in \confusing(\epsilonflat; \model)}
            \braces*{
                \sum_{\pair \in \pairs}
                \imeasure(\pair)
                \KL(\model(\pair)||\model^\dagger(\pair))
            }
            \ge
            1.
        \end{cases}
        \hspace{-0.33em}
    \end{equation*}
\end{definition}

Observe that $\regretlb_{\EPSILON}(\model)$ is the solution of a convex optimization problem, where the objective function is $\epsilonreg$-strongly convex and the constraints spawn a closed convex cone of $\RR_+^\pairs$. 
Using these properties, we show in \Cref{appendix_regretlb_bound,appendix_regretlb_existence} that $\regretlb_\EPSILON (\model) < \infty$ and that there exists an optimizer in $\imeasures(\epsilonunif; \model)$.
When the convexification is strict with $\epsilonreg > 0$, that optimizer is unique, called the \strong{$\EPSILON$-optimal exploration measure} and denoted $\imeasure^*_\EPSILON$.

\subsection{Properties of the regularized lower bound}
\label{section_properties_regularized}

In this paragraph, we overview the different properties of the regularized regret lower bound.
In \Cref{proposition_regretlb_approximation}, we state that $\regretlb_\EPSILON (\model)$ converges to $\regretlb(\model)$ as $\EPSILON \to 0$, meaning that the regularized lower bound is a good approximation of the true lower bound when the regularizer is small enough. 
In \Cref{proposition_regretlb_convergence_solution}, we state that the associated optimizer $\imeasure^*_\EPSILON$ converges to a special optimizer of $\regretlb(\model)$ (in the formulation of \eqref{equation_definition_regretlb}) as the components of the regularizer $\EPSILON$ vanish at the right speed; In particular, the uniformization regularizer must vanish faster than the convexification regularizer. 

\begin{proposition}[Approximation of $\regretlb(\model)$]
\label{proposition_regretlb_approximation}
    Let $\models$ be a space of Markov decision processes. 
    For all $\model \in \models$ that is communicating, we have
    \begin{equation*}
        \regretlb(\model) 
        \le \regretlb_\EPSILON (\model)
        \le \regretlb(\model) + \OH\parens*{\norm{\epsilon}}
        \quad \mathrm{as} \quad
        \norm{\EPSILON} \to 0.
    \end{equation*}
\end{proposition}

\begin{proposition}[Convergence of $\imeasure^*_\EPSILON$]
\label{proposition_regretlb_convergence_solution}
    Let $\models$ be a space of Markov decision processes and let $\model \in \models$ be a communicating model.
    Let $\imeasure^*(\model) \in \imeasures(\model)$ be the (unique) optimizer of $\regretlb(\model)$ in the sense of \eqref{equation_definition_regretlb} that further minimizes $\norm{\imeasure}_2$ among optimizers. 
    We have
    \begin{equation*}
        \norm*{\imeasure^*_\EPSILON(\model) - \imeasure^*(\model)}_2 \to 0
        \quad\mathrm{as}\quad
        \frac{\epsilonunif}{\epsilonreg} + \epsilonreg + \epsilonflat \to 0
        .
    \end{equation*}
\end{proposition}

More precise statements of \Cref{proposition_regretlb_approximation} and \Cref{proposition_regretlb_convergence_solution}, together with proofs, can be bound in \Cref{appendix_regretlb_continuity}, see \Cref{proposition_regularized_lowerbound_approximates} and \Cref{proposition_convergence_optimal_measure} respectively. 

Regarding the continuity properties of $\regretlb_\EPSILON(\model)$ with respect to $\model$, we add further assumptions on the structure of the ambient space of Markov decision processes $\models$.
Firstly, we assume that $\models$ is a space of models with Bernoulli rewards (\Cref{assumption_bernoulli}).
Doing so, the simple stared norm $\snorm{-}$ from \Cref{definition_stared_distance} is adequate and we do not run into topological troubles---the result is probably easy to generalize to single parameter exponential families, but we have chosen to keep everything as simple as possible. 
Secondly, we assume that $\models$ is in product form $\models \equiv \product_{\pair \in \pairs} (\kernels'_\pair \times \rewards'_\pair)$ (\Cref{assumption_space}), where $\kernels'_\pair \subseteq \probabilities(\states)$ and $\rewards'_\pair = (0, 1)$.
That is, $\models$ is a product and all rewards are $(0, 1)$. 
Under these two assumptions, we show in \Cref{proposition_regretlb_continuity} that $\regretlb_\EPSILON(\model)$ and $\imeasure^*_\EPSILON(\model)$ are both continuous relatively to the topology induced by $\snorm{-}$. 

\begin{assumption}
\label{assumption_bernoulli}
    For all $\model \in \models$, $\model$ has Bernoulli rewards.
\end{assumption}

\begin{assumption}
\label{assumption_space}
    The space $\models$ is of the form $\models \equiv \product_{\pair \in \pairs} (\probabilities'_\pair \times (0, 1))$ where $\probabilities'_\pair \subseteq \probabilities(\states)$. 
\end{assumption}

\begin{proposition}[Continuity of $\regretlb_\EPSILON(\model)$ and $\imeasure^*_\EPSILON$]
\label{proposition_regretlb_continuity}
    Let $\models$ be a space of Markov decision processes and let $\model \in \models$ be a communicating model.
    Assume that rewards are Bernoulli (\Cref{assumption_bernoulli}) and that $\models$ is in product form (\Cref{assumption_space}). 
    Let $\EPSILON \equiv (\epsilonflat, \epsilonunif, \epsilonreg) \in (\RR_+^*)^3$ be a positive regularization hyperparameter.
    If $\epsilonreg < \gaingap(\model)$, then, in $\models$, we have
    \begin{equation*}
        \regretlb_\EPSILON(\model) =
        \lim_{\snorm{\model' - \model} \to 0}
        \regretlb_\EPSILON(\model')
        \quad\text{and}\quad
        \imeasure^*_\EPSILON (\model) =
        \lim_{\snorm{\model' - \model} \to 0}
        \imeasure^*_\EPSILON (\model')
    \end{equation*}
\end{proposition}

A more precise statement can be found in \Cref{appendix_regretlb_continuity}, see \Cref{proposition_continuity_regularized_lower_bound}. 

Combining \Cref{proposition_regretlb_approximation,proposition_regretlb_convergence_solution,proposition_regretlb_continuity}, it means that given an approximation $\hat{\model}$ of $\model$ and a regularizer $\EPSILON \ll 1$ such that $\snorm{\hat{\model} - \model} \ll \norm{\EPSILON}$, we have $\regretlb_\EPSILON (\hat{\model}) \approx \regretlb_\EPSILON(\model) \approx \regretlb(\model)$ and $\imeasure^*_\EPSILON(\hat{\model}) \approx \imeasure^*_\EPSILON(\model) \approx \imeasure^*(\model)$.
Accordingly, we can estimate the lower bound $\regretlb(\model)$ and a good exploration policy out of empirical data. 

Now, we can provide a complete description of \texttt{ECoE}.

    \section{\texttt{ECoE*}, an algorithm with asymptotically optimal regret}
    \label{section_final_ecoe}

For the clarity of the exposition, we focus on an instance of \texttt{ECoE} that ignores the computational weight of its subroutines. 
Ideally, \texttt{ECoE} eventually estimate the regret lower bound $\regretlb(\model)$, an optimal exploration measure $\imeasure^*$ and check dynamically when information is lacking. 
All three problems are computationally difficult (NP-hard in some cases) and solving them requires a dedicate care that is too heavy to carry for a first regret analysis. 
That is why, for this ``oracle-based'' version of \texttt{ECoE} called \texttt{ECOE*}, we shall assume that we are able to compute the regret lower bound $\regretlb(-)$, optimal exploration measures $\imeasure^*$ and to run exploration tests \eqref{equation_exploration_idea} just by snapping fingers.

The pseudo-code of \texttt{ECoE*} is provided in \Cref{algorithm_ecoe_annotated}.

\begin{algorithm}[ht]
    \strong{Parameters:}

    A family of regularization parameters functions $\epsilonflat, \epsilonunif, \epsilonreg, \epsilontest$;

    \begin{algorithmic}[1]
        \STATE Initialize $\stimes^-, \stimes^\pm, \stimes^+, \stimes^+_0$ and $\stimes^!$ to $\emptyset$;
        \STATE Initialize $t \gets 1$;
        \FOR {$k = 1, 2, \ldots$}
            \STATE Set $t_k \gets t$;
            \STATE Update floored hyperparameters $\EPSILON'(t)$, see \eqref{equation_floored_regularization};
            \STATE Update maximum likelihood estimator $\hat{\model}_t$ for $\model$;
            \STATE Update co-exploration structure $\flatme{\optpairs}{\iepsilonflat'(t)}(\hat{\model}_t)$, split the structure into components $\structure_t := (\pairs_t^1, \ldots, \pairs_t^{c(t)})$, and let $i(t)$ be the component containing $\State_t$ (current component);%
            \STATE Update exploitation policy $\policy_{t_k}^+$ as given by \eqref{equation_exploitation_policy};
            \STATE Update exploration measure $\imeasure_{t_k} \gets \imeasure_{\EPSILON'(t)}^*(\hat{\model}_t)$ by computing $\regretlb_{\EPSILON'(t)}(\hat{\model}_t)$ using an oracle;%
            \STATE Update exploration policy by setting $\policy^-_{t_k}(\action|\state) \propto \imeasure_{t_k}(\state, \action)$;
            \STATE Update skeleton $\skeleton_t \gets \braces{\pair \in \pairs: \visits_t (\pair) \ge \log^2(t)}$;

            \IF {$\State_t$ is not recurrent under $\policy_t^+$ in $\hat{\model}_t$ or $\glrtest(\History_t)$ holds, see \eqref{equation_glr_exploration}}

                \STATE Play $\Action_t \sim \policy_t^- (-|\State_t)$;
                    \hfill $\triangleright$ \textit{vanilla exploration}
                \STATE Increment $t \gets t + 1$, add $t - 1$ in $\stimes^-$;
            \ELSIF {
                \eqref{equation_square_trick}
                $
                    ~
                    (\min \braces{
                        \visits_t (\pair) : \pair \in \pairs_t^{i(t)}
                    })^{1/2}
                    \ge
                    \min \braces{
                        \visits_t (\pair) : \pair \in \flatme{\optpairs}{\iepsilonflat'(t)}(\hat{\model}_t)
                    }
                $
            }
                \STATE Play $\Action_t \sim \policy_t^- (-|\State_t)$;
                    \hfill $\triangleright$ \textit{co-exploration}
                \STATE Increment $t \gets t + 1$, add $t - 1$ in $\stimes^-$ and in $\stimes^\pm$;
            \ELSE
                \STATE Add $t$ in $\stimes_0^+$;
                    \hfill $\triangleright$ \textit{exploitation until regeneration}
                \REPEAT
                    \STATE Play $\Action_t \sim \policy_t^+(-|\State_t)$;
                    \STATE Increment $t \gets t + 1$;
                    \STATE \strong{if} $\State_t \notin \states(\pairs_t^{i(t)})$ \strong{then} add $t - 1$ in $\stimes^!$ and \strong{break};
                    \hfill $\triangleright$ \textit{transition discovery}
                    \STATE \strong{else} add $t - 1$ in $\stimes^+$;
                \UNTIL {
                    $\State_t = \State_{t_k}$; 
                    \hfill
                    $\triangleright$ \textit{regeneration}
                }
            \ENDIF
        \ENDFOR
    \end{algorithmic}
    \caption{
    \label{algorithm_ecoe_annotated}
        \texttt{ECoE*}, an instantiation of \ECOE{} (\Cref{algorithm_ecoe_informal}) 
    }
\end{algorithm}

\subsection{Overview of \texttt{ECoE*}}

First, \texttt{ECoE*} requires a regularization hyperparameter $\EPSILON \equiv (\epsilonflat, \epsilonunif, \epsilonreg, \epsilontest) : \N \to \R_+^4$ that is a slowly decreasing function.
This hyperparameter is essential to the regularization of the lower bound $\regretlb(\model)$, that is discontinuous in $\model$ and that cannot be estimated under noise. 
In comparison to the regularization of $\regretlb(\model)$ in \Cref{definition_regularized_regret_lower_bound}, note the presence of a fourth hyperparameter $\epsilontest$, called the \strong{test hyperparameter}, that controls the power of exploration tests (see \Cref{section_exploration}).
The choice of $\EPSILON$ is discussed in \Cref{appendix_choice_regularization_hyperparameter}.

\Cref{algorithm_ecoe_annotated} works in phases $k = 1, 2, 3, \ldots$ that are very short.

At the beginning of a phase, it first updates the regularization parameter to a floored regularization $\epsilon'$ described in \Cref{appendix_choice_regularization_hyperparameter} and \eqref{equation_floored_regularization}.
Then, it updates the maximum likelihood estimator for $\model$, denoted $\hat{\model}_t$, as described in \Cref{section_estimator} and \eqref{equation_estimator}, and computes the leveled optimal pairs $\optpairs_{\iepsilonflat'}(\hat{\model}_t)$ using the leveling threshold of $\epsilonflat'(t)$ and deduces an exploitation policy $\policy_t^+$ and an exploration policy $\policy_t^-$. 
As motivated in \Cref{section_coexploration}, \texttt{ECoE*} chooses its exploitation policy as the policy playing leveled optimal pairs uniformly at random, if any, and actions uniformly at random if no element of $\optpairs_{\iepsilonflat'}(\hat{\model}_t)$ can be played right away, i.e., 
\begin{equation}
\label{equation_exploitation_policy}
    \policy_t^+ (-|\state) 
    :=
    \begin{cases}
        \mathrm{Unif}\parens*{\braces{
            \action \in \actions(\state)
            : 
            (\state, \action) \in \flatme{\optpairs}{\iepsilonflat'(t)}(\hat{\model}_t)
        }}
        & \text{if $\state \in \states(\flatme{\optpairs}{\iepsilonflat'(t)}(\hat{\model}_t))$}
        \\
        \mathrm{Unif}(\actions(\state))
        & \text{if $\state \notin \states(\flatme{\optpairs}{\iepsilonflat'(t)}(\hat{\model}_t))$}
    \end{cases}
\end{equation}
where we wrote $\states(\pairs') := \braces{ \state \in \states: \exists \action \in \actions(\state), (\state, \action) \in \pairs' }$ for $\pairs' \subseteq \pairs$. 
The exploration policy of \texttt{ECoE*} is the policy induced by the measure achieving $\regretlb_{\EPSILON(t)}(\hat{\model}_t)$, i.e.,
\begin{equation}
\label{equation_exploration_policy}
    \policy_t^- (\action|\state)
    := \frac{
        \imeasure_{\EPSILON'(t)}^* (\state, \action; \hat{\model}_t)
    }{
        \sum_{\action' \in \actions(\state)}
        \imeasure_{\EPSILON'(t)}^* (\state, \action'; \hat{\model}_t)
    }
\end{equation}
where $\imeasure^*_{\EPSILON'(t)}$ is the unique probability invariant measure of $\hat{\model}_t$ achieving $\regretlb_{\EPSILON'(t)}(\hat{\model}_t)$. 
Then, the algorithm decides whether 
\textcolor{\explorationcolor}{(1)} it \textcolor{\explorationcolor}{explores}, i.e., plays $\policy_t^-$ once to gather information on presumed sub-optimal pairs; or
\textcolor{\coexplorationcolor}{(2)} it \textcolor{\coexplorationcolor}{coexplores}, i.e., plays $\policy_t^-$ to reach a sub-visited components of nearly optimal pairs; or
\textcolor{\exploitationcolor}{(3)} it \textcolor{\exploitationcolor}{exploits}, i.e., plays $\policy_t^+$ to score as much as possible. 
More precisely:

\textcolor{\explorationcolor}{(0) \textsc{By-default exploration $(t \in \stimes^-$)}}:
If the algorithm is not on a component of the set of leveled optimal pairs $\optpairs_{\iepsilonflat'}(\hat{\model}_t)$, it iterates $\policy_t^-$ and ends the phase.
Otherwise, the algorithm moves to the exploration GLR test.

\textcolor{\explorationcolor}{(1) \textsc{Exploration $(t \in \stimes^-$)}}:
Following the ideas in \Cref{section_exploitation}, the algorithm checks  first if there is a lack of information on sub-optimal pairs by running a \strong{generalized log-likelihood ratio test} (GLR) that is truncated to a \strong{skeleton} $\skeleton_t$, that contains the pairs visited more than $\log^2(t)$ times. 
The GLR test for exploration is $\glrtest(\History_t)$ given by 
\begin{equation}
\label{equation_glr_exploration}
    \glrtest(\History_t)
    :
    \exists \model^\dagger\in \confusing(\hat{\model}_t)
    \text{~s.t}
    \begin{cases}
        \model^\dagger = \hat{\model}_t \text{~on~} \skeleton_t \cup \flatme{\optpairs}{\epsilonflat'(t)}(\hat{\model}_t)
        ; \text{~and~}
        \\ 
        \displaystyle
        \sum_{\pair \in \pairs}
        \visits_t(\pair)
        \KL(\hat{\model}_t(\pair)||\model^\dagger(\pair))
        \le 
        (1 + \epsilontest'(t)) \log(t)
    \end{cases}
    \!
\end{equation}
where $\epsilontest : \NN \to \RR_+$ is a well-chosen vanishing function.
If \eqref{equation_glr_exploration} is true, the algorithm iterates $\policy_t^-$ once and the phase ends.
Otherwise, the algorithm deals with co-exploration. 

\textcolor{\coexplorationcolor}{(2) \textsc{CoExploration $(t \in \stimes^\pm \cap \stimes^-$)}}:
As motivated in \Cref{section_coexploration}, to avoid missing information on nearly optimal pairs, the algorithms synchronizes the visit counts on all the communicating components of leveled optimal pairs $\optpairs_{\iepsilonflat'}(\hat{\model}_t)$.
Following the square trick \eqref{equation_square_trick}, if one component is visited less than the square root of the current component, the algorithm decides that this sub-visited component should be reached and attempts to do so by explorating: It iterates $\policy_t^-$ once and the phase ends.
The idea is to explore by the meantime and to reconsider the exploration-exploitation trade-off once another component of $\flatme{\optpairs}{\iepsilonflat'}(\hat{\model}_t)$ has been reached.
Otherwise, the amount of information on components of $\optpairs_{\iepsilonflat'}(\hat{\model}_t)$ is considered to be balanced enough, and the algorithm moves to exploitation.

\textcolor{\exploitationcolor}{(3) \textsc{Exploitation} $(t \in \stimes^+)$}:
The algorithm iterates $\policy_t^+$ until regeneration.

(!) \textsc{Panic $(t \in \stimes^!)$}: 
When the algorithm iterates the policy $\policy_t^+$ until regeneration, the regeneration may never come if the algorithm is wrong about the support of transition kernels.
In that case, at some point, the algorithm will observe a transition $(\State_t, \Action_t, \State_{t+1})$ that it has never seen before, with $\State_{t+1}$ outside of the current component of estimated optimal pairs.
We call these time-instants \strong{panic times}.
The algorithm deals with them by aborting the current phase right away.

\subsection{The regularization hyperparameter \texorpdfstring{$\EPSILON$}{eps}}
\label{appendix_choice_regularization_hyperparameter}

The regularization hyperparameter $\EPSILON \equiv (\epsilonflat, \epsilonunif, \epsilonreg, \epsilontest) : \NN \to (\RR_+)^4$ regularizes the estimation of optimal pairs, the computation of the exploration policy and the exploration GLR tests. 
Every component of $\EPSILON$ has a specific purpose and although they are all maps $\NN \to \RR_+$, they do not take the same input.
\begin{itemize}
    \item
        $\epsilonflat$ is the \strong{leveling} parameter: It  determines the threshold beyond which a sub-optimal state-action pair of $\hat{\model}_t$ is considered optimal because it is near-optimal enough;
        It is a function of the global time $t$;
    \item
        $\epsilonunif$ is the \strong{uniformization} parameter: It makes sure that exploration is slightly uniform and is key bounding the probability that \texttt{ECoE*} blunders the estimation of the hidden model, or that \texttt{ECoE*} purposely avoids regions of the environment that should be explored, but that aren't because the empirical data is completely off;
        It is a function of the exploration time $\abs{\stimes^-}$;
    \item 
        $\epsilonreg$ is the \strong{convexification} parameter: It comes from the way the regret lower bound is being regularized (\Cref{definition_regularized_regret_lower_bound}) and makes sure that the estimated exploration measure converges;
        It is a function of the exploration time $\abs{\stimes^-}$;
    \item
        $\epsilontest$ is the \strong{test power} parameter: It controls the power of the exploration test that regulates second order errors in exploration tests. 
        It is a function of the global time $t$;
\end{itemize}
Note that not all regularizers take the same input, as some are functions of the global time $t$ in \Cref{algorithm_ecoe_annotated} and others are functions of the number of exploration phases $\abs{\stimes^-}$.
Also, the regularization is \strong{floored} into $\EPSILON'$ during play to stabilize the algorithm:
\begin{equation}
\label{equation_floored_regularization}
\begin{gathered}
    \epsilonflat'(t) := \epsilonflat(t)
    \qquad
    \epsilontest'(t) := \epsilontest(t)
    \\
    \epsilonunif'(t) := \epsilonunif\parens*{2^{\floor{\log_2 \abs{\stimes^-}}}},
    \qquad
    \epsilonreg'(t) := \epsilonreg\parens*{2^{\floor{\log_2 \abs{\stimes^-}}}}
    .
\end{gathered}
\end{equation}
Moreover, for \texttt{ECoE*} to be asymptotically optimal, the regularization hyperparameter $\EPSILON$ must be carefully be tuned. 
We move the technical discussion on the set of assumptions (\Cref{assumption_regularization_hyperparameter}) that $\EPSILON$ must satisfy to the appendix.
For now, we will assume that the regularization hyperparameters are chosen as follows:
\begin{equation}
\label{equation_choice_regularization}
\begin{gathered}
    \epsilonflat(t) := \frac 1{\max\parens*{1, \log\log(t)}},
    \qquad
    \epsilontest(t) := \frac 1{\max\parens*{1, \log\log(t)}},
    \\
    \epsilonunif(m) := \frac 1{\max\parens*{1, \log\log(m)}},
    \qquad
    \epsilonreg(m) := \frac 1{\max\parens*{1, \log(m)}}
    .
\end{gathered}
\end{equation}
We can finally provide the regret guarantees of \texttt{ECoE*}.

\subsection{Maximum likelihood estimator of $\model$ and structure of the ambient space}
\label{section_estimator}

The {maximum likelihood estimator} of $\model$ under history $\History_t$ is the unique model $\hat{\model}_t \equiv (\pairs, \hat{\kernel}_t, \hat{\rewarddistribution}_t)$ that maximizing the likelihood of $\History_t$.
Formally,
\begin{equation}
\label{equation_estimator}
    \hat{\model}_t 
    \in
    \argmax_{\model' \in \closure(\models)}
    \product_{i=1}^{t-1}
    \hat{\kernel}_t (\State_{i+1}|\Pair_i)
    \hat{\kerneldistribution}_t (\Reward_i|\Pair_i)
    .
\end{equation}
It is preferable for the optimizer in \eqref{equation_estimator} to be unique and easy to compute. 
Also, the regret analysis of \texttt{ECoE*} requires $\hat{\model}_t$ to be well-behaved relatively to $\model$, with $\support(\hat{\kernel}_t) \subseteq \support(\kernel)$ for instance. 
For that purpose, we will strengthen our set of assumptions (\Cref{assumption_space}) on the structure of the ambient space of models $\models$, see \Cref{assumption_space_strong} below.

\begin{assumption}
\label{assumption_space_strong}
    The space $\models$ is of the form $\models \equiv \product_{\pair \in \pairs} (\kernels_\pair \times (0, 1))$ where $\kernels_\pair = \probabilities(\states)$ or $\kernels_\pair = \braces{\kernel(\pair)}$ for some $\kernel(\pair) \in \probabilities(\states)$.
    We write $\closure(\models) := \product_{\pair \in \pairs} (\kernels_\pair \times [0, 1])$. 
\end{assumption}

Under \Cref{assumption_space_strong} and for $\model \in \models$, the maximum likelihood estimator $\hat{\model}_t$ is the model with reward function and transition kernel given by
\begin{equation*}
    \hat{\reward}_t (\pair)
    =
    \frac {
        \sum_{i=1}^{t-1} 
        \Reward_i 
        \indicator{\Pair_i = \pair}
    }{
        \visits_t (\pair)
    }
    \quad\text{and}\quad
    \hat{\kernel}_t (\state|\pair)
    = \begin{cases}
        \displaystyle
        \frac {
            \sum_{i=1}^{t-1} 
            \indicator{\Pair_i = \pair, \State_{i+1} = \state}
        }{
            \visits_{t-1}(\pair)
        }
        & \text{if $\kernels_\pair = \probabilities(\states)$;}
        \\
        \displaystyle
        \kernel(\pair)
        & \text{if $\kernels_\pair = \braces{\kernel(\pair)}$.}
    \end{cases}
\end{equation*}
That is, $\hat{\reward}_t$ is the empirical mean of the rewards observed when playing $\pair$ and $\hat{\kernel}_t (\state|\pair)$ is the number of transitions to $\state$ observed when playing $\pair$. 
Note that $\hat{\model}_t (\pair) \ll \model (\pair)$ for all $t \ge 1$ {a.s.}~as soon as $\visits_t (\pair) \ge 1$. 

\subsection{The regret guarantees of \texttt{ECoE*}}
\label{appendix_regret_analysis_overview}

\texttt{ECoE*} tuned with $\EPSILON$ chosen as in \Cref{equation_choice_regularization} is asymptotically optimal.

\begin{theorem}[Asymptotic optimality]
\label{theorem_upper_bound}
    Assume that the regularization hyperparameter $\EPSILON$ is chosen as in \eqref{equation_choice_regularization}, and that the ambient space $\models$ has Bernoulli rewards (\Cref{assumption_bernoulli}) and strong product structure (\Cref{assumption_space_strong}).
    For all $\model \in \models$ communicating and all initial state $\state_0 \in \states$, \texttt{ECoE*} satisfies
    \begin{equation*}
        \Reg(T; \model, \texttt{\upshape ECoE*}, \state_0) 
        \le
        \regretlb(\model) \log(T)
        + \oh(\log(T))
    \end{equation*}
    as $T \to \infty$.
    Accordingly, \texttt{ECoE*} is asymptotically optimal.
\end{theorem}

The proof of \Cref{theorem_lower_bound} is difficult and is deferred to \Cref{appendix_regret}.
We bound the expected regret $\Reg(T; \model, \learner, \state_0)$ by upper-bounding the pseudo regret $\sum_{t=1}^{T} \ogaps(\Pair_t)$ in expectation. 
As $\ogaps(\Pair_t) > 0$ only for $\Pair_t \notin \optpairs(\model)$, this is about upper-bounding the number of visits $\visits_T (\pair)$ in expectation for sub-optimal pairs $\pair \notin \optpairs(\model)$.
We show that there exists a special optimizer $\imeasure^* \in \imeasures(\model)$ of \eqref{equation_definition_regretlb} such that, for all sub-optimal pairs $\pair \notin \optpairs(\model)$, we have
\begin{equation*}
    \EE\brackets*{
        \visits_T (\pair)
    }
    \le
    \imeasure^*(\pair) \log(T)
    + \oh(\log(T))
\end{equation*}
when $T \to \infty$. 
Because $\imeasure^*$ is an optimizer of the optimization problem \eqref{equation_definition_regretlb} associated to $\regretlb(\model)$, it immediately follows that \texttt{ECoE*} is asymptotically optimal. 

Therefore, the regret lower bound of \Cref{theorem_lower_bound} is tight.

    \section{Open directions and future research}
    
In this paper, we have shown that under reasonable assumptions on the space of ambient Markov decision processes $\models$, then
\begin{equation*}
    \min_{\learner \text{~consistent}}
    \sup_{
        \scriptstyle \model \in \models 
        \atop 
        \scriptstyle \text{communicating}
    }
    \limsup_{T \to \infty}
    \braces*{
        \frac{\Reg(T; \model, \learner)}{\log(T)}
        - \regretlb(\model)
    }
    = 0.
\end{equation*}
In other words, we have provided a learning algorithm $\learner$ with $\Reg(T; \model, \learner) \le \regretlb(\model) \log(T) + \oh(\log(T))$ for all communicating model $\model \in \models$ simultaneously, while every reasonable (i.e., consistent) learning algorithm $\learner'$ satisfies $\Reg(T; \model, \learner') \ge \regretlb(\model) \log(T) + \oh(\log(T))$. 
Accordingly, our algorithm is universally asymptotically optimal with $\Reg(T; \model, \learner) = \regretlb(\model) \log(T) + \oh(\log(T))$ and the regret lower bound of \cite{boone_regret_2025} is tight. 

From there, many directions open up. 

\paragraph{Tractable algorithms.}
Our algorithm \texttt{ECoE*} is not easy to implement, because the computation of $\regretlb(\model)$ is difficult.
In fact, even the exploration tests \eqref{equation_glr_exploration} are computationaly difficult in general, see \cite{boone_regret_2025}. 
Nonetheless, we believe that this computational difficulties can be avoided and that an asymptotically optimal version of \ECOE{} that runs in reasonable time is doable---if the size of the pair space stays small, at least. 

\paragraph{More general settings.}
To simplify the exposition, we have assumed that rewards are Bernoulli (\Cref{assumption_bernoulli}). 
This assumption can be dropped for another.
Our proof is likely to hold for reward distributions among a single parameter exponential family, but this is yet to prove. 
More interestingly, there is the question of dropping the communicating assumption (\Cref{assumption_communicating}) and of dropping the a priori knowledge on the structure of the finite pair space $\pairs$ (\Cref{assumption_finite}). 
Indeed, not only \ECOE{} is only designed to work for communicating Markov decision processes, but some a priori knowledge on $\pairs$ is hard-coded in $\models$, because every element of $\models$ has pair space $\pairs$. 
This raises the following question:
\begin{quote}
    \itshape
    What can we say about the regret lower bound when $\models$ is the space of all Markov decision processes with Bernoulli rewards and \emph{arbitrary} finite state space?
\end{quote}
We believe that the above is essentially the same problem as dealing with weakly communicating Markov decision processes with fixed finite pair space. 

All this is left for future work.

    \section*{Acknowledgements}
    This research was supported in part by the French National Research Agency (ANR) in the framework of the PEPR IA FOUNDRY project (ANR-23-PEIA-0003).

    \clearpage
    \appendix

    \numberwithin{equation}{section}
    \numberwithin{definition}{section}
    \numberwithin{theorem}{section}

    \clearpage
    \section{Additional notions on Markov decision processes}
    In this appendix, we provide various complementary results on Markov decision processes.
In \Cref{appendix_bellman_optimality}, we provide a refresher on Bellman optimality that comes back for the proof of \Cref{proposition_continuity_near_optimality}.
In \Cref{appendix_worst_diameter_gain_gap}, we define the worst diameter and the gain gap.
The last consequent \Cref{appendix_invariant_measures} is dedicated to the analysis of invariant measures (\Cref{definition_invariant_measure}) of Markov decision processes. 

\begin{definition}[Definite minimum of a model]
\label{definition_definite_minimum}
    Let $\model \equiv (\pairs, \kernel, \rewardd)$ be a Markov decision process with Bernoulli rewards. 
    Its \strong{definite minimum} is $\dmin(\model) := \min \braces{\dmin(\kernel), \dmin(\rewardd)}$
\end{definition}

\subsection{Refresher on gain, Bellman and bias optimalities}
\label{appendix_bellman_optimality}

At several places in this paper, we will use the notion of Bellman optimality.
Bellman optimality sits in-between gain and bias optimalities. 
We recall its definition below.

\begin{definition}[Bellman optimal]
\label{definition_bellman_optimal}
    Let $\model \equiv (\pairs, \kernel, \reward)$ be a communicating Markov decision process.
    A policy is said \strong{Bellman optimal}, written $\policy \in \boptpolicies(\model)$, if $\vecspan{\gainof{\policy}(\model)} = 0$ and $\policy$ satisfies the first order Bellman equations:
    \begin{equation}
    \label{equation_bellman_equations}
        \forall \state \in \states,
        \quad
        \gainof{\policy}(\state; \model)
        + \biasof{\policy}(\state; \model)
        =
        \max_{\action \in \actions(\state)} \braces[\Big]{
            \reward(\state, \action)
            + 
            \kernel(\state, \action) \biasof{\policy}(\model)
        }
        .
    \end{equation}
\end{definition}

In other words, Bellman optimal policies are fixpoints of the Policy Iteration algorithm, see \cite[§8.6]{puterman_markov_2014}.
It is linked to gain and bias optimalities as follows.

\begin{proposition}
\label{proposition_optimalities}
    In the space of communicating Markov decision processes, every Bellman optimal policy is gain optimal, and every bias optimal policy is Bellman optimal;
    The converse statements are wrong in general.
\end{proposition}

A proof of \Cref{proposition_optimalities} can be found in \cite[Theorem~I.3]{boone_thesis_2024}.

\subsection{Worst diameter and gain-gap of a communicating Markov decision process}
\label{appendix_worst_diameter_gain_gap}

At several places in this work, we rely on the \strong{worst diameter} and the \strong{gain-gap} of a communicating Markov decision process. 
They are defined in this paragraph for further reference.
The worst diameter $\worstdiameter(\model)$ is involved in gain deviations inequalities (see \Cref{appendix_mdpart}) that are uniform over the space of policies.
The gain-gap is related to how small the leveling threshold must be so that the leveling transform $\flatmodel{\model}{\epsilon}$ is well-behaved, see \Cref{appendix_regularized} and \Cref{proposition_continuity_near_optimality} more specifically.

\begin{definition}[Worst diameter]
\label{definition_worst_diameter}
    The \strong{worst diameter} $\worstdiameter(\model) \in \RR_+$ of a communicating Markov decision process $\model$ is the largest policy diameter (\Cref{definition_policy_diameter}) of its deterministic stationary policies, visually, 
    \begin{equation*}
        \worstdiameter(\model) 
        := 
        \max_{\policy \in \policies}
        \diameter(\policy; \model)
        .
    \end{equation*}
\end{definition}

\begin{definition}[Gain gap]
\label{definition_gain_gap}
    The \strong{gain-gap} $\gaingap(\model) \in \RR_+$ of a communicating Markov decision process $\model$ is the difference between the optimal gain and the gain of the best sub-optimal policy, visually,
    \begin{equation}
        \gaingap(\model)
        :=
        \min_{\policy \notin \optpolicies(\model)}
        \norm{\optgain(\model) - \gainof{\policy}(\model)}_\infty
        .
    \end{equation}
\end{definition}

\subsection{Invariant measures of Markov decision processes}
\label{appendix_invariant_measures}

In this section, we provide a collection of results on the invariant measures of a Markov decision process.
Recall that $\imeasure$ is an \strong{invariant measure} of $\model \equiv (\pairs, \kerneld, \rewardd)$ if
\begin{equation*}
    \forall \state \in \states,
    \quad
    \sum_{\action \in \actions(\state)}
    \imeasure(\state, \action)
    =
    \sum_{\pair \in \pairs}
    \kernel(\state|\pair) \imeasure(\pair)
    .
\end{equation*}
We write $\imeasure \in \imeasures(\model)$. 
The set of \strong{probability invariant measures} is written
\begin{equation}
\label{equation_definition_probability_invariant_measures}
    \pimeasures(\model) 
    := 
    \imeasures(\model) \cap \probabilities(\pairs)
    .
\end{equation}
An invariant measure $\imeasure \in \imeasures(\model)$ is said \strong{$\epsilon$-uniform} (\Cref{definition_uniform_invariant_measure}) if it is the (unique) invariant measure of some $\epsilon$-uniform policy, i.e., if $\imeasure(\state, \action) \ge \epsilon \sum_{\action' \in \actions(\state)} \imeasure(\state, \action')$ for all $(\state, \action) \in \pairs$.
We write $\imeasure \in \imeasures(\epsilon; \model)$ their space and $\pimeasures(\epsilon; \model) := \imeasures(\epsilon; \model) \cap \probabilities(\pairs)$.

\paragraph{Outline.}
In \Cref{appendix_algorithms_link_invariant_measures}, we draw the link between invariant measures and the asymptotic behavior of learning algorithms by showing that the normalized visit vector $\frac 1t \visits_t$ converges to the space of invariant measures as $t \to \infty$. 
Using this property, we show that there exists a stationary randomized policy that covers all pairs uniformly fast with respect to the diameter in \Cref{appendix_covering_policy}.
In \Cref{appendix_policies_link_invariant_measures}, we link fully supported randomized policies to fully supported probability invariant measures by describing the natural one-to-one mapping between $\fullyrandomizedpolicies$ and $\bigcup_{\epsilon > 0} \pimeasures(\epsilon; \model)$. 
In \Cref{appendix_approximation_invariant_measures}, we show that elements of $\imeasures(\model) \equiv \imeasures(0; \model)$ can be approximated with arbitrary precision by elements of $\imeasures(\epsilon; \model)$ as $\epsilon \to 0$.
In \Cref{appendix_policies_invariant_measures_diameter_support}, we provide properties on the minimum and the diameter of $\epsilon$-uniform invariant measures, then deduce a bound of the variations of $\pimeasures(\model)$ when $\model$ vary.
In the last \Cref{appendix_decomposition_invariant_measures}, we show that every element of $\imeasures(\model)$ can be written as a convex combination of invariant measures of unichain policies and provide several useful consequences of this result. 

\subsubsection{Links between invariant measures and learning algorithms}
\label{appendix_algorithms_link_invariant_measures}

We begin this appendix by showing that the average visit counts $\frac 1t \visits_t$ converges to the space of invariant measures almost surely, whatever the learning algorithm is. 

\begin{proposition}[Asymptotic flow property of $\visits_t$]
\label{proposition_flow}
    Let $\model \equiv (\pairs, \kernel, \rewardd)$ be a communicating Markov decision process and let $\learner$ be a learning algorithm.
    Then $\frac 1t \visits_t \to \imeasures(\model)$ a.s.~when $t \to \infty$, i.e., every limit point of $(\frac 1t \visits_t)_{t \ge 1}$ is an invariant measure with probability one. 
\end{proposition}

\begin{proof}
    This follows from a direct computation.
    Denote $\hat{\imeasure}_t := \frac 1t \visits_t$ and let $\imeasure^*$ be a limit point of $\hat{\imeasure}_t$ when $t \to \infty$. 
    There exists a sequence $t_n \to \infty$ such that $\lim_{n \to \infty} \hat{\imeasure}_{t_n} = \imeasure^*$. 
    Let $\state \in \states$.
    We check that $\sum_{\action \in \actions(\state)} \imeasure^*(\state, \action) = \sum_{\pair \in \pairs} \imeasure^*(\pair) \kernel(\state|\pair)$ as follows:
    \begin{align*}
        \sum_{\action \in \actions(\state)}
        \imeasure^* (\state, \action)
        & \overset{(\dagger)}=
        \frac 1{t_n} \parens*{
            \sum_{\action \in \actions(\state)}
            \visits_{t_n} (\state, \action)
            + \oh(t_n)
        }
        \\
        & \overset{(\ddagger)}=
        \frac 1{t_n} \parens*{
            \sum_{\pair \in \pairs}
            \visits_{t_n} (\state|\pair)
            + \OH(1)
            + \oh(t_n)
        }
        \\
        & \overset{(\S)}=
        \frac 1{t_n} \parens*{
            \sum_{\pair \in \pairs}
            \parens*{
                \hat{\kernel}_{t_n} (\state|\pair)
                - \kernel (\state|\pair)
            }
            \visits_{t_n} (\pair)
            + \sum_{\pair \in \pairs}
            \kernel (\state|\pair)
            \visits_{t_n} (\pair)
            + \oh(t_n)
        }
        \\
        & \overset{(\$)}=
        \frac 1{t_n} \parens*{
            \sum_{\pair \in \pairs}
            \kernel (\state|\pair)
            \visits_{t_n} (\pair)
            + \oh(t_n)
        }
        \\
        & \overset{(\#)}=
        \sum_{\pair \in \pairs}
        \imeasure^*(\pair) \kernel(\state|\pair)
    \end{align*}
    where 
    $(\dagger)$ holds by assumption;
    $(\ddagger)$ uses the notation $\visits_t (\state|\pair) := \sum_{i=1}^{t-1} \indicator{\Pair_i = \pair, \State_{i+1} = \state}$ and uses the quasi-flow property of $\visits_{t_n}$, which is $\abs{\sum_\action \visits_{t_n} (\state, \action) - \sum_\pair \visits_{t_n} (\state|\pair)} \le 2$; 
    $(\S)$ uses that the empirical transition kernel satisfies $\hat{\visits}_{t_n} (\state|\pair) \visits_{t_n} (\pair) = \visits_{t_n} (\state|\pair)$ by definition;
    $(\$)$ follows from the strong law of large numbers; and
    $(\#)$ holds by definition of $(t_n)$. 
\end{proof}

\begin{corollary}
\label{lemma_flow}
    Let $\model \equiv (\pairs, \kernel, \rewardd)$ be a communicating Markov decision process.
    Let $\learner$ a learning algorithm such that $\frac 1t \visits_t (\pair) \to \imeasure(\pair)$ a.s., for all $\pair \in \pairs$.
    Then $\imeasure \in \imeasures(\model)$. 
\end{corollary}

\subsubsection{A convenient fast covering policy}
\label{appendix_covering_policy}

In this paragraph, we show that there exists a policy $\policy$ that covers all pairs uniformly fast, with $\min(\imeasureof{\policy}) \ge \abs{\pairs}^{-1} \diameter(\model)^{-1}$. 
This policy is obtained as the limit of average behavior of a non-stationary policy that navigates as fast as possible between pairs in a round-robin fashion, see \Cref{lemma_policy_exploration_diameter} below.
The existence of that limit average behavior is guaranteed by the flow properties of the visit vector shown upstream, see \Cref{lemma_flow}. 

This covering policy plays an important part in the approximation of invariant measures by $\epsilon$-uniform invariant measures when $\epsilon \to 0$, see \Cref{theorem_invariant_measures_approximate_with_covering}.
It also provides a simple bound on $\regretlb(\model)$ in terms of $\diameter(\model)$, see \Cref{theorem_regretlb_bound}. 

\begin{lemma}[Fast covering policy]
\label{lemma_policy_exploration_diameter}
    Let $\models \equiv (\pairs, \kernel, \rewardd)$ a communicating Markov decision process.
    There exists a stationary fully-randomized policy $\policy \in \randomizedpolicies$ such that its unique probability invariant measure $\imeasureof{\policy}$ (\Cref{proposition_measures_policies_correspondence}) satisfies
    \begin{equation*}
        \min(\imeasureof{\policy}) 
        \ge 
        \frac 1{\abs{\pairs} \diameter(\model)}
        .
    \end{equation*}
\end{lemma}

\begin{proof}
    Enumerate $\pairs$ as $\pairs = \braces{\pair_0, \ldots, \pair_{m-1}}$. 
    Given $\pair_i \in \pairs$, pick $\policy_i \in \policies$ a policy achieving
    \begin{equation*}
        \min_{\policy \in \policies}
        \max_{\state \in \states}
        \EE_{\state}^{\model, \policy} \brackets*{
            \inf \braces*{
                t \ge 1 : \Pair_t = \pair_i
            }
        }
        .
    \end{equation*}
    By definition, $\EE_{\state}^{\model, \policy_i} \brackets{\inf\braces{t \ge 1 : \Pair_t = \pair_i}} \le \diameter(\model)$ for $i = 0, \ldots, m-1$. 
    Consider the non-stationary policy $\learner$ given by the following pseudo-code:
    \begin{algorithmic}[1]
        \STATE Set $t \gets 1$;
        \FOR{$n=0, 1, 2, \ldots$}
            \STATE Set $t_n \gets t$; 
            \STATE Let $i \gets n~\mathrm{mod}~m$ and $\policy_n \gets \policy_i$;
            \STATE Iterate $\policy_i$ until $\Pair_t = \pair_i$;
        \ENDFOR
    \end{algorithmic}
    Assume throughout that we iterate $\learner$ on $\model$.
    
    First, we argue that $\frac 1t \visits_t (\pair)$ converges almost-surely as $t \to \infty$ for all $\pair \in \pairs$. 

    Fix $\pair \in \pairs$.
    The idea is that, by design of $\learner$, the random states $\State_{t_{km + i}}$ and $\State_{t_{k'm+i}}$ are identically distributed for $k, k' \ge 1$ and $i \in \braces{0, \ldots, m-1}$. 
    Denote $\imeasure_i$ their common distribution.
    Then, for all $k \ge 1$ and regardless of $\state_0 \in \states$, we have
    \begin{equation*}
        \EE^{\model, \learner}_{\state_0} [t_{km+i+1} - t_{km+i}]
        =
        \EE^{\model, \policy_i}_{\imeasure_i}\brackets*{
            \inf \braces*{t \ge 1: \Pair_t = \pair_i}
        }
        =:
        T_i
        \in
        \RR
        .
    \end{equation*}
    Denoting $T := \sum_{i=0}^{m-1} T_i$, we deduce from the Strong Law that $t_{n} \sim \frac{nT}m$ as $n \to \infty$. 
    Let 
    \begin{equation*}
        \alpha_i (\pair)
        := 
        \EE^{\model, \learner}
        \brackets*{
            \sum_{t=t_{km+i}}^{t_{km+1+i}-1} 
            \indicator{\Pair_t = \pair}
        }
    \end{equation*}
    the expected number of times the pair $\pair$ is played when $\learner$ iterates $\policy_i$ over the episode $n = km + i$ for $k \ge 1$.
    Denothing $\alpha(\pair) := \sum_{i=0}^{m-1} \alpha_i (\pair)$, we deduce from the Strong Law that $\visits_{t_n} (\pair) \sim \frac {n \alpha(\pair)}m$ as $n \to \infty$.
    All together, we obtain
    \begin{equation*}
        \visits_t (\pair) 
        \sim
        \frac{t m} T \cdot \frac {\alpha(\alpha)} m
        = \frac {\alpha(\pair) t} T
        \quad
        \text{a.s.}
    \end{equation*}
    as $t \to \infty$. 
    Hence $\frac 1t \visits_t (\pair) \sim \frac{\alpha(\pair)}T =: \imeasure(\pair)$. 

    Second, we argue that 
    (I) $\imeasure(\pair) \ge \abs{\pairs}^{-1} \diameter(\model)^{-1}$ and that 
    (II) $\imeasure \in \imeasures(\model) \cap \probabilities(\pairs)$. 
    The assertion (I) is immediate from the observation that $\alpha_i (\pair_i) \ge 1$ and that $T_i \le \diameter(\model)$.
    The assertion (II) is a consequence of the flow property of the visit vector $\visits_t$, see \Cref{lemma_flow}. 

    Conclude using that the randomized policy induced by $\imeasure$, which is given by $\policy^{\imeasure}(\action|\state) := \imeasure(\state, \action) (\sum_{\action' \in \actions(\state)} \imeasure(\state, \action'))^{-1}$, is fully randomized (because $\imeasure$ is fully supported) and has unique probability invariant measure $\imeasure$ (\Cref{proposition_measures_policies_correspondence}).
\end{proof}

\subsubsection{Links between fully supported policies and fully supported measures}
\label{appendix_policies_link_invariant_measures}

In this paragraph, we show that fully supported invariant measures can be thought as fully supported policies, because $\bigcup_{\epsilon > 0} \pimeasures(\epsilon; \model) \cong \fullyrandomizedpolicies$.
Indeed, in \Cref{proposition_measures_policies_correspondence} below, we state that from a fully randomized probability invariant measure $\imeasure$, we can construct a policy $\policy^\imeasure$ and that by iterating that policy, expected visit rates converge to $T \imeasure(\pair) + \oh(T)$. 
This further encourages the computation of $\regretlb(\model)$ via \eqref{equation_regretlb_2} in the first place, because a near optimal solution $\policy \in \fullyrandomizedpolicies$ provides a policy that, if iterated, makes visit rates converge to the optimal rates in the sense of \eqref{equation_definition_regretlb}, hence $\policy$ gathers information near optimally. 

For later convenience, we provide a notation for the set of invariant measures of a policy.

\begin{definition}[Invariant measures of a policy]
\label{definition_invariant_measure_policy}
    Let $\model$ a Markov decision process. 
    The set of invariant measures of $\policy \in \policies$ in $\model$, denoted $\imeasures(\policy; \model)$, is the collection of $\imeasure \in \imeasures(\model)$ such that $\imeasure(\state, \action) = \policy(\action|\state) \sum_{\action' \in \actions(\state)} \imeasure(\state, \action')$ for all $(\state, \action) \in \pairs$.
\end{definition}

It is to be noted that $\imeasures(\policy; \model)$ is isomorphic to $\imeasures(\model_\policy)$ where $\model_\policy \equiv (\states, \kernel_\policy, \reward_\policy)$ is the single-action Markov decision process (hence is a Markov reward process) with kernel $\kernel_\policy (\state) := \sum_{\action \in \actions(\state)} \policy(\action|\state) \kernel(\state, \action)$ and reward function $\reward_\policy (\state) := \sum_{\action \in \actions(\state)} \policy(\action|\state) \reward(\state, \action)$, where the isomorphism maps $\imeasure \in \imeasures(\policy; \model)$ to $\imeasure' \in \RR^\states$ given by $\imeasure' (\state) := \imeasure(\state, \policy(\state))$. 

\begin{proposition}[Randomized policies and invariant measures]
\label{proposition_measures_policies_correspondence}
    Assume that $\model$ is a communicating Markov chain. 
    There is a one-to-one correspondence between probability invariant measures of full support and fully supported randomized policies $\fullyrandomizedpolicies$, with
    \begin{enum}
        \item A policy $\policy \in \fullyrandomizedpolicies$ is associated to a unique probability invariant measure $\imeasureof{\policy}$, given by $\imeasureof{\policy}(\pair) := \lim \EE_{\state_0}^{\model, \policy}\brackets{\frac 1T \sum_{t=1}^T \indicator{\Pair_t = \pair}}$, regardless of the initial state $\state_0 \in \states$;
            In other words, $\imeasures(\policy; \model) = \RR_+ \imeasureof{\policy}$;
        \item A fully supported invariant measure $\imeasure$ is associated to a unique fully randomized policy $\policy^\imeasure$, given by $\policy^\imeasure(\action|\state) := \imeasure(\state, \action) (\sum_{\action' \in \actions(\state)} \imeasure(\state, \action'))^{-1}$;
        \item We have $\policy^{\imeasureof{\policy}} = \policy$ for all $\policy \in \fullyrandomizedpolicies$.
    \end{enum}
\end{proposition}

The correspondence described in \Cref{proposition_measures_policies_correspondence} cannot be generalized easily to non-fully supported invariant measures.
If every probability invariant measure is a probability invariant measure of some policy, it is not the unique one when that policy is multi-chain; And as matter of fact, the iterates of a multi-chain policy $\policy$ converge to a specific invariant measure of $\policy$ that depend on the initial state. 
When policies and invariant measures are of full support, we don't need to take such considerations into account. 

\bigskip
\def\proofname{Proof of \Cref{proposition_measures_policies_correspondence}}
\begin{proof}
    Assertion (1) is a result about Markov chains.
    If $\policy$ is fully randomized and $\model$ is communicating, then it is easy to see that the Markov chain that $\policy$ induces on $\pairs$ is irreducible, so that it has a unique probability invariant measure from classical results, see \cite[Corollary 1.17]{levin_markov_2017}.
    The fact that this measure is found as the asymptotic presence probability is also standard.

    Assertion (2) is immediate, because $\policy^\imeasure$ satisfies $\policy^\imeasure(\action|\state) \ge \epsilon$ by definition of $\imeasures(\epsilon; \model)$.
    Then, it is easy to check that $\imeasure$ is an invariant measure of $\policy^\imeasure$; It is the unique one, because the Markov chain induced by $\policy$ on $\pairs$ is irreducible since $\policy^\imeasure$ is fully supported and $\model$ is communicating. 

    For Assertion (3), we have $\imeasureof{\policy}(\state,\action) \propto \policy(\action|\state)$ so $\policy^{\imeasureof{\policy}}(\action|\state) = \policy(\action|\state)$ is immediate.
\end{proof}
\def\proofname{Proof}

\subsubsection{Approximation of invariant measures with uniform invariant measures}
\label{appendix_approximation_invariant_measures}

In this paragraph, we show that uniform invariant measures can approximate invariant measures with arbitrary precision.
In \Cref{theorem_invariant_measures_approximate_with_covering} below, we show that provided that the required uniformization $\epsilon$ is small enough, an invariant measure can be made $\epsilon$-uniform by blending it with the invariant measure of the fast covering policy of \Cref{lemma_policy_exploration_diameter}.

\begin{theorem}[Approximation of invariant measures]
\label{theorem_invariant_measures_approximate_with_covering}
    Let $\imeasure \in \pimeasures(\model)$ be a probability invariant measure. 
    For all $\epsilon < \abs{\pairs}^{-1} \diameter(\model)^{-1}$, there exists $\imeasure_\epsilon \in \pimeasures(\epsilon; \model)$ such that
    \begin{equation*}
        \norm{\imeasure_\epsilon - \imeasure}_\infty 
        \le 
        \abs{\pairs} \diameter(\model)
        \epsilon
        .
    \end{equation*}
\end{theorem}

\begin{proof}
    Fix $\imeasure \in \pimeasures(\model)$ a probability invariant measure.
    Let $\imeasure_c$ be the (unique) probability invariant measure of the fast covering policy given by \Cref{lemma_policy_exploration_diameter}, satisfying $\min(\imeasure_c) \ge \abs{\pairs}^{-1} \diameter(\model)^{-1}$. 
    As $\pimeasures(\model)$ is obtained as the intersection of finitely many hyperplanes, it is convex.
    In particular, for every $\lambda \in [0, 1]$, $(1 - \lambda) \imeasure + \lambda \imeasure_c \in \pimeasures(\model)$ and $(1 - \lambda) \imeasure + \lambda \imeasure_c \in \RR_+^\pairs$.
    It follows that
    \begin{equation*}
        \min \parens*{(1-\lambda) \imeasure + \lambda \imeasure_c}
        \ge
        \lambda \min(\imeasure_c)
        .
    \end{equation*}
    Let $\lambda_\epsilon := \min(\imeasure_c)^{-1}\epsilon$.
    Since $\epsilon < \min(\imeasure_c)$, $\imeasure_\epsilon := (1 - \lambda_\epsilon) \imeasure + \lambda_\epsilon \imeasure_c$ is a probability invariant measure.
    Moreover, we see that for all $(\state, \action) \in \imeasure_\epsilon$, we have
    \begin{equation*}
        \imeasure_\epsilon (\state, \action)
        \ge
        \lambda_\epsilon \min(\imeasure_c)
        \ge \epsilon 
        \ge \epsilon \sum_{\action' \in \actions(\state)} \imeasure_\epsilon (\state, \action')
    \end{equation*}
    so that $\imeasure_\epsilon$ is $\epsilon$-uniform. 
    Moreover, we have $\norm{\imeasure_\epsilon - \imeasure}_\infty = \lambda_\epsilon \norm{\imeasure - \imeasure_c}_\infty \le \lambda_\epsilon$. 
\end{proof}

\subsubsection{Properties of policies induced by uniform invariant measures}
\label{appendix_policies_invariant_measures_diameter_support}

In this paragraph, we provide a few properties on uniform probability invariant measures.
In \Cref{lemma_uniform_measures_support_diameter}, we provide a lower bound of the minimums of $\epsilon$-uniform invariant measures as well as the diameters of their induced policies.
In \Cref{lemma_uniform_measures_perturbations}, we show that the projection of $\imeasure \in \pimeasures(\epsilon; \model)$ onto $\pimeasures(\epsilon; \model')$ is close to $\imeasure$, leading to a bound on the Hausdorff distance between $\pimeasures(\epsilon; \model)$ and $\pimeasures(\epsilon; \model')$ in \Cref{corollary_uniform_measures_hausdorff_distance}.
In turn, \Cref{corollary_uniform_measures_hausdorff_distance} plays an important part in proving that $\regretlb_\EPSILON (\model') \to \regretlb_\EPSILON (\model)$ when $\snorm{\model' - \model} \to 0$, see \Cref{proposition_continuity_regularized_lower_bound}.

\begin{lemma}
\label{lemma_uniform_measures_support_diameter}
    Let $\model \in \models$ be a communicating Markov decision process and $\epsilon > 0$.
    For all $\epsilon$-uniform invariant measure $\imeasure \in \imeasures(\epsilon; \model)$, the following holds:
    \begin{enum}
        \item We have $\min(\imeasure) \ge \epsilon^{\abs{\states}-1} \dmin(\kernel)^{\abs{\states}-1} \norm{\imeasure}_1$;
        \item Its induced policy has small diameter with $\diameter(\policy_\imeasure; \model) \le \epsilon^{1 - \abs{\states}} \dmin(\kernel)^{1 - \abs{\states}} (\abs{\states} - 1)$.
    \end{enum}
\end{lemma}
\begin{proof}
    Up to normalizing $\imeasure$, we assume that $\norm{\imeasure}_1 = 1$.

    We start with assertion (1).
    Pick $\pair \in \pairs$ and let $\state \in \states$ maximizing $\sum_{\action \in \actions(\state)} \imeasure(\state, \action)$.
    In particular, we have $\sum_{\action \in \actions(\state)} \imeasure(\state, \action) \ge \abs{\states}^{-1}$. 
    Because $\model$ is communicating, there exists a path from $\state$ to $\pair$ of length at most $\abs{\states}$, i.e., $(\state_1, \action_1, \ldots, \state_n, \action_n)$ such that 
    (I) $\state_1 = \state$, 
    (II) $(\state_n, \action_n) = \pair$,
    (III) $\kernel(\state_{i+1}|\state_i,\action_i) > 0$ for $i = 1, \ldots, n-1$, and
    (IV) $n \le \abs{\states}$.
    Then,
    \begin{align*}
        \imeasure(\pair)
        & \ge
        \parens*{
            \product_{i=1}^{n-1} 
            \frac{
                \kernel(\state_{i+1}|\state_i, \action_i)
                \imeasure(\state_i, \action_i)
            }{
                \sum_{\action'_i \in \actions(\state_i)}
                \imeasure(\state_i, \action'_i)
            }
        }
        \sum_{\action' \in \actions(\state)}
        \imeasure(\state, \action')
        \\
        & \overset{(\dagger)}=
        \parens*{
            \product_{i=1}^{n-1} 
            \kernel(\state_{i+1}|\state_i, \action_i)
            \policy_\imeasure (\action_i|\state_i)
        }
        \sum_{\action' \in \actions(\state)}
        \imeasure(\state, \action')
        \\
        & \overset{(\ddagger)}\ge
        \parens*{
            \product_{i=1}^{n-1} 
            \kernel(\state_{i+1}|\state_i, \action_i)
            \epsilon
        }
        \frac 1{\abs{\states}}
        \ge \frac{
            \dmin(\kernel)^{n-1}
            \epsilon^{n-1} 
        }{\abs{\states}}
        \ge \frac{
            \dmin(\kernel)^{\abs{\states}-1}
            \epsilon^{\abs{\states}-1} 
        }{\abs{\states}}
    \end{align*}
    where 
    $(\dagger)$ follows by definition of the induced policy, and
    $(\ddagger)$ uses that $\policy_\imeasure$ is $\epsilon$-uniform and $\sum_{\action' \in \actions(\state)} \imeasure(\state, \action') \ge \abs{\states}^{-1}$. 

    We continue with assertion (2).
    Let $\state, \state' \in \states$ and denote $T(\state, \state') := \E_{\state}^{\policy_\imeasure, \model}[\inf \set{t \ge 0: \State_t = \state'}]$ the reaching time to $\state'$ from $\state$.
    Since $\policy_\imeasure$ is recurrent, by definition, we have $\diameter(\policy_\imeasure; \model) = \max_{\state \ne \state'} T(\state, \state')$.
    Let $(\state, \state') \in \states^2$ achieving the maximum of $T$.
    Using Markov's property, we find that
    \begin{align*}
        T(\state, \state')
        & \le 
        \abs{\states} - 1 
        + \parens*{
            1 - \Pr_\state^{\policy_\imeasure, \model} \parens*{
                \exists t \le \abs{\states}
                :
                \State_t = \state'
            }
        } T(\state, \state')
        \\
        & \le
        \abs{\states} - 1
        + \parens*{
            1 - \dmin(\kernel)^{\abs{\states}-1} \epsilon^{\abs{\states}-1}
        } T(\state, \state')
    \end{align*}
    where $(\dagger)$ follows with the same argument as for assertion (1).
    Solve in $T(\state, \state')$.
\end{proof}

\begin{lemma}
\label{lemma_uniform_measures_perturbations}
    Let $\model, \model' \in \models$ be a pair of communicating Markov decision processes and fix $\epsilon > 0$.
    For all $\imeasure \in \pimeasures(\epsilon; \model)$, there is $\imeasure' \in \pimeasures(\epsilon; \model')$ such that
    \begin{equation*}
        \norm{\imeasure' - \imeasure}_\infty
        \le
        \min \braces[\big]{
            \diameter(\policy_\imeasure; \model),
            \diameter(\policy_\imeasure; \model') 
        }
        \norm{\kernel' - \kernel}_\infty
        .
    \end{equation*}
\end{lemma}
\begin{proof}
    As $\imeasure$ is a $\epsilon$-uniform invariant measure, its induced policy $\policy_\imeasure$ is $\epsilon$-uniform.
    Moreover, $\model'$ is communicating, so $\policy_\imeasure$ is recurrent on $\model'$.
    Let $\imeasure'$ be the unique probability invariant measure of $\policy_\imeasure$ in $\model'$.
    By \Cref{lemma_unichain_invariant_measure_variations}, we have
    \begin{equation*}
        \norm{\imeasure' - \imeasure}_\infty
        \le
        \min \braces[\big]{
            \diameter(\policy_\imeasure; \model),
            \diameter(\policy_\imeasure; \model') 
        }
        \norm{\kernel' - \kernel}_\infty
    \end{equation*}
    hence the result.
\end{proof}

Combining \Cref{lemma_uniform_measures_support_diameter} and \Cref{lemma_uniform_measures_perturbations}, we can express the variations of the set of $\epsilon$-uniform probability invariant measures $\pimeasures(\epsilon)$ in Hausdorff distance as follows.

\begin{corollary}
\label{corollary_uniform_measures_hausdorff_distance}
    Let $\model, \model' \in \models$ be a pair of communicating Markov decision processes and fix $\epsilon > 0$.
    The Hausdorff distance (in $\ell_\infty$-norm) between $\pimeasures(\epsilon; \model)$ and $\pimeasures(\epsilon; \model')$ is bounded by
    \begin{equation*}
        d_\infty (\pimeasures(\epsilon; \model'), \pimeasures(\epsilon; \model))
        \le
        \frac {
            \abs{\states}
            \cdot
            \norm{\kernel' - \kernel}_\infty
        }{
            \epsilon^{\abs{\states}-1}
            \dmin(\kernel)^{\abs{\states}-1}
        }
        .
    \end{equation*}
\end{corollary}

\subsubsection{Decomposition of invariant measures with unichain policies}
\label{appendix_decomposition_invariant_measures}

In this paragraph, we show that every invariant measure can be expressed as a convex combination of invariant measures of unichain policies, see \Cref{proposition_decomposition_invariant_measures}.
This result is analogous to the decomposition of flow into simple cycles in Graph Theory. 
\Cref{proposition_decomposition_invariant_measures} has several interesting consequences.
The first that we present is a bound on the variations (in Hausdorff distance) of the space of probability invariant measures $\pimeasures(\model) := \imeasures(\model) \cap \probabilities(\pairs)$ as $\model$ varies, see \Cref{lemma_invariant_measures_hausdorff_distance} downstream. 
Combined with \Cref{theorem_invariant_measures_approximate_with_covering,theorem_invariant_measures_approximate_with_covering}, we further obtain a bound on the variations (in Hausdorff distance) of the space of $\epsilon$-uniform probability invariant measures $\pimeasures(\epsilon; \model) := \imeasures(\epsilon; \model) \cap \probabilities(\pairs)$ as $\model$ varies, see \Cref{lemma_invariant_measures_uniform_hausdorff_distance}---A bound that can be compared with \Cref{corollary_uniform_measures_hausdorff_distance}.
\Cref{proposition_decomposition_invariant_measures} is also used to prove that optimal exploration measures exist, see \Cref{theorem_optimal_measures_existence}.

\begin{proposition}
\label{proposition_decomposition_invariant_measures}
    Let $\model$ be a communicating Markov decision process. 
    Let $\policies_u$ be the set of unichain policies in $\model$.
    We have $\imeasures(\model) = \convexhull(\braces{\imeasures(\policy; \model) : \policy \in \policies_u})$, where $\convexhull(-)$ denotes the convex hull.
\end{proposition}

\begin{proof}
    Because $\imeasures(\policy; \model) \subseteq \imeasures(\model)$ and that $\imeasures(\model)$ is convex, we have $\convexhull(\braces{\imeasures(\policy; \model): \policy \in \policies_u}) \subseteq \imeasures(\model)$.

    We show the reverse inclusion.

    Let $\imeasure \in \imeasures(\model)$. 
    If $\imeasure = 0$, there is nothing to prove, so we assume $\support(\imeasure) \ne \emptyset$. 
    Let $(\state_1, \action_1, \state_2, \action_2, \ldots, \state_n) \in \pairs^{n-1} \times \states$ a closed path in $\model$ such that 
    (I) $\kernel(\state_{i+1}|\state_i, \action_i) > 0$,
    (II) $\imeasure(\state_i, \action_i) > 0$,
    (III) $\state_1 = \state_n$ and
    (IV) $n \ge 2$.
    The condition (III) states that the path is closed and (IV) that the path is non-trivial.
    Up to simplifying that path, we can assume that (V) it is simple in the sense that $\state_i \ne \state_j$ for all $i \ne j < n$. 
    Then, the set of pairs $\pairs_0 := \braces{(\state_i, \action_i): i < n}$ is the set of recurrent pairs of the policy given by
    \begin{equation*}
        \policy_0 (\action|\state)
        :=
        \begin{cases}
            1 & \text{if~$\state = \state_i$ and $\action = \action_i$;}
            \\
            0 & \text{if~$\state = \state_i$ and $\action \ne \action_i$;}
            \\
            \abs{\actions(\state)}^{-1} & \text{if $\state \ne \state_i$.}
        \end{cases}
    \end{equation*}
    Note that $\policy_0$ is unichain. 
    Let $\imeasure_0$ the unique probability invariant measure of $\policy_0$. 
    We have $\support(\imeasure_0) = \pairs_0$.
    By construction of the path, we have $\pairs_0 \subseteq \support(\imeasure)$, see (II).
    Therefore, the set $\braces{\lambda > 0: \lambda \imeasure_0 < \imeasure}$ is non-empty; Let $\lambda_0 > 0$ be its supremum. 
    Consider the measure $\imeasure'$ given by
    \begin{equation*}
        \imeasure' := \imeasure - \lambda_0 \imeasure_0.
    \end{equation*}
    By choice of $\lambda_0$, we see that $\imeasure' \in \RR_+^\pairs$.
    Note that $\imeasure' \in \imeasures(\model)$.
    Indeed, $\imeasures(\model)$ is the intersection
    \begin{equation*}
        \imeasures(\model)
        =
        \RR_+^\pairs
        \cap
        \mathcal{F}
        \quad\text{where}\quad
        \mathcal{F}
        :=
        \bigcap_{\state \in \states}
        \braces*{
            \imeasure'' \in \RR^\pairs 
            :
            \sum_{\action \in \actions(\state)}
            \imeasure''(\state, \action)
            = 
            \sum_{\pair \in \pairs}
            \imeasure''(\pair)
            \kernel(\state|\pair)
        }
    \end{equation*}
    and $\mathcal{F}$ is a linear space, so that $\imeasure' := \imeasure - \lambda_0 \imeasure_0 \in \mathcal{F}$. 
    As $\imeasure' \in \RR_+^\pairs$ as explained above, we have $\imeasure' \in \imeasures(\model)$.
    But by choice of $\lambda_0$, the support of $\imeasure'$ is strictly smaller than the support of $\imeasure$, because $\support(\imeasure') \subseteq \support(\imeasure)$ and there must be at least one $\pair_0 \in \pairs_0$ such that $\pair_0 \notin \support(\imeasure_0)$.

    So, by induction on the size of the support of $\imeasure$, we find a sequence $(\lambda_i, \imeasure_i, \policy_i)_{i \le m}$ with $\lambda_i > 0$, $\imeasure_i \in \imeasures(\policy_u; \model)$ and $\policy_i \in \policies_u$ such that
    \begin{equation*}
        \imeasure 
        = 
        \lambda_0 \imeasure_0 + \ldots + \lambda_m \imeasure_m
        .
    \end{equation*}
    Note that we can ensure $m + 1 \le \abs{\pairs}$.
    This concludes the proof. 
\end{proof}

Follow \Cref{proposition_decomposition_invariant_measures}, we show that $\pimeasures(\model)$ is locally $\worstdiameter(\model)$-Lipschitz continuous where $\worstdiameter(\model)$ is the worst diameter (\Cref{definition_worst_diameter}).

\begin{lemma}[Variations of $\pimeasures(\model)$]
\label{lemma_invariant_measures_hausdorff_distance}
    Let $\model \equiv (\pairs, \kernel, \reward), \model' \equiv (\pairs, \kernel', \reward') \in \models$ be a pair of communicating Markov decision processes with the same kernel supports, i.e., $\kernel \sim \kernel'$. 
    The Hausdorff distance (in $\ell_\infty$-norm) between $\pimeasures(\model)$ and $\pimeasures(\model')$ is bounded by
    \begin{equation*}
        d_\infty (\pimeasures(\model), \pimeasures(\model'))
        \le
        \worstdiameter(\model)
        \norm{\kernel' - \kernel}_\infty
    \end{equation*}
    where $\worstdiameter(\model)$ is the worst diameter (\Cref{definition_worst_diameter}). 
\end{lemma}

\begin{proof}
    Because $\model$ and $\model'$ have the same kernel supports, they have the same space of unichain policies $\policies_u$. 
    Given $\policy \in \policies_u$, we write $\imeasure_\policy$ and $\imeasure'_\policy$ the unique invariant probability measure of $\policy$ in $\model$ and $\model'$ respectively. 
    For all $\policy \in \policies_u$, we bound $\norm{\imeasure'_\policy - \imeasure_\policy}_\infty$ using \Cref{lemma_unichain_invariant_measure_variations} as
    \begin{equation}
    \label{equation_invariant_measures_hausdorff_1}
        \norm{\imeasure'_\policy - \imeasure_\policy}_\infty
        \le
        \diameter(\policy; \model)
        \norm{\kernel' - \kernel}_\infty
        \le 
        \worstdiameter(\model)
        \norm{\kernel' - \kernel}_\infty
    \end{equation}
    where the second inequality follows by definition of the worst diameter. 

    Let $\imeasure \in \pimeasures(\model)$.
    By \Cref{proposition_decomposition_invariant_measures}, there exists $\lambda \in \probabilities(\policies_u)$ such that $\imeasure = \sum_{\policy \in \policies_u} \lambda_\policy \imeasure_\policy$. 
    Introduce $\imeasure' := \sum_{\policy \in \policies_u} \lambda_\policy \imeasure'_\policy$.
    By \Cref{proposition_decomposition_invariant_measures}, we have $\imeasure' \in \pimeasures(\model')$.
    So, using \eqref{equation_invariant_measures_hausdorff_1}, we deduce that
    \begin{align*}
        \inf_{\imeasure'' \in \pimeasures(\model')}
        \norm{\imeasure'' - \imeasure}_\infty
        \le
        \norm{\imeasure' - \imeasure}_\infty
        \le 
        \sum_{\policy \in \policies_u}
        \lambda_\policy
        \norm{\imeasure'_\policy - \imeasure_\policy}_\infty
        \le 
        \worstdiameter(\model)
        \norm{\kernel' - \kernel}_\infty
        .
    \end{align*}
    Similarly, we show that for all $\imeasure' \in \pimeasures(\model)$, we have $\inf_{\imeasure'' \in \pimeasures(\model)} \norm{\imeasure'' - \imeasure'}_\infty \le \worstdiameter(\model) \norm{\kernel' - \kernel}_\infty$. 
    This concludes the proof.
\end{proof}

\begin{lemma}[Variations of $\pimeasures(\epsilon; \model)$]
\label{lemma_invariant_measures_uniform_hausdorff_distance}
    Let $\model \equiv (\pairs, \kernel, \reward), \model' \equiv (\pairs, \kernel', \reward') \in \models$ be a pair of communicating Markov decision processes with the same kernel supports, i.e., $\kernel \sim \kernel'$. 
    Let $\epsilon < \frac 12 \abs{\pairs}^{-1} \diameter(\model)^{-1}$ be the required uniformization. 
    The Hausdorff distance (in $\ell_\infty$-norm) between $\pimeasures(\epsilon; \model)$ and $\pimeasures(\epsilon; \model')$ is bounded by
    \begin{equation*}
        d_\infty (\pimeasures(\epsilon; \model), \pimeasures(\epsilon; \model'))
        \le
        \worstdiameter(\model)
        \norm{\kernel' - \kernel}_\infty
        + 2 \abs{\pairs} \diameter(\model) \epsilon
    \end{equation*}
    where $\worstdiameter(\model)$ is the worst diameter (\Cref{definition_worst_diameter}). 
\end{lemma}

\begin{proof}
    Since $\pimeasures(\epsilon; -) \subseteq \probabilities(\pairs)$, we have $d_\infty (\pimeasures(\epsilon; \model), \pimeasures(\epsilon; \model')) \le 1$ trivially. 
    So, the bound of \Cref{lemma_invariant_measures_uniform_hausdorff_distance} is only interesting when $\norm{\kernel' - \kernel}_\infty \le \worstdiameter(\model)^{-1}$.
    By \Cref{proposition_diameter_and_worst_diameter}, we have $\diameter(\model) \le \worstdiameter(\model)$ so that we can assume that $\norm{\kernel' - \kernel}_\infty \le \diameter(\model)^{-1}$.
    So, by \Cref{lemma_variations_diameter}, we have 
    \begin{equation}
        \label{equation_hausdorff_1}
        \diameter(\model') \le 2 \diameter(\model)
        .
    \end{equation}
    Let $\imeasure \in \pimeasures(\epsilon; \model)$.
    By \Cref{lemma_invariant_measures_hausdorff_distance} and because $\pimeasures(\epsilon; \model')$ is compact, there exists $\imeasure' \in \pimeasures(\epsilon; \model')$ such that $\norm{\imeasure' - \imeasure}_\infty \le \worstdiameter(\model) \norm{\kernel' - \kernel}_\infty$. 
    Following \eqref{equation_hausdorff_1}, we have $\epsilon < \abs{\pairs}^{-1} \diameter(\model')^{-1}$. 
    So, by uniformizing $\imeasure'$ into $\imeasure'_\epsilon \in \pimeasures(\epsilon; \model')$ via \Cref{theorem_invariant_measures_approximate_with_covering}, we have $\norm{\imeasure' - \imeasure_\epsilon}_\infty \le \epsilon \abs{\pairs} \diameter(\model')$.
    So, by triangular inequality, we get
    \begin{equation*}
        \norm{\imeasure' - \imeasure_\epsilon}
        \le
        \worstdiameter(\model) \norm{\kernel' - \kernel}_\infty
        +
        \epsilon \abs{\pairs} \diameter(\model')
        \overset{(\dagger)}\le
        \worstdiameter(\model) \norm{\kernel' - \kernel}_\infty
        +
        2 \abs{\pairs} \diameter(\model) \epsilon
    \end{equation*}
    where $(\dagger)$ follows from \eqref{equation_hausdorff_1}.

    When starting with $\imeasure' \in \pimeasures(\epsilon; \model')$, we proceed similarly.
\end{proof}

Note that \Cref{corollary_uniform_measures_hausdorff_distance} and \Cref{lemma_invariant_measures_uniform_hausdorff_distance} bound the same quantity, $d_\infty (\pimeasures(\epsilon; \model), \pimeasures(\epsilon; \model'))$.
Their respective bounds are
\begin{equation*}
    \frac {
        \abs{\states}
        \cdot
        \norm{\kernel' - \kernel}_\infty
    }{
        \epsilon^{\abs{\states}-1}
        \dmin(\kernel)^{\abs{\states}-1}
    }
    \quad\text{V.S.}\quad
    \worstdiameter(\model) \norm{\kernel' - \kernel}_\infty
    +
    2 \abs{\pairs} \diameter(\model) \epsilon
    .
\end{equation*}
It is to be noted that the second is \emph{not} better than the first; The two bounds are rather of a different nature.
Since $\diameter(\model), \worstdiameter(\model) \le \abs{\states} \dmin(\kernel)^{1 - \abs{\states}}$, it is clear that the second bound is much better than the first when $\epsilon$ and $\norm{\kernel' - \kernel}_\infty$ are of the same order. 
However, when $\norm{\kernel' - \kernel}_\infty \ll \epsilon^{\abs{\states}}$ as in the proof of \Cref{proposition_regularized_lowerbound_approximates}, the first bound is better.
As a matter of fact, the second introduces a flat error $2 \abs{\pairs} \diameter(\model) \epsilon$ that does not vanish when $\norm{\kernel' - \kernel}_\infty \to 0$, although $d_\infty(\pimeasures(\epsilon; \model), \pimeasures(\epsilon; \model')) \to 0$.

    \clearpage
    \section{Regularized lower bound and approximate optimalities}
    \label{appendix_regularized}

In this appendix, we provide proof details on approximate optimalities, leveled Bellman gaps and near optimal pairs, as well as the regularized regret lower bound.

\paragraph{Outline.}
\Cref{appendix_near_optimality} is dedicated to the proof of \Cref{proposition_continuity_near_optimality}, that establishes the most important properties of the leveling transform (\Cref{definition_leveled_mdp}). 
In the remaining \Cref{appendix_information_value,appendix_regretlb_continuity}, we investigate the continuity properties of the regularized regret lower bound (\Cref{definition_regularized_regret_lower_bound}). 
In \Cref{appendix_information_value}, we introduce and study the properties of the concept of information value (\Cref{definition_information_value}), that is central to the GLR exploration tests \eqref{equation_glr_exploration} and to the regret lower bound.
In the last \Cref{appendix_regretlb_continuity}, we focus on the properties of the regularized regret lower bound and establish \Cref{proposition_regretlb_approximation,proposition_regretlb_convergence_solution,proposition_regretlb_continuity}.

\subsection{Properties of leveled optimality and proof of \Cref{proposition_continuity_near_optimality}}
\label{appendix_near_optimality}

The goal of this paragraph is to provide a proof of \cref{proposition_continuity_near_optimality} in the main text (the statement is recalled below for better readability).
\Cref{proposition_continuity_near_optimality} states that if $\snorm{\model' - \model}$ is small with respect to the leveling threshold $\epsilon > 0$, then the $\epsilon$-leveled transform of $\model'$ (\Cref{definition_leveled_mdp}) has the same gain optimal policies and optimal pairs as $\model$.
Moreover, the required ratio between $\snorm{\model' - \model}$ and $\epsilon$ is shown to be rational in the gain-gap $\gaingap(\model)$ (\Cref{definition_gain_gap}) and the worst diameter $\worstdiameter(\model)$ (\Cref{definition_worst_diameter}).

\bigskip
\noindent
\textbf{\cref{proposition_continuity_near_optimality}.}
\textit{
    For every communicating Markov decision process $\model \equiv (\pairs, \kerneld, \rewardd)$, there exists a quantity $C(\model) < \infty$, that is a rational function of $\gaingap(\model)$ and $\worstdiameter(\model)$, for all $\epsilon > 0$ and $\model' \in \models$ satisfying
    \begin{equation}
        \epsilon + 2 C (\model) \snorm{\model' - \model} < \gaingap(\model)
        \quad\text{and}\quad
        C (\model)
        \snorm{\model' - \model}
        < \epsilon
    \end{equation}
    we have
    \begin{enum}
        \item $\flatmodel{\model'}{\epsilon}$ and $\model$ have the same gain optimal policies;
        \item $\flatme{\optpairs}{\epsilon}(\model) = \optpairs(\model)$; 
    \end{enum}
}

\paragraph{Proof outline.}
The remaining of this paragraph is dedicated to the proof of \Cref{proposition_continuity_near_optimality}.
In \Cref{lemma_leveled_bias_is_bellman}, we show that bias optimal policies of $\model'$ are Bellman optimal (\Cref{definition_bellman_optimal}) in $\flatmodel{\model'}{\epsilon}$.
In turn, we prove \Cref{lemma_leveled_same_gain_optimal}, that consists in the assertion (1) of \Cref{proposition_continuity_near_optimality}: by leveling a noisy model, we recover the set of gain optimal policies. 
Once (1) is established, the other assertion (2) is immediate by definition of $\optpairs(\model)$. 

\begin{lemma}
\label{lemma_leveled_bias_is_bellman}
    Let $\model' \in \models$ and $\epsilon > 0$. 
    Every bias optimal policy of $\model'$ is Bellman optimal (\Cref{definition_bellman_optimal}) in $\flatmodel{\model'}{\epsilon}$, i.e., for all $\policy \in \policies$ that is bias optimal policy in $\model'$, we have
    \begin{enum}
        \item 
            $\vecspan{\gainof{\policy}(\flatmodel{\model'}{\epsilon})} = 0$; and 
        \item 
            $\flatmodel{\reward'}{\epsilon}(\state, \action) + \flatmodel{\kernel'}{\epsilon}(\state, \action) \biasof\policy(\flatmodel{\model'}{\epsilon}) \le \gainof{\policy}(\state; \flatmodel{\model'}{\epsilon}) + \biasof{\policy}(\state; \flatmodel{\model'}{\epsilon})$ for all $(\state, \action) \in \pairs$.
    \end{enum}
    In particular, $\optpolicies(\model') \subseteq \optpolicies(\flatmodel{\model'}{\epsilon})$ and $\optgain(\model') = \optgain(\flatmodel{\model'}{\epsilon})$. 
\end{lemma}
\begin{proof}
    Let $\optpolicy$ be a bias optimal policy of $\model'$, so that $\gainof{\optpolicy}(\model') = \optgain(\model')$ and $\biasof{\optpolicy}(\model') = \optbias(\model')$.
    By construction, we have $\flatmodel{\reward'}{\epsilon}(\pair) = \reward'(\pair) + \indicator{\ogaps(\pair; \model') < \epsilon} \ogaps(\pair; \model')$ for $\pair \in \pairs$, so
    \begin{equation*}
        \gainof{\optpolicy}(\flatmodel{\model'}{\epsilon})
        = \gainof{\optpolicy}(\model') 
        = \optgain(\model')
        \quad\text{and}\quad
        \biasof{\optpolicy}(\flatmodel{\model'}{\epsilon})
        = \biasof{\optpolicy}(\model')
        = \optbias(\model')
        .
    \end{equation*}
    Therefore, $\vecspan{\gainof{\optpolicy}(\flatmodel{\model'}{\epsilon})} = 0$ and $\optpolicy$ satisfies the Bellman equations \eqref{equation_bellman_equations} in $\flatmodel{\model'}{\epsilon}$, i.e., it is Bellman optimal (\Cref{definition_bellman_optimal}) in $\flatmodel{\model'}{\epsilon}$;
    Hence the assertions (1) and (2).
\end{proof}

In \Cref{appendix_mdpart}, we investigate the sensibility of the gain and bias functions to changes of reward function and transition kernel. 
Following \cref{lemma_multichain_gain_variations} and \cref{lemma_multichain_bias_variations}, for every policy $\policy \in \policies$ and provided that $\snorm{\kernel' - \kernel}_\infty \le {\worstdiameter(\model)}^{-1}$, we have
\begin{equation}
\label{equation_gain_bias_deviations}
\begin{aligned}
    \norm*{
        \gainof{\policy}(\model') - \gainof{\policy}(\model)
    }_\infty
    & \le
    C_\gain(\model) \snorm{\model' - \model}
    \\
    \norm*{
        \biasof{\policy}(\model') - \biasof{\policy}(\model)
    }_\infty
    & \le
    C_\bias(\model) \snorm{\model' - \model}
    \\
    \norm*{\gapsof{\policy}(\model') - \gapsof{\policy}(\model)}_\infty
    & \le 
    C_\gaps(\model) \snorm{\model' - \model}
\end{aligned}
\end{equation}
where $C_\gain(\model), C_\bias(\model), C_\gaps(\model)$ are polynomial in $\worstdiameter(\model)$ and of respective orders $\worstdiameter(\model)$, $\worstdiameter(\model)^2$, $\worstdiameter(\model)^2$.

\begin{lemma}
\label{lemma_leveled_same_gain_optimal}
    Fix $\epsilon > 0$ a leveling parameter.
    For every pair of communicating Markov decision processes $\model, \model' \in \models$ such that 
    \begin{equation}
    \label{equation_lemma_leveled_same_gain_optimal_condition}
        \epsilon + 2 C_\gain (\model) \snorm{\model' - \model} < \gaingap(\model)
        \quad\text{and}\quad
        C_\gaps (\model) \worstdiameter(\model)
        \snorm{\model' - \model}
        < \epsilon
    \end{equation}
    we have $\optpolicies(\flatmodel{\model'}{\epsilon}) = \optpolicies(\model)$. 
\end{lemma}
\begin{proof}
    We proceed by double inclusion. 

    Assume that $\policy \in \optpolicies(\flatmodel{\model'}{\epsilon})$. 
    By \cref{lemma_leveled_bias_is_bellman}, bias optimal policies of $\model'$ are Bellman optimal (\Cref{definition_bellman_optimal}) in $\flatmodel{\model'}{\epsilon}$.
    In particular, they are gain optimal.
    In the mean time, bias optimal policies are supported in weakly optimal actions, so by construction of $\flatmodel{\model'}{\epsilon}$, the gain of bias optimal policies of $\model'$ is the same in $\model'$ and $\flatmodel{\model'}{\epsilon}$.
    Combined, we conclude that $\optgain(\flatmodel{\model'}{\epsilon}) = \optgain(\model')$.
    Now, since $\model'$ and $\flatmodel{\model'}{\epsilon}$ only differ in their reward function with $\norm{\reward' - \flatmodel{\reward'}{\epsilon}}_\infty \le \epsilon$, it follows that
    \begin{equation*}
        \gainof{\policy}(\model') 
        \ge 
        \gainof{\policy}(\flatmodel{\model'}{\epsilon}) - \epsilon \unit
        =
        \optgain(\model') - \epsilon \unit
    \end{equation*}
    where $\unit$ is the unitary vector.
    By \eqref{equation_gain_bias_deviations}, we further have $\gainof{\policy}(\model) \ge \gainof{\policy}(\model') - C_\gain(\model) \snorm{\model' - \model} \unit$ and $\optgain(\model') \ge \optgain(\model) - C_\gain(\model) \snorm{\model' - \model} \unit$. 
    All together,
    \begin{equation*}
        \gainof{\policy}(\model) \ge \optgain(\model) - \parens[\big]{\epsilon + 2 C_\gain(\model) \snorm{\model' - \model}} \unit
    \end{equation*}
    so $\policy \in \optpolicies(\model)$ if $\epsilon + 2 C_\gain(\model) \snorm{\model' - \model} < \gaingap(\model)$.
    This is indeed the case by assumption.

    Conversely, let $\policy \in \optpolicies(\model)$.
    Let $\pairs_1, \ldots, \pairs_m$ be the recurrent components of $\policy$ and let $\imeasure_i$ be the unique probability invariant measure of $\policy$ supported in $\pairs_i$.
    Fix $\policy'$ a bias optimal policy of $\model'$, so that $\gapsof{\policy'}(\model') \ge 0$.
    So, 
    \begin{equation}
        \label{equation_proof_lemma_leveled_same_gain_optimal_1}
        \gapsof{\policy'}(\pair; \model) 
        \ge
        - C_\gaps(\model) \snorm{\model' - \model}
    \end{equation}
    for every pair $\pair \in \pairs$.
    We have already shown that $\optpolicies(\flatmodel{\model'}{\epsilon}) \subseteq \optpolicies(\model)$. 
    Since $\optpolicies(\model') \subseteq \optpolicies(\flatmodel{\model'}{\epsilon})$ by \Cref{lemma_leveled_bias_is_bellman}, we have $\optpolicies(\model') \subseteq \optpolicies(\model)$. 
    So, $\policy' \in \optpolicies(\model)$. 

    We deduce that $\gainof{\policy'}(\model) = \gainof{\policy}(\model)$.
    Now, for $i \in \braces{1, \ldots, m}$, we have $\gainof{\policy}(\model) = \gainof{\policy'}(\model) - \sum_{\pair \in \pairs} \imeasure_i (\pair) \gapsof{\policy'}(\pair; \model)$, so together with $\gainof{\policy}(\model) = \gainof{\policy'}(\model)$, we get $\sum_{\pair \in \pairs} \pospart{\imeasure_i(\pair) \gapsof{\policy'}(\pair; \model)} = \sum_{\pair \in \pairs} \negpart{\imeasure_i (\pair) \gapsof{\policy'}(\pair; \model)}$ for all $i \in \braces{1, \ldots, m}$. 
    Combined with \eqref{equation_proof_lemma_leveled_same_gain_optimal_1}, we obtain 
    \begin{equation}
    \label{equation_proof_lemma_leveled_same_gain_optimal_2}
        \max_{\pair \in \pairs}
        \imeasure(\pair) 
        \gapsof{\policy'}(\pair; \model)
        \le
        C_\gaps(\model) 
        \snorm{\model' - \model}
        .
    \end{equation}
    By \Cref{lemma_diameter_invariant_measures}, we have $\min(\imeasure_i) \ge \diameter(\policy; \model|_{\pairs_i})^{-1}$.
    Together with $\diameter(\policy; \model|_{\pairs_i}) \ge \diameter(\policy; \model)^{-1}$ and \eqref{equation_proof_lemma_leveled_same_gain_optimal_2}, we conclude that 
    \begin{equation*}
        \gapsof{\policy'}(\pair; \model) 
        \le
        {C_\gaps(\model)}{\diameter(\policy; \model)}
        \snorm{\model' - \model}
        \le
        {C_\gaps(\model)}{\worstdiameter(\model)}
        \snorm{\model' - \model}
    \end{equation*}
    for all $\pair \in \pairs$ that is recurrent under $\policy$. 
    So, provided that $C_\gaps(\model) \worstdiameter(\model) \snorm{\model' - \model} < \epsilon$---which is indeed the case by assumption, we have $\gapsof{\policy'}(\pair; \flatmodel{\model}{\epsilon}) = 0$ for every $\pair \in \pairs$ that is recurrent under $\policy$ in $\model$. 
    Now, note that the recurrent pairs of $\policy$ are the same on $\model$, $\model'$, $\flatmodel{\model'}{\epsilon}$.
    So, again using the formula $\gainof{\policy}(\flatmodel{\model}{\epsilon}) = \gainof{\policy'}(\flatmodel{\model}{\epsilon}) - \sum_{\pair \in \pairs} \imeasure(\pair) \gapsof{\policy'}(\pair; \flatmodel{\model}{\epsilon})$ that holds for every probability invariant measure of $\policy$ in $\flatmodel{\model'}{\epsilon}$, we conclude that $\gainof{\policy}(\flatmodel{\model}{\epsilon}) = \gainof{\policy'}(\flatmodel{\model}{\epsilon})$.\footnote{To see this, note that this already holds for $\imeasure = \imeasure_i$ with $i \in \braces{1, \ldots, m}$ and use that every probability invariant measure of $\policy$ is a convex combination of $\imeasure_1, \ldots, \imeasure_m$.}
    By \Cref{lemma_leveled_bias_is_bellman}, $\policy'$ is Bellman optimal (\Cref{definition_bellman_optimal}) in $\flatmodel{\model'}{\epsilon}$ hence is gain optimal in $\flatmodel{\model'}{\epsilon}$ in particular. 
    We conclude that $\gainof{\policy}(\flatmodel{\model}{\epsilon}) = \optgain(\model)$, i.e., $\policy \in \optpolicies(\model)$.
\end{proof}

With \Cref{lemma_leveled_same_gain_optimal} in hand, we can prove \Cref{proposition_continuity_near_optimality}.

\medskip
\def\proofname{Proof of \Cref{proposition_continuity_near_optimality}}
\begin{proof}
    Assume that $\model' \in \models$ satisfies $\epsilon + 2 C_\gain (\model) \snorm{\model' - \model} < \gaingap(\model)$. 
    By \Cref{lemma_leveled_same_gain_optimal}, we have $\optpolicies(\flatmodel{\model'}{\epsilon}) = \optpolicies(\model)$, hence proving (1).
    Together with the fact that the transition kernels of $\model$ and $\flatmodel{\model'}{\epsilon}$ have the same support, it follows that $\optpairs(\model) = \optpairs(\flatmodel{\model'}{\epsilon})$, hence proving (2). 
    This concludes the proof.
\end{proof}
\def\proofname{Proof}

\subsection{Information value of measures and critical models}
\label{appendix_information_value}

In this section, we discuss the notion of the \textbf{information value} (\Cref{definition_information_value}), that quantifies how efficient is a measure on $\pairs$ to reject the leveled confusing set. 
Information values provide an alternative view on information constraints, are key in the derivation of the continuity properties of the regularized lower bound.  
In association with the information value, we introduce \strong{critical models} (\Cref{definition_critical_model}) as the confusing Markov decision processes that are the hardest to reject under a measure on $\pairs$. 

\paragraph{Additional notations.}
In this section, we write $\probabilities_\delta (\pairs)$ the set of probability distributions $\imeasure$ on $\pairs$ such that $\min(\imeasure) \ge \delta$. 
We write $\unit := \sum_{\pair \in \pairs} \unit_\pair$ the unit measure on $\pairs$ and $(\unit_\pair)_{\pair \in \pairs}$ the canonical basis of $\R^\pairs$. 
For $\models' \subseteq \models$, we denote $\closure(\models')$ the closure of $\models'$ for the topology induced by $\norm{-}$. 

\begin{definition}[Information value]
\label{definition_information_value}
\label{definition_leveled_information_value}
    Fix $\epsilon > 0$ a leveling parameter. 
    The \strong{$\epsilon$-leveled information value} of a measure $\imeasure \in \R^\pairs$ in $\model$ is given by
    \begin{equation*}
        \ivalue_\epsilon (\imeasure, \model)
        :=
        \inf_{\model^\dagger \in \confusing_\epsilon(\model)}
        \braces*{
            \sum_{\pair \in \pairs} \imeasure(\pair) \KL (\model(\pair)||\model^\dagger(\pair))
        }
        \in [0, \infty]
        .
    \end{equation*}
\end{definition}

Note that the regret lower bound can be written as
\begin{equation*}
    \regretlb(\model) 
    =
    \inf_{\imeasure \in \imeasures(\model)}
    \braces*{
        \frac{
            \sum_{\pair \in \pairs} \imeasure(\pair) \ogaps(\pair; \model)
        }{
            \ivalue_0 (\imeasure, \model)
        }
    }
    .
\end{equation*}
As a matter of fact, given $\imeasure \in \imeasures(\model)$, the quantity $\ivalue_0(\imeasure, \model)^{-1}$ is the scalar $\lambda \in [0, \infty]$ with which $\imeasure$ must be normalized so that it properly rejects all confusing models, in the sense of $\inf_{\model^\dagger \in \confusing(\model)} \sum_{\pair \in \pairs} \lambda \imeasure(\pair) \cdot \KL(\model(\pair)||\model^\dagger(\pair)) \ge 1$. 

\begin{definition}[Critical models]
\label{definition_critical_model}
    Fix $\epsilon > 0$ a leveling parameter and $\imeasure \in \R^\pairs$ a measure.
    A Markov decision process $\model^\dagger \in \closure(\confusing_\epsilon(\model))$ is said \strong{$(\imeasure, \epsilon, \model)$-critical} if $\model^\dagger$ achieves the $\epsilon$-leveled information value of $\imeasure$ in $\model$, i.e., 
    \begin{equation*}
        \ivalue_\epsilon (\imeasure, \model)
        =
        \sum_{\pair \in \pairs} 
        \imeasure(\pair)
        \KL(\model(\pair)||\model^\dagger(\pair))
        .
    \end{equation*}
\end{definition}

For $\epsilon = 0$, we talk more simply of information value and $(\imeasure, \model)$-critical models.
Unfortunately, the case $\epsilon = 0$ is not well-behaved relatively to $\model$ in general.
It is among the reasons why the regret lower bound is discontinuous with respect to $\model$.
This is quite inconvenient when one has only access to noisy copies of $\model$, as it is the case in learning. 
Thankfully, when the leveling parameter satisfies $\epsilon > 0$ and when $\imeasure$ is fully supported, the leveled information value is continuous in both $\model$ (\Cref{lemma_information_value_continuous_model}) and $\imeasure$ (\Cref{lemma_information_value_continuous_measure}).
Accordingly, it can be correctly estimated even in the presence of noise. 

\subsubsection{Properties of the confusing set and of critical models}

In this paragraph, we provide a few useful properties of leveled confusing sets and critical models. 
In \Cref{lemma_property_confusing}, we rephrase the definition of the $\epsilon$-confusing set in terms of the gain function. 
In \Cref{lemma_property_critical}, we show that all critical models have small bias span and that their gain optimal policies are unichain. 

\begin{lemma}[Characterization of confusing models]
\label{lemma_property_confusing}
    Let $\model \in \models$ be a communicating model and $\epsilon > 0$.
    Fix $\model' \gg \model$ such that $\model'(\pair) = \model(\pair)$ for all $\pair \in \flatme{\optpairs}{\epsilon}(\model)$. 
    The following assertions are equivalent:
    \begin{enum}
        \item $\model^\dagger \in \confusing_\epsilon(\model)$, i.e., $\optpolicies(\model^\dagger) \cap \optpolicies(\model) = \emptyset$;
        \item $\optgain(\model^\dagger) > \gainof{\policy}(\model^\dagger)$ for all $\policy \in \optpolicies(\model)$;
        \item $\optgain(\model^\dagger) > \optgain(\model)$.
    \end{enum}
\end{lemma}
\begin{proof}
    Let $\model' \gg \model$ be such that $\model'(\pair) = \model(\pair)$ for all $\pair \in \flatme{\optpairs}{\epsilon}(\model)$. 
    Because $\model' \gg \model$ and $\model$ is communicating, $\model'$ is communicating as well.

    Let $\policy \in \optpolicies(\model)$. 
    If $\pairs_\policy$ denotes the recurrent pairs of $\policy$ in $\model$, then since $\kernel' \gg \kernel$, we have $\Pr_{\state}^{\policy, \model'} \parens{\forall m, \exists n \ge m : \Pair_n \in \pairs_{\policy}} = 1$ regardless of the initial state $\state \in \states$.
    In other words, the iterates of $\policy$ hit $\pairs_\policy$ infinitely often in $\model'$. 
    But $\model' = \model$ on $\optpairs(\model)$ and $\pairs_\policy \subseteq \optpairs(\model)$ since $\policy \in \optpolicies(\model)$; So, every recurrent pair of $\policy$ in $\model'$ belongs to $\optpairs(\model)$ and in particular, 
    \begin{equation}
    \label{equation_proof_property_confusing}
        \gainof{\policy}(\model') = \gainof{\policy}(\model) = \optgain(\model)
        .
    \end{equation}
    Following \eqref{equation_proof_property_confusing}, the equivalence between the assertions (2) and (3) in \Cref{lemma_property_confusing} is immediate; and $\optpolicies(\model^\dagger) \cap \optpolicies(\model) = \emptyset$ as soon as $\policy \notin \optpolicies(\model^\dagger)$, i.e., as soon as $\optgain(\model^\dagger) > \gainof{\policy}(\model)$. 
\end{proof}

\begin{lemma}[Simplification of confusing models]
\label{lemma_confusing_simplification}
    Assume that $\models$ is in product form, i.e., that $\models = \product_{\pair \in \pairs} (\rewards_\pair \times \kernels_\pair)$. 
Let $\model \in \models$ a communicating model and fix $\epsilon \ge 0$.
    For every $\model^\dagger \in \confusing_\epsilon (\model)$, there exists $\model^\ddagger \in \confusing_\epsilon (\model)$ such that 
    \begin{enum}
        \item $\model^\ddagger$ is communicating;
        \item $\vecspan{\optbias(\model^\ddagger)} \le \diameter(\model)$;
        \item $\model^\ddagger (\pair) = \model(\pair)$ for all $\pair \in \flatme{\optpairs}{\epsilon}(\model) \cup \optpairs(\model^\ddagger)^c$;
        \item For all $\pair \in \pairs$, $\KL(\model(\pair)||\model^\ddagger(\pair)) \le \KL(\model(\pair)||\model^\dagger(\pair))$;
        \item $\norm{\model^\ddagger - \model} \le \norm{\model^\dagger - \model}$.
    \end{enum}
\end{lemma}

\begin{proof}
    We consider the extended Markov decision process $\model' = \model^\dagger \cup \model$ where, from the state $\state \in \states$, a choice of action consists in choosing whether the transition is done in $\model$ or in $\model^\dagger$, i.e., choosing $\action \in \actions(\state)$, then a reward among $\braces{\rewardd(\state, \action), \rewardd^\dagger(\state, \action)}$ and a transition kernel among $\braces{\kerneld(\state, \action), \kerneld^\dagger(\state, \action)}$. 
    Note that $\model'$ is communicating with $\diameter(\model') \le \diameter(\model)$. 
    Moreover, 
    \begin{equation*}
        \optgain(\model') \ge \optgain(\model)
        \text{\quad and \quad}
        \optgain(\model') \ge \optgain(\model^\dagger)
        .
    \end{equation*}
    An (extended) policy achieving optimal bias in $\model'$ defines a policy $\policy'$ on $\model$ together with a model $\model^\ddagger$ such that $\optgain(\model^\ddagger) = \optgain(\model') = \gainof{\policy'}(\model^\ddagger)$ and $\vecspan{\optbias(\model^\dagger)} = \vecspan{\optbias(\model')} \le \diameter(\model') \le \diameter(\model)$, see \Cref{lemma_bias_diameter}. 
    So $\model^\ddagger$ satisfies (1). 
    By construction, we also have 
    \begin{equation}
    \label{equation_proof_property_critical_2}
        \sum_{\pair \in \pairs} 
        \imeasure(\pair) 
        \KL(\model(\pair)||\model^\ddagger(\pair))
        \le
        \sum_{\pair \in \pairs} 
        \imeasure(\pair) 
        \KL(\model(\pair)||\model^\dagger(\pair))
        .
    \end{equation}
    To conclude, we show that $\model^\ddagger \in \confusing_\epsilon(\model)$ as follows.
    Because $\model^\dagger \in \confusing_\epsilon(\model)$, it follows that $\model^\ddagger$ coincides with $\model$ on $\flatme{\optpairs}{\epsilon}(\model)$. 
    In particular, every gain optimal policy $\policy \in \optpolicies(\model)$ has at least one recurrent component within $\optpairs(\model) \subseteq \flatme{\optpairs}{\epsilon}(\model)$ hence satisfies $\gainof{\policy}(\state; \model^\ddagger) = \optgain(\state; \model)$ for some $\state \in \states$.
    But $\optgain(\model^\ddagger) \ge \optgain(\model^\dagger) > \optgain(\model)$ by \Cref{lemma_property_confusing}, so by \Cref{lemma_property_confusing} again, we have $\model^\ddagger \in \confusing_\epsilon(\model)$. 
\end{proof}

Following \Cref{lemma_confusing_simplification}, we obtain that critical models have small bias span, see \Cref{lemma_property_critical} below.

\begin{corollary}[Structure of critical models]
\label{lemma_property_critical}
    Assume that $\models$ is in product form, i.e., that $\models = \product_{\pair \in \pairs} (\rewards_\pair \times \kernels_\pair)$. 
    Let $\model \in \models$ a communicating model, $\epsilon > 0$ and fix $\imeasure \in (\R^*_+)^\pairs$ a fully supported measure.
    Every $(\imeasure, \epsilon, \model)$-critical model $\model^\dagger$ is such that
    \begin{enum}
        \item $\vecspan{\optbias(\model^\dagger)} \le \diameter(\model)$;
        \item $\model^\dagger (\pair) = \model(\pair)$ for all $\pair \in \flatme{\optpairs}{\epsilon}(\model) \cup \optpairs(\model^\dagger)^c$.
    \end{enum}
\end{corollary}

Another consequence of \Cref{lemma_confusing_simplification} is that every model is bounded away (in norm) from its confusing set relatively to its gain-gap and its diameter, see \Cref{lemma_confusing_separated} below.

\begin{lemma}[Distance to confusing set]
\label{lemma_confusing_separated}
    Every communicating model $\model \in \models$ is bounded away from its confusing set (in norm), i.e.,
    \begin{equation*}
        \inf_{\model^\dagger \in \confusing(\model)}
        \norm{\model^\dagger - \model}
        \ge
        \frac{\gaingap(\model)}{4 \diameter(\model)}
        >
        0
    \end{equation*}
    where $\gaingap(\model)$ is the gain-gap of $\model$, see \Cref{definition_gain_gap}.
\end{lemma}

\begin{proof}
    Let $\model^\dagger \in \confusing(\model)$.
    By \Cref{lemma_confusing_simplification}, there exists $\model^\ddagger \in \confusing(\model)$ such that $\norm{\model^\ddagger - \model} \le \norm{\model^\dagger - \model}$ and $\vecspan{\optbias(\model^\dagger)} \le \diameter(\model)$. 
    By \Cref{lemma_bias_diameter}, we also have $\vecspan{\optbias(\model)} \le \diameter(\model)$. 
    
    Both $\model$ and $\model^\ddagger$ are communicating, so $\vecspan{\optgain(\model)} = \vecspan{\optgain(\model^\ddagger)} = 0$, so by \Cref{lemma_unichain_gain_variations}, 
    \begin{equation}
    \label{equation_proof_alternative_separated}
    \begin{gathered}
        \norm{\gainof{\policy}(\model^\ddagger) - \gainof{\policy}(\model)}_\infty
        \le
        \norm{\reward^\ddagger - \reward}_\infty
        + \frac 12 \diameter(\model) \norm{\kernel^\ddagger - \kernel}_\infty
        ;
        \\
        \norm{\gainof{\policy^\ddagger}(\model^\ddagger) - \gainof{\policy^\ddagger}(\model)}_\infty
        \le
        \norm{\reward^\ddagger - \reward}_\infty
        + \frac 12 \diameter(\model) \norm{\kernel^\ddagger - \kernel}_\infty
    \end{gathered}
    \end{equation}
    where $\policy$ and $\policy^\ddagger$ are bias optimal policies of $\model$ and $\model^\ddagger$ respectively.

    By assumption, we have $\policy^\ddagger \notin \optpolicies(\model)$, so
    \begin{align*}
        \gaingap(\model)
        & \le 
        \norm{ \gainof{\policy^\ddagger}(\model) - \gainof{\policy}(\model) }_\infty
        =
        \max \parens*{ \gainof{\policy}(\model) - \gainof{\policy^\ddagger}(\model) }
        \\
        & \le 
        \max \parens*{ \gainof{\policy}(\model) - \gainof{\policy}(\model^\ddagger) }
        +
        \max \parens*{ \gainof{\policy}(\model^\ddagger) - \gainof{\policy^\ddagger}(\model^\ddagger) }
        +
        \max \parens*{ \gainof{\policy}(\model^\ddagger) - \gainof{\policy^\ddagger}(\model) }
        \\
        & \overset{(\dagger)}\le
        \max \parens*{ \gainof{\policy}(\model) - \gainof{\policy}(\model^\ddagger) }
        +
        \max \parens*{ \gainof{\policy}(\model^\ddagger) - \gainof{\policy^\ddagger}(\model) }
        \\
        & \overset{(\ddagger)}\le 
        2 \parens*{
            \norm{\reward^\ddagger - \reward}_\infty
            + \diameter(\model) \norm{\kernel^\ddagger - \kernel}_\infty
        }
        \le 
        4 \max \braces{1, \diameter(\model)} \norm{\model^\ddagger - \model}
    \end{align*}
    where 
    $(\dagger)$ follows from $\policy^\ddagger \in \optpolicies(\model^\ddagger)$ and
    $(\ddagger)$ follows from \eqref{equation_proof_alternative_separated}. 
    Note that $\diameter (\model) \ge 1$ by definition.
    We conclude by solving the above in $\norm{\model^\ddagger - \model}$ and using that $\norm{\model^\dagger - \model} \ge \norm{\model^\ddagger - \model}$.
\end{proof}

For the last result of this paragraph, we provide a variant of \Cref{lemma_confusing_separated}, see \Cref{lemma_alternative_separated_weak}.
This result generalized the lower bound of \Cref{lemma_confusing_separated} from confusing set to a weak version of the \strong{alternative set}. 
The alternative set of $\model$ is the collection of $\model^\dagger \gg \model$ such that $\optpolicies(\model^\dagger) \cap \optpolicies(\model) = \emptyset$. 
In \Cref{lemma_alternative_separated_weak} below, the absolute continuity condition ``$\model^\dagger \gg \model$'' is strenghtened to a mutual absolute continuity condition on supports, i.e., to ``$\kernel^\dagger \sim \kernel$''.

\begin{lemma}[Distance to alternative set, weak version]
\label{lemma_alternative_separated_weak}
    Given a communicating model $\model \equiv (\pairs, \kernel, \reward) \in \models$, if $\model^\dagger \equiv (\pairs, \kernel^\dagger, \reward^\dagger)$ is such that 
    (1) $\kernel^\dagger \sim \kernel$ and
    (2) $\optpolicies(\model^\dagger) \setminus \optpolicies(\model) \ne \emptyset$, then
    \begin{equation*}
        \norm{\model^\dagger - \model}
        \ge
        \frac{\gaingap(\model)}{4 \diameter(\model)}
    \end{equation*}
    where $\gaingap(\model)$ is the gain-gap of $\model$, see \Cref{definition_gain_gap}.
\end{lemma}

\begin{proof}
    By (2), $\kernel^\dagger$ and $\kernel$ have the same support, so $\model^\dagger$ is communicating as well.
    Assume that $\snorm{\kernel^\dagger - \kernel}_\infty \le \diameter(\model)^{-1}$.
    Then, by \Cref{lemma_variations_diameter}, we have $\diameter(\model^\dagger) \le 2 \diameter(\model)$ so $\vecspan{\optbias(\model^\dagger)} \le 2 \diameter(\model)$.
    By \Cref{lemma_bias_diameter}, we also have $\vecspan{\optbias(\model)} \le \diameter(\model)$. 

    Let $\policy^\dagger \in \optpolicies(\model^\dagger) \setminus \optpolicies(\model)$.
    In particular, $\gainof{\policy^\dagger}(\model) < \optgain(\model)$ so $\policy^\dagger$ has a recurrent class of pairs $\pairs^\dagger \subseteq \pairs$ (in $\model$) on which it has sub-optimal gain (in $\model$).
    By (2), $\kernel^\dagger$ and $\kernel$ have the same support, so the recurrent classes of policies are the same under $\model$ and $\model^\dagger$.
    In particular, $\pairs^\dagger$ is also a recurrent class of $\policy^\dagger$ in $\model^\dagger$.
    Now, up to an infinitesimal perturbation of $\reward^\dagger$ of the form $\reward^\dagger (\pair) + \epsilon \indicator{\pair \in \pairs^\dagger}$ with $\epsilon > 0$, we can assume that $\pairs^\dagger$ is the unique optimal recurrent class of $\model^\dagger$; In other words, that all gain optimal policies of $\model^\dagger$ are unichain with recurrent class $\pairs^\dagger$.
    In particular, all bias optimal policies of $\model^\dagger$ with unique recurrent class $\pairs^\dagger$, so are sub-optimal in $\model$. 
    Let $\policy^\dagger$ a bias optimal policy of $\model^\dagger$.

    From then, the proof is mostly similar to the one of \Cref{lemma_confusing_separated}.
    
    Both $\model$ and $\model^\dagger$ are communicating, so $\vecspan{\optgain(\model)} = \vecspan{\optgain(\model^\dagger)} = 0$, so by \Cref{lemma_unichain_gain_variations}, 
    \begin{equation}
    \label{equation_proof_alternative_separated}
    \begin{gathered}
        \norm{\gainof{\policy}(\model^\dagger) - \gainof{\policy}(\model)}_\infty
        \le
        \norm{\reward^\dagger - \reward}_\infty
        + \frac 12 \diameter(\model) \norm{\kernel^\dagger - \kernel}_\infty
        ;
        \\
        \norm{\gainof{\policy^\dagger}(\model^\dagger) - \gainof{\policy^\dagger}(\model)}_\infty
        \le
        \norm{\reward^\dagger - \reward}_\infty
        + \diameter(\model) \norm{\kernel^\dagger - \kernel}_\infty
    \end{gathered}
    \end{equation}
    where $\policy$ and $\policy^\dagger$ are bias optimal policies of $\model$ and $\model^\dagger$ respectively.

    As argued upstream, we have $\policy^\dagger$ has sub-optimal gain in $\model$, so
    \begin{align*}
        \gaingap(\model)
        & \le 
        \norm{ \gainof{\policy^\dagger}(\model) - \gainof{\policy}(\model) }_\infty
        =
        \max \parens*{ \gainof{\policy}(\model) - \gainof{\policy^\dagger}(\model) }
        \\
        & \le 
        \max \parens*{ \gainof{\policy}(\model) - \gainof{\policy}(\model^\dagger) }
        +
        \max \parens*{ \gainof{\policy}(\model^\dagger) - \gainof{\policy^\dagger}(\model^\dagger) }
        +
        \max \parens*{ \gainof{\policy}(\model^\dagger) - \gainof{\policy^\dagger}(\model) }
        \\
        & \overset{(\dagger)}\le
        \max \parens*{ \gainof{\policy}(\model) - \gainof{\policy}(\model^\dagger) }
        +
        \max \parens*{ \gainof{\policy}(\model^\dagger) - \gainof{\policy^\dagger}(\model) }
        \\
        & \overset{(\dagger)}\le 
        2 \parens*{
            \norm{\reward^\dagger - \reward}_\infty
            + \diameter(\model) \norm{\kernel^\dagger - \kernel}_\infty
        }
        \le 
        4 \max \braces{1, \diameter(\model)} \norm{\model^\dagger - \model}
    \end{align*}
    where 
    $(\dagger)$ follows from $\policy^\dagger \in \optpolicies(\model^\dagger)$ and
    $(\dagger)$ follows from \eqref{equation_proof_alternative_separated}. 
    Conclude using $\diameter(\model) \ge 1$.
\end{proof}

\subsubsection{Regularity properties of the information value}

In this paragraph, we show that for $\epsilon > 0$ and when working away from the boundary of $\probabilities(\pairs)$, the information value $\ivalue_\epsilon(\imeasure, \model)$ is bounded relatively to the reference value $\ivalue_\epsilon(\unit, \model)$ (\Cref{lemma_information_value_bounded}) that itself is bounded below (\Cref{lemma_information_value_positive}), and is continuous relatively to $\imeasure \in \probabilities_\delta(\pairs)$ (\Cref{lemma_information_value_continuous_measure}) and $\model \in \models$ (\Cref{lemma_information_value_continuous_model}). 

\begin{lemma}
\label{lemma_information_value_bounded}
    Fix $\model \in \models$, a leveling parameter $\epsilon \ge 0$ and $\delta > 0$. 
    The leveled information value of $\imeasure \in \probabilities_\delta (\pairs)$ is lower and upper bounded as
    \begin{equation*}
        \delta~\ivalue_\epsilon (\unit, \model)
        \le
        \ivalue_\epsilon (\imeasure, \model)
        \le 
        \ivalue_\epsilon (\unit, \model)
        .
    \end{equation*}
\end{lemma}
\begin{proof}
    By definition, $\min(\imeasure) \ge \delta$ for all $\imeasure \in \probabilities_\delta (\pairs)$.
    So,
    \begin{align*}
        \ivalue_\epsilon (\imeasure, \model)
        & :=
        \inf_{\model^\dagger \in \confusing_\epsilon(\model)} 
        \braces*{
            \sum_{\pair \in \pairs} 
            \imeasure(\pair) 
            \KL(\model(\pair)||\model^\dagger (\pair))
        }
        \\
        & \ge
        \delta 
        \inf_{\model^\dagger \in \confusing_\epsilon(\model)} 
        \braces*{
            \sum_{\pair \in \pairs} 
            1 \cdot
            \KL(\model(\pair)||\model^\dagger (\pair))
        }
        = \delta~\ivalue_\epsilon (\unit, \model)
        .
    \end{align*}
    Using that $\max(\imeasure) \le 1$, the upper bound follows from a similar computation. 
\end{proof}

\begin{lemma}[Gap bound of information value]
\label{lemma_information_value_positive}
    For all $\model \in \models$ and whatever the leveling parameter $\epsilon \ge 0$, we have
    \begin{equation*}
        \ivalue_\epsilon (\unit, \model)
        \ge
        \parens*{
            \frac{\gaingap(\model)}{
                4 \diameter(\model)
            }
        }^2
        .
    \end{equation*}
\end{lemma}
\begin{proof}
    The result follows from a straight forward computation.
    We have
    \begin{align*}
        \ivalue_\epsilon (\model) 
        & =
        \inf_{\model^\dagger \in \confusing_\epsilon(\model)}
        \braces*{
            \sum_{\pair \in \pairs}
            \KL(\model(\pair)||\model^\dagger (\pair))
        }
        \\
        & \overset{(\dagger)}\ge
        \inf_{\model^\dagger \in \confusing(\model)}
        \braces*{
            \sum_{\pair \in \pairs}
            \KL(\model(\pair)||\model^\dagger (\pair))
        }
        \\
        & \ge
        \inf_{\model^\dagger \in \confusing(\model)}
        \max_{\pair \in \pairs} \braces*{
            \KL(\model(\pair)||\model^\dagger (\pair))
        }
        \\
        & \overset{(\ddagger)}\ge
        \inf_{\model^\dagger \in \confusing(\model)}
        \max_{\pair \in \pairs} \braces*{
            \parens*{
                \norm{\reward^\dagger(\pair) - \reward(\pair)}_\infty
                + \norm{\kernel^\dagger(\pair) - \kernel(\pair)}_1
            }^2
        }
        \\
        & =
        \inf_{\model^\dagger \in \confusing(\model)}
        \norm{\model^\dagger - \model}^2
        \\
        & \overset{(\S)}\ge
        \parens*{
            \frac{\gaingap(\model)}{
                4 \diameter(\model)
            }
        }^2
    \end{align*}
    where 
    $(\dagger)$ follows from $\confusing(\epsilon; \model) \subseteq \confusing(\model)$;
    $(\ddagger)$ follows from Pinsker's inequality; and
    $(\S)$ follows from \Cref{lemma_confusing_separated}.
\end{proof}

\begin{lemma}[Continuity in $\imeasure$]
\label{lemma_information_value_continuous_measure}
    Fix $\model \in \models$, a leveling parameter $\epsilon \ge 0$ and $\delta > 0$. 
    For all $\imeasure, \imeasure' \in \probabilities_\delta (\pairs)$ such that $\norm{\imeasure' - \imeasure}_\infty < \frac 12 \delta$, we have
    \begin{equation*}
        \abs*{
            \ivalue_\epsilon (\imeasure', \model)
            -
            \ivalue_\epsilon (\imeasure', \model)
        }
        \le
        \frac{2 C_\epsilon(\imeasure, \model)}{\delta}
        \cdot
        \norm{\imeasure' - \imeasure}_\infty
        .
    \end{equation*}
\end{lemma}
\begin{proof}
    Let $\imeasure, \imeasure' \in \probabilities_\delta (\pairs)$. 
    By \Cref{lemma_information_value_bounded}, $\ivalue_\epsilon (\imeasure, \model)$ and $\ivalue_\epsilon (\imeasure', \model)$ are either both finite or infinite. 
    If $\confusing(\model) = \emptyset$, then these quantities are infinite and there is nothing to be said; Let us assume that $\ivalue_\epsilon (\imeasure, \model), \ivalue_\epsilon (\imeasure', \model) < \infty$. 
    Accordingly, there exist $\model^\dagger_\imeasure$ and $\model^\dagger_{\imeasure'}$ that are a $(\imeasure, \epsilon, \model)$-critical and a $(\imeasure', \epsilon, \model)$-critical model respectively. 
    Note that since $\norm{\imeasure' - \imeasure}_\infty < \frac 12 \delta$ with $\imeasure \in \probabilities_\delta(\pairs)$, we can write $\imeasure' = \imeasure + u \imeasure$ with $\norm{u}_\infty \le \frac 12 \delta^{-1} \norm{\imeasure' - \imeasure}_\infty$. 
    Therefore,
    \begin{align*}
        \ivalue_\epsilon (\imeasure', \model) 
        & := 
        \inf_{\model^\dagger \in \confusing_\epsilon(\model)}
        \braces*{
            \sum_{\pair \in \pairs} 
            \imeasure'(\pair)
            \KL(\model(\pair)||\model^\dagger(\pair))
        }
        \\
        & \le 
        \sum_{\pair \in \pairs}
        \imeasure'(\pair)
        \KL(\model(\pair)||\model^\dagger_{\imeasure}(\pair))
        \\
        & \le
        (1 + \norm{u}_\infty)
        \sum_{\pair \in \pairs}
        \imeasure(\pair)
        \KL(\model(\pair)||\model^\dagger_{\imeasure}(\pair))
        \le
        \parens*{
            1 + \frac{\norm{\imeasure' - \imeasure}_\infty}{\delta}
        }
        \ivalue_\epsilon (\imeasure, \model)
        .
    \end{align*}
    As $\imeasure$ and $\imeasure'$ play a symmetric role in the proof, they can be interchanged. 
    This immediately provides a lower bound of $\ivalue_\epsilon (\imeasure', \model)$ with respect to $\ivalue_\epsilon (\imeasure, \model)$.
    We conclude by using $(1 + \frac{\norm{\imeasure' - \imeasure}_\infty}{\delta})^{-1} \ge 1 - \frac{2 \norm{\imeasure' - \imeasure}_\infty}{\delta}$, that holds because $\norm{\imeasure' - \imeasure}_\infty < \frac 12 \delta$. 
\end{proof}

\begin{lemma}[Continuity in $\model$]
\label{lemma_information_value_continuous_model}
    Let $\models$ be an ambient space with Bernoulli rewards (\Cref{assumption_bernoulli}) in product form (\Cref{assumption_space}).
    Fix $\model \in \models$ a communicating model.
    There exist functions $\alpha, \beta_1, \beta_2, \beta : \models \to \R_+^*$ that are polynomial in $\abs{\pairs}, \diameter, \dmin(\model)$ and $\ivalue(\unit, \model)$, such that, for all leveling parameter $\epsilon > 0$ with $\confusing_\epsilon (\model) = \confusing(\model)$ and all $\delta > 0$, if $\model' \in \models$ satisfies
    \begin{equation}
    \label{equation_information_value_continuous_model_condition}
        \snorm{\model' - \model}
        \le
        \min \braces*{
            \frac{\epsilon}{\alpha(\model)}
            ,
            \exp \parens*{
                - \frac 1\delta \beta_1(\model) - \beta_2(\model)
            }
        }
    \end{equation}
    then for all $\imeasure \in \probabilities_\delta (\model)$, we have
    \begin{equation*}
        \abs*{
            \ivalue_\epsilon (\imeasure, \model')
            -
            \ivalue_\epsilon (\imeasure, \model)
        }
        \le 
        \exp\parens*{\beta(\model)}
        \snorm{\model' - \model}
        .
    \end{equation*}
\end{lemma}

\paragraph{Proof idea.}
The proof is difficult. 
We start by introducing $\models_\epsilon$, a well-behaved neighborhood of $\model$ for the support-aware norm $\snorm{-}$, given by
\begin{equation}
\label{equation_proof_information_value_continuous_model_0}
    \models_\epsilon := \braces*{
        \model'
        :
        \snorm{\model' - \model} 
        < \min \braces*{
            \frac 1{\diameter(\model)},
            \frac {\gaingap(\model)}{4 \max \braces{1, \diameter(\policy_u; \model)}},
            \frac 12 \dmin(\kernel),
            \frac {\epsilon}{\alpha (\model)}
        }
    }
\end{equation}
where 
(I) the condition with $\diameter(\model)^{-1}$ is to make sure that $\diameter(\model') \le 2 \diameter(\model)$, see \Cref{lemma_variations_diameter};
(II) the condition with $\frac 14 \diameter(\model)^{-1} \gaingap(\model)$ is to make sure that $\model' \notin \confusing(\model)$, see \Cref{lemma_alternative_separated_weak};
(III) the condition with $\frac12 \dmin(\kernel)$ is a bit technical, and is to obtain uniform bounds on escaping probabilities; and
(IV) the condition with $\frac \epsilon{\alpha(\model)}$ follows from \Cref{proposition_continuity_near_optimality}, stating that there exists $\alpha(\model) < \infty$ such that if the leveling parameter $\epsilon > 0$ is small enough, every model $\model' \in \models$ that satisfies 
\begin{equation}
\label{equation_proof_information_value_continuous_model_1}
    \snorm{\model' - \model} \le \frac{\epsilon}{\alpha(\model)}
\end{equation}
is such that $\optpairs(\flatmodel{\model'}{\epsilon}) = \optpairs(\model)$. 
Note that by \Cref{lemma_property_confusing}, such a model $\model'$ satisfies
\begin{equation}
\label{equation_proof_information_value_continuous_model_2}
    \confusing_\epsilon(\model') 
    = 
    \braces*{
        \model'^\dagger \gg \model 
        : 
        \model'^\dagger = \model' \text{~on~} \optpairs(\model) 
        \text{~and~} 
        \optgain(\model'^\dagger) > \optgain(\model')
    }
    .
\end{equation}
Fix $\hat{\model}, \widetilde{\model} \in \models_\epsilon$ generic Markov decision processes in the neighborhood of $\model$.
The goal is to upper-bound the information value of $\hat{\model}$ relatively to the information value of $\widetilde{\model}$. 
We will conclude by instantiating $\hat{\model}$ and $\widetilde{\model}$ as $\model = \hat{\model}, \model' = \widetilde{\model}$ and $\model = \widetilde{\model}, \model' = \hat{\model}$.

Let $\widetilde{\model}^\dagger$ be a $(\imeasure, \epsilon, \widetilde{\model})$-critical model. 
We introduce $\hat{\model}^\dagger$ the model given by
\begin{equation}
\label{equation_proof_information_value_continuous_model_3}
    \hat{\model}^\dagger(\pair)
    :=
    \begin{cases}
        \hat{\model}(\pair) & \text{if~} \pair \in \optpairs(\model);
        \\
        \widetilde{\model}^\dagger(\pair) & \text{if~} \pair \notin \optpairs(\model).
    \end{cases}
\end{equation}
In general, we have $\hat{\model}^\dagger \notin \confusing_\epsilon (\hat{\model})$ because the condition $\optgain(\hat{\model}^\dagger) > \optgain(\hat{\model})$ is false---barely false, but false nonetheless.
But if it is false, then because $\optgain(\widetilde{\model}^\dagger) > \optgain(\widetilde{\model})$ and $\hat{\model} \approx \widetilde{\model}$, there is a slight modification of $\hat{\model}^\dagger$ that belongs to $\confusing_\epsilon(\hat{\model})$.
In the first place, the condition ``$\optgain(\hat{\model}^\dagger) > \optgain(\hat{\model})$'' may be false because by going from $\widetilde{\model}^\dagger$ to $\hat{\model}^\dagger$, the optimal pairs of $\widetilde{\model}^\dagger$ may be degraded while the (leveled) optimal pairs of $\widetilde{\model}$ may be improved. 
Naturally, the idea of the proof is to \strong{repair} $\hat{\model}^\dagger$ into $\hat{\model}^\ddagger \approx \hat{\model}^\dagger$ with $\hat{\model}^\ddagger \in \confusing_\epsilon(\model)$ by increasing the reward function at well-chosen pairs. 

Given $\eta > 0$, we introduce the repaired model as 
\begin{equation}
\label{equation_repaired_model}
    \hat{\model}^\ddagger_\eta
    \equiv
    \hat{\model}^{\ddagger \eta}
    \equiv 
    \parens{
        \pairs,
        \hat{\reward}^\dagger + \eta~\unit_{\pairs^\dagger},
        \hat{\kernel}^\dagger
    }
    \text{\quad where \quad}
    \pairs^\dagger := \optpairs(\widetilde{\model}^\dagger) \setminus \optpairs(\model)
    ,
\end{equation}
i.e., as the copy of $\hat{\model}^\dagger$ where the reward function is increased by $\eta$ at every pair of $\pairs^\dagger$. 
The difficulty of the proof consists in choosing $\eta > 0$ correctly. 

\medskip
\def\proofname{Proof of \Cref{lemma_information_value_continuous_model}}
\begin{proof}
    \def\proofname{Proof}
    We proceed with the Markov decision processes $\hat{\model}, \widetilde{\model}, \widetilde{\model}^\dagger, \hat{\model}^\dagger$ and $\hat{\model}^\ddagger_\eta \equiv \hat{\model}^{\ddagger \eta}$ defined upstream. 
    The proof is organized as follows.

    We begin by the end.
    We show in \STEP{1} that if $\hat{\model}^{\ddagger\eta}$ is an $\epsilon$-confusing model and if $\eta$ is small enough, then $\ivalue_\epsilon (\imeasure, \hat{\model})$ is upper bounded with respect to $\ivalue_\epsilon (\imeasure, \model)$; and that is the desired conclusion.
    The remaining steps consist in tuning $\eta > 0$.
    In \STEP{2}, we show that $\hat{\model}^{\ddagger \eta}$ is well-defined if $\eta$ is small enough relatively to $\model$ and $\epsilon$. 
    In \STEP{3} and \STEP{4}, we detail how the optimal gain of $\hat{\model}^{\ddagger \eta}$ is influenced by the choice of $\eta$.
    The point is to derive, in \STEP{5}, a lower bound $c(\model, \epsilon) > 0$ on $\eta$ that is sufficient to guarantee that $\hat{\model}^{\ddagger \eta} \in \confusing_\epsilon (\model)$.
    The final result is obtained by choosing $\eta = c(\model, \epsilon)$, that makes sure that $\hat{\model}^{\ddagger \eta}$ is well-defined, and to inject this expression in \STEP{1}.

    \medskip
    \par
    \noindent
    \STEP{1}
    \textit{
        For every $\eta > 0$ with $\eta < \min\braces{1 - \max(\widetilde{\reward}^\dagger), \frac 12 \max(\widetilde{\reward}^\dagger)}$ and $\hat{\model}^\ddagger_\eta \in \confusing_\epsilon(\model)$, we have
        \begin{equation}
            \ivalue_\epsilon(\imeasure, \hat{\model})
            \le
            \ivalue_\epsilon (\imeasure, \widetilde{\model})
            + 2 \beta_0 \parens*{
                \ivalue_\epsilon (\imeasure, \widetilde{\model})
                +
                2 \log(2e \abs{\states})
            }
        \end{equation}
        where $\beta_0 := \max \braces{\max(\widetilde{\reward}^\dagger)^{-1} \eta, \dmin(\model)^{-1} \snorm{\hat{\model} - \widetilde{\model}}}$.
    }
    \medskip
    \begin{subproof}
        Unfolding the definition of $\ivalue_\epsilon (\imeasure, \hat{\model})$, we have
        \begin{align*}
            C_\epsilon (\imeasure, \hat{\model})
            & :=
            \inf_{\hat{\model}' \in \confusing_\epsilon(\hat{\model})}
            \braces*{
                \sum_{\pair \in \pairs} 
                \imeasure(\pair)
                \KL(\hat{\model}(\pair)||\hat{\model}'(\pair))
            }
            \\
            & \overset{(\dagger)}\le
            \sum_{\pair \in \pairs} 
            \imeasure(\pair)
            \KL(\hat{\model}(\pair)||\hat{\model}^\ddagger_\eta(\pair))
            \\
            & \overset{(\ddagger)}=
            \sum_{\pair \in \pairs}
            \imeasure(\pair)
            \KL(\hat{\kerneld}(\pair)||\hat{\kerneld}^\ddagger_\eta (\pair))
            +
            \sum_{\pair \in \pairs}
            \imeasure(\pair)
            \KL(\hat{\rewardd}(\pair)||\hat{\rewardd}^\ddagger_\eta (\pair))
        \end{align*}
        where 
        $(\dagger)$ uses that $\hat{\model}_\eta^{\ddagger} \in \confusing_\epsilon(\hat{\model})$.
        Both sums in $(\ddagger)$ are reworked using \Cref{lemma_variations_kullback_leibler}.

        We begin with the sum involving kernels. 
        Note that for $\gamma := \dmin(\hat{\kernel})^{-1} \snorm{\widetilde{\kernel} - \hat{\kernel}}_\infty$, we have $(1 - \gamma) \hat{\kernel}(\state|\pair) \le \widetilde{\kernel}(\state|\pair) \le (1 + \gamma) \hat{\kernel}(\state|\pair)$ for all $\pair \in \pairs$ and $\state \in \support(\widetilde{\kernel}(\pair))$. 
        By assumption, we further have $\dmin(\model)^{-1} \snorm{\widetilde{\model} - \hat{\model}} < \frac 12$, so we can bound the varaitions of $\KL(-||-)$ by invoking \Cref{lemma_variations_kullback_leibler} to obtain
        \begin{align*}
            \sum_{\pair \in \pairs}
            \imeasure(\pair)
            \KL(\hat{\kernel}(\pair)||\hat{\kernel}_\eta^\ddagger(\pair))
            & \overset{(\dagger)}=
            \sum_{\pair \notin \optpairs(\model)}
            \imeasure(\pair)
            \KL(\hat{\kernel}(\pair)||\hat{\kernel}^\dagger(\pair))
            \\
            & \overset{(\ddagger)}=
            \sum_{\pair \notin \optpairs(\model)}
            \imeasure(\pair)
            \KL(\hat{\kernel}(\pair)||\widetilde{\kernel}^\dagger(\pair))
            \\
            & \overset{(\S)}\le
            \parens*{
                1 + \beta
            }
            \sum_{\pair \in \pairs}
            \imeasure(\pair)
            \KL(\widetilde{\kernel}(\pair)||\widetilde{\kernel}^\dagger(\pair))
            +
            \beta \log\parens*{e \abs{\pairs}}
        \end{align*}
        where
        $(\dagger)$ uses that $\hat{\kernel}^{\ddagger\eta} = \hat{\kernel}^\dagger$ by definition and that $\hat{\model}$ and $\hat{\model}^{\ddagger\eta}$ only differ outside of $\optpairs(\model)$;
        $(\ddagger)$ uses that $\hat{\kernel}^\dagger (\pair) = \widetilde{\kernel}^\dagger (\pair)$ for $\pair \notin \optpairs(\model)$; and 
        $(\S)$ introduces $\beta := 2~\dmin(\model)^{-1} \snorm{\widetilde{\model} - \hat{\model}}$ and invokes \Cref{lemma_variations_kullback_leibler}.

        Continuing with the sum involving rewards, we note that for $\gamma := \dmin(\reward)^{-1} \snorm{\widetilde{\reward} - \reward}_\infty$, we have $(1 - \gamma) \hat{\rewardd}(i|\pair) \le \widetilde{\rewardd}(i|\pair) \le (1 + \gamma) \hat{\rewardd}(i|\pair)$ for all $\pair \in \pairs$ and $i = 0, 1$. 
        Introduce the shorthand $\alpha := \max\braces{\max(\widetilde{\reward}^\dagger)^{-1} \eta, \dmin(\model)^{-  1} \snorm{\widetilde{\model} - \hat{\model}}}$.
        Note that $\alpha < \frac 12$.
        We have
        \begin{align*}
            \sum_{\pair \in \pairs} 
            \imeasure(\pair)
            \KL(\hat{\rewardd}(\pair)||\hat{\rewardd}^\ddagger_\eta (\pair))
            & =
            \sum_{\pair \in \pairs}
            \imeasure(\pair)
            \kl\parens*{
                \hat{\reward}(\pair)
                \middle\|
                \hat{\reward}^\dagger(\pair) + \eta \indicator{\pair \in \pairs^\dagger}
            }
            \\
            & \overset{(\dagger)}=
            \sum_{\pair \notin \optpairs(\model)}
            \imeasure(\pair)
            \kl\parens*{
                \hat{\reward}(\pair)
                \middle\|
                \hat{\reward}^\dagger(\pair) + \eta \indicator{\pair \in \pairs^\dagger}
            }
            \\
            & \overset{(\ddagger)}=
            \sum_{\pair \notin \optpairs(\model)}
            \imeasure(\pair)
            \kl\parens*{
                \hat{\reward}(\pair)
                \middle\|
                \widetilde{\reward}^\dagger(\pair) + \eta \indicator{\pair \in \pairs^\dagger}
            }
            \\
            & \overset{(\S)}\le
            \sum_{\pair \notin \optpairs(\model)}
            \imeasure(\pair)
            \kl\parens*{
                \hat{\reward}(\pair)
                \middle\|
                \parens*{
                    1 + \frac{\eta}{\max(\widetilde{\reward}^\dagger)}
                }
                \widetilde{\reward}^\dagger(\pair) 
            }
            \\
            & \overset{(\$)}\le
            (1 + 2\alpha) 
            \sum_{\pair \in \pairs}
            \imeasure(\pair)
            \KL(\widetilde{\rewardd}(\pair)||\widetilde{\rewardd}^\dagger(\pair))
            + 2 \alpha \log (2e)
        \end{align*}
        where 
        $(\dagger)$ follows from $\optpairs(\model) \cap \pairs^\dagger = \emptyset$;
        $(\ddagger)$ follows by construction of $\hat{\model}^\dagger$;
        $(\S)$ follows by monotonicity of $v \mapsto \kl(\hat{\reward}(\pair)||v)$ for $v \ge \hat{\reward}(\pair)$; and 
        $(\$)$ follows by \Cref{lemma_variations_kullback_leibler}. 

        Conclude by combining everything together, with $\beta \le \alpha$ and $1 \le \abs{\states}$. 
    \end{subproof}

    \par
    \noindent
    \STEP{2}
    \textit{
        We have
        \begin{equation*}
            \max(\widetilde{\reward}^\dagger) 
            \le
            1 - \exp \parens*{
                - \frac{
                    \frac{\ivalue_\epsilon(\unit, \widetilde{\model})}{\delta}
                    + \log(2)
                }{
                    1 - \max({\reward}) - \norm{\widetilde{\reward} - \reward}_\infty
                }
            }
            .
        \end{equation*}
        In particular, $\hat{\model}_\eta^\ddagger$ is well-defined for all $\eta$ satisfying
        \begin{equation*}
            0
            < \eta
            < \exp \parens*{
                - \frac {
                    \frac{\ivalue_\epsilon(\unit, \widetilde{\model})}{\delta}
                    + \log(2)
                }{
                    1 - \max({\reward}) - \norm{\widetilde{\reward} - \reward}_\infty
                }
            }
            - \norm{\hat{\reward} - \widetilde{\reward}}_\infty
        .
        \end{equation*}
    }
    \begin{subproof}
        Using the bound on $\ivalue_\epsilon (\unit, \widetilde{\model})$ from \Cref{lemma_information_value_bounded}, we obtain
        \begin{align*}
            \ivalue_\epsilon (\imeasure, \widetilde{\model})
            & \ge
            \delta~\ivalue_\epsilon (\unit, \widetilde{\model})
            \\
            & \ge
            \delta 
            \max_{\pair \in \pairs} \braces*{
                \KL(\widetilde{\model}(\pair)||\widetilde{\model}^\dagger(\pair))
            }
            \\
            & \ge
            \delta
            \max_{\pair \in \pairs} \braces*{
                \KL(\widetilde{\rewardd}(\pair)||\widetilde{\rewardd}^\dagger(\pair))
            }
            \\
            & \ge
            \delta
            \max_{\pair \in \pairs} \braces*{
                - \entropy(\widetilde{\reward}(\pair))
                + \widetilde{\reward}(\pair) \log \parens*{\frac 1{\widetilde{\reward}^\dagger(\pair)}}
                + (1 - \widetilde{\reward}(\pair)) \log \parens*{\frac 1{1 - \widetilde{\reward}^\dagger(\pair)}}
            }
            .
        \end{align*}
        In particular, for all $\pair \in \pairs$, $(1 - \widetilde{\reward}(\pair)) \log \parens*{\frac 1{1 - \widetilde{\reward}^\dagger(\pair)}} \le \entropy(\widetilde{\reward}(\pair)) + \frac 1\delta \ivalue_\epsilon(\unit, \model)$.
        We obtain the desired result by solving in $\hat{\reward}^\dagger(\pair)$ and using $\max(\widetilde{\reward}) \le \max(\reward) + \norm{\widetilde{\reward} - \reward}_\infty$. 
    \end{subproof}

    \par
    \noindent
    \STEP{3}
    \textit{
        Assume that $\snorm{\widetilde{\model} - \model} < \diameter(\model)^{-1}$.
        Let $\widetilde{\policy}^\dagger$ be a bias optimal policy of $\widetilde{\model}^\dagger$.
        For all $\policy \in \optpolicies({\model})$, we have
        \begin{align}
        \label{equation_proof_information_value_continuous_model_4}
            \gainof{\widetilde{\policy}^\dagger}(\hat{\model}_\eta^\ddagger)
            & \ge 
            \optgain(\widetilde{\model}^\dagger)
            - (1 + 2 \diameter({\model})) \snorm{\hat{\model} - \widetilde{\model}} \unit_\pairs 
            + \eta \cdot \gainof{\widetilde{\policy}^\dagger}(\unit_{\pairs^\dagger}, \hat{\kernel}^\dagger)
            ;
            \\
        \label{equation_proof_information_value_continuous_model_5}
            \gainof{\policy}(\hat{\model}_\eta^\ddagger) 
            & \le 
            \optgain(\widetilde{\model}^\dagger)
            + (1 + 2 \diameter({\model})) \snorm{\hat{\model} - \widetilde{\model}}
            .
        \end{align}%
    }
    \begin{subproof}
        Because $\snorm{\widetilde{\model} - \model} < \diameter(\model)^{-1}$, we have $\diameter(\widetilde{\model}) \le 2 \diameter(\model)$ by \Cref{lemma_variations_diameter}.

        For $\policy \in \policies$, the Cesàro-limit of $\kernel_\policy^t$ is denoted $\imeasure_\policy := \lim T^{-1} \sum_{t=0}^{T-1} \kernel_\policy^t$.
        We further denote $\widetilde{\reward}_\policy^\dagger, \hat{\reward}_\policy^\dagger$ and $\hat{\reward}_\policy^{\ddagger\eta}$ the reward functions of $\policy$ in $\widetilde{\model}^\dagger, \hat{\model}^\dagger$ and $\hat{\model}^\ddagger_\eta$ respectively. 

        Starting with the first inequality, writing $\policy \equiv \widetilde{\policy}^\dagger$ for short, we have
        \begin{align*}
            \gainof{\policy} (\hat{\model}_\eta^\ddagger)
            = \hat{\imeasure}_\policy^{\ddagger \eta} \cdot \hat{\reward}_{\policy}^{\ddagger \eta}
            & \overset{(\dagger)}= 
            \hat{\imeasure}_\policy^{\dagger} \cdot
            \parens*{\hat{\reward}_\policy^\dagger + \eta ~ \unit_{\pairs^\dagger}}
            \\
            & \overset{(\ddagger)}\ge
            \gainof{\policy}(\widetilde{\model})
            - \parens*{
                1 + \frac 12 \vecspan{\biasof{\policy}(\widetilde{\model}^\dagger)}
            } \snorm{\hat{\model}^\dagger - \widetilde{\model}^\dagger}
            + \eta ~ \gainof{\policy}(\unit_{\pairs^\dagger}, \hat{\kernel}^\dagger)
            \\
            & \ge
            \gainof{\policy}(\widetilde{\model})
            - \parens*{
                1 + \frac 12 \vecspan{\biasof{\policy}(\widetilde{\model}^\dagger)}
            } \snorm{\hat{\model} - \widetilde{\model}}
            + \eta ~ \gainof{\policy}(\unit_{\pairs^\dagger}, \hat{\kernel}^\dagger)
            \\
            & \overset{(\S)}\ge 
            \gainof{\policy}(\widetilde{\model})
            - \parens*{
                1 + \diameter(\widetilde{\model})
            } \snorm{\hat{\model} - \widetilde{\model}}
            + \eta ~ \gainof{\policy}(\unit_{\pairs^\dagger}, \hat{\kernel}^\dagger)
            \\
            & \overset{(\$)}\ge
            \gainof{\policy}(\widetilde{\model})
            - \parens*{1 + 2 \diameter({\model})} \snorm{\hat{\model} - \widetilde{\model}}
            + \eta ~ \gainof{\policy}(\unit_{\pairs^\dagger}, \hat{\kernel}^\dagger)
        \end{align*}
        where 
        in $(\dagger)$ we wrote $\unit_{\pairs^\dagger} := \sum_{\pair \in \pairs^\dagger} \unit_\pair$; 
        $(\ddagger)$ follows from \Cref{lemma_unichain_gain_variations};
        $(\S)$ follows from $\vecspan{\biasof{\policy}(\widetilde{\model}^\dagger)} \le \diameter(\widetilde{\model})$, because $\policy \equiv \widetilde{\policy}^\dagger$ is a bias optimal in $\widetilde{\model}^\dagger$ which is a critical model, hence $\vecspan{\biasof{\policy}(\widetilde{\model}^\dagger)} = \vecspan{\optbias(\widetilde{\model}^\dagger)} \le \diameter(\widetilde{\model})$ by \Cref{lemma_property_critical}, assertion (1);
        and $(\$)$ follows from $\diameter(\widetilde{\model}) \le 2 \diameter(\model)$.

        For the second inequality, first note that since $\norm{\widetilde{\model} - \model} < \frac 14 (\max\braces{1, \diameter(\policy_u; \model)})^{-1} \gaingap(\model)$ where $\policy_u$ is the uniform policy, all gain optimal policies of $\widetilde{\model}$ are gain optimal in $\model$ by \Cref{lemma_confusing_separated}.
        As $\snorm{\widetilde{\model} - \model} < \infty$, we further have $\widetilde{\model} \sim \model$ so $\optpairs(\widetilde{\model}) \subseteq \optpairs(\model)$.
        So, for all $\policy \in \optpolicies({\model})$, we have
        \begin{align*}
            \gainof{\policy}(\hat{\model}_\eta^\ddagger)
            =
            \hat{\imeasure}_\policy^{\ddagger \eta}
            \cdot
            \hat{\reward}_\policy^{\ddagger \eta}
            & \overset{(\dagger)}=
            \hat{\imeasure}_\policy \cdot \hat{\reward}_\policy 
            \\
            & \overset{(\ddagger)}\ge
            \gainof{\policy}(\widetilde{\model})
            - \parens*{1 + \frac 12 \vecspan{\optbias(\widetilde{\model})}} \snorm{\hat{\model} - \widetilde{\model}}
            \\
            & \overset{(\S)}\ge
            \gainof{\policy}(\widetilde{\model})
            - \parens*{1 + \diameter(\widetilde{\model})} \snorm{\hat{\model} - \widetilde{\model}}
            \\
            & \overset{(\$)}\ge
            \gainof{\policy}(\widetilde{\model})
            - \parens*{1 + 2 \diameter({\model})} \snorm{\hat{\model} - \widetilde{\model}}
            ,
        \end{align*}
        where
        $(\dagger)$ follows from $\hat{\model}^{\ddagger\eta} \sim \widetilde{\model}$ and $\hat{\model}^{\ddagger\eta}(\pair) = \hat{\model}(\pair)$ for $\pair \in \optpairs(\widetilde{\model})$;
        $(\ddagger)$ follows from the gain deviation inequality of \Cref{lemma_unichain_gain_variations};
        $(\S)$ follows from \Cref{lemma_bias_diameter} and
        $(\$)$ follows from $\diameter(\widetilde{\model}) \le 2\diameter(\model)$.
    \end{subproof}

    \par
    \noindent
    \STEP{4}
    \textit{
        Assume that $\hat{\kernel}(\state'|\state, \action) \ge \frac 12 \kernel(\state'|\state, \action)$ for all $\state, \action, \state'$. 
        Writing $\unit := \sum_{\state \in \states} \unit_\state$, we have
        \begin{equation*}
            \gainof{\widetilde{\policy}^\dagger} (\unit_{\pairs^\dagger}, \hat{\kernel}^\dagger)
            \ge
            \frac{\dmin(\hat{\kernel})^{\abs{\states}-1}}{\abs{\states}} \unit
            \ge
            \frac{\dmin(\kernel)^{\abs{\states}-1}}{2^{\abs{\states}-1}\abs{\states}} \unit
            .
        \end{equation*}
    }
    \begin{subproof}
        For conciseness, we write $\policy \equiv \widetilde{\policy}^\dagger$.

        Recall that 
        (1) $\pairs^\dagger = \optpairs(\widetilde{\model}^\dagger) \setminus \optpairs(\model)$,
        (2) $\policy \equiv \widetilde{\policy}^\dagger$ is a bias optimal policy of $\widetilde{\model}^\dagger$; 
        and that (3) $\widetilde{\model}$ and $\widetilde{\model}^\dagger$ are identical on $\optpairs(\model)$ by \eqref{equation_proof_information_value_continuous_model_2} and $\widetilde{\model}^\dagger \in \confusing_\epsilon(\widetilde{\model})$.
        By mutual absolute continuity of $\widetilde{\model}^\dagger$ and $\hat{\model}^\dagger$, recurrent components of $\policy$ are the same in $\widetilde{\model}^\dagger$ and $\hat{\model}^\dagger$.
        They are denoted $\pairs_1, \ldots, \pairs_m$.
        Let $\hat{\imeasure}^\dagger_i$ be the unique probability invariant measure of $\policy$ in $\hat{\model}^\dagger$ that is supported in $\pairs_i$.
        Given $\state \in \states$, note that $\gainof{\widetilde{\policy}^\dagger}(\state; \unit_{\pairs^\dagger}, \hat{\kernel}^\dagger)$ is a convex combination of $(\hat{\imeasure}_i^\dagger \cdot \unit_{\pairs^\dagger})_{1 \le i \le m}$.
        Accordingly, it is enough to show that for all $i = 1, \ldots, m$, we have
        \begin{equation}
        \label{equation_information_value_continuity_1}
            \sum_{\pair \in \pairs^\dagger}
            \hat{\imeasure}_i^\dagger (\pair)
            \ge
            \frac{\dmin(\hat{\kernel})^{\abs{\states}-1}}{\abs{\states}}
            .
        \end{equation}
        We focus on showing \eqref{equation_information_value_continuity_1} for an arbitrary $i \in \braces{1, \ldots, m}$.

        Let $(\state_0, \action_0) \in \pairs_i$ maximizing $\hat{\imeasure}_i^\dagger(\state_0)$, where we use the writing shorthand $\hat{\imeasure}^\dagger_i(\state) := \sum_{\action \in \actions(\state)} \hat{\imeasure}_i^\dagger(\state, \action)$;
        In particular, $\hat{\imeasure}_i(\state_0) \ge \abs{\states}^{-1}$.

        If $(\state_0, \action_0) \in \pairs^\dagger$, there is nothing left to prove. 

        Let us assume otherwise, i.e., $(\state_0, \action_0) \notin \pairs^\dagger$.
        Because $\policy$ is gain optimal in $\widetilde{\model}^\dagger \in \confusing(\widetilde{\model})$, it is not gain optimal in $\model$, so $\pairs_i \cap \pairs^\dagger \ne \emptyset$.
        By considering all the paths in $\pairs_i$ of length $n = 1, \ldots, \abs{\states}$ starting from $\state_0$, at least one must hit $\pairs^i \cap \pairs^\dagger$.
        So, there exist $(\state'_0, \action'_0) \in \pairs_i \cap \pairs^\dagger$ together with a path $(\state_1, \action_1, \ldots, \state_n)$ of length $n \le \abs{\states}$ such that $\state_1 = \state_0$, $\state_n = \state'_0$, $\kernel(\state_{j+1}|\state_j,\action_j) > 0$ and $(\state_j, \action_j) \in \pairs_i \setminus \pairs^\dagger$.
        In particular, $(\state_j, \action_j) \in \optpairs(\widetilde{\model})$ for $j \le n - 1$, hence $\hat{\kernel}^\dagger (\state_j, \action_j) = \hat{\kernel}(\state_j, \action_j)$ by definition of $\hat{\model}^\dagger$.
        We get
        \begin{align*}
            \hat{\imeasure}_i^\dagger (\state', \action'_0)
            & \ge 
            \policy (\action'_0|\state'_0)
            \parens*{
                \product_{j=1}^{n-1}
                \policy(\action_j|\state_j) 
                \hat{\kernel}^\dagger(\state_{j+1}|\state_j, \action_j)
            } \hat{\imeasure}_i^\dagger (\state_0)
            \\
            & \ge 
            \policy (\action'_0|\state'_0)
            \parens*{
                \product_{j=1}^{n-1}
                \policy(\action_j|\state_j) 
                \hat{\kernel} (\state_{j+1}|\state_j, \action_j)
            } \hat{\imeasure}_i^\dagger (\state_0)
            \overset{(\dagger)}\ge
            \frac{\dmin(\hat{\kernel})^{n-1}}{\abs{\states}}
            \overset{(\ddagger)}\ge
            \frac{\dmin({\kernel})^{n-1}}{2^{n-1}\abs{\states}}
        \end{align*}
        where 
        $(\dagger)$ follows from the observation that $\policy$ is deterministic and 
        $(\ddagger)$ that $\hat{\kernel}(\state'|\state,\action) \ge \frac 12 \kernel(\state'|\state, \action)$ for all $\state, \action, \state'$.
        Using that $n \le \abs{\states}$, this shows \eqref{equation_information_value_continuity_1} and concludes the proof. 
    \end{subproof}

    \par 
    \noindent
    \STEP{5}
    \textit{
        Assume that $\hat{\kernel}(\state'|\state, \action) \ge \frac 12 \kernel(\state'|\state, \action)$ for all $\state, \action, \state'$ and that $\snorm{\widetilde{\model} - \model} < \diameter(\model)^{-1}$.
        We have $\hat{\model}_\eta^\ddagger \in \confusing_\epsilon(\hat{\model})$ for every 
        \begin{equation*}
            \eta 
            >
            \frac{
                2^{\abs{\states}} (1 + 2 \diameter(\model)) 
                \snorm{\hat{\model} - \widetilde{\model}}
            }{
                \dmin(\kernel)^{\abs{\states}-1} 
                \abs{\states}
            }
            .
        \end{equation*}
    }
    \begin{subproof}
        By definition, $\hat{\model}^{\ddagger \eta} \in \confusing_\epsilon(\hat{\model})$ if, and only if 
        (I)~$\hat{\model}^{\ddagger \eta} = \hat{\model}$ on $\flatme{\optpairs}{\epsilon}(\model)$ and
        (II)~$\hat{\model}^{\ddagger \eta}$ belongs to the confusing set of $\hat{\model}$, i.e., $\optpolicies(\hat{\model}^{\ddagger \eta}) \cap \optpolicies(\hat{\model}) = \emptyset$.
        
        For (I), note that $\flatme{\optpairs}{\epsilon}(\hat{\model}) = \optpairs(\model)$ because $\snorm{\hat{\model} - \model} \le \alpha(\model)^{-1} \epsilon$, see \eqref{equation_proof_information_value_continuous_model_1};
        And $\hat{\model}^{\ddagger \eta} = \hat{\model}$ on $\optpairs(\model)$ by construction, see \eqref{equation_proof_information_value_continuous_model_3} and \eqref{equation_repaired_model}.

        We derive a sufficient condition for (II) following \STEP{3} and \STEP{4}.
        By \Cref{lemma_property_confusing}, $\hat{\model}^{\ddagger \eta} \in \confusing_\epsilon(\model)$ if, and only if $\optgain(\hat{\model}^{\ddagger \eta}) > \optgain(\hat{\model})$.

        Because $\snorm{\hat{\model} - \model} < \frac 14 \diameter(\model)^{-1} \gaingap(\model)$, all gain optimal policies of $\hat{\model}$ are gain optimal in $\model$ by \Cref{lemma_alternative_separated_weak}.
        In \STEP{3} \Cref{equation_proof_information_value_continuous_model_5}, we provide a bound of $\gainof{\policy}(\hat{\model}^{\ddagger \eta})$ for all $\policy \in \optpolicies(\model)$ and as a matter of fact, we have $\gainof{\policy}(\hat{\model}^\dagger) = \gainof{\policy}(\hat{\model})$ for all $\policy \in \optpolicies(\model)$. 
        Indeed, $\hat{\kernel} \sim \kernel$ so the recurrent classes of policies are the same on $\hat{\model}$ and $\model$; And $\hat{\kernel}^\dagger \gg \hat{\kernel}$ so gain optimal policies of $\model$ have their recurrent classes contained in $\optpairs(\model)$ in $\hat{\model}^{\ddagger \eta}$, on which $\hat{\model}^{\ddagger \eta} = \hat{\model}$. 
        Together with \eqref{equation_proof_information_value_continuous_model_5}, we find:
        \begin{equation*}
            \optgain(\hat{\model}) 
            \le
            \optgain(\widetilde{\model}^\dagger)
            + (1 + 2\diameter(\model)) \snorm{\hat{\model} - \widetilde{\model}}
            .
        \end{equation*}
        So, by \STEP{3} \Cref{equation_proof_information_value_continuous_model_4}, $\widetilde{\policy}^\dagger$ satisfies $\gainof{\widetilde{\policy}^\dagger}(\hat{\model}^{\ddagger \eta}) > \optgain(\hat{\model})$ when
        \begin{equation*}
            \eta 
            > \frac{
                2 (1 + 2 \diameter(\model)) 
                \snorm{\hat{\model} - \widetilde{\model}}
            }{
                \gainof{\widetilde{\policy}^\dagger}(\unit_{\pairs^\dagger}, \hat{\kernel}^\dagger)
            }
            \overset{(\dagger)}\ge
            \frac{
                2^{\abs{\states}} (1 + 2 \diameter(\model)) 
                \snorm{\hat{\model} - \widetilde{\model}}
            }{
                \dmin(\kernel)^{\abs{\states}-1} 
                \abs{\states}
            }
        \end{equation*}
        in particular, where $(\dagger)$ follows from \STEP{4}.
        This concludes the proof.
    \end{subproof}

    We may now conclude.
    We proceed by upper bounding the information value of $\hat{\model}$ with respect to the information value of $\widetilde{\model}$. 
    First, we pick some $\eta > 0$ such that (I) $\hat{\model}^{\ddagger \eta}$ is well-defined and (II)~$\hat{\model}^{\ddagger \eta} \in \confusing_\epsilon (\model)$.
    \STEP{2} and \STEP{5} provide sufficient conditions so that (I)~and (II) are satisfied respectively. 
    For the conditions given by \STEP{2} and \STEP{5} to be compatible, we need 
    \begin{equation*}
        \frac{
            2^{\abs{\states}} (1 + 2 \diameter(\model)) 
            \snorm{\hat{\model} - \widetilde{\model}}
        }{
            \dmin(\kernel)^{\abs{\states}-1} 
            \abs{\states}
        }
        <
        \exp \parens*{
            - \frac {
                \frac{\ivalue_\epsilon(\unit, \widetilde{\model})}{\delta}
                + \log(2)
            }{
                1 - \max({\reward}) - \norm{\widetilde{\reward} - \reward}_\infty
            }
        }
        - \norm{\hat{\reward} - \widetilde{\reward}}_\infty
        .
    \end{equation*}
    Provided that $\norm{\widetilde{\reward} - \reward}_\infty \le \frac 12 (1 - \max(\reward))$ and using the upper bound $\norm{\hat{\reward} - \widetilde{\reward}}_\infty \le \snorm{\hat{\model} - \widetilde{\model}}$, we solve the above in $\snorm{\hat{\model} - \widetilde{\model}}$ and find the sufficient condition:
    \begin{equation*}
        \snorm{\hat{\model} - \widetilde{\model}}
        <
        \exp \parens*{
            - \frac {
                2 \parens*{
                    \frac{\ivalue_\epsilon(\unit, \widetilde{\model})}{\delta}
                    + \log(2)
                }
            }{
                1 - \max({\reward})
            }
            +
            \log \parens*{
                \frac{
                    \dmin(\kernel)^{\abs{\states}-1} 
                    \abs{\states}
                }{
                    \dmin(\kernel)^{\abs{\states}-1} 
                    \abs{\states}
                    +
                    2^{\abs{\states}} (1 + 2 \diameter(\model)) 
                }
            }
        }
        .
    \end{equation*}
    Together with the condition $\hat{\model}, \widetilde{\model} \in \models_\epsilon$ as given by \eqref{equation_proof_information_value_continuous_model_0}, we retrieve indeed a condition in the form as given by \eqref{equation_information_value_continuous_model_condition} in \Cref{lemma_information_value_continuous_model}.
    Conclude with \STEP{1} by taking $\eta = \dmin(\kernel)^{1 - \abs{\states}} \abs{\states}^{-1} 2^{\abs{\states}} (1 + 2 \diameter(\model)) \snorm{\hat{\model} - \widetilde{\model}}$ as allowed by \STEP{5}, to obtain an bound of the form
    \begin{equation}
        \ivalue_\epsilon(\imeasure, \hat{\model}) 
        \le
        \ivalue_\epsilon(\imeasure, \widetilde{\model})
        + 
        \exp(\beta(\model)) \snorm{\hat{\model} - \widetilde{\model}}
    \end{equation}
    where $\beta(\model)$ is polynomial in $\abs{\pairs}, \diameter(\model), \dmin(\model), \gaingap(\model)$ and $\ivalue(\unit, \model)$. 
    
    Take $\model = \hat{\model}, \model' = \widetilde{\model}$ to upper bound the information value of $\model$ relatively to $\model'$, and $\model = \widetilde{\model}, \model' = \hat{\model}$ to lower bound the information value of $\model$ relatively to $\model'$.  
\end{proof}
\def\proofname{Proof}

\subsection{Properties of the regularized lower bound}
\label{appendix_regretlb_continuity}

In this section, we study the properties of the regularized lower bound $\regretlb_\EPSILON (\model)$ (\Cref{definition_regularized_regret_lower_bound}) which is given as the solution of the optimization problem below:
\begin{equation*}
    \min 
    \sum_{\pair \in \pairs} 
    \imeasure(\pair) 
    \ogaps(\pair; \model)
    + \epsilonreg \norm{\imeasure}_2^2
    \quad \text{s.t.} \quad
    \begin{cases}
        \imeasure \in \imeasures({\epsilonunif}; \model)
        \\
        \displaystyle
        \inf_{
            \model^\dagger \in \confusing(\epsilonflat; \model)
        }
        \sum_{\pair \in \pairs} 
        \imeasure(\pair)
        \KL(\model(\pair)||\model^\dagger(\pair))
        \ge
        1
    \end{cases}
\end{equation*}
where $\EPSILON \equiv (\epsilonflat, \epsilonunif, \epsilonreg) \in \R_+^3$ is the regularization hyperparameter.
The goal of this section is to show that the above has a solution $\imeasure^*_\EPSILON (\model)$ called an $\EPSILON$-optimal exploration measures, and study the continuity and convergence properties of $\regretlb_\EPSILON (\model)$ and $\imeasure^*_\EPSILON (\model)$ as $\EPSILON$ and $\model$ vary. 

\paragraph{Outline.}
We begin by providing a simple bound on $\regretlb(\model)$ and $\regretlb_\EPSILON (\model)$ in \Cref{appendix_regretlb_bound}.
In \Cref{appendix_regretlb_existence}, we show that $\EPSILON$-optimal exploration measures exist, even when $\EPSILON = 0$, and that $\imeasure^*_\EPSILON$ is unique when the convexification $\epsilonreg$ is strict.
The remaining of the appendix is dedicated to the continuity properties of $\regretlb_\EPSILON (\model)$ and $\imeasure^*_\EPSILON (\model)$. 
In \Cref{appendix_regularized_convenient_form}, we present an alternative way of writing $\regretlb_\EPSILON(\model)$ that is adequate for that purpose. 
In \Cref{appendix_regularized_convergence}, we show that the sequence of $\EPSILON$-optimal exploration measures $(\imeasure^*_\EPSILON)$ converges to a special optimal exploration $\imeasure^*$ when $\EPSILON \to 0$ at the adequate rate (\Cref{proposition_convergence_optimal_measure}).
In \Cref{appendix_regularized_approximates}, we show that as $\EPSILON \to 0$, we have $\regretlb_\EPSILON(\model) \downto \regretlb(\model)$ (\Cref{proposition_regularized_lowerbound_approximates}), meaning that the regularized lower bound can approximates the true lower bound with arbitrary precision as the regularization vanishes. 
Then, in \Cref{appendix_regularized_continuous}, we show that if $\snorm{\model' - \model}$ is small relatively to $\EPSILON$, then $\regretlb_\EPSILON(\model')$ is close to $\regretlb_\EPSILON (\model)$.
Qualitatively, it means that 
\begin{equation}
\label{equation_regularized_qualitative_convergence}
    \regretlb(\model) 
    =
    \lim_{\EPSILON \to 0} 
    \lim_{\snorm{\model' - \model} \to 0}
    \regretlb_\EPSILON (\model')
    .
\end{equation}
The precise rates of convergence in \eqref{equation_regularized_qualitative_convergence} are given in \Cref{proposition_continuity_regularized_lower_bound}.

\subsubsection{A simple bound on the regret lower bound}
\label{appendix_regretlb_bound}

In this paragraph, we provide an upper-bound of $\regretlb(\model)$ of order $\abs{\pairs} \diameter(\model)^3 \gaingap(\model)^{-2}$, see \Cref{theorem_regretlb_bound} where $\diameter(\model)$ is the diameter and $\gaingap(\model)$ is the gain-gap (\Cref{definition_gain_gap}).
Our bound is of the same kind as \cite[Theorem~4]{auer_near_optimal_2009}, that indirectly provide a bound of order $\abs{\pairs} \abs{\states} \diameter(\model)^2 \gaingap(\model)^{-1}$ by showing that \texttt{UCRL2} has regret growing as $\abs{\pairs} \abs{\states} \diameter(\model)^2 \gaingap(\model)^{-1} \log(T)$ in the asymptotic.
Their bound is better than our own in most cases---as soon as $\abs{\states} \le \diameter^2(\model) \gaingap(\model)^{-1}$ in fact, but requires the construction and the analysis of a complete learning algorithm. 

Overall, we believe that bounds on $\regretlb(\model)$ in terms of $\diameter(\model)$ and $\gaingap(\model)$ can be improved up to $\abs{\pairs} \diameter(\model)^2 \gaingap(\model)^{-1}$.
This is left for future work. 

\begin{theorem}[Simple upper bound of $\regretlb(\model)$]
\label{theorem_regretlb_bound}
    Let $\models$ be an ambient space with Bernoulli rewards (\Cref{assumption_bernoulli}).
    For every communicating model $\model \equiv (\pairs, \kernel, \rewardd) \in \models$, the regret lower bound is upper-bounded as
    \begin{equation*}
        \regretlb(\model) 
        \le 
        \frac{16 \abs{\pairs} \diameter(\model)^3}{\gaingap(\model)^2}
        .
    \end{equation*}
    where $\diameter(\model)$ is the diameter of $\model$ (\Cref{assumption_communicating}) and $\gaingap(\model)$ its gain-gap (\Cref{definition_gain_gap}).
    In particular, $\regretlb(\model) < \infty$. 
\end{theorem}

\begin{proof}
    Let $\policy \in \randomizedpolicies$ the fast covering policy provided by \Cref{lemma_policy_exploration_diameter}, of which the unique probability invariant measure satisfies $\min(\imeasureof{\policy}) \ge \abs{\pairs}^{-1} \diameter(\model)^{-1}$.
    By combining \Cref{lemma_information_value_positive} and \Cref{lemma_information_value_bounded}, the information value of $\imeasure$ is lower bounded as
    \begin{equation}
    \label{equation_regretlb_bound_1}
        \ivalue(\imeasureof{\policy}, \model)
        \ge \frac{
            \ivalue(\unit, \model)
        }{
            \abs{\pairs}
            \diameter(\model)
        }
        \ge \frac{
            \gaingap(\model)^2
        }{
            16 \abs{\pairs}
            \diameter(\model)^3
        }
        .
    \end{equation}
    Using the formulation of the regret lower bound in terms of policies (\Cref{proposition_regretlb_policies}, see \eqref{equation_regretlb_2}), we deduce that
    \begin{equation*}
        \regretlb(\model)
        \le
        \frac{\optgain(\model) - \gainof{\policy}(\model)}{\ivalue(\imeasureof{\policy}, \model)}
        \overset{(\dagger)}\le
        \frac{16 \abs{\pairs} \diameter(\model)^3}{\gaingap(\model)^2}
    \end{equation*}
    where 
    $(\dagger)$ invokes \eqref{equation_regretlb_bound_1} and uses that rewards are $[0, 1]$, so that $\optgain(\model) - \gainof{\policy}(\model) \le 1$. 
\end{proof}

\subsubsection{Existence of optimal and \texorpdfstring{$\EPSILON$}{eps}-optimal exploration measures}
\label{appendix_regretlb_existence}

In this paragraph, we show that $\regretlb(\model)$ and $\regretlb_\EPSILON(\model)$ are achieved by invariant measures. 
More specifically, we show that there exists $\imeasure^* \in \imeasures(\model)$ with $\sum_{\pair \in \pairs} \imeasure(\pair) \KL(\model(\pair)||\model^\dagger (\pair)) \ge 1$ for all $\model^\dagger \in \confusing(\model)$ and $\sum_{\pair \in \pairs} \imeasure^*(\pair) \ogaps(\pair) = \regretlb(\model)$---with a similar result for the regularized lower bound. 
This shows that the concepts of optimal exploration measure (\Cref{section_lower_bound}) and $\EPSILON$-optimal exploration measure (\Cref{section_definition_regularized}) are well-defined.

\begin{theorem}[Existence of optimal exploration measures]
\label{theorem_optimal_measures_existence}
    Let $\models$ be an arbitrary ambient space of Markov decision processes.
    Let $\model \in \models$ be a communicating model. 
    Then the infemum of \eqref{equation_definition_regretlb} is a minimum, i.e., $\regretlb(\model)$ is equal to
    \begin{equation*}
        \min \braces*{
            \sum_{\pair \in \pairs}
            \imeasure(\pair)
            \ogaps(\pair)
            :
            \imeasure \in \imeasures(\model)
            \mathrm{~and~}
            \inf_{\model^\dagger \in \confusing(\model)}
            \braces*{
                \sum_{\pair \in \pairs}
                \imeasure(\pair)
                \KL(\model(\pair)||\model^\dagger(\pair))
            }
            \ge
            1
        }
        .
    \end{equation*}
\end{theorem}

\begin{proof}
    The idea of the proof is to write $\regretlb(\model)$ as the solution of an optimization problem over a compact region. 
    This task is not that easy because invariant measures contain irrelevant information; For instance, it can be shown that if $\imeasure^* \in \imeasures(\model)$ is an optimizer of \eqref{equation_definition_regretlb} achieving $\regretlb(\model)$, then for every gain optimal unichain policy $\policy^*$ and every of its invariant measure $\imeasure' \in \imeasures(\policy^*; \model)$, the invariant measure $\imeasure^* + \imeasure'$ is also an optimizer of \eqref{equation_definition_regretlb}.
    This makes the set of optimizers of \eqref{equation_definition_regretlb} unbounded.

    To remove the irrelevant information contained by invariant measures, we write $\regretlb(\model)$ in an alternative form.

    Let $\policies_u$ be the set of policies that are unichain in $\model$ and denote $\policies_u^- := \policies_u \setminus \optpolicies(\model)$. 
    We denote $\imeasure_\policy$ the unique probability invariant measure of $\policy \in \policies_u$ in $\model$. 
    By \Cref{proposition_decomposition_invariant_measures}, every invariant measure $\imeasure \in \imeasures(\model)$ can be written as $\imeasure = \sum_{\policy \in \policies_u} \lambda_\policy \imeasure_\policy$ for some $(\lambda_\policy) \in \RR_+^{\policies_u}$. 
    Now, note the information value of $\imeasure = \sum_{\policy \in \policies_u} \lambda_\policy \imeasure_\policy$ satisfies
    \begin{align*}
        \ivalue(\imeasure, \model)
        & := 
        \inf_{\model^\dagger \in \confusing(\model)}
        \braces*{
            \sum_{\policy \in \policies_u}
            \sum_{\pair \in \pairs}
            \lambda_\policy
            \imeasure_\policy (\pair)
            \KL(\model(\pair)||\model^\dagger (\pair))
        }
        \\
        & \overset{(\dagger)}=
        \inf_{\model^\dagger \in \confusing(\model)}
        \braces*{
            \sum_{\policy \in \policies_u^-}
            \sum_{\pair \in \pairs}
            \lambda_\policy
            \imeasure_\policy (\pair)
            \KL(\model(\pair)||\model^\dagger (\pair))
        }
    \end{align*}
    where $(\dagger)$ holds following from the observation that $\support(\imeasure_\policy) \subseteq \optpairs(\model)$ for all $\policy \in \policies_u \cap \optpolicies(\model)$ and that every confusing model coincide with $\model$ on $\optpairs(\model)$ by definition. 
    Accordingly, in the optimization problem of \eqref{equation_definition_regretlb}, the optimization region $\imeasures(\model)$ can be changed as
    \begin{align*}
        & \regretlb(\model)
        \\
        & =
        \inf_{\lambda \in \RR_+^{\policies_u^-}}
        \braces*{
            \sum_{\policy \in \policies_u^-}
            \sum_{\pair \in \pairs}
            \lambda_\policy
            \imeasure_\policy (\pair)
            \ogaps(\pair)
            :
            \inf_{\model^\dagger \in \confusing(\model)}
            \braces*{
                \sum_{\policy \in \policies_u^-}
                \sum_{\pair \in \pairs}
                \lambda_\policy
                \imeasure_\policy (\pair)
                \KL(\model(\pair)||\model^\dagger (\pair))
            }
            \ge
            1
        }
        \\
        & \overset{(\dagger)}=
        \inf_{\lambda \in \RR_+^{\policies_u^-}}
        \braces*{
            \sum_{\policy \in \policies_u^-}
            \lambda_\policy
            \parens*{\optgain - \gainof{\policy}}
            :
            \inf_{\model^\dagger \in \confusing(\model)}
            \braces*{
                \sum_{\policy \in \policies_u^-}
                \sum_{\pair \in \pairs}
                \lambda_\policy
                \imeasure_\policy (\pair)
                \KL(\model(\pair)||\model^\dagger (\pair))
            }
            \ge
            1
        }
    \end{align*}
    where $(\dagger)$ follows from the observvation that $\sum_{\pair \in \pairs} \imeasure_\policy (\pair) \ogaps(\pair) = \optgain - \gainof{\policy}$ for every unichain policy $\policy \in \policies_u$, which can be shown using that $\EE_{\state}^{\policy}[\sum_{t=1}^{T} \indicator{\Pair_t = \pair}] = T \imeasure_\policy (\pair) + \OH(1)$ for all $\pair \in \pairs$ and initial state $\state \in \states$. 

    We now restrict $(\RR_+)^{\policies_u^-}$ to a compact sub-region as follows.
    By \Cref{theorem_regretlb_bound}, we know that $\regretlb(\model) \le 16 \abs{\pairs} \diameter(\model)^3 \gaingap(\model)^{-2}$. 
    As every policy $\policy \in \policies_u^-$ is sub-optimal, we have $\optgain - \gainof{\policy} \ge \gaingap(\model)$. 
    So, if $(\lambda_\policy) \in (\RR_+)^{\policies_u^-}$ is such that $\norm{\lambda}_\infty > 16 \abs{\pairs} \diameter(\model)^3 \gaingap(\model)^{-3}$, we know that the invariant measure $\sum_{\policy \in \policies_u^-} \lambda_\policy \imeasure_\policy$ does not achieve $\regretlb(\model)$.
    In other words, setting 
    \begin{equation*}
        \mathcal{F} 
        :=
        \braces*{
            \lambda \in (\RR_+)^{\policies_u^-}
            :
            {
                \norm{\lambda}_\infty 
                \le 
                16 \abs{\pairs} \diameter(\model)^3 \gaingap(\model)^{-3}
                \mathrm{~and~}
                \atop
                \forall \model^\dagger \in \confusing(\model),
                \sum_{\policy \in \policies_u^-}
                \sum_{\pair \in \pairs}
                \lambda_\policy
                \imeasure_\policy (\pair)
                \KL(\model(\pair)||\model^\dagger (\pair))
                \ge
                1
            }
        },
    \end{equation*}
    we have
    \begin{equation}
    \label{equation_regretlb_existence_1}
        \regretlb(\model)
        =
        \inf_{\lambda \in \mathcal{F}}
        \braces*{
            \sum_{\policy \in \policies_u^-}
            \lambda_\policy
            \parens*{\optgain - \gainof{\policy}}
        }
        .
    \end{equation}
    Because $\mathcal{F}$ is compact and that the objective function in \eqref{equation_regretlb_existence_1} is continuous, the infemum of \eqref{equation_regretlb_existence_1} is a minimum.
    Let $\lambda^* \in \mathcal{F}$ achieving that minimum.
    The invariant measure $\imeasure^* := \sum_{\policy \in \policies_u^-} \lambda^*_\policy \imeasure_\policy$ is an optimal invariant measure of $\model$, concluding the proof. 
\end{proof}

With the same proof, the result is generalized to $\regretlb_\EPSILON (\model)$. 

\begin{theorem}[Existence of $\EPSILON$-optimal exploration measures]
\label{theorem_near_optimal_measure_existence}
    Let $\models$ be an arbitrary ambient space of Markov decision processes.
    Let $\model \in \models$ be a communicating model and $\EPSILON \equiv (\epsilonflat, \epsilonunif, \epsilonreg) \in \RR_+^3$ be a regularization hyperparameter.
    Then the infemum of the optimization problem associated to $\regretlb_\EPSILON(\model)$ (see \Cref{definition_regularized_regret_lower_bound}) is a minimum, i.e., $\regretlb_\EPSILON(\model)$ is equal to
    \begin{equation*}
        \min
        \sum_{\pair \in \pairs}
        \imeasure(\pair)
        \ogaps(\pair)
        + \epsilonreg \norm{\imeasure}_2^2
        \quad \mathrm{s.t.} \quad
        \begin{cases}
            & \displaystyle
            \imeasure \in \imeasures(\epsilonunif; \model)
            , \quad \mathrm{and}
            \\
            & \displaystyle
            \inf_{\model^\dagger \in \confusing(\epsilonflat; \model)}
            \braces*{
                \sum_{\pair \in \pairs}
                \imeasure(\pair)
                \KL(\model(\pair)||\model^\dagger(\pair))
            }
            \ge
            1.
        \end{cases}
        \hspace{-0.33em}
    \end{equation*}
    Moreover, if the convexification is strict with $\epsilonreg > 0$, then the minimum $\imeasure^*_\EPSILON$ is unique. 
\end{theorem}

\begin{proof}
    The existence of the minimum is shown similarly to \Cref{theorem_optimal_measures_existence}. 
    The uniqueness of $\imeasure^*_\EPSILON$ is guaranteed by strong convexity of the objective function. 
\end{proof}

\subsubsection{A convenient form for the \texorpdfstring{$\EPSILON$}{eps}-regularized lower bound}
\label{appendix_regularized_convenient_form}

\begin{figure}[ht]
    \centering
    \begin{tikzpicture}
        \node at (1.5, 0) {model $\model_\theta$};
        \node[state] (1) at (0, 0) {$1$};
        \node[state] (2) at (3, 0) {$2$};
        \draw[transition] (1) to[bend left] node[midway, above] {$0$} (2);
        \draw[transition] (2) to[bend left] node[midway, below] {$0$} (1);
        \draw[transition] (1) to[loop] node[midway, above] {$0.5+\theta$} (1);
        \draw[transition] (2) to[loop] node[midway, above] {$0.5$} (2);
    \end{tikzpicture}%
    \hspace{-2.5em}%
    \begin{tikzpicture}
        \node at (1.5, 0) {$\ogaps$, $\theta > 0$};
        \node[state] (1) at (0, 0) {$1$};
        \node[state] (2) at (3, 0) {$2$};
        \draw[transition] (1) to[bend left] node[midway, above] {$1+2\theta$} (2);
        \draw[transition] (2) to[bend left] node[midway, below] {$\mathbf{0}$} (1);
        \draw[transition] (1) to[loop] node[midway, above] {$0$} (1);
        \draw[transition] (2) to[loop] node[midway, above] {$\theta$} (2);
    \end{tikzpicture}%
    \hspace{-2.5em}%
    \begin{tikzpicture}
        \node at (1.5, 0) {$\ogaps$, $\theta = 0$};
        \node[state] (1) at (0, 0) {$1$};
        \node[state] (2) at (3, 0) {$2$};
        \draw[transition] (1) to[bend left] node[midway, above] {$1$} (2);
        \draw[transition] (2) to[bend left] node[midway, below] {$\mathbf{1}$} (1);
        \draw[transition] (1) to[loop] node[midway, above] {$0$} (1);
        \draw[transition] (2) to[loop] node[midway, above] {$0$} (2);
    \end{tikzpicture}
    \caption{
    \label{figure_example_discontinuous_gaps}
        A discontinuity of $\ogaps(\model)$.
        A class of Bernoulli reward models with deterministic transitions parameterized by $\theta \in \Theta \equiv [-\frac 12, \frac 12]$. 
        Arrows are choices of actions that deterministically lead to the pointed state and labels are mean rewards. 
    }
\end{figure}
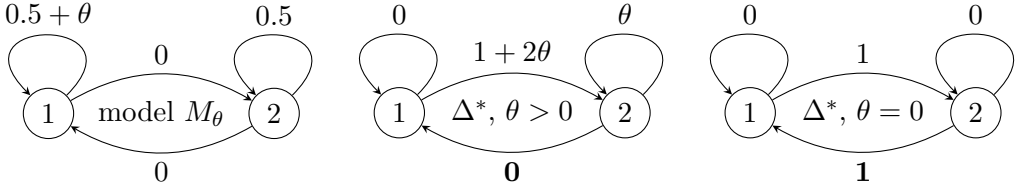

To prepare the analysis of the continuity properties of $\regretlb_\EPSILON (\model)$ and $\imeasure^*_\EPSILON (\model)$, we start by rewriting the $\EPSILON$-regularized lower bound with respect to information values as follows:
\begin{equation}
\label{equation_regularized_lower_bound_rewrote}
    \regretlb_\EPSILON(\model)
    =
    \inf_{\imeasure \in \pimeasures(\epsilonunif; \model)}
    \braces*{
        \frac{
            \sum_{\pair \in \pairs}
            \imeasure(\pair)
            \ogaps(\pair; \model)
        }{
            \ivalue_{\iepsilonflat}(\imeasure, \model)
        }
        +
        \frac{
            \epsilonreg
            \norm{\imeasure}_2^2
        }{
            \ivalue_{\iepsilonflat}(\imeasure, \model)^2
        }
    }
\end{equation}
where $\pimeasures(\epsilonunif; \model) := \imeasures(\epsilonunif; \model) \cap \probabilities(\pairs)$ is the space of $\epsilonunif$-uniform probability invariant measures of $\model$. 
The issue in \eqref{equation_regularized_lower_bound_rewrote} is that Bellman gaps $\ogaps(\pair; \model)$ are discontinuous in $\model$ even regarding the support-aware norm $\snorm{-}$. 
A simple example of discontinuity point of $\ogaps(-)$ is provided in \Cref{figure_example_discontinuous_gaps}.

This discontinuity is actually artificial.
If $\ogaps(-)$ is discontinuous indeed, the inner product between an invariant measure and the Bellman gaps $\sum_{\pair \in \pairs} \imeasure(\pair) \ogaps(\pair; \model)$ is not, because it can be rewritten as $\sum_{\pair \in \pairs} \imeasure(\pair) (\optgain(\model) - \reward(\pair))$ using \Cref{lemma_invariant_measures_and_gaps} below, and $\optgain - \reward$ can be shown to be a continuous function of $\model$. 

\begin{lemma}
\label{lemma_invariant_measures_and_gaps}
    For every communicating Markov decision process $\model \equiv (\pairs, \kernel, \reward)$ and all invariant measure $\imeasure \in \imeasures(\model)$, we have
    \begin{equation*}
        \sum_{\pair \in \pairs}
        \imeasure(\pair)
        \ogaps(\pair; \model)
        =
        \sum_{\pair \in \pairs}
        \imeasure(\pair)
        \parens*{\optgain(\model) - \reward(\pair)}
        .
    \end{equation*}
\end{lemma}
\begin{proof}
    Let $\imeasure \in \imeasures(\model)$.
    This follows from the direct computation:
    \begin{align*}
        & \sum_{\pair \in \pairs} 
        \imeasure(\pair) 
        \parens*{
            \optgain(\model) - \reward(\pair)
        }
        \\
        & \overset{(\dagger)}=
        \sum_{\pair \equiv (\state, \action) \in \pairs} 
        \imeasure(\pair) 
        \parens[\big]{
            \ogaps(\pair; \model)
            + \parens*{
                \unit_\state - \kernel(\pair)
            } \optbias(\model)
        }
        \\
        & = 
        \sum_{\pair \in \pairs}
        \imeasure(\pair) 
        \ogaps(\pair; \model)
        + 
        \sum_{\state \in \states}
        \optbias(\state; \model)
        \sum_{\action \in \actions(\state)}
        \imeasure(\state, \action)
        -
        \sum_{\state' \in \states}
        \optbias(\state'; \model)
        \sum_{\pair \in \pairs}
        \imeasure(\pair) \kernel(\state'|\pair)
        \\
        & \overset{(\ddagger)}=
        \sum_{\pair \in \pairs}
        \imeasure(\pair) 
        \ogaps(\pair; \model)
        + 
        \sum_{\state \in \states}
        \optbias(\state; \model)
        \sum_{\action \in \actions(\state)}
        \imeasure(\state, \action)
        -
        \sum_{\state' \in \states}
        \optbias(\state'; \model)
        \sum_{\action' \in \actions(\state')}
        \imeasure(\state', \action')
        \\
        & =
        \sum_{\pair \in \pairs}
        \imeasure(\pair) \ogaps(\pair; \model)
    \end{align*}
    where 
    $(\dagger)$ follows by definition of $\ogaps(\pair; \model)$ and
    $(\ddagger)$ holds because $\imeasure \in \imeasures(\model)$.
\end{proof}

\subsubsection{Convergence of regularized optimal exploration measures}
\label{appendix_regularized_convergence}

In this section, we provide convergence guarantees for the unique solution $\imeasure^*_\EPSILON$ of the regularized lower bound $\regretlb_{\EPSILON}(\model)$.
Informally, we show that the optimal $\EPSILON$-regularized exploration measure $\imeasure^*_\EPSILON$ converges to the optimal exploration measure $\imeasure^*$ that further minimized the $\ell_2$-norm, provided that $\epsilonunif/\epsilonreg + \epsilonreg \to 0$, see \Cref{proposition_convergence_optimal_measure} below.
Under the same conditions, this shows that the optimal $\EPSILON$-regularized exploration policy converges to a unique randomized policy when $\imeasure^*$ puts positive mass on every state $\sum_{\action \in \actions(\state)} \imeasure^*(\state, \action) > 0$ for all $\state \in \states$---or, equivalently, if there exists an optimal exploration policy that is state-recurrent.

\begin{proposition}
\label{proposition_convergence_optimal_measure}
    Let $\model \in \models$ be a communicating Markov decision process.
    Given a regularization hyperparameter $\EPSILON \equiv (\epsilonflat, \epsilonunif, \epsilonreg)$, we write $\imeasure^*_{\EPSILON}(\model)$ the optimal $\EPSILON$-regularized exploration measure. 
    Let $\imeasure^* \in \imeasures(\model)$ be the (unique) optimizer of $\regretlb(\model)$ as in \eqref{equation_definition_regretlb} that further minimizes $\norm{\imeasure}_2^2$ among optimizers. 
    There exists a function $\psi : \RR_+ \to \RR_+$ such that
    \begin{equation}
    \label{equation_convergence_optimal_measure}
        \forall \EPSILON, 
        \quad
        \norm{\imeasure^*_\EPSILON - \imeasure^*}_2^2
        \le
        \psi\parens*{
            \frac{\epsilonunif}{\epsilonreg} + \epsilonreg
            + \epsilonflat
        }
    \end{equation}
    with $\psi(x) = \OH(x)$ as $x \to 0$.
\end{proposition}

Therefore, \Cref{proposition_convergence_optimal_measure} shows that one optimal exploration measure of $\model$ is a bit more special than the others regarding our regularization scheme, that we call the \strong{central} optimal exploration measure, see \Cref{definition_central_exploration_measure}.
We use the terminology ``central'' for that measure maximizes the support over optimal exploration measures. 

\begin{definition}[Central measure]
\label{definition_central_exploration_measure}
    The \strong{central} optimal exploration probability measure of $\model$ is the unique optimal exploration probability measure $\imeasure^* \in \probabilities(\pairs)$ that achieves $\regretlb(\model)$ and that further minimizes $\norm{\imeasure}_2^2$ among optimizers of $\regretlb(\model)$. 
\end{definition}

\begin{proof}
    For $\EPSILON \ge 0$, we write $f_\EPSILON (\imeasure) := \sum_{\pair \in \pairs} \imeasure(\pair) \ogaps(\pair; \model) + \epsilonreg \norm{\imeasure}_2^2$ the objective function and $\mathcal{F}_\EPSILON$ the feasible set associated to $\regretlb_\EPSILON (\model)$ seen as an optimization problem, i.e., 
    \begin{equation*}
        \mathcal{F}_\EPSILON := \braces*{
            \imeasure \in \imeasures(\epsilonunif; \model)
            :
            \forall \model^\dagger \in \confusing(\epsilonflat; \model),
            \sum_{\pair \in \pairs}
            \imeasure(\pair) 
            \KL(\model(\pair)||\model^\dagger(\pair))
            \ge
            1
        }.
    \end{equation*}
    Assume throughout that $\epsilonflat$ is small enough so that $\confusing(\epsilonflat; \model) = \confusing(\model)$, say $\epsilonflat < \gaingap(\model)$ for e.g., see \Cref{proposition_continuity_near_optimality}.

    Let $\tilde{\imeasure}_\EPSILON^*$ be the optimizer of $f_\EPSILON$ in $\mathcal{F}_0$ and let $\tilde{\imeasure}'_\EPSILON$ be its projection on $\mathcal{F}_\EPSILON$. 
    By \Cref{theorem_invariant_measures_approximate_with_covering} and provided that $\epsilonunif < \abs{\pairs}^{-1} \diameter(\model)^{-1}$, we have
    \begin{equation}
    \label{equation_convergence_optimal_measure_1}
        \norm*{
            \tilde{\imeasure}'_\EPSILON - \tilde{\imeasure}_\EPSILON^* 
        }_\infty
        \le
        \abs{\pairs} \diameter(\model)
        \epsilonunif \norm{\tilde{\imeasure}^*_\EPSILON}_1
        .
    \end{equation}
    For conciseness, denote $\alpha := \abs{\pairs} \diameter(\model)$.
    Now, we have
    \begin{align*}
        & f_\EPSILON (\imeasure_\EPSILON^*)
        \\
        & \overset{(\dagger)}\ge
        \sum_{\pair \in \pairs}
        \tilde{\imeasure}_\EPSILON^* (\pair) 
        \ogaps(\pair) 
        + \epsilonreg \norm{\tilde{\imeasure}^*_\EPSILON}_2^2
        \\
        & \overset{(\ddagger)}\ge
        \sum_{\pair \in \pairs}
        \tilde{\imeasure}'_\EPSILON (\pair) 
        \ogaps(\pair) 
        + \epsilonreg \norm{\tilde{\imeasure}'_\EPSILON}_2^2
        - \alpha {\epsilonunif \norm{\tilde{\imeasure}'_\EPSILON}_1} \parens[\Big]{
            \norm{\ogaps}_1 
            + \epsilonreg \parens*{
                2 \norm{\tilde{\imeasure}^*_\EPSILON}_1
                + \alpha {\epsilonunif \abs{\pairs} \norm{\tilde{\imeasure}^*_\EPSILON}_1}
            }
        }
        \\
        & \overset{(\S)}\ge
        f_\EPSILON(\imeasure_\EPSILON^*)
        + \epsilonreg \norm{\tilde{\imeasure}_\EPSILON' - \imeasure_\EPSILON^*}_2^2
        - \alpha {\epsilonunif \norm{\tilde{\imeasure}'_\EPSILON}_1} \parens[\Big]{
            \norm{\ogaps}_1 
            + \epsilonreg \parens*{
                2 \norm{\tilde{\imeasure}^*_\EPSILON}_1
                + \alpha {\epsilonunif \abs{\pairs} \norm{\tilde{\imeasure}^*_\EPSILON}_1}
            }
        }
    \end{align*}
    where 
    $(\dagger)$ follows by definition of $\tilde{\imeasure}^*_\EPSILON$ and $\imeasure_\EPSILON^* \in \mathcal{F}_{\EPSILON} \subseteq \mathcal{F}_0$;
    $(\ddagger)$ follows from \eqref{equation_convergence_optimal_measure_1} and a bit of algebra; and
    $(\S)$ holds because $f_\EPSILON (\imeasure) \ge f_\EPSILON(\imeasure_\EPSILON^*) + \epsilonreg \norm{\imeasure - \imeasure_\EPSILON^*}_2^2$ for all $\imeasure \in \mathcal{F}_\EPSILON$.
    Together with \eqref{equation_convergence_optimal_measure_1}, we conclude that 
    \begin{equation}
    \label{equation_convergence_optimal_measure_2}
        \norm{\tilde{\imeasure}_\EPSILON^* - \imeasure_\EPSILON^*}
        \le
        \frac{\epsilonunif \norm{\tilde{\imeasure}^*_\EPSILON}_1}{1/\alpha}
        + \frac{\epsilonunif \norm{\tilde{\imeasure}'_\EPSILON}_1}{\epsilonreg / \alpha} \parens*{
            \norm{\ogaps}_1 
            + \epsilonreg \parens*{
                2 \norm{\tilde{\imeasure}^*_\EPSILON}_1
                + \frac{\epsilonunif \abs{\pairs} \norm{\tilde{\imeasure}^*_\EPSILON}_1}{1/\alpha}
            }
        }
    \end{equation}
    Note that $\norm{\tilde{\imeasure}'_\EPSILON}_1 = \norm{\tilde{\imeasure}^*_\EPSILON}_1 + \OH(\epsilonunif)$ by \eqref{equation_convergence_optimal_measure_1}.

    We finish the proof by showing that $\imeasure^* = \tilde{\imeasure}_\EPSILON^*$ when $\epsilonreg$ is small enough. 
    
    In the remaining of the proof, we more simply write $\epsilon$ in place of $\epsilonreg$.
    Consider the relaxation of the optimization problem $\min_{\imeasure \in \mathcal{F}_0} \norm{\imeasure}_2^2$ s.t.~$f_0(\imeasure) \le \regretlb(\model)$ given by 
    \begin{equation*}
        \min_{\imeasure \in \mathcal{F}_0}
        \norm{\imeasure}_2^2 + \lambda (f_0(\imeasure) - \regretlb(\model))
    \end{equation*}
    for $\lambda \ge 0$.
    Note that Slater's condition is satisfied.
    Indeed, the constraints ``$\imeasure \in \imeasures(\model)$'' and ``$f_0(\imeasure) = \regretlb(\model)$'' are polytopes, while the constraint ``$\sum_{\pair \in \pairs} \imeasure(\pair) \KL(\model(\pair)||\model^\dagger(\pair)) \ge 1$ for all $\model^\dagger \in \confusing(\model)$'' is equivalently written in the form of ``$\psi(\imeasure) \le 0$'' where $\psi(\imeasure) := 1 - \sup_{\model^\dagger \in \confusing(\model)} \sum_{\pair \in \pairs} \imeasure(\pair) \KL(\model(\pair)||\model^\dagger(\pair))$ that is convex. 
    By taking $\imeasure_0 \in \probabilities(\pairs)$ the unique probability invariant measure of the uniform policy $\policy_0 (\action|\state) := \abs{\actions(\state)}^{-1}$ and considering $\lambda_0 \imeasure_0 \in \imeasures(\model)$ for $\lambda_0 > 0$ large enough such that $\sum_{\pair \in \pairs} \lambda_0 \imeasure_0 (\pair) \KL(\model(\pair)||\model^\dagger(\pair)) \ge 2$ for all $\model^\dagger \in \confusing(\model)$, we obtain a point in the relative interior of the feasible region $\mathcal{F}_0$.
    So strong duality holds, see \cite[§5.2.3]{boyd_convex_2004}.
    We deduce that 
    \begin{equation}
    \label{equation_convergence_optimal_measure_3}
        \max_{\lambda \ge 0}
        \min_{\imeasure \in \mathcal{F}_0}
        \parens*{
            \norm{\imeasure}_2^2 + \lambda (f_0(\imeasure) - \regretlb(\model))
        }
        =
        \min_{\imeasure \in \mathcal{F}_0}
        \max_{\lambda \ge 0}
        \parens*{
            \norm{\imeasure}_2^2 + \lambda (f_0(\imeasure) - \regretlb(\model))
        }
        .
    \end{equation}
    By definition, $\imeasure^*$ is the unique solution of $\min_{\imeasure \in \mathcal{F}_0} \norm{\imeasure}_2^2$ s.t.~$f_0(\imeasure) \le \regretlb(\model)$.
    Let $\lambda^* \equiv \frac 1{\epsilon^*}$ be the associated Lagrange multiplier. 
    By definition again, $\tilde{\imeasure}^*_\epsilon$ is the unique solution of $\min_{\imeasure \in \mathcal{F}_0} f_0(\imeasure) + \epsilon \norm{\imeasure}_2^2$, so $\tilde{\imeasure}^*_\epsilon$ is the unique solution of $\min_{\imeasure \in \mathcal{F}_0} \norm{\imeasure}_2^2 + \frac 1\epsilon (f_0(\imeasure) - \regretlb(\model))$. 
    Let $\epsilon \in (0, \epsilon^*]$. 
    Note that for $\imeasure \in \mathcal{F}_0$, we have $\norm{\imeasure}_2^2 + \frac 1{\epsilon}(f_0(\imeasure) - \regretlb(\model)) \ge \norm{\imeasure}_2^2 + \frac 1{\epsilon^*}(f_0(\imeasure) - \regretlb(\model))$, so that taking the minimum for $\imeasure \in \mathcal{F}_0$, we obtain
    \begin{equation*}
        \norm{\tilde{\imeasure}^*_\epsilon}_2^2 
        + \frac 1{\epsilon} \parens*{
            f_0(\tilde{\imeasure}^*_\epsilon) - \regretlb(\model)
        }
        \ge
        \norm{\tilde{\imeasure}^*_{\epsilon^*}}_2^2 
        + \frac 1{\epsilon^*} \parens*{
            f_0(\tilde{\imeasure}^*_{\epsilon^*}) - \regretlb(\model)
        }
        =
        \norm{\tilde{\imeasure}^*_{\epsilon^*}}_2^2 
        =
        \norm{\imeasure^*}_2^2
        .
    \end{equation*}
    It follows that $(\tilde{\imeasure}^*_\epsilon, \frac 1\epsilon)$ is a solution of \eqref{equation_convergence_optimal_measure_3}, so $\tilde{\imeasure}^*_\epsilon = \imeasure^*$ by uniqueness of $\imeasure^*$.

    Accordingly, we have $\tilde{\imeasure}^*_\EPSILON = \imeasure^*$ for $\epsilonreg \le \epsilon^*$.
    It further follows that $\norm{\tilde{\imeasure}'_\EPSILON}_1 = \norm{\imeasure^*}_1 + \OH(\epsilonunif) + \OH(\epsilonreg)$. 
    Combined with \eqref{equation_convergence_optimal_measure_2}, we conclude.
\end{proof}

\subsubsection{The regularized lower bound approximates the lower bound}
\label{appendix_regularized_approximates}

In this paragraph, we prove that the regularized lower bound $\regretlb_\EPSILON(\model)$ converges to the true lower bound $\regretlb(\model)$ as $\EPSILON \to 0$ and provide the rate of convergence at first order, see \Cref{proposition_regularized_lowerbound_approximates} below.
This result is a straight forward consequence of the approximation properties of the leveling operation $\flatmodel{-}{\epsilon}$ described by \Cref{proposition_continuity_near_optimality} and of the possibility of approximating invariant measures by uniform invariant measures, see \Cref{theorem_invariant_measures_approximate_with_covering}.

\begin{proposition}
\label{proposition_regularized_lowerbound_approximates}
    Let $\model \in \models$ be a communicating Markov decision process.
    Then, as the regularization hyperparameter $\EPSILON \equiv (\epsilonflat, \epsilonunif, \epsilonreg)$ vanishes, we have
    \begin{equation*}
        \regretlb(\model) 
        \le \regretlb_\EPSILON(\model) 
        \le \regretlb(\model) + \OH(\norm{\EPSILON}_\infty)
        .
    \end{equation*}
    More precisely, if $\epsilonflat < \gaingap(\model)$ and $\epsilonunif < \frac 12 \dmin(\imeasure^*) \abs{\pairs} \diameter(\model) \norm{\imeasure^*}_1^{-1}$ where $\imeasure^*$ is an arbitrary optimizer of $\regretlb(\model)$, then 
    \begin{align*}
        \regretlb(\model)
        & \le
        \regretlb_\EPSILON(\model)
        \\
        & \le
        \regretlb(\model)
        + \OH \parens*{
            \norm{\imeasure^*}_1 \abs{\pairs} \diameter(\model)
            \parens*{
                \frac{\regretlb(\model)}{\dmin(\imeasure^*)}
                + 
                \norm{\ogaps(\model)}_\infty
            } \epsilonunif
            +
            \norm{\imeasure^*}_2^2 \epsilonreg
            + \epsilonflat
        }
        .
    \end{align*}
\end{proposition}

\begin{proof}
    We begin with the upper bound.
    Following \Cref{proposition_continuity_near_optimality}, if $\epsilonflat < \gaingap(\model)$, then $\flatme{\optpairs}{\iepsilonflat}(\model) = \optpairs(\model)$ and in particular $\confusing(\epsilonflat; \model) = \confusing(\model)$.
    Moreover, we have $\imeasures(\epsilonunif; \model) \subseteq \imeasures(\model)$.
    So, if $\imeasure_\EPSILON^*$ is the optimum of $\regretlb_\EPSILON(\model)$, then $\imeasure_\EPSILON^* \in \imeasures(\model)$ and $\sum_{\pair \in \pairs} \imeasure_\EPSILON^* (\pair) \KL(\model(\pair)||\model^\dagger(\pair)) \ge 1$ for all $\model^\dagger \in \confusing(\model)$.
    Accordingly, $\imeasure_\EPSILON^*$ is within the feasible region of the optimization problem associated to $\regretlb(\model)$, so
    \begin{align*}
        \regretlb(\model)
        & \le 
        \sum_{\pair \in \pairs} 
        \imeasure_\EPSILON^*(\pair)
        \ogaps(\pair; \model)
        \le 
        \sum_{\pair \in \pairs} 
        \imeasure_\EPSILON^*(\pair)
        \ogaps(\pair; \model)
        + \epsilonreg \norm{\imeasure_\EPSILON^*}_2^2
        = \regretlb_\EPSILON^*(\model)
        .
    \end{align*}

    We continue with the lower bound. 
    Let $\imeasure^*$ be an optimum of $\regretlb(\model)$. 
    As $\imeasure^* \norm{\imeasure^*}_1^{-1} \in \imeasures(\model) \cap \probabilities(\pairs)$, by \Cref{theorem_invariant_measures_approximate_with_covering}, there exists $\imeasure_\EPSILON \in \imeasures(\epsilonunif; \model)$ such that
    \begin{equation}
        \label{equation_proof_proposition_regularized_lowerbound_approximates_1}
        \norm{
            \imeasure^* - \imeasure_\EPSILON
        }_\infty
        \le
        \abs{\pairs} \diameter(\model) 
        {\epsilonunif \norm{\imeasure^*}_1}
        .
    \end{equation}
    By \Cref{proposition_continuity_near_optimality} again, if $\epsilonflat < \gaingap(\model)$, then $\flatme{\optpairs}{\iepsilonflat}(\model) = \optpairs(\model)$, so $\confusing(\epsilonflat; \model) = \confusing(\model)$.  
    Introduce the quantity $\alpha := \dmin(\imeasure^*)^{-1} \abs{\pairs} \diameter(\model) \norm{\imeasure^*}_1 > 0$ and note that $\imeasure_\EPSILON \ge (1 - \alpha \epsilonunif) \imeasure^*$.
    So, if $\alpha \epsilonunif < 1$, we have $(1 - \alpha \epsilonunif)^{-1} \imeasure_\EPSILON \ge \imeasure^*$.
    Following $\confusing(\epsilonflat; \model) = \confusing(\model)$, we find that $(1 - \alpha \epsilonunif)^{-1} \imeasure_\EPSILON$ is within the feasible region of the optimization problem associated to $\regretlb_\EPSILON(\model)$.
    So, writing $\beta := \min(\imeasure_0)^{-1} \norm{\imeasure^*}_1$, we have
    \begin{align*}
        \regretlb_\EPSILON(\model)
        & \le
        \frac 1{1 - \alpha \epsilonunif}
        \sum_{\pair \in \pairs} 
        \imeasure_\EPSILON(\pair)
        \ogaps(\pair; \model)
        + \frac{
            \epsilonreg
            \norm{\imeasure_\EPSILON}_2^2
        }{(1 - \alpha \epsilonunif)^2}
        \\
        & \overset{(\dagger)}\le
        \frac 1{1 - \alpha \epsilonunif}
        \sum_{\pair \in \pairs} 
        \parens*{
            \imeasure^*(\pair) + \beta \epsilonunif
        }
        \ogaps(\pair; \model)
        + \frac{
            2\epsilonreg
            \parens*{
                \norm{\imeasure^*}_2^2 
                + \beta^2 \epsilonunif^2 \abs{\pairs}^2
            }
        }{(1 - \alpha \epsilonunif)^2}
        \\
        & =
        \regretlb(\model)
        + \OH \parens*{\norm{\EPSILON}_\infty}
    \end{align*}
    where 
    $(\dagger)$ invokes \eqref{equation_proof_proposition_regularized_lowerbound_approximates_1} together with $\norm{u + v}_2^2 \le 2(\norm{u}_2^2 + \norm{v}_2^2)$.
\end{proof}
\def\proofname{Proof}

\subsubsection{Properties of the regularized lower bound}
\label{appendix_regularized_continuous}

The most difficult part are continuity properties.
In this paragraph, we show that for a fixed regularization hyperparameter $\EPSILON \in (\RR_+^*)^3$, we have $\regretlb_\EPSILON(\model') \to \regretlb(\model)$ and $\imeasure^*_\EPSILON(\model') \to \imeasure^*_\EPSILON(\model)$ as $\snorm{\model' - \model} \to 0$.
In other words, we show that both the regularized lower bound $\regretlb_\EPSILON(-)$ and its optimizer $\imeasure^*_\EPSILON(-)$ are continuous with respect to the support-aware norm $\snorm{-}$.  

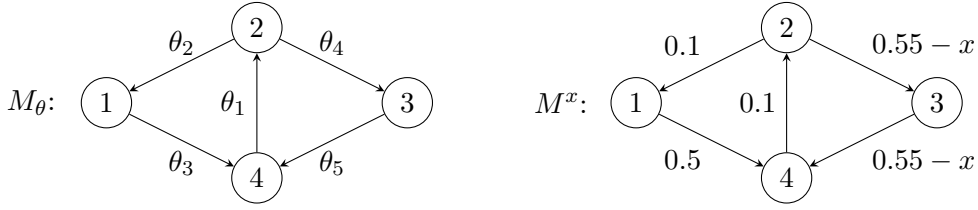
\begin{figure}[ht]
    \centering
    \begin{tikzpicture}
        \node at (-1, 0) {$\model_\theta$:};
        \node[state] (s1) at (0, 0) {$1$};
        \node[state] (s2) at (2, 1) {$2$};
        \node[state] (s3) at (4, 0) {$3$};
        \node[state] (s4) at (2,-1) {$4$};
        \draw[transition] (s4) to node[midway, left] {$\theta_1$} (s2);
        \draw[transition] (s2) to node[midway, above] {$\theta_2$} (s1);
        \draw[transition] (s1) to node[midway, below] {$\theta_3$} (s4);
        \draw[transition] (s2) to node[midway, above] {$\theta_4$} (s3);
        \draw[transition] (s3) to node[midway, below] {$\theta_5$} (s4);
    \end{tikzpicture}
    \hspace{2.5em}
    \begin{tikzpicture}
        \node at (-1, 0) {$\model^x$:};
        \node[state] (s1) at (0, 0) {$1$};
        \node[state] (s2) at (2, 1) {$2$};
        \node[state] (s3) at (4, 0) {$3$};
        \node[state] (s4) at (2,-1) {$4$};
        \draw[transition] (s4) to node[midway, left] {$0.1$} (s2);
        \draw[transition] (s2) to node[midway, anchor=south east] {$0.1$} (s1);
        \draw[transition] (s1) to node[midway, anchor=north east] {$0.5$} (s4);
        \draw[transition] (s2) to node[midway, anchor=south west] {$0.55-x$} (s3);
        \draw[transition] (s3) to node[midway, anchor=north west] {$0.55-x$} (s4);
    \end{tikzpicture}
    \caption{
    \label{figure_regretlb_discontinuous}
        A discontinuity of $\regretlb_\EPSILON (\model)$ when the product form (\Cref{assumption_space}) is dropped.
        A set of Bernoulli reward models with deterministic transitions parameterized by $\theta \in \Theta \equiv [0, 1]^5$ (to the left). 
        Arrows are choices of actions that deterministically lead to the pointed state and labels are mean rewards. 
    }
\end{figure}

Our proof requires a few technical assumptions on the structure of the ambient space $\models$. 
The first is that $\models$ is a space of Bernoulli reward Markov decision processes (\Cref{assumption_bernoulli}).
This assumption can probably be generalized to Gaussian rewards or, more generally, to rewards among a single parameter exponential family. 
The second is that $\models$ is in product form (\Cref{assumption_space}).
Without this assumption, the set of leveled confusing models $\confusing_\EPSILON (\model)$ can be discontinuous in $\model$ and information values (\Cref{definition_information_value}) can be discontinuous, provoking discontinuities of $\regretlb_\EPSILON (\model)$ that cannot be addressed easily. 

A simple example of discontinuity is provided in \Cref{figure_regretlb_discontinuous}. 
Consider the ambient spaces $\models := \braces{\model_\theta: \theta \in \Theta}$ (in product form) and $\models' := \braces{\model_\theta: \theta_1 = \theta_2 \mathrm{~and~} \theta_4 = \theta_5}$ (not in product form) and the model $\model^x \in \models \cap \models'$ as given to the right of \Cref{figure_regretlb_discontinuous}.
Fix $\EPSILON \ge 0$ a small enough regularizer (possibly $0$).
As soon as $\abs{x}$ is small enough, we always have $\confusing_\EPSILON (\model^x; \models) \ne \emptyset$ by considering the confusing model with $\theta_2 \approx \theta_3 \approx 1$, so that $\regretlb_\EPSILON (\model^x; \models) > 0$---In fact, $\regretlb_\EPSILON (\model^x; \models)$ is a continuous function of $x$ as shown by \Cref{proposition_continuity_regularized_lower_bound} below.
When the ambient space is $\models'$ however, this confusing model doesn't exist in $\models'$ and we find
\begin{equation*}
    \confusing_\EPSILON (\model^x; \models') 
    =
    \begin{cases}
        \emptyset & \text{if $x \ge 0$};
        \\
        \braces[\Big]{
            \model_\theta
            :
            \theta_1 = \theta_2 = 0.1,
            \theta_4 = \theta_5 = 0.55-x
            \mathrm{~and~}
            \theta_3 \ge 1 - 2x
        } & \text{if $x < 0$}.
    \end{cases}
\end{equation*}
So, we have $\inf_{x < 0} \regretlb_\EPSILON (\model^x; \models') > 0$ while $\regretlb_\EPSILON (\model^x; \models') = 0$ if $x \ge 0$, so $\regretlb_\EPSILON (\model^x; \models')$ is not continuous at $x = 0$. 
Accordingly, the product form assumption (\Cref{assumption_space}) on the ambient space cannot be dropped in general. 

\begin{proposition}
\label{proposition_continuity_regularized_lower_bound}
    Let $\models$ be an ambient space with Bernoulli rewards (\Cref{assumption_bernoulli}) in product form (\Cref{assumption_space}).
    Let $\model \in \models$ be a communicating model. 
    The regularized lower bound is continuous in the follow sense: There exist functions $\psi_\regretlb, \psi_\imeasure : \RR_+ \to [0, \infty]$ such that for all regularization hyperparameter $\EPSILON \equiv (\epsilonflat, \epsilonunif, \epsilonreg) \in (\RR_+^*)^3$ and all model $\model' \in \models$, the errors $\abs{\regretlb_{\EPSILON}(\model') - \regretlb_{\EPSILON}(\model)}$ and $\norm{\imeasure^*_{\EPSILON}(\model') - \imeasure^*_{\EPSILON}(\model)}_\infty$ are respectively bounded by
    \begin{equation*}
    \begin{gathered}
        \psi_\regretlb \parens*{
            \frac{\snorm{\model' - \model}}{(\epsilonunif)^{3 \abs{\states}}}
            + \eqindicator{
                \norm{\model' - \model}
                >
                \beta_1 
                \min \braces*{
                    \epsilonflat,
                    \exp \parens*{
                        - \frac{\beta_2}{(\epsilonunif)^{\abs{\states}}}
                    }
                }
            }
        }
        \\
        \mathrm{and}
        \\
        \psi_\imeasure\parens*{
            \sqrt{
                \frac{\snorm{\model' - \model}}{(\epsilonunif)^{3 \abs{\states}} \epsilonreg}
            }
            + \eqindicator{
                \norm{\model' - \model}
                >
                \beta_1 
                \min \braces*{
                    \epsilonflat,
                    \exp \parens*{
                        - \frac{\beta_2}{(\epsilonunif)^{\abs{\states}}}
                    }
                }
            }
        }
    \end{gathered}
    \end{equation*}
    with $\psi_\regretlb(x) = \OH(x)$ and $\psi_\imeasure(x) = \OH(x)$ when $x \to 0$, and $\beta_1, \beta_2 \in \RR_+$ depend on $\model$.
\end{proposition}

\bigskip
\begin{proof}
    Let $\model_1, \model_2$ be in the neighborhood of $\model$, denoting $\model_i = (\pairs, \kernel_i, \reward_i)$ for $i = 1, 2$, with $\model \in \braces{\model_1, \model_2}$.
    The result will follow by upper-bounding $\regretlb_\EPSILON(\model_2)$ relatively to $\regretlb_\EPSILON(\model_1)$.
    Using \Cref{lemma_invariant_measures_and_gaps} and \Cref{equation_regularized_lower_bound_rewrote}, for $i = 1, 2$, the $\EPSILON$-regularized regret lower bound of $\model_i$ is rewritten with respect to information values as follows:
    \begin{equation*}
        \regretlb_\EPSILON(\model_i)
        =
        \inf_{\imeasure \in \pimeasures(\epsilonunif; \model_i)}
        \braces*{
            \frac{
                \sum_{\pair \in \pairs}
                \imeasure(\pair)
                (\optgain(\model) - \reward_i(\pair))
            }{
                \ivalue_{\iepsilonflat}(\imeasure, \model)
            }
            +
            \frac{
                \epsilonreg
                \norm{\imeasure}_2^2
            }{
                \ivalue_{\iepsilonflat}(\imeasure, \model)^2
            }
        }
        .
    \end{equation*}
    The dot product $\sum_{\pair \in \pairs} \imeasure(\pair) (\optgain(\model) - \reward(\pair))$ will be more concisely written $\imeasure (\optgain - \reward)$.
    Let $\imeasure_1^* \in \pimeasures(\epsilonunif; \model)$ the optimum of $\regretlb_\EPSILON(\model_1)$ and let $\imeasure_2$ be its projection onto $\pimeasures(\epsilonunif; \model_2)$ as given by \Cref{theorem_invariant_measures_approximate_with_covering}.
    We introduce the following short-hand notations:
    \begin{equation}
    \label{equation_proof_proposition_continuity_regularized_lower_bound_0}
    \begin{aligned}
        \dd \imeasure & := \imeasure_2 - \imeasure^*_1
        \\
        \dd \gaps & := \optgain(\model_2) + \reward(\model_1) - \optgain(\model_1) - \reward(\model_2)
        \\
        \dd \ivalue & := 
            \ivalue_{\iepsilonflat}(\imeasure_2, \model_2)
            - \ivalue_{\iepsilonflat}(\imeasure_1, \model_1)
        \\
        \dd \model & := \snorm{\model_2 - \model_1}
    \end{aligned}
    \end{equation}
    Using these notations, we have
    \begin{align}
    \notag
        \regretlb_\EPSILON(\model_2)
        & \le 
        \frac{
            \imeasure_2
            (\optgain_2 - \reward_2)
        }{
            \ivalue_{\epsilonflat}(\imeasure_2, \model_2)
        }
        +
        \frac{
            \epsilonreg
            \norm{\imeasure_2}_2^2
        }{
            \ivalue_{\epsilonflat}(\imeasure_2, \model_2)^2
        }
        \\
    \label{equation_proof_proposition_continuity_regularized_lower_bound_1}
        & =
        \frac{
            (\imeasure_1 + \dd \imeasure)
            (\optgain_1 - \reward_1 + \dd \gaps)
        }{
            \ivalue_{\epsilonflat}(\imeasure_1, \model_1)
            + \dd \ivalue
        }
        +
        \frac{
            \epsilonreg
            \norm{\imeasure_1 + \dd \imeasure}_2^2
        }{
            (\ivalue_{\epsilonflat}(\imeasure_1, \model_1) + \dd \ivalue)^2
        }
        .
    \end{align}
    Roughly speaking, when $\dd \model \ll \dmin(\EPSILON)$, then all the variational quantities $\dd \imeasure$, $\dd \gaps$ and $\dd \ivalue$ are of order $\dd \model$.
    By injecting that in \eqref{equation_proof_proposition_continuity_regularized_lower_bound_1}, we find $\regretlb_\EPSILON(\model_2) \le \regretlb_\EPSILON (\model_1) + \OH(\dd \model)$ and conclude accordingly. 
    The point is that to make this argument precise. 

    \par
    \bigskip
    \noindent
    \STEP{1}
    \textsc{
        Bounding the variational quantities
    }

    \par
    \medskip
    \noindent
    All variational quantities of \eqref{equation_proof_proposition_continuity_regularized_lower_bound_0} are bounded relatively to $\dd \model$.
    We find:
    \begin{equation}
    \label{equation_proof_proposition_continuity_regularized_lower_bound_2}
    \begin{aligned}
        \norm{\dd \imeasure}_\infty
        & \le \frac{
            \abs{\states}
        }{
            (\epsilonunif)^{\abs{\states}-1}
            \dmin(\kernel)^{\abs{\states}-1}
        }
        \dd \model
        \\
        \norm{\dd \gaps}_\infty
        & \le 
        (1 + 2\diameter(\model)) \dd \model
        \\
        \abs{\dd \ivalue}
        & \le 
        \parens*{
            \frac{
                2 \parens{
                    \ivalue_\epsilonflat(\unit, \model)
                    + {e^{\beta(\model)}} \dd \model
                } \abs{\states}
            }{
                (\epsilonunif)^{\abs{\states}}
                \dmin(\kernel)^{\abs{\states}-1}
            } 
            + e^{\beta(\model)}
        }
        \dd \model
        .
    \end{aligned}
    \end{equation}
    The various upper-bounds in \eqref{equation_proof_proposition_continuity_regularized_lower_bound_2} require conditions on $\dd \model$ and $\EPSILON$.
    The bound on $\norm{\dd \imeasure}_\infty$ is obtained directly using \Cref{corollary_uniform_measures_hausdorff_distance} and holds uniformly for $\epsilonunif$ and $\dd \model$.
    The bound on $\norm{\dd \gaps}_\infty$ is obtained with a combination of \Cref{lemma_optimal_gain_variations,lemma_bias_diameter}; Indeed, by \Cref{lemma_optimal_gain_variations},
    \begin{align*}
        \norm{\dd \gaps}_\infty
        & \le 
        2 \norm{\reward' - \reward}_\infty
        + \frac 12 \min \braces*{
            \vecspan{\optbias(\model_1)},
            \vecspan{\optbias(\model_2)}
        } \norm{\kernel' - \kernel}
        \\
        & \overset{(\dagger)}\le 
        2 \norm{\reward' - \reward}_\infty
        + \frac 12 \vecspan{
            \optbias(\model)
        } \norm{\kernel' - \kernel}
        \\
        & \overset{(\ddagger)}\le 
        2 \norm{\reward' - \reward}_\infty
        + \frac 12 \diameter(\model) \norm{\kernel' - \kernel}
        \le 
        \parens*{2 + \diameter(\model)} \dd \model
    \end{align*}
    where 
    $(\dagger)$ follows from $\model \in \braces{\model_1, \model_2}$ and 
    $(\ddagger)$ follows from \Cref{lemma_bias_diameter}.
    The resulting bound holds uniformly for $\dd \model$ and $\EPSILON$.

    Lastly, the bound on $\dd \ivalue$ in \eqref{equation_proof_proposition_continuity_regularized_lower_bound_2} is obtained with a combination of \Cref{lemma_information_value_bounded,lemma_information_value_continuous_model,lemma_information_value_continuous_measure} as follows:
    \begin{align*}
        \abs{\dd \ivalue}
        & \overset{(\dagger)}=
        \abs*{
            \ivalue_{\iepsilonflat} (\imeasure', \model') 
            - \ivalue_{\iepsilonflat} (\imeasure, \model)
        }
        \\
        & \le 
        \abs*{
            \ivalue_{\iepsilonflat} (\imeasure', \model')
            - \ivalue_{\iepsilonflat} (\imeasure', \model)
        }
        + \abs*{
            \ivalue_{\iepsilonflat} (\imeasure', \model)
            - \ivalue_{\iepsilonflat} (\imeasure, \model)
        }
        \\
        & \overset{(\ddagger)}\le
        e^{\beta(\model)} \dd \model
        + \frac {
            2 \ivalue_{\iepsilonflat}(\imeasure, \model) \dd \imeasure
        }{\epsilonunif}
        \\
        & \overset{(\S)}\le
        e^{\beta(\model)} \dd \model
        + \frac {
            2 \ivalue_{\iepsilonflat}(\unit, \model) \dd \imeasure
        }{\epsilonunif}
        \\
        & \overset{(\$)}\le
        \parens*{
            e^{\beta(\model)} 
            + \frac {
                2 \ivalue_{\iepsilonflat}(\unit, \model) 
                \abs{\states}
            }{
                (\epsilonunif)^{\abs{\states}}
                \dmin(\kernel)^{\abs{\states}-1}
            }
        } \dd \model
    \end{align*}
    where 
    $(\dagger)$ rewrite $\ivalue_{\iepsilonflat}(\imeasure_2, \model_2) - \ivalue_{\iepsilonflat}(\imeasure_1, \model_1)$ by using that $\model \in \braces{\model_1, \model_2}$ and introducing $\model' \in \braces{\model_1, \model_2} \setminus \braces{\model}$;
    $(\ddagger)$ bound the first term using \Cref{lemma_information_value_continuous_model} and the second using \Cref{lemma_information_value_continuous_measure};
    $(\S)$ bounds $\ivalue_{\iepsilonflat}(\imeasure, \model) \le \ivalue_{\iepsilonflat}(\unit, \model)$ using \Cref{lemma_information_value_bounded}; and
    $(\$)$ unfolds the upper bound of $\dd \imeasure$.
    Note that for \Cref{lemma_information_value_continuous_measure,lemma_information_value_continuous_model} to be applicable in $(\ddagger)$, we need $\dd \imeasure \le \frac 12 \epsilonunif$, $\confusing(\epsilonflat; \model) = \confusing(\model)$ and $\dd \model \le \min\braces{\epsilonflat \alpha(\model)^{-1}, \exp( - (\epsilonunif)^{-1} \beta_1(\model) - \beta_2(\model)}$ where $\alpha, \beta_1, \beta_2$ are polynomial functions $\abs{\pairs}$, $\diameter$, $\dmin (\model)$ and $\ivalue(\unit, \model)$ as given in \Cref{lemma_information_value_continuous_model}.
    We obtain the conditions
    \begin{equation}
    \label{equation_proof_proposition_continuity_regularized_lower_bound_3}
    \begin{gathered}
        \epsilonflat + \alpha (\model) \dd \model < \gaingap(\model)
        \\
        \dd \model \le \min \braces*{
            \frac {\epsilonflat}{\alpha (\model)},
            \exp \parens*{
                - \frac{\beta_1(\model)}{(\epsilonunif)^{\abs{\states}}}
                - \beta_2(\model)
            },
            \frac{
                (\epsilonunif)^{\abs{\states}} 
                \dmin(\kernel)^{\abs{\states}-1}
            }{
                2 \abs{\states}
            }
        }
    \end{gathered}
    \end{equation}
    Under \eqref{equation_proof_proposition_continuity_regularized_lower_bound_3}, all the bounds of \eqref{equation_proof_proposition_continuity_regularized_lower_bound_2} hold and in particular, the variational quantities $\dd \imeasure$, $\dd \gaps$ and $\dd \ivalue$ are of order $\OH(\dd \model)$; Hence are vanishing with $\dd \model$.

    \par
    \bigskip
    \noindent
    \STEP{2}
    \textsc{
        Bounding the variations of $\regretlb_\EPSILON$
    }

    \par
    \medskip
    \noindent
    When $\dd \model$ is vanishing, only keeping the terms of first order, we find 
    \begin{align*}
        \regretlb_\EPSILON(\model_2)
        & \le 
        \frac 1{1 + \frac{\dd \ivalue}{\ivalue_{\iepsilonflat}(\imeasure_1, \model_1)}}
        \frac{
            (\imeasure_1 + \dd \imeasure)
            (\optgain_1 - \reward_1 + \dd \gaps)
        }{
            \ivalue_{\iepsilonflat}(\imeasure_1, \model_1)
        }
        +
        \parens*{
            \frac 1{1 + \frac{\dd \ivalue}{{\ivalue_{\iepsilonflat}} (\imeasure_1, \model_1)}}
        }^2
        \frac{
            \epsilonreg
            \norm{\imeasure_1 + \dd \imeasure}_2^2
        }{
            \ivalue_{\iepsilonflat}(\imeasure_1, \model_1)^2 
        }
        \\
        & \overset{(\dagger)}\lesssim
        \regretlb_\EPSILON(\model_1)
        + \frac{
            (\optgain_1 - \reward_1) \dd \imeasure 
            + \imeasure_1 \dd \gaps
            + (1 + 2 \epsilonreg) \dd \ivalue
        }{
            \ivalue_{\iepsilonflat}(\imeasure_1, \model_1)
        }
        + \frac{
            2 \epsilonreg \imeasure_1 \dd \imeasure
        }{
            \ivalue_{\iepsilonflat}(\imeasure_1, \model_1)^2
        }
        \\
        & \overset{(\ddagger)}\le 
        \regretlb_\EPSILON(\model_1)
        + \frac{
            (\optgain_1 - \reward_1) \dd \imeasure 
            + \imeasure_1 \dd \gaps
            + (1 + 2 \epsilonreg) \dd \ivalue
        }{
            f(\epsilonunif) \ivalue_{\iepsilonflat}(e, \model_1)
        }
        + \frac{
            2 \epsilonreg \imeasure_1 \dd \imeasure
        }{
            f(\epsilonunif)^2 \ivalue_{\iepsilonflat}(e, \model_1)^2
        }
        \\
        & \overset{(\S)}\lesssim
        \regretlb_\EPSILON(\model_1)
        + \frac{
            \norm{\dd \imeasure}_1
            + \norm{\dd \gaps}_\infty
            + 3 \abs{\dd \ivalue}
        }{
            f(\epsilonunif) \ivalue_{\iepsilonflat}(e, \model)
        }
        + \frac{
            2 \epsilonreg \norm{\dd \imeasure}_\infty
        }{
            f(\epsilonunif)^2 \ivalue_{\iepsilonflat}(e, \model)^2
        }
        \\
        & \overset{(\$)}\le
        \regretlb_\EPSILON(\model_1)
        + \OH \parens*{
            \frac{1}{
                f(\epsilonunif)
            }
            \parens*{
                \frac{
                    \abs{\states}\parens*{
                        1 
                        + \frac{\epsilonreg}{\ivalue_{\iepsilonflat}(\unit, \model) f(\epsilonunif)}
                    }
                }{\ivalue_{\iepsilonflat}(\unit, \model) f(\epsilonunif)}
                + \frac 1{f(\epsilonunif)}
                + e^{\beta(\model)}
            }
            \dd \model
        }
        \\ 
        & \overset{(\#)}=
        \regretlb_\EPSILON(\model_1)
        + \OH \parens*{
            \frac{\abs{\states}}{
                f(\epsilonunif)
            }
            \parens*{
                \frac{
                    1 
                    + f(\epsilonunif) \ivalue_{\iepsilonflat}(\unit, \model)^2
                }{
                    \ivalue_{\iepsilonflat}(\unit, \model)^2
                    f(\epsilonunif)^2
                }
                + e^{\beta(\model)}
            }
            \dd \model
        }
    \end{align*}
    where 
    $(\dagger)$ is obtained by seeing every variational quantity $\dd \imeasure$, $\dd \gaps$ and $\dd \ivalue$ as an infinitesimal quantity and developping at first order---Note that to do so, we need $\ivalue_{\iepsilonflat}(\imeasure_i, \model)$ to be bounded below in a neighborhood of $\model$, which is indeed the case by combining \Cref{lemma_information_value_bounded,lemma_information_value_positive};
    $(\ddagger)$ follows by combining \Cref{lemma_uniform_measures_support_diameter} assertion (1) and \Cref{lemma_information_value_bounded} while introducing the shorthand $f(\epsilon) := \epsilon^{\abs{\states}-1} \dmin(\kernel)^{\abs{\states}-1}$;
    $(\S)$ follows by relating $\ivalue_{\iepsilonflat} (e, \model)$ and $\ivalue_{\iepsilonflat} (e, \model_1)$ using \Cref{lemma_information_value_continuous_model} and expanding at first order; and
    $(\$$) follows by expanding everything using \eqref{equation_proof_proposition_continuity_regularized_lower_bound_2} and $\diameter(\model) \le \abs{\states} f(\epsilonunif)^{-1}$; and
    $(\#)$ follows by rearranging terms. 

    In the end, by keeping only the dominant terms, we find a bound of the form
    \begin{equation*}
        \regretlb_{\EPSILON}(\model_2)
        \le
        \regretlb_{\EPSILON}(\model_1)
        + \OH \parens*{
            \frac{
                \snorm{\model_2 - \model_1}
            }{
                (\epsilonunif)^{3 \abs{\states}}
            }
        }
    \end{equation*}
    under a condition of the form
    \begin{equation}
        \label{equation_proof_proposition_continuity_regularized_lower_bound_3_alt}
        \norm{\model' - \model} 
        \le
        \beta_1 \min \braces*{
            \epsilonflat, 
            \exp \parens*{
                - \frac{\beta_2}{(\epsilonunif)^{\abs{\states}}}
            }
        }
    \end{equation}
    which is obtained by keeping the dominant terms in \eqref{equation_proof_proposition_continuity_regularized_lower_bound_3}.
    Hence the bound on the variations of $\regretlb_\EPSILON$ described in \Cref{proposition_continuity_regularized_lower_bound}.

    \par
    \bigskip
    \noindent
    \STEP{3}
    \textsc{
        Deducing the variations of $\imeasure^*_\EPSILON$
    }

    \par
    \medskip
    \noindent
    By symmetry of the previous argument, $\regretlb_{\EPSILON}(\model_2)$ is also upper-bounded by $\regretlb_{\EPSILON}(\model_1)$. 
    For short, introduce $\regretlb_\EPSILON (\imeasure; \model) := \sum_{\pair \in \pairs} \imeasure(\pair) \ogaps(\pair; \model) + \epsilonreg \norm{\imeasure}_2^2$ the objective function in the optimization problem associated with $\regretlb_\EPSILON(\model)$.
    Setting $\imeasure_2' = \ivalue_\EPSILON(\imeasure_2, \model_2) \imeasure_2$ the re-scaled version of $\imeasure_2$, we have shown that $\regretlb_\EPSILON (\model_2) \le \regretlb_\EPSILON (\imeasure_2'; \model_2) \le \regretlb_\EPSILON (\model_2) + \mathrm{B}$ where $\imeasure_2'$ is a feasible point of $\regretlb_\EPSILON (\model_2)$ and $\mathrm{B}$ is of order
    \begin{equation*}
        \mathrm{B}
        = \OH \parens*{
            \frac{\abs{\states}}{
                f(\epsilonunif)
            }
            \parens*{
                \frac{
                    1 
                    + f(\epsilonunif) \ivalue_{\iepsilonflat}(\unit, \model)^2
                }{
                    \ivalue_{\iepsilonflat}(\unit, \model)^2
                    f(\epsilonunif)^2
                }
                + e^{\beta(\model)}
            }
            \dd \model
        }
    \end{equation*}
    where $f(\epsilon) := \dmin(\kernel)^{\abs{\states}} \epsilon^{\abs{\states}}$.

    A direct computation shows $\regretlb_\EPSILON (\imeasure'; \model) = \regretlb_\EPSILON(\imeasure; \model) + \nabla_\imeasure \regretlb_\EPSILON (\imeasure; \model)(\imeasure' - \imeasure) + \epsilonreg \norm{\imeasure' - \imeasure}_2^2$ for all $\imeasure, \imeasure' \in \RR^\pairs$.
    As $\imeasure_2'$ is in the feasible region of $\regretlb_\EPSILON(\model_2)$ and $\imeasure_\EPSILON^*(\model_2)$ is the associated minimizer, we have $\nabla_\imeasure \regretlb_\EPSILON (\imeasure_\EPSILON^*(\model_2); \model_2) (\imeasure_2' - \imeasure_\EPSILON(\model_2)) \ge 0$, so
    \begin{equation}
    \label{equation_proof_proposition_continuity_regularized_lower_bound_4}
        \norm{\imeasure_2' - \imeasure_\EPSILON^* (\model_2)}_2^2
        \le 
        \frac{\regretlb_\EPSILON(\imeasure'_2; \model_2) - \regretlb_\EPSILON(\model_2)}{\epsilonreg}
        \le
        (\epsilonreg)^{-1} \mathrm{B}
        .
    \end{equation}
    Now, at first order, we have 
    \begin{align}
    \notag
        \norm{\imeasure'_2 - \imeasure_\EPSILON^* (\model_1)}_\infty
        & =
        \norm{
            \ivalue_\EPSILON (\imeasure_2, \model_2) (\imeasure_1^* + \dd \imeasure)
            - \ivalue_\EPSILON (\imeasure_1^*, \model_1) \imeasure_1^*
        }_\infty
        \\
    \label{equation_proof_proposition_continuity_regularized_lower_bound_5}
        & \lesssim
        \ivalue_\EPSILON (\unit, \model_1) \norm{\dd \imeasure}_\infty
        + \abs{\dd \ivalue}
        .
    \end{align}
    We conclude by combining \eqref{equation_proof_proposition_continuity_regularized_lower_bound_4} and \eqref{equation_proof_proposition_continuity_regularized_lower_bound_5} and by invoking the bounds of \eqref{equation_proof_proposition_continuity_regularized_lower_bound_2}.
    Then, this bound can be simplified as 
    \begin{equation*}
        \norm*{
            \imeasure_\EPSILON^* (\model_2)
            - \imeasure_\EPSILON^* (\model_1)
        }_\infty
        = \OH \parens*{
            \sqrt{
                \frac{
                    \snorm{\model_2 - \model_1}
                }{
                    (\epsilonunif)^{3 \abs{\states}} \epsilonreg
                }
            }
        }
    \end{equation*}
    under same condition \eqref{equation_proof_proposition_continuity_regularized_lower_bound_3} that we simplify as \eqref{equation_proof_proposition_continuity_regularized_lower_bound_3_alt} as well. 
    Hence the bound on the variations of $\imeasure^*_\EPSILON$ described in \Cref{proposition_continuity_regularized_lower_bound}.
\end{proof}

\subsubsection{Auxiliary results}

In this paragraph, we provide a few instrumental results, starting with the sensibility of the optimal gain function to perturbation of the reward vector and of the transition kernel. 

\begin{lemma}[Variations of the optimal gain]
\label{lemma_optimal_gain_variations}
    Let $\model \equiv (\pairs, \kernel, \reward)$ and $\model' \equiv (\pairs, \kernel', \reward')$ be two Markov decision processes (possibly of infinite diameter). 
    If $\vecspan{\optgain(\model)} = 0$, then 
    \begin{equation*}
        \norm{
            \optgain(\model')
            - \optgain(\model)
        }_\infty
        \le
        \norm{\reward' - \reward}_\infty
        + \frac 12 \vecspan{\optbias(\model)} \norm{\kernel' - \kernel}_\infty
        .
    \end{equation*}
\end{lemma}

\begin{proof}
    Let $\optpolicy$ be a bias optimal policy of $\model$.
    We have $\vecspan{\gainof{\optpolicy}(\model)} = 0$, so by \Cref{lemma_unichain_gain_variations}, 
    \begin{equation}
        \label{equation_proof_lemma_optimal_gain_variations_1}
        \norm{\gainof{\optpolicy}(\model') - \optgain(\model)}_\infty
        =
        \norm{\gainof{\optpolicy}(\model') - \gainof{\optpolicy}(\model)}_\infty
        \le
        \norm{\reward' - \reward}_\infty
        + \frac 12 \vecspan{\optbias(\model)} \norm{\kernel' - \kernel}
        .
    \end{equation}
    Now, consider $\policy \in \policies$ an arbitrary policy and fix $\state \in \states$. 
    \begin{align*}
        \gainof{\policy}(\state; \model')
        & =
        \lim_{T \to \infty}
        \frac 1T 
        \E_{\state}^{\policy, \model'} \brackets*{
            \sum_{t=1}^T 
            \reward'(\Pair_t)
        }
        \\
        & \le
        \lim_{T \to \infty}
        \frac 1T 
        \E_{\state}^{\policy, \model'} \brackets*{
            \sum_{t=1}^T 
            \reward(\Pair_t)   
        }
        + \norm{\reward' - \reward}_\infty
        \\
        & \overset{(\dagger)}=
        \lim_{T \to \infty}
        \frac 1T 
        \E_{\state}^{\policy, \model'} \brackets*{
            \sum_{t=1}^T 
            \parens[\big]{
                \optgain(\model)
                + \parens*{\unit_{\State_t} - \kernel(\Pair_t)} \optbias(\model)
                - \ogaps(\Pair_t; \model)
            }
        }
        + \norm{\reward' - \reward}_\infty
        \\
        & \overset{(\ddagger)}\le
        \lim_{T \to \infty}
        \frac 1T 
        \E_{\state}^{\policy, \model'} \brackets*{
            \sum_{t=1}^T 
            \parens[\big]{
                \unit_{\State_t} - \kernel(\Pair_t)
            }
            \optbias(\model)
        }
        + \optgain(\model)
        + \norm{\reward' - \reward}_\infty
        \\
        & \overset{(\S)}=
        \lim_{T \to \infty}
        \frac 1T 
        \E_{\state}^{\policy, \model'} \brackets*{
            \sum_{t=1}^T 
            \parens[\big]{
                \kernel'(\Pair_t) - \kernel(\Pair_t)
            }
            \optbias(\model)
        }
        + \optgain(\model)
        + \norm{\reward' - \reward}_\infty
        \\
        & \overset{(\$)}\le
        \optgain(\model)
        + \norm{\reward' - \reward}_\infty
        + \frac 12 \vecspan{\optbias(\model)} \norm{\kernel' - \kernel}_\infty
    \end{align*}
    where 
    $(\dagger)$ is obtained by definition of $\ogaps(\model)$ and Doob's optional stopping theorem;
    $(\ddagger)$ uses that since $\vecspan{\optgain(\model)} = 0$, we have $\ogaps(\model) \ge 0$;
    $(\S)$ uses that $\E_{\state}^{\policy, \model'}[\sum_{t=1}^T (\unit_{\State_t} - \kernel'(\Pair_t)) \optbias(\model)] = \E_{\state}^{\policy, \model'}[\optbias(\state; \model) - \optbias(\State_{T+1}; \model)] \le \vecspan{\optbias(\model)}$; and
    $(\$)$ follows from $\abs{(\kernel'(\pair) - \kernel(\pair)) \optbias(\model)} \le \frac 12 \vecspan{\optbias(\model)} \norm{\kernel'(\pair) - \kernel(\pair)}_1$ for $\pair \in \pairs$.

    We conclude by combining the above with \eqref{equation_proof_lemma_optimal_gain_variations_1}.
\end{proof}

In \Cref{lemma_bias_diameter} below, we recall a classical result linking the bias span to the diameter. 

\begin{lemma}[\cite{bartlett_regal_2009,fruit_exploration_exploitation_2019}]
\label{lemma_bias_diameter}
    Let $\model \equiv (\pairs, \kernel, \reward)$ be a communicating Markov decision process.
    Then $\vecspan{\optbias(\model)} \le \vecspan{\reward} \diameter(\model)$
\end{lemma}
\begin{proof}
    This result is well-known.
    We provide a proof for self-containedness.
    Let $\state \ne \state'$ and let $\policy \in \policies$ be a policy such that $\EE_{\state}^{\model, \policy}[\tau_{\state'}] < \infty$, where $\tau_{\state'} := \inf \braces{t \ge 1 : \State_t = \state'}$. 
    We have
    \begin{align*}
        \EE_{\state}^{\model, \policy} \brackets*{
            \sum_{t=1}^{\tau_{\state'}-1}
            (\optgain - \Reward_t) 
        }
        & \overset{(\dagger)}=
        \EE_{\state}^{\model, \policy} \brackets*{
            \sum_{t=1}^{\tau_{\state'}-1}
            \parens*{
                \kernel(\Pair_t) \optbias - \optbias(\State_t) 
                + \ogaps(\Pair_t)
            }
        }
        \\
        & \overset{(\ddagger)}=
        \EE_{\state}^{\model, \policy} \brackets*{
            \sum_{t=1}^{\tau_{\state'}-1}
            \parens*{
                \optbias(\State_{t+1}) - \optbias(\State_t)
            }
        }
        +
        \EE_{\state}^{\model, \policy} \brackets*{
            \sum_{t=1}^{\tau_{\state'}-1}
            \ogaps(\Pair_t)
        }
        \\
        & \overset{(\S)}\ge
        \optbias(\state') - \optbias(\state)
    \end{align*}
    where 
    $(\dagger)$ follows by definition of $\ogaps$;
    $(\ddagger)$ uses that $\EE[\unit_{\State_{t+1}}|\Pair_t] = \kernel(\Pair_t)$; and
    $(\S)$ follows from the Bellman equation $\ogaps(\Pair_t) \ge 0$. 
    On the other hand, we see that $\EE_{\state}^{\model, \policy} \brackets{ \sum_{t=1}^{\tau_{\state'}-1} (\optgain - \Reward_t) } \le \vecspan{\reward} \EE[\tau_{\state'}]$. 
    Taking the infimum for $\policy \in \policies$, we conclude. 
\end{proof}

In \Cref{lemma_variations_kullback_leibler} below, we bound the sensibility of $\KL(p||q)$ to perturbations of $p$ and $q$.

\begin{lemma}
\label{lemma_variations_kullback_leibler}
    Let $p \ll q \in \probabilities[m]$ and $p' \ll q' \in \probabilities[m]$ be two pairs of categorical distributions over $m$ elements.
    Assume that $\support(p) = \support(p')$ and that, for all $i \in \support(p)$, we have $(1 - \epsilon) p'_i \le p_i \le (1 + \epsilon) p'_i$ and $(1 - \epsilon) q'_i \le q_i \le (1 + \epsilon) q'_i$ for some $\epsilon \le \frac 12$.
    Then
    \begin{align*}
        (1 - \epsilon) \KL(p'||q') - 2 \epsilon \log(e m)
        \le
        \KL(p||q) 
        & \le 
        (1 + \epsilon) \KL(p'||q') + 2 \epsilon \log(e m)
        \\
        (1 - \epsilon) \KL(p||q) - 2 \epsilon \log(e m)
        \le
        \KL(p'||q')
        & \le
        (1 + 2 \epsilon) \KL(p||q) + 2 \epsilon \log(e m)
    \end{align*}
\end{lemma}
\begin{proof}
    Using that $\log(1 + x) \le x$ and $\log(1 - x) > - x$ for $x \in [0, \frac 12]$, we get on the one hand:
    \begin{align*}
        \sum_{i=1}^m p_i \log\parens*{\frac{p_i}{q_i}}
        & =
        - \sum_{i=1}^m p_i \log \parens*{\frac 1{p_i}}
        + \sum_{i=1}^m p_i \log \parens*{\frac 1{q_i}}
        \\
        & \le
        - (1 - \epsilon) \sum_{i=1}^m p'_i \log \parens*{\frac 1{(1+\epsilon) p'_i}}
        + (1 + \epsilon) \sum_{i=1}^m p'_i \log \parens*{\frac 1{(1-\epsilon) q'_i}}
        \\
        & = 
        (1 + \epsilon) \KL(p'||q') 
        + 2 \epsilon \entropy(p') 
        + (1-\epsilon') \log(1+\epsilon) - (1+\epsilon) \log(1-\epsilon)
        \\
        & \le 
        (1 + \epsilon) \KL(p'||q') + 2 \epsilon \log (e m)
        .
    \end{align*}
    On the other hand, 
    \begin{align*}
        \sum_{i=1}^m p_i \log \parens*{\frac{p_i}{q_i}}
        & = 
        - \sum_{i=1}^m p_i \log \parens*{\frac 1{p_i}}
        + \sum_{i=1}^m p_i \log \parens*{\frac 1{q_i}}
        \\
        & \ge 
        - (1 + \epsilon) \sum_{i=1}^m p'_i \log \parens*{\frac 1{(1-\epsilon) p'_i}}
        + (1 - \epsilon) \sum_{i=1}^m p'_i \log \parens*{\frac 1{(1+\epsilon) q'_i}}
        \\
        & \ge 
        (1 - \epsilon) \KL(p'||q') 
        - 2 \epsilon \entropy(p') 
        + (1 + \epsilon) \log(1-\epsilon) - (1 - \epsilon) \log(1+\epsilon)
        \\
        & \ge
        (1 - \epsilon) \KL(p'||q') - 2 \epsilon \log (e m)
        .
    \end{align*}
    This concludes the proof. 
    The second pair of inequalities are dervied from the first two. 
\end{proof}

    \clearpage
    \section{Regret analysis of \texttt{ECoE*}}
    \label{appendix_regret}

In this section, we provide a regret analysis of \texttt{ECoE*} (\Cref{algorithm_ecoe_annotated}) and we prove \Cref{theorem_upper_bound}, that shows that $\Reg(T; \model, \texttt{ECoE*}) \le \regretlb(\model) \log(T) + \oh(\log(T))$. 

As \texttt{ECoE*} is the only algorithm of interest in the sequel, we drop the dependence in $\learner \equiv \texttt{ECoE*}$ in notations. 
Similarly, the underlying Markov decision process will be always equal to $\model$ throughout, so that we will drop the dependent in $\model$ as well. 
We end up writing $\EE[-]$ and $\Pr(-)$ in place of $\EE^{\model, \texttt{ECoE*}}[-]$ and $\Pr^{\model, \texttt{ECoE*}}(-)$

The starting point of the regret analysis is to write the expected regret as 
\begin{equation*}
    \Reg(T) 
    = 
    \EE \brackets*{
        \sum_{\pair \in \pairs} \visits_T (\pair) \ogaps(\pair)
    }
    + \OH(1)
\end{equation*}
As $\ogaps(\pair) \ge 0$, the natural strategy is to upper-bound $\EE^{\model, \learner}[\visits_T (\pair)]$ for pairs with $\ogaps(\pair) > 0$. 
Such pairs corresponds in particulars to non-optimal pairs $\pair \notin \optpairs(\model)$, see \Cref{section_pairs}.
In \Cref{theorem_visit_suboptimal_precise} below, we upper-bound $\EE[\visits_T(\pair)]$ for a large panel of choices for the regularization hyperparameter of \texttt{ECoE*}.
The set of assumptions that the regularization must satisfy is technical and detailed in \Cref{appendix_assumption_regularizer}, see \Cref{assumption_regularization_hyperparameter}.

\begin{theorem}[Visit rates of \texttt{ECoE*}, general statement]
\label{theorem_visit_suboptimal_precise}
    Assume that the regularization hyperparameter $\EPSILON$ satisfies \Cref{assumption_regularization_hyperparameter}, and that the ambient space $\models$ has Bernoulli rewards (\Cref{assumption_bernoulli}) and strong product structure (\Cref{assumption_space_strong}).
    Fix $\model \in \models$ a communicating Markov decision process.
    Let $\imeasure^*$ the central optimal exploration probability measure of $\model$ as given by \Cref{definition_central_exploration_measure}. 
    For every non-optimal pair $\pair \notin \optimalpairs(\model)$, its expected number of visits under \texttt{ECoE*} satisfies
    \begin{equation*}
        \EE\brackets*{
            \visits_T (\pair)
        }
        \le
        \frac{\imeasure^*(\pair) \log(T)}{\ivalue(\imeasure^*, \model)}
        + \OH \parens*{
            \frac 1{\epsilonflat(T)^6}
            + \sum_{t=1}^{T-1} \exp\parens*{
                - \parens*{1 + \frac 12 \epsilontest(t)} \log(t)
            }
        }
        + \oh(\log(T))
    \end{equation*}
    when $T \to \infty$. 
\end{theorem}

\Cref{theorem_upper_bound} is an immediate corollary of \Cref{theorem_visit_suboptimal_precise} for the choice of regularizer described in \eqref{equation_choice_regularization}, because the second order term grows as $\epsilonflat(T)^{-6} + \integral^T (\frac 1t)^{1 + \epsilontest(t)} \dd t = \oh(\log(T))$. 

\paragraph{Outline.}
The proof is tedious. 
The idea is to decompose the number of visits of $\pair \notin \optpairs(\model)$ along with the decomposition of time instants that is inherent to \Cref{algorithm_ecoe_annotated}: There are exploration times $t \in \stimes^-$, co-exploration times $t \in \stimes^\pm$,  exploitation times $\stimes^+$ and panic times $\stimes^!$. 
As $\stimes^\pm \subseteq \stimes^-$, we decompose $\visits_T (\pair)$ as
\begin{equation}
\label{equation_decomposition_visits}
\begin{aligned}
    \visits_T (\pair) 
    & =
    \sum_{t=1}^{T-1}
    \indicator{\Pair_t = \pair, t \in \stimes^-}
    + \sum_{t=1}^{T-1}
    \indicator{\Pair_t = \pair, t \in \stimes^+}
    + \sum_{t=1}^{T-1}
    \indicator{\Pair_t = \pair, t \in \stimes^!}
    .
\end{aligned}
\end{equation}
That is, we distinguish three cases in \eqref{equation_decomposition_visits}.
Either $\Pair_t = \pair$ is played because
(I) the algorithm is explorating to collect information on presumably sub-optimal pairs; or
(II) because it is exploiting the (wrongly) presumed optimal policy;
(III) the algorithm just panicked by discovering a transition unseen so far.

We begin by (II) in \Cref{appendix_wrong_coexploration}, i.e., by bounding the number of times a sub-optimal pair may be played during exploitation and show that it is of order $\epsilonflat(T)^{-6} + \integral^T \exp(-(1 + \epsilontest(t)) \log(t)) \dd t$.
From there, simple algebra provides conditions on how small $\epsilonflat$ and $\epsilontest$ must be so that this term is sub-logarithmic, leading to \Cref{assumption_regularization_hyperparameter}. 
In \Cref{appendix_exploration}, we show that the dominant is due to exploration (I), for a total number of visits of order $\imeasure^*(\pair) \ivalue(\imeasure^*, \model)^{-1} \log(T)$, where $\imeasure^*$ is central optimal exploration measure of $\model$. 
Lastly, the number of panic times (III) is upper-bounded by $\abs{\pairs} \abs{\states}$ in \Cref{appendix_panic_times}. 

\subsection{Required assumptions on the regularization hyperparameter}
\label{appendix_assumption_regularizer}

In the regret analysis, we assume that the regularization hyparameter $\EPSILON$ satisfies \Cref{assumption_regularization_hyperparameter}.
These assumptions are technical and we invite the reader to only take them into consideration when necessary.
It is easy to check that \Cref{assumption_regularization_hyperparameter} is satisfied for $\EPSILON$ chosen as in \eqref{equation_choice_regularization} in the main text. 

\begin{assumption}[Properties of regularization hyperparameter]
\label{assumption_regularization_hyperparameter}
    The regularization hyperparameter $\EPSILON \equiv (\epsilonflat, \epsilonunif, \epsilonreg, \epsilontest) : \NN \to (0, 1]$ satisfies
    \begin{itemize}[leftmargin=2.56em] 
        \item[\upshape(\strong{A0})]
            The map $\EPSILON$ is vanishing, i.e., $\EPSILON(m) \to 0$ as $m \to \infty$;
        \item[\upshape(\strong{A1})]
            The map $\EPSILON$ is non-increasing for the product order $\le$, i.e., $\EPSILON(n) \le \EPSILON(m)$ for $n \ge m$;
        \item[\upshape(\strong{A2})]
            For all $\alpha > 0$, $\epsilonflat(T) = \omega(\log(T)^{-\alpha})$ when $T \to \infty$;
        \item[\upshape(\strong{A3})]
            For all $\alpha > 0$, $\epsilonreg(m) = \omega(m^{-\alpha})$ when $m \to \infty$;
        \item[\upshape(\strong{A4})]
            For all $\alpha > 0$, $\epsilonunif(m) = \omega(\log(m)^{-\alpha})$ when $m \to \infty$;
        \item[\upshape(\strong{A5})]
            We have $\epsilonunif(m) = \oh(\epsilonreg(m))$ when $m \to \infty$;
        \item[\upshape(\strong{A6})]
            For all $(\alpha_i) \ge 0$, the function $\varphi_\alpha(m) := \epsilonreg(m)^{\alpha_1} \epsilonunif(m)^{\alpha_2} \exp(- \alpha_3 \epsilonunif(m)^{-\alpha_4}) m$ is non-decreasing asymptotically, i.e., $\exists m_\alpha, \forall \ell' \ge \ell \ge m_\alpha, \varphi_\alpha (\ell) \le \varphi_\alpha (\ell')$;
        \item[\upshape(\strong{A7})]
            $\epsilontest(T) = \omega\parens*{\log\log(T) \log(T)^{-1}}$ when $T \to \infty$;
    \end{itemize}
\end{assumption}

\subsection{Amount of wrong exploitation}
\label{appendix_wrong_coexploration}

In this section, we upper-bound the number of ``wrong'' exploitation times, i.e., the amount of time spent playing the approximately optimal empirical policy $\policy_t^+$ while $\policy_t^+$ is a sub-optimal policy of the true hidden model $\model$.

\begin{lemma}
\label{lemma_wrong_coexploration}
    For all $\pair \notin \optpairs(\model)$, we have
    \begin{equation}
    \label{equation_wrong_coexploration}
        \EE \brackets*{
            \sum_{t=1}^{T-1}
            \indicator{\Pair_t = \pair, t \in \stimes^+}
        }
        = \OH \parens*{
            \frac 1{\epsilonflat(T)^6}
            + \sum_{t=1}^{T-1} \exp\parens*{
                - \parens*{1 + \frac 12 \epsilontest(t)} \log(t)
            }
        }
    \end{equation}
\end{lemma}

All throughout, we fix a non-optimal pair $\pair_0 \notin \optpairs(\model)$.

\subsubsection{Idea of the proof}
\label{appendix_wrong_coexploration_good_events}

To begin with, note that the algorithm only takes decisions at \emph{initial} exploitation times.
Therefore, it is only relevant to decompose $t \in \stimes^+$ according to good events if $t \in \stimes^+_0$. 
Thankfully, the number of exploitation times $\abs{\stimes^+}$ and the number of initial exploitation times $\abs{\stimes^+_0}$ are proportional to one another, because exploitation phases are either done up to regeneration or killed by the discovery of an unseen transition; Both happen quickly in average and exploitation phases are short. 
This is formalized in \Cref{lemma_wrong_coexploration_initial}.

\begin{lemma}
\label{lemma_wrong_coexploration_initial}
    For all $\pair_0 \notin \optpairs(\model)$, we have
    \begin{equation*}
        \E \brackets*{
            \sum_{t=1}^{T-1}
            \indicator{\Pair_t = \pair_0, t \in \stimes^+}
        }
        =
        \Theta \parens*{
            \E \brackets*{
                \sum_{t=1}^{T-1}
                \indicator{
                    \pair_0 \in
                    \flatme{\optpairs}{\iepsilonflat(t)}(\hat{\model}_t)
                    ,
                    t \in \stimes^+_0
                }
            }
        }
        \quad\text{as~$T \to \infty$.}
    \end{equation*}
\end{lemma}
The proof of \Cref{lemma_wrong_coexploration_initial} relies on the notion of \strong{co-exploration structure} and is postponed to \Cref{appendix_wrong_coexploration_inital}.

Introduce the good events
\begin{equation}
\label{equation_coexploration_events}
\begin{aligned}
    \event_t & := \parens*{
        \forall \pair \in \skeleton_t,
        \kappa(\model) \snorm{\hat{\model}_t(\pair) - \model(\pair)} 
        < \epsilonflat(t)
        < \frac 13 \gaingap(\model)
    }
    \text{;~and}
    \\
    \event'_t & := \parens*{
        \forall \pair \in \flatme{\optpairs}{\iepsilonflat(t)}(\hat{\model}_t),
        \kappa(\model) \snorm{\hat{\model}_t(\pair) - \model(\pair)} 
        < \epsilonflat(t)
        < \frac 13 \gaingap(\model)
    }
\end{aligned}
\end{equation}
where $\kappa(\model) := \max \braces{C(\model), 1 + \worstdiameter(\model)} < \infty$ is the maximum between the constant $C(\model)$ given by \Cref{proposition_continuity_near_optimality} and the worst diameter (\Cref{definition_worst_diameter}); and $\gaingap(\model)$ is the gain-gap of $\model$ (\Cref{definition_gain_gap}).
The first event $\event_t$ states that the data is approximately correct on the skeleton and the second $\event'_t$ that it is approximately correct on leveled optimal pairs. 
Note that under $\event_t \cap \event_t'$, we have $\snorm{\hat{\model}_t(\pair) - \model(\pair)} < \infty$ for every pair $\pair \in \skeleton_t \cup \flatme{\optpairs}{\iepsilonflat}(\hat{\model}_t)$, so that $\hat{\model}_t (\pair) \sim \model(\pair)$. 

\paragraph{Decomposition in good events.}
Following \Cref{lemma_wrong_coexploration_initial}, we are left to bound the number of initial exploitation times with $\flatme{\optpairs}{\iepsilonflat}(\hat{\model}_t) \setminus \optpairs(\model) \ne \emptyset$.
Under the event $(\flatme{\optpairs}{\iepsilonflat}(\hat{\model}_t) \setminus \optpairs(\model) \ne \emptyset)$, leveled optimal pairs are wrong and the algorithm may play sub-optimal pairs during exploitation. 
Yet, if the algorithm exploits at time $t \in \stimes^+_0$, then it has passed the exploration GLR test \eqref{equation_glr_exploration}, meaning that the algorithm considers that every leveling-aware confusing model has been rejected with satisfying certainty.
It means that the GLR exploration test has failed to detect a lack of information, and we account for three different scenarios for that failure, as given by the decomposition of $\indicator{\pair_0 \in \flatme{\optpairs}{\iepsilonflat}(\hat{\model}_t), t \in \stimes^+_0}$ below:
\begin{equation}
\label{equation_coexploration_visits_decomposition}
    \eqindicator{
        \pair_0 \in \flatme{\optpairs}{\iepsilonflat}(\hat{\model}_t)
        \atop
        t \in \stimes^+_0
    }
    \le 
    \eqindicator{
        \pair_0 \in \flatme{\optpairs}{\iepsilonflat}(\hat{\model}_t)
        \atop
        t \in \stimes^+_0,
        \event_t,
        \event'_t
    }
    + 
    \eqindicator{
        \pair_0 \in \flatme{\optpairs}{\iepsilonflat}(\hat{\model}_t)
        \atop
        t \in \stimes^+_0,
        (\event_t')^c
    }
    + \indicator{\event_t^c}
    .
\end{equation}
where $\pair_0$ is some fixed pair $\pair_0 \notin \optpairs(\model)$.
The first term $\indicator{\pair_0 \in \flatme{\optpairs}{\iepsilonflat}(\hat{\model}_t), t \in \stimes^+_0, \event_t, \event'_t}$ accounts for exploitation times where the empirical data is approximately correct on both the skeleton $\skeleton_t$ and leveled optimal pairs $\flatme{\optpairs}{\iepsilonflat}(\hat{\model}_t)$; Such cases boil down to the power of the GLR exploration test \eqref{equation_glr_exploration} and the tuning of $\epsilontest(t)$, for a total expected aggregate error of order $\integral (\frac 1t)^{1 + \epsilontest(t)} \dd t$, see \Cref{lemma_wrong_coexploration_sanov}.
The third term $\indicator{\event_t^c}$ quantifies how likely it is for the empirical data to be wrong on the skeleton.
Because visit counts are high on the skeleton by definition, the aggregate error due to this term is bounded in expectation, see \Cref{lemma_wrong_coexploration_skeleton}.
The dominant term is the second one, $\indicator{\pair_0 \in \flatme{\optpairs}{\iepsilonflat}(\hat{\model}_t), t \in \stimes^+, (\event')^c_t}$, and accounts for the speed at which information is gathered on leveled optimal pairs $\flatme{\optpairs}{\iepsilonflat}(\hat{\model}_t)$ during exploitation.
This term can be made sub-logarithmic thanks to the co-exploration mechanism of \texttt{ECoE*}.
In expectation, the associated expected aggregate error term is shown to be of order $\epsilonflat(T)^{-6}$, see \Cref{lemma_wrong_coexploration_correct_structure,lemma_wrong_coexploration_wrong_structure}.

\paragraph{The co-exploration structure $\structure_t^+$.}
When a exploitation phase begins, it lasts until regeneration or until regeneration is compromised by the discovery of a new transition that provokes a panic time. 
The possibility that the support of $\hat{\model}_t$ is wrong and that components of $\flatme{\optpairs}{\iepsilonflat}(\hat{\model}_t)$ may not be closed in $\model$ has to be taken into consideration.
Indeed, in the early phases of \texttt{ECoE*}, some transitions $(\State_t, \Action_t, \State_{t+1})$ have never been observed yet.
If they are first observed during an exploitation phase, it is possible that their discovery make \texttt{ECoE*} exit $\flatme{\optpairs}{\iepsilonflat}(\hat{\model}_t)$ and compromise the regenerative requirement of exploitation phases. 
In these regards, we introduce below the notion of \strong{co-exploration structure} and we partition exploitation times $\stimes^+$ depending on it. 

By design, the algorithm splits $\flatme{\optpairs}{\iepsilonflat}(\hat{\model}_t)$ into components $\pairs_t^1, \ldots, \pairs_t^{c(t)}$ that are the communicating components of $\flatme{\optpairs}{\iepsilonflat}(\hat{\model}_t)$ under the empirically observed model $\hat{\model}_t$.
The decomposition itself is denoted $\structure_t^+ := (\pairs_t^1, \ldots, \pairs_t^{c(t)})$ and is called the \strong{co-exploration structure} at time $t$. 
Given a co-exploration structure $\structure_0$, we introduce $\stimes^+ (\structure_0)$ the exploitation times when the exploitation structure is $\structure_0$ at the associated exploitation time, i.e.,
\begin{equation}
\label{equation_coexploration_times_structure}
    \stimes^+(\structure_0)
    :=
    \bigcup \braces*{
        \braces*{t_k, \ldots, t_{k+1}-1}
        :
        k \ge 1, 
        t_k \in \stimes^+_0
        \text{~and~}
        \structure_{t_k}^+ = \structure_0
    }
\end{equation}
and further write $\stimes^+_0(\structure_0) = \stimes^+(\structure_0) \cap \stimes^+_0$ the associated initial exploitation times.
Depending on whether the co-exploration structure $\structure_0$ is closed in $\model$ (i.e., every $\pairs_0 \in \structure_0$ is communicating in $\model$) or not (see \Cref{definition_closed_structure} below), the analysis is different, see \Cref{lemma_wrong_coexploration_correct_structure,lemma_wrong_coexploration_wrong_structure}.

\begin{definition}[Closed structure]
\label{definition_closed_structure}
    Given a Markov decision process $\model \equiv (\pairs, \kernel, \rewardd)$, a set of pairs $\pairs' \subseteq \pairs$ is said \strong{closed} in $\model$ if $\pairs'$ is equal to its unique communicating component (\Cref{definition_component}), i.e.., if the model $\model|_{\pairs'} \equiv (\pairs', \kernel|_{\pairs'}, \rewarddistribution|_{\pairs'})$ is well-defined and communicating;
    A structure $\structure \equiv (\pairs_1, \ldots, \pairs_c)$ is said \strong{closed} in $\model$ if every $\pairs_i \subseteq \pairs$ is closed in $\model$. 
\end{definition}

\subsubsection{Proof of \Cref{lemma_wrong_coexploration_initial}: Initial exploitation and exploitation times}
\label{appendix_wrong_coexploration_inital}

All throughout the section, we fix $\pair_0 \notin \optpairs(\model)$.

\medskip
\begin{proof}
    Let $t_{k} \in \stimes^+_0(\structure_0)$ be an initial exploitation time with co-exploration structure $\structure_0$ such that $\pair_0 \in \bigcup \structure_0 \equiv (\pairs_0^1, \ldots, \pairs_0^m)$ where $\pairs_0^i$ is the $i$-th communicating component of $\structure_0$ in $\hat{\model}_{t_k}$.
    Denote $\pairs_0^u$ the component containing $\State_{t_k}$. 
    We introduce the reward function 
    \begin{equation*}
        f_{t_k}(\state, \action) 
        \equiv f(\state, \action) 
        := 
        \indicator{\state = \State_{t_k}} 
        + \indicator{\state \notin \states(\pairs_0^u)}
    \end{equation*}
    that tracks visits to the initial state and states outside of $\pairs_0^u$.
    Let $\gainof{f}$, $\biasof{f}$ be the gain and bias functions obtained by iterating the uniform policy $\policy^u$ on $\pairs_0^u$, extended to the uniform policy outsie of $\states(\pairs_0^u)$ for the reward function $f$.
    Let $\gapsof{f}(\state, \action) = \gainof{f}(\state) + \biasof{f}(\state) - f(\state, \action) - \kernel(\state, \action) \biasof{f}$ be the associated gap function.

    If $\pairs_0^u$ is closed in $\model$, then we see that the iterates of $\policy^u$ stating from $\State_{t_k}$ remain almost surely on $\pairs^u_0$, so $\vecspan{\gainof{f}|{\scriptstyle \states(\pairs_0^i)}} = 0$;
    If $\pairs_0^u$ is not closed in $\model$, then $\pairs_0^u$ is either transient or recurrent under $\policy^u$ in $\model$, and in either cases $\vecspan{\gainof{f}} = 0$.
    Overall, we have $\vecspan{\gainof{f}|{\scriptstyle \states(\pairs_0^u)}} = 0$, that $\alpha_{t_k} := \min(\gainof{f}|{\scriptstyle \states(\pairs_0^u)}) > 0$ and that $\beta_{t_k} := \max\braces{\vecspan{\biasof{f}}, \norm{\gapsof{f}}_\infty} < \infty$.
    Let $0 < \alpha$, $\beta < \infty$ be the respective minimum and maximum values that $\alpha_{t_k}$, $\beta_{t_k}$ can take for all finitely many possible values for $\structure_0$, $u$ and $\State_{t_k}$.

    Now, during the phase $\braces{t_k, \ldots, t_{k+1}-1}$, note that $f(\Pair_t) = 1$ holds only for $t = t_k$ by construction of $f$ and of exploitation phases.
    Therefore,
    \begin{align*}
        1 
        & =
        \sum_{t=t_k}^{t_{k+1}-1}
        f_{t_k} (\Pair_t)
        \\
        & \overset{(\dagger)}=
        \sum_{t=t_k}^{t_{k+1}-1}
        \parens*{
            \gainof{f_{t_k}}(\State_t)
            + \parens*{\unit_{\State_t} - \kernel(\Pair_t)} \biasof{f_{t_k}}
            - \gapsof{f_{t_k}}(\Pair_t)
        }
        \\
        & \overset{(\ddagger)}\ge
        \alpha \parens*{t_{k+1} - t_k}
        - \beta \indicator{\State_{t_{k+1}} \ne \State_{t_k}}
        + \sum_{t=t_k}^{t_{k+1}-1}
        \parens*{
            \parens*{\unit_{\State_{t+1}} - \kernel(\Pair_t)} \biasof{f_{t_k}}
            - \gapsof{f_{t_k}}(\Pair_t)
        }
    \end{align*}
    where 
    $(\dagger)$ follows by definition of $\gapsof{f}(\Pair_t)$; and
    $(\ddagger)$ follows from a telescopic sum. 
    Note that the right-most term is a martingale, hence its expectation is zero.
    Indeed, $\EE[\unit_{\State_{t+1}}|\Pair_t] = \kernel(\Pair_t)$ and by Poisson's equation, $\EE[\gapsof{f}(\Pair_t)|\State_t] = \sum_{\action \in \actions(\State_t)} \policy^u (\action|\State_t) \gapsof{f}(\State_t, \action_t) = 0$. 

    Summing for $t_k \in \stimes_0^+$, we obtain
    \begin{align*}
        & \EE \brackets*{
            \sum_{t=1}^{T-1}
            \indicator{
                \pair_0 \in \flatme{\optpairs}{\iepsilonflat(t)}(\hat{\model}_t),
                t \in \stimes^+_0
            }
        }
        \\
        & \overset{(\dagger)}\equiv
        \EE \brackets*{
            \sum_{t=1}^{T-1}
            \indicator{
                \event^0_t,
                t \in \stimes^+_0
            }
        }
        \\
        & \ge
        \EE \brackets*{
            \sum_{k=1}^\infty
            \indicator{
                t_k < T,
                \event^0_t,
                t_k \in \stimes^+_0
            } \parens[\Big]{
                \alpha \parens*{
                    \min(T, t_{k+1}) - t_k
                }
                - \beta \indicator{\State_{t_{k+1}} \ne \State_{t_k}}
            }
        }
        \\
        & \ge
        \alpha \EE \brackets*{
            \sum_{t=1}^{T-1}
            \indicator{
                \event^0_t,
                t \in \stimes^+
            }
        }
        - \beta \EE \brackets*{
            \sum_{k=1}^\infty
            \indicator{
                \event^0_t,
                t_k \in \stimes^+_0,
                \State_{t_{k+1}} \ne \State_{t_k}
            } 
        }
        \\
        & =
        \alpha \EE \brackets*{
            \sum_{t=1}^{T-1}
            \indicator{
                \event^0_t,
                t \in \stimes^+
            }
        }
        - \beta 
        \sum_{\structure_0}
        \EE \brackets*{
            \sum_{k=1}^\infty
            \indicator{
                \event^0_t,
                t_k \in \stimes^+_0(\structure_0),
                \State_{t_{k+1}} \ne \State_{t_k}
            } 
        }
        \\
        & \overset{(\ddagger)}\ge
        \alpha \EE \brackets*{
            \sum_{t=1}^{T-1}
            \indicator{
                \event^0_t,
                t \in \stimes^+
            }
        }
        - \beta \abs{\states}\abs{\pairs}
    \end{align*}
    where 
    $(\dagger)$ introduces the short-hand $\event_t^0 \equiv (\pair_0 \in \flatme{\optpairs}{\iepsilonflat}(\hat{\model}_t))$; and 
    $(\ddagger)$ relies on the observation that the number of exploitation phases killed by a panic is at most $\abs{\states} \abs{\pairs}$, the maximal number of times a transition may be discovered. 
    We conclude accordingly.
\end{proof}

\subsubsection{Wrong exploitation with correct data on leveled optimal pairs}
\label{appendix_wrong_coexploration_sanov}

We start by controlling the error term due to exploitation times for which $\Pair_t \notin \optpairs(\model)$ while $\event_t$ and $\event'_t$ hold, see \eqref{equation_coexploration_events}. 
In other words: The algorithm exploits a non-optimal pair, meaning that the optimal policy is wrongly estimated while the data is correct on the skeleton and on leveled optimal pairs because $\event_t$ and $\event'_t$ hold. 

\begin{lemma}
\label{lemma_wrong_coexploration_sanov}
    Let $\pair_0 \notin \optpairs(\model)$ and let $\event_t$, $\event_t'$ be the events as given by \eqref{equation_coexploration_events}.
    Then
    \begin{equation*}
        \EE \brackets*{
            \sum_{t=1}^{T-1}
            \indicator{
                \pair_0 \in \flatme{\optpairs}{\iepsilonflat(t)}(\hat{\model}_t),
                t \in \stimes^+_0,
                \event_t, \event'_t
            }
        }
        \le
        \sum_{t=1}^{T-1}
        \exp \parens*{
            - \parens*{1 + \frac 12 \epsilontest(t)} \log(t)
        }
        .
    \end{equation*}
\end{lemma}

The idea of the proof is that if the exploration policy is not optimal, then the empirical data $\hat{\model}_t(\pair)$ is significatively incorrect for some $\pair \in \pairs$. 
Yet, under $\event_t \cap \event_t'$, the data is approximately correct on $\skeleton_t \cup \flatme{\optpairs}{\iepsilonflat(t)}(\hat{\model}_t)$; Yet, the algorithm cannot exploits unless the GLR exploration test of \eqref{equation_glr_exploration} has been passed, that is supposed to check the correctness of empirical data outside of $\skeleton_t \cup \flatme{\optpairs}{\iepsilonflat(t)}(\hat{\model}_t)$.
So, the error term bounded by \Cref{lemma_wrong_coexploration_sanov} is directly related to the correctness of the exploration test.

\medskip
\begin{proof}
    Introduce $\model_t^\dagger$ the model given as follows:
    \begin{equation}
    \label{equation_wrong_coexploration_sanov_step_0}
        \model_t^\dagger (\pair)
        := 
        \begin{cases}
            \hat{\model}_t(\pair) 
            & \text{if $\pair \in \skeleton_t \cup \flatme{\optpairs}{\iepsilonflat}(\hat{\model}_t)$;}
            \\
            \model(\pair)
            & \text{if $\pair \notin \skeleton_t \cup \flatme{\optpairs}{\iepsilonflat}(\hat{\model}_t)$.}
        \end{cases}
    \end{equation}
    Note that since $\hat{\model}_t \ll \model$, we have $\hat{\model}_t \ll \model^\dagger_t$ as well.

    \par
    \medskip
    \noindent
    \STEP{1}
    \textit{
        Under the event $\event_t \cap \event'_t \cap (\pair_0 \in \flatme{\optpairs}{\iepsilonflat}(\hat{\model}_t))$, the model $\model^\dagger_t$ as given by \eqref{equation_wrong_coexploration_sanov_step_0} satisfies $\model_t^\dagger \in \confusing(\hat{\model}_t)$.
    }
    \medskip
    \begin{subproof}
        Under $\event_t \cap \event'_t$ and following \Cref{proposition_continuity_near_optimality}, we have $\optpairs(\model^\dagger_t) \subseteq \optpairs(\model)$. 
        Now, under the event $(\pair_0 \in \flatme{\optpairs}{\iepsilonflat}(\hat{\model}_t))$, there exists a policy $\policy \in \optpolicies(\flatmodel{\hat{\model}_t}{\iepsilonflat(t)})$ that is (\emph{i}) unichain in $\hat{\model}_t$ (\emph{ii}) making $\pair_0$ recurrent in $\hat{\model}_t$ such that 
        \begin{equation}
        \label{equation_wrong_coexploration_correct_data_step_1_a}
            \gainof{\policy}(\hat{\model}_t) 
            \overset{(\dagger)}\ge
            \optgain(\flatmodel{\hat{\model}_t}{\iepsilonflat(t)}) - \epsilonflat(t)
            \overset{(\ddagger)}=
            \optgain(\hat{\model}_t) - \epsilonflat(t)
        \end{equation}
        where 
        $(\dagger)$~follows by definition of the leveling transform $\flatmodel{-}{\epsilon}$, see \Cref{definition_leveled_mdp}; and
        $(\ddagger)$~follows by \Cref{lemma_leveled_bias_is_bellman}.
        Under $\event_t \cap \event'_t$, we have $\hat{\model}_t(\pair) \sim \model(\pair)$ for $\pair \in \skeleton_t \cup \flatme{\optpairs}{\iepsilonflat}(\hat{\model}_t)$ and $\hat{\model}_t(\pair) \ll \model(\pair)$ for other pairs, so $\policy$ is unichain and $\pair_0$ is recurrent under $\policy$ in $\model$ as well. 
        Together with $\pair_0 \notin \optpairs(\model)$, we deduce that $\gainof{\policy}(\state; \model) \le \optgain(\state; \model) - \gaingap(\model)$ for all $\state \in \states$.
        Now, by design of $\kappa(\model)$ in \eqref{equation_coexploration_events}, 
        \begin{align}
        \notag
            \gainof{\policy}(\state; \model_t^\dagger)
            & \overset{(\dagger)}\le 
            \gainof{\policy}(\state; \model)
            + \kappa(\model) \snorm{\model_t^\dagger - \model}
            \\
        \notag
            & \overset{(\ddagger)}\le 
            \gainof{\optpolicy}(\state; \model) 
            - \gaingap(\model)
            + \kappa(\model) \snorm{\model_t^\dagger - \model}
            \\
        \label{equation_wrong_coexploration_correct_data_step_1_b}
            & \overset{(\S)}\le
            \gainof{\optpolicy}(\state; \model_t^\dagger) 
            - \gaingap(\model)
            + 2 \kappa(\model) \snorm{\model_t^\dagger - \model}
        \end{align}
        for all $\state \in \states$, where in the above
        $(\dagger)$~follows by \Cref{lemma_unichain_gain_variations} and by $\kappa(\model) \ge 1 + \wdiameter(\model)$;
        $(\ddagger)$~introduces $\optpolicy$ an arbitrary bias optimal policy of $\model$ and $\gaingap(\model)$ is the gain-gap (\Cref{definition_gain_gap}); and
        $(\S)$~invokes \Cref{lemma_unichain_gain_variations} and $\kappa(\model) \ge 1 + \wdiameter(\model)$ again.
        So, if $\hat{\policy} \in \optpolicies(\hat{\model}_t)$, then for all $\state \in \states$, we have
        \begin{align*}
            \gainof{\hat{\policy}}(\state; \model_t^\dagger)
            & \overset{(\dagger)}=
            \gainof{\hat{\policy}}(\state; \hat{\model}_t)
            \\
            & \overset{(\ddagger)}\le
            \gainof{\policy}(\state; \hat{\model}_t) + \epsilonflat(t)
            \\
            & \overset{(\S)}\le
            \gainof{\policy}(\state; \model^\dagger_t) + \epsilonflat(t)
            \overset{(\$)}\le
            \gainof{\optpolicy}(\state; \model_t^\dagger) 
            - \gaingap(\model)
            + 2 \kappa(\model) \snorm{\model_t^\dagger - \model}
            + \epsilonflat(t)
        \end{align*}
        where 
        $(\dagger)$ follows from $\model_t^\dagger = \hat{\model}_t$ on $\flatme{\optpairs}{\iepsilonflat}(\hat{\model}_t)$;
        $(\ddagger)$ follows from \eqref{equation_wrong_coexploration_correct_data_step_1_a}; 
        $(\S)$ follows from $\model_t^\dagger = \hat{\model}_t$ on $\flatme{\optpairs}{\iepsilonflat}(\hat{\model}_t)$ again; and
        $(\$)$ follows from \eqref{equation_wrong_coexploration_correct_data_step_1_b}.
        As $\snorm{\model^\dagger_t - \model} \le \snorm{\hat{\model}_t - \model}$, we conclude that if $2 \kappa(\model) \snorm{{\model}^\dagger_t - \model} + \epsilonflat(t) < \gaingap(\model)$, then $\hat{\policy} \notin \optpolicies(\model_t^\dagger)$.

        Hence $\model_t^\dagger \in \confusing(\hat{\model}_t)$.
    \end{subproof}

    \noindent
    \STEP{2}
    \textit{
        For all $\pair_0 \notin \optpairs(\model)$, we have
        \begin{equation}
        \label{equation_wrong_coexploration_sanov_step_2}
            \Pr \parens*{
                \pair_0 \in \flatme{\optpairs}{\iepsilonflat}(\hat{\model}_t),
                t \in \stimes^+_0,
                \event_t,
                \event'_t
            }
            \le
            \exp \parens*{
                - \parens*{1 + \frac 12 \epsilontest(t)} \log(t)
            }
            .
        \end{equation}
        \vspace{-1em}
    }
    \begin{subproof}
        By definition of initial exploitation times, if $t \in \stimes_0^+$ then the algorithm has passed its exploration test of \eqref{equation_glr_exploration}, meaning that for all $\model^\dagger \in \confusing(\hat{\model}_t)$ that coincides with $\hat{\model}_t$ on $\skeleton_t \cup \flatme{\optpairs}{\iepsilonflat}(\hat{\model}_t)$, we have
        \begin{equation}
        \label{equation_wrong_coexploration_sanov_step_2_a}
            \psi_{\visits_t}(\hat{\model}_t||\model^\dagger)
            :=
            \sum_{\pair \in \pairs}
            \visits_t (\pair) 
            \KL(\hat{\model}_t(\pair)||\model^\dagger(\pair))
            \ge
            (1 + \epsilontest(t)) \log(t)
            .
        \end{equation}
        Let $\model_t^\dagger$ be the model as given by \eqref{equation_wrong_coexploration_sanov_step_0}.
        On $\event_t \cap \event'_t \cap (t \in \stimes_0^+) \cap (\pair_0 \in \flatme{\optpairs}{\iepsilonflat}(\hat{\model}_t)$ and by \STEP{1}, we have $\model_t^\dagger \in \confusing(\hat{\model}_t)$ so \eqref{equation_wrong_coexploration_sanov_step_2_a} holds in particular for $\model^\dagger \equiv \model^\dagger_t$ and
        \begin{align}
            (1 + \epsilontest(t)) \log(t)
            & \overset{(\dagger)}\le
        \notag
            \psi_{\visits_t} (\hat{\model}_t||\model^\dagger_t)
            \\
        \label{equation_wrong_coexploration_sanov_step_2_b}
            & \overset{(\ddagger)}=
            \sum_{\pair \notin \skeleton_t}
            \visits_t(\pair) 
            \KL(\hat{\model}_t(\pair)||\model(\pair))
            \overset{(\S)}\equiv 
            \psi_{\visits'_t} (\hat{\model}_t||\model)
        \end{align}
        where 
        $(\dagger)$ holds on $\event_t \cap \event_t' \cap (t \in \stimes_0^+) \cap (\pair_0 \notin \flatme{\optpairs}{\iepsilonflat}(\hat{\model}_t))$ following \eqref{equation_wrong_coexploration_sanov_step_2_a}; 
        $(\ddagger)$ follows by construction of $\model^\dagger_t$; and
        $(\S)$ introduces $\visits'_t (\pair) := \indicator{\pair \notin \skeleton_t} \visits_t (\pair)$.
        We conclude the proof with a combinatorial argument invoking Sanov's theorem.
        For $n \in \N^\pairs$, we denote $\hat{\model}_{(n)}$ the observed model under the deterministic number of visits $\visits_t = n$; Note that $\hat{\model}_t = \hat{\model}_{(\visits_t)}$ by definition. 
        Let $\models_{(n)}$ the discrete space of Markov decision processes where rewards and kernels at $\pair \in \pairs$ are all the possible empirical distributions that can be obtained after $n(\pair)$ samples.
        Further denote $[\log^2(t)] := \braces{0, \ldots, \floor{\log^2(T)}}$. 
        We have
        \begin{align*}
            & \EE \brackets*{
                \indicator{
                    \pair_0 \in \flatme{\optpairs}{\iepsilonflat}(\hat{\model}_t),
                    t \in \stimes^+_0,
                    \event_t,
                    \event'_t
                }
            }
            \\
            & \overset{(\dagger)}\le
            \EE \brackets*{
                \eqindicator{
                    \psi_{\visits'_t} (\hat{\model}_t||\model)
                    \ge (1 + \epsilontest(t)) \log(t)
                }
            }
            \\
            & \overset{(\ddagger)}=
            \sum_{n \in [\log^2(t)]^\pairs}
            \EE \brackets*{\eqindicator{
                \psi_n (\hat{\model}_t||\model)
                \ge
                (1 + \epsilontest(t)) \log(t),
                \atop
                (\forall \pair \notin \skeleton_t, \visits_t (\pair) = n (\pair)),
                (\forall \pair \in \skeleton_t, n(\pair) = 0)
            }}
            \\
            & =
            \sum_{n \in [\log^2(t)]^\pairs}
            \sum_{\model' \in \models_{(n)}}
            \EE \brackets*{
                \indicator{\psi_n(\model'||\model) \ge (1 + \epsilontest(t)) \log(t)}
                \indicator{\hat{\model}_{(n)} = \model'}
            }
            \\
            & =
            \sum_{n \in [\log^2(t)]^\pairs}
            \sum_{\model' \in \models_{(n)}}
            \indicator{\psi_n(\model'||\model) \ge (1 + \epsilontest(t)) \log(t)}
            \Pr \parens*{
                \hat{\model}_{(n)} = \model'
            }
            \\
            & \overset{(\S)}\le
            \sum_{n \in [\log^2(t)]^\pairs}
            \sum_{\model' \in \models_{(n)}}
            \eqindicator{\psi_n(\model'||\model) \ge (1 + \epsilontest(t)) \log(t)}
            \exp \parens[\Big]{
                - \psi_n (\model'||\model)
            }
            \\
            & \le 
            \sum_{n \in [\log^2(t)]^\pairs}
            \sum_{\model' \in \models_{(n)}}
            \exp \parens[\Big]{
                - (1 + \epsilontest(t)) \log(t)
            }
            \\
            & \overset{(\$)}\le
            \parens*{
                1 + \floor{\log^2(t)}
            }^{\abs{\pairs}(3 + \abs{\states})}
            \parens*{
                \frac 1t
            }^{1 + \iepsilontest(t)}
            \\
            & \overset{(\#)}\le
            \exp \parens*{
                - \parens*{1 + \frac 12 \epsilontest(t)}
                \log(t)
            }
            .
        \end{align*}
        In the above,
        $(\dagger)$ is obtained by invoking \eqref{equation_wrong_coexploration_sanov_step_2_b};
        $(\ddagger)$ follows by construction of $\visits'_t$;
        $(\S)$ is a consequence of an all-time Sanov theorem (\Cref{lemma_sanov_all_time});
        $(\$)$ follows from classic combinatorial bounds (\Cref{lemma_combinatorial_bounds}); and
        $(\#)$ is obtained by using $\epsilontest(t) \ge 4 \abs{\pairs}(3 + \abs{\states}) \log \log(t) \log(t)^{-1}$ (see \Cref{assumption_regularization_hyperparameter}).
    \end{subproof}

    We conclude the proof of \Cref{lemma_wrong_coexploration_sanov} by summing \eqref{equation_wrong_coexploration_sanov_step_2} of \STEP{2} for $t \ge 1$.
\end{proof}

\subsubsection{Uniform exploitation: Fundamental property of co-exploration}

Before moving on with the control of the dominant in the decomposition \eqref{equation_coexploration_visits_decomposition}, we establish a fundamental property of the exploitation mechanism of \texttt{ECoE}, that guarantees that all components of a co-exploration structure $\structure_0$ are explored uniformly often relatively to the number of initial exploitation times with the so-called structure, see \Cref{lemma_coexploration_structure_uniform}.
This result holds whether the structure is closed or not.

\begin{lemma}
\label{lemma_coexploration_structure_uniform}
    Let $\structure_0$ be a co-exploration structure. 
    Let $k(i)$ be the $i$-th phase with co-exploration structure $\structure_0$, i.e., $\stimes_0^+(\structure_0) = \braces{t_{k(1)}, t_{k(2)}, \ldots}$. 
    There exists $\alpha > 0$ such that
    \begin{equation}
        \Pr \parens*{
            \exists \pair \in {\textstyle \bigcup \structure_0},
            ~
            \visits_{t_{k(i)}} (\pair) < \alpha \sqrt{i}
        }
        = \oh \parens*{i^{-2}}
        \quad \text{as $i \to \infty$.}
    \end{equation}
\end{lemma}

\begin{proof}
    The co-exploration structure is decomposed into its communicating components (in $\hat{\model}_t$) as $\structure_0 \equiv (\pairs_0^1, \ldots, \pairs_0^c)$.
    Remark that $\stimes^+_0 (\structure_0) = \braces{t_{k(i)} : i \ge 1}$ and that $\stimes^+ (\structure_0) = \bigcup_{i=1}^\infty \braces{t_{k(i)}, \ldots, t_{k(i)+1}-1}$.
    Given a component $\pairs_0^u$ of $\structure_0$, we write $\mathcal{I}^u := \braces{i : \State_{t_{k(i)}} \in \states(\pairs_0^u)}$ the indices of initial exploitation times starting in the component $u = 1, \ldots, c$.

    \medskip
    \par
    \noindent
    \STEP{1} 
    \textit{
        There is $\alpha > 0$ such that for all component $u \in \braces{1, \ldots, m}$ and $\pair_u \in \pairs_0^u$, we have
        \begin{equation}
           \label{equation_proof_lemma_wrong_coexploration_wrong_structure_1}
            \Pr \parens*{
                \visits_{t_{k(i)+1}}(\pair_u) < \alpha \ell
                \text{~and~}
                \ell \le \abs{\mathcal{I}^u \cap \braces{1, \ldots, i}}
            }
            = 
            \oh \parens{\ell^{-2}}
            \quad\text{as~$\ell \to \infty$}
        \end{equation}%
        \vspace{-1em}
    }%
    \begin{subproof}
        The proof shares many similarities with \Cref{lemma_wrong_coexploration_initial}.

        Fix a component $u \in \braces{1, \ldots, c}$ and $\pair_u \in \pairs_0^u$.
        Introduce the reward function $f(\state, \action) := \indicator{(\state, \action) = \pair_u} + \indicator{\state \notin \states(\pairs_0^u)}$.
        Let $\gainof{f}$, $\biasof{f}$ be the gain and bias functions for the reward function $f$ obtained by iterating the uniform policy on $\pairs_0^u$, extended to the uniform policy outside of $\states(\pairs_0^u)$.
        Let $\gapsof{f}(\state, \action) = \gainof{f}(\state) + \biasof{f}(\state) - f(\state, \action) - \kernel(\state, \action) \biasof{f}$ be the associated gap function.
        Note that $\vecspan{\gainof{f}} = 0$ regardless of whether $\pairs_0^u$ is closed (\Cref{definition_closed_structure}) or not. 
        Denote $\alpha := \min(\gainof{f}) > 0$ and $\beta := \max \braces{\vecspan{\biasof{f}}, \norm{\gapsof{f}}_\infty} < \infty$.
        Denoting $[i] := \braces{1, \ldots, i}$, we have
        \begin{align*}
            & \visits_{t_{k(i)+1}}(\pair_u)
            \\
            & \ge
            \sum_{j \in \mathcal{I}^u \cap [i]}
            \sum_{t=t_{k(j)}}^{t_{k(j)+1}-1}
            f (\Pair_t)
            \\
            & \overset{(\dagger)}=
            \sum_{j \in \mathcal{I}^u \cap [i]}
            \sum_{t=t_{k(j)}}^{t_{k(j)+1}-1}
            \parens*{
                \gainof{f}(\State_t) 
                + \parens*{
                    \unit_{\State_t} - \kernel(\Pair_t)
                } \biasof{f}
                - \gapsof{f}(\Pair_t)
            }
            \\
            & \overset{(\ddagger)}\ge
            \alpha \parens*{
                \sum_{j \in \mathcal{I}^u \cap [i]}
                \parens*{t_{k(j)+1} - t_{k(j)}}
            } 
            - \beta
            + \sum_{j \in \mathcal{I}^u \cap [i]}
            \sum_{t=t_{k(j)}}^{t_{k(j)+1}-1}
            \parens*{
                \parens*{
                    \unit_{\State_{t+1}} - \kernel(\Pair_t)
                } \biasof{f}
                - \gapsof{f}(\Pair_t)
            }
            \\
            & \overset{(\S)}\ge
            \alpha T_i^u - 2 \beta \parens*{
                1 + \sqrt{T_i^u \log \parens*{\frac{1 + T_i^u}\delta}}
            }
        \end{align*}
        where
        $(\dagger)$ follows by $\gainof{f}(\state) + \biasof{f}(\state) = f(\state, \action) + \kernel(\state, \action) \biasof{f} + \gapsof{f}(\state, \action)$;
        $(\ddagger)$ uses that by design of panic times, at most one episode $k(j)$ isn't regenerative, i.e., is such that $\State_{t_{k(j)}} \ne \State_{t_{k(j)+1}}$; and 
        $(\S)$ bounds the RHS martingale with a time-uniform Azuma-Hoeffding inequality (\Cref{lemma_azuma_time_uniform}) with probability $1 - \delta$, and $T_i^u$ is a short-hand for $\sum_{j \in \mathcal{I}^u \cap [i]} (t_{k(j)+1} - t_{k(j)})$.
        Remark that
        \begin{equation*}
            \beta \sqrt{
                T_i^u \log\parens*{\frac{1 + T_i^u}\delta}
            }
            <
            \frac 14 \alpha T_i^u
            \iff
            \delta > \exp \parens*{
                - \frac{\alpha \sqrt{T_i^u}}{4 \beta}
                + \log(1 + T_i^u)
            }
            \overset{(\dagger)}=
            \oh \parens*{
                \abs{\mathcal{I}^u \cap [i]}^{-2}
            }
        \end{equation*}
        where $(\dagger)$ holds after $T_i^u \ge \abs{\mathcal{I}^u \cap [i]}$.
        We conclude accordingly.
    \end{subproof}

    \par
    \noindent
    \STEP{2} \textit{
        There exists a constant $\alpha > 0$ such that
        \begin{equation}
           \label{equation_proof_lemma_wrong_coexploration_wrong_structure_2}
            \forall i \ge 1,
            \forall \pair \in \bigcup \structure_0,
            \quad
            \Pr \parens*{
                \visits_{t_{k(i)}} < \alpha \sqrt{i}
            }
            =
            \oh \parens*{i^{-2}}
            .
        \end{equation}
        \vspace{-1em}
    }
    \begin{subproof}
        Since $\structure_0$ has $c$ components, the more visited component $u \in \braces{1, \ldots, c}$ is such that $\abs{\mathcal{I}^u \cap [i]} \ge \ceil{\frac ic}$. 
        In particular and following from the square trick \eqref{equation_square_trick}, for some $j < i$, the component $u$ has been chosen to start a exploitation phase, i.e., there exists $j \in \braces{1, \ldots, i-1}$ such that
        \begin{equation}
        \label{equation_proof_lemma_wrong_coexploration_correct_structure_3}
            \min \braces*{
                \visits_{t_{k(j)}}(\pair) : \pair \in \pairs_0^u
            }
            \le
            \parens*{
                \min \braces*{
                    \visits_{t_{k(j)}}(\pair) : \pair \in \bigcup \structure_0
                }
            }^2
        \end{equation}
        and such that $\abs{\mathcal{I}^u \cap [j]} \ge \ceil{\frac ic} - 1$. 
        Meanwhile, by combining $\abs{\mathcal{I}^u \cap [j]} \ge \ceil{\frac ic} - 1 \ge \frac{i - c}c$ with \eqref{equation_proof_lemma_wrong_coexploration_wrong_structure_1} from \STEP{1}, we readily get
        \begin{equation}
        \label{equation_proof_lemma_wrong_coexploration_correct_structure_4}
            \Pr \parens*{
                \log \min \braces*{
                    \visits_{t_{k(j)+1}}(\pair) : \pair \in \pairs_0^u
                }
                \ge
                \log \parens*{
                    \alpha \cdot \frac{i - m}m
                }
            }
            = 1 - \oh \parens*{i^{-2}}
            .
        \end{equation}
        Because visit counts are monotone with respect to time, combining \eqref{equation_proof_lemma_wrong_coexploration_correct_structure_3} and \eqref{equation_proof_lemma_wrong_coexploration_correct_structure_4}, we obtain that with probability $1 - \oh(i^{-2})$, we have
        \begin{align*}
            \min \braces*{
                \visits_{t_{k(i)}}(\pair) : \pair \in \bigcup \structure_0
            }
            & \ge
            \min \braces*{
                \visits_{t_{k(j)+1}}(\pair) : \pair \in \bigcup \structure_0
            }
            \\
            & \ge 
            \exp \parens*{
                \frac 12 \log \parens*{
                    \alpha \cdot \frac{i-m}m
                }
            }
            = \Omega \parens*{\sqrt{i}}
            .
        \end{align*}
        This proves the claim.
    \end{subproof}

    This concludes the proof of \Cref{lemma_coexploration_structure_uniform}.
\end{proof}

\subsubsection{Wrong exploitation with correct empirical kernel supports}
\label{appendix_wrong_coexploration_correct_structure}

We continue with the dominant term of wrong exploitation, corresponding to exploitation times with $\pair_0 \in \flatme{\optpairs}{\iepsilonflat}(\hat{\model}_t)$, wrong data on leveled optimal pairs $\flatme{\optpairs}{\iepsilonflat}(\hat{\model}_t)$ yet $t$ an initial exploitation time with closed co-exploration structure.
Because the co-exploration mechanism guarantees that all components of $\structure_0$ are explored nearly uniformly fast (up to a square root), the empirical data on $\flatme{\optpairs}{\iepsilonflat}(\hat{\model}_t)$ becomes uniformly approximately correct quickly.
Therefore, the amount of such initial exploitation times is determined by the required precision of the empirical data on $\flatme{\optpairs}{\iepsilonflat}(\hat{\model}_t)$ in $\event'_t$, which is of order $\epsilonflat(t)$.

We end up with a bound of order $\epsilonflat(T)^{-6}$, see \Cref{lemma_wrong_coexploration_correct_structure} thereafter.

\begin{lemma}
\label{lemma_wrong_coexploration_correct_structure}
    Let $\pair_0 \notin \optpairs(\model)$.
    Let $\event_t, \event'_t$ be the good events as given by \eqref{equation_coexploration_events}. 
    For all closed co-exploration structure $\structure_0$, we have
    \begin{equation*}
        \EE \brackets*{
            \sum_{t=1}^{T-1} 
            \eqindicator{
                \pair_0 \in \flatme{\optpairs}{\iepsilonflat(t)}(\hat{\model}_t),
                \atop
                t \in \stimes^+_0 (\structure_0), 
                (\event_t')^c
            }
        }
        =
        \OH \parens*{ 
            \frac 1{
                \epsilonflat(T)^6
            }
        }
        .
    \end{equation*}
\end{lemma}

\medskip
\begin{proof}
    Fix $\structure_0 \equiv (\pairs_0^1, \ldots, \pairs_0^c)$ a closed co-exploration structure.
    Recall that $\stimes^+_0 (\structure_0) := \stimes^+ (\structure_0) \cap \stimes^+_0$ denotes the initial exploitation times with co-exploration structure $\structure_0$.
    By definition, see \eqref{equation_coexploration_times_structure}, $\stimes^+(\structure_0)$ is a union of phases $\braces{t_k, \ldots, t_{k+1}-1}$ and we denote $k(i)$ the $i$-th phase for which $t_{k(i)} \in \stimes^+(\structure_0)$. 
    We have
    \begin{equation*}
        \stimes^+_0 (\structure_0)
        = \braces{t_{k(i)} : i \ge 1}
        \text{\quad and \quad}
        \stimes^+ (\structure_0)
        = \bigcup_{i=1}^\infty \braces{t_{k(i)}, \ldots, t_{k(i)+1}-1}
        .
    \end{equation*}
    Given a component $\pairs_0^u$ of $\structure_0$, we write $\mathcal{I}^u := \braces{i : \State_{t_{k(i)}} \in \states(\pairs_0^u)}$ the indices of initial exploitation times starting in the component $u = 1, \ldots, c$.
    
    \par
    \medskip
    \noindent
    \STEP{1}
    \textit{
        There exists a constant $\alpha > 0$ such that, as $T \to \infty$, we have
        \begin{equation}
        \notag
            \EE \brackets*{
                \sum_{t=1}^{T-1}
                \eqindicator{
                    t \in \stimes^+_0 (\structure_0), 
                    \atop
                    (\event'_t)^c
                }
            }
            \le
            \EE \brackets*{
                \sum_{j=1}^\infty
                \sum_{\pair \in \bigcup \structure_0}
                \eqindicator{
                    C (\model) \snorm{\hat{\model}_{t_{k(j)}}(\pair) - \model(\pair)} \ge \epsilonflat(T),
                    \atop
                    \visits_{t_{k(j)}}(\pair) > \alpha \sqrt{j}
                }
            }
            + \frac 1\alpha
            .
        \end{equation}
        \vspace{-1em}
    }
    \begin{subproof}
        This follows from the following direction computation:
        \begin{align*}
            & \EE \brackets*{
                \sum_{t=1}^{T-1}
                \indicator{
                    t \in \stimes^+_0(\structure_0), 
                    (\event_t')^c
                }
            }
            \\
            & = 
            \EE \brackets*{
                \sum_{j=1}^\infty
                \indicator{
                    t_{k(j)} < T,
                    (\event_{t_{k(j)}}')^c
                }
            }
            \\
            & \le
            \EE \brackets*{
                \sum_{j=1}^\infty
                \eqindicator{
                    t_{k(j)} < T,
                    (\event_{t_{k(j)}}')^c,
                    \atop
                    \parens{
                        \forall \pair \in \bigcup\structure_0,
                        \visits_{t_{k(j)}} (\pair) \ge \alpha \sqrt{j}
                    }
                }
            }
            + \EE \brackets*{
                \sum_{j=1}^\infty
                \indicator{
                    \exists \pair \in {\textstyle \bigcup \structure_0},
                    \visits_{t_{k(j)}} (\pair) < \alpha \sqrt{j}
                }
            }
            \\
            & \overset{(\dagger)}=
            \EE \brackets*{
                \sum_{j=1}^\infty
                \eqindicator{
                    t_{k(j)} < T,
                    {
                        \parens{
                            \exists \pair \in \bigcup\structure_0,
                            \kappa(\model) \snorm{\hat{\model}_{t_{k(j)}}(\pair) - \model(\pair)}
                            \ge
                            \epsilonflat(t_{k(j)})
                        }
                        \atop
                        \parens{
                            \forall \pair \in \bigcup\structure_0,
                            \visits_{t_{k(j)}} (\pair) \ge \alpha \sqrt{j}
                        }
                    }
                }
            }
            + \sum_{j=1}^\infty \oh\parens*{\frac 1{j^2}}
            \\
            & \overset{(\ddagger)}\le
            \EE \brackets*{
                \sum_{j=1}^\infty
                \eqindicator{
                    t_{k(j)} < T,
                    {
                        \parens{
                            \exists \pair \in \bigcup\structure_0,
                            \kappa(\model) \snorm{\hat{\model}_{t_{k(j)}}(\pair) - \model(\pair)}
                            \ge
                            \epsilonflat(T)
                        }
                        \atop
                        \parens{
                            \forall \pair \in \bigcup\structure_0,
                            \visits_{t_{k(j)}} (\pair) \ge \alpha \sqrt{j}
                        }
                    }
                }
            }
            + \OH(1)
            \\
            & \le 
            \EE \brackets*{
                \sum_{j=1}^\infty
                \sum_{\pair \in \bigcup\structure_0}
                \eqindicator{
                    \kappa(\model) \snorm{\hat{\model}_{t_{k(j)}}(\pair) - \model(\pair)} \ge \epsilonflat(T)
                    \atop
                    \visits_{t_{k(j)}}(\pair) > \alpha \sqrt{j}
                }
            }
            + \OH(1)
        \end{align*}
        where 
        $(\dagger)$ unfolds the definition of $(\event_{\tau_j}')^c$ and invokes \Cref{lemma_coexploration_structure_uniform}; and
        $(\ddagger)$ follows by monotonicity of $\epsilonflat$.
        This proves the claim.
    \end{subproof}

    Continuing from \STEP{1}, we prove \Cref{lemma_wrong_coexploration_correct_structure}.
    We write
    \begin{align*}
        & \EE \brackets*{
            \sum_{t=1}^{T-1}
            \indicator{
                t \in \stimes^+_0 (\structure_0), 
                (\event'_t)^c
            }
        }
        \\
        & \overset{(\dagger)}\le
        \EE \brackets*{
            \sum_{j=1}^\infty
            \sum_{\pair \in \bigcup \structure_0}
            \eqindicator{
                C (\model) \snorm{\hat{\model}_{t_{k(j)}}(\pair) - \model(\pair)} \ge \epsilonflat(T),
                \atop
                \visits_{t_{k(j)}}(\pair) > \alpha \sqrt{j}
            }
        }
        + \OH(1)
        \\
        & \overset{(\ddagger)}=
        \sum_{\pair \in \bigcup \structure_0}
        \sum_{j=1}^\infty 
        \Pr \parens*{
            C (\model) \norm{\hat{\model}_{t_{k(j)}}(\pair) - \model(\pair)} \ge \epsilonflat(T),
            \atop
            \visits_{t_{k(j)}}(\pair) > \alpha \sqrt{j}
        }
        + \OH(1)
        \\
        & \overset{(\S)}\le
        \sum_{\pair \in \bigcup \structure_0}
        \sum_{j=1}^\infty 
        \min \braces*{
            1,
            \exp \parens*{
                - \frac{2 \alpha \epsilonflat(T)^2 \sqrt{j}}{\kappa(\model)^2}
                + \frac 12 \log\parens*{1 + \alpha \sqrt{j}}
                + (\abs{\states}+1) \log(2)
            }
        }
        + \OH(1)
    \end{align*}
    where
    $(\dagger)$ follows from \STEP{1};
    $(\ddagger)$ converts $\snorm{-}$ to $\norm{-}$ using \Cref{lemma_convert_snorm_to_norm}; and 
    $(\S)$ follows by \Cref{corollary_threshold_concentration}.
    Setting $\lambda = 2 \epsilonflat(T)^2 \kappa(\model)^{-2}$, we have to upper-bound a term of the form 
    \begin{equation*}
        B
        :=
        \sum_{n=1}^\infty
        \min \braces*{
            1,
            \parens*{1 + \alpha \sqrt{n}}
            \exp \parens*{ - \lambda \sqrt{n} }
        }
    \end{equation*}
    We have
    \begin{align*}
        B
        & \le 
        \frac 2{\alpha}
        \sum_{n=1}^\infty
        \min \braces*{
            \sqrt{n}
            \exp\parens*{- \lambda \sqrt{n}}
            , 1
        }
        \\
        & \le
        \frac 2 \alpha \parens*{
            \ceil*{\frac 1{\lambda^2}}
            + 
            \sum_{n > \ceil{\lambda^{-2}}}
            \sqrt{n} \exp \parens*{- \lambda \sqrt{n}}
        }
        \\
        & \overset{(\ddagger)}\le
        \frac 2\alpha \parens*{
            \ceil*{\frac 1{\lambda^2}}
            + 
            \integral_0^\infty \sqrt{x} \exp\parens*{-\lambda \sqrt{x}} \dd x
        }
        =
        \frac 2\alpha \parens*{
            \ceil*{\frac 1{\lambda^2}}
            + \frac 4{\lambda^3}
        }
        = \OH \parens*{
            \frac 1{\epsilonflat(T)^6}
        }
    \end{align*}
    where $(\dagger)$ follows by a sum-integral comparison, using that $f(x) := \sqrt{x} \exp(-\lambda \sqrt{x})$ is decreasing on $[\lambda^{-2}, \infty)$.
    This concludes the proof.
\end{proof}

\subsubsection{Wrong exploitation with wrong empirical kernel supports}
\label{appendix_wrong_coexploration_wrong_structure}

The proof of \Cref{lemma_wrong_coexploration_correct_structure} can be generalized to non-closed structures.
For non-closed structures however, we can provide a better bound than $\epsilonflat(T)^{-6}$ and show that such structures are all rejected in finite time. 
The idea is simple: If $\structure_0$ is not closed in $\model$, it has a non-closed component $\pairs_0$ and every time the algorithm exploits this component, it has positive probability to escape from it and to panic accordingly---and thanks to the co-exploration mechanism, (\Cref{lemma_coexploration_structure_uniform}), this component is visited about as many times as the algorithm exploits with co-exploration structure $\structure_0$.
See \Cref{lemma_wrong_coexploration_wrong_structure} below.

\begin{lemma}
\label{lemma_wrong_coexploration_wrong_structure}
    Let $\pair_0 \notin \optpairs(\model)$.
    Let $\event_t, \event'_t$ be the good events as given by \eqref{equation_coexploration_events}. 
    For all non-closed co-exploration structure $\structure_0$, we have
    \begin{equation*}
        \EE \brackets*{
            \sum_{t=1}^\infty
            \eqindicator{
                \pair_0 \in \flatme{\optpairs}{\iepsilonflat(t)}(\hat{\model}_t),
                \atop
                t \in \stimes^+_0 (\structure_0), 
                (\event_t')^c
            }
        }
        <
        \infty
        .
    \end{equation*}
\end{lemma}

\begin{proof}
    Fix $\structure_0 \equiv (\pairs_0^1, \ldots, \pairs_0^c)$ a \emph{non} closed exploration structure, meaning that there is $\pairs_0^u \in \structure_0$ together with some $\pair \in \pairs_0^u$ such that $\support(\kernel(\pair)) \not \subseteq \states(\pairs_0^u)$, i.e., playing $\pair \in \pairs_0^u$ makes one exit the states spawned by $\pairs_0^u$ with positive probability. 
    Let $\stimes_0^+ (\structure_0) := \stimes^+ (\structure_0) \cap \stimes_0^+$ be the initial exploitation times with co-exploration structure $\structure_0$. 
    By definition, see \eqref{equation_coexploration_times_structure}, $\stimes^+(\structure_0)$ is a union of phases $\braces{t_k, \ldots, t_{k+1}-1}$ and we denote $k(i)$ the $i$-th phase for which $t_{k(i)} \in \stimes^+(\structure_0)$. 
    We have
    \begin{equation*}
        \stimes^+_0 (\structure_0)
        = \braces{t_{k(i)} : i \ge 1}
        \text{\quad and \quad}
        \stimes^+ (\structure_0)
        = \bigcup_{i=1}^\infty \braces{t_{k(i)}, \ldots, t_{k(i)+1}-1}
        .
    \end{equation*}
    Given a component $\pairs_0^u$ of $\structure_0$, we write $\mathcal{I}^u := \braces{i : \State_{t_{k(i)}} \in \states(\pairs_0^u)}$ the indices of initial exploitation times starting in the component $u = 1, \ldots, c$.

    \par
    \medskip
    \noindent
    \STEP{1} 
    \textit{
        Let $\pairs_0^u \in \structure_0$ be a component that is not closed.
        There exists $C > 0$ such that, for all $x \ge 0$, we have
        \begin{equation}
           \label{equation_proof_lemma_wrong_coexploration_wrong_structure_4}
            \Pr \parens*{
                \sum_{i \in \mathcal{I}^u}
                \parens*{
                    t_{k(i)+1} - t_{k(i)}
                }
                \ge
                x
            } 
            \le
            \exp \parens*{
                - C x + \frac 1C
            }
        \end{equation}
        \vspace{-0.66em}
    }
    \begin{subproof}
        Let $T_i^u := \sum_{j \in \mathcal{I}^u \cap [i]} (t_{k(j)+1} - t_{k(j)})$ the time spent on component $\pairs_0^u$.
        Consider the reward function $f(\state, \action) := \indicator{\state \notin \states(\pairs_0^u)}$ and let $\gainof{f}$, $\biasof{f}$ be the gain and bias functions for ther reward function $f$ obtained by iterating the uniform policy $\policy_u$ on $\pairs_0^u$ extended to the uniform policy on $\actions$ outside of $\states(\pairs_0^u)$.
        As usual, observe that $\vecspan{\gainof{f}} = 0$ and that $0 < \alpha := \min(\gainof{f}), \beta := \vecspan{\biasof{f}} < \infty$.
        By design of panic times, we cannot have $f(\Pair_t) = 1$ for $t = t_{k(i)}, \ldots, t_{k(i)+1}-1$ if $i \in \mathcal{I}^u$.
        We deduce that
        \begin{align*}
            0
            & =
            \sum_{j \in \mathcal{I}^u \cap [i]}
            \sum_{t=t_{k(j)}}^{t_{k(j)+1}-1}
            f (\Pair_t)
            \\
            & \overset{(\dagger)}=
            \sum_{j \in \mathcal{I}^u \cap [i]}
            \sum_{t=t_{k(j)}}^{t_{k(j)+1}-1}
            \parens*{
                \gainof{f}(\State_t)
                + \parens*{\unit_{\State_t} - \kernel^{\policy_u}(\State_t)} \biasof{f}
            }
            \\
            & \overset{(\ddagger)}\ge 
            \alpha T_i^u 
            - \beta 
            + 
            \sum_{j \in \mathcal{I}^u \cap [i]}
            \sum_{t=t_{k(j)}}^{t_{k(j)+1}-1}
            \parens*{\unit_{\State_{t+1}} - \kernel^{\policy_u}(\State_t)} \biasof{f}
            \overset{(\S)}\ge
            \alpha T_i^u - \beta \parens*{
                1 + \sqrt{T_i^u \log \parens*{\frac{1+ T_i^u}\delta}}
            }
        \end{align*}
        where 
        $(\dagger)$ follows by the Poisson equation $\gainof{f}(\state) + \biasof{f}(\state) = f^{\policy_u}(\state) + \kernel^{\policy_u}(\state) \biasof{f}$;
        $(\ddagger)$ uses that by design of panic times, at most one episodes $k(j)$ isn't regenerative; 
        and $(\S)$ bounds the RHS martingale with a time-uniform Azuma-Hoeffding's inequality (\Cref{lemma_azuma_time_uniform}) and holds with probability $1 - \delta$.
        By letting $i$ go to infinity, we see that $T^u := \sum_{i \in \mathcal{I}^u} (t_{k(i)+1} - t_{k(i)})$ satisfies
        \begin{equation*}
            \forall \delta \ge 0,
            \quad
            \Pr \parens*{
                T^u 
                \le 
                \frac \alpha \beta \parens*{
                    1 + \sqrt{
                        T^u \log \parens*{
                            \frac{1 + T^u}\delta
                        }
                    }
                }
            }
            \le
            \delta
        \end{equation*}
        and we conclude with straight-forward algebra that $T^u$ has sub-exponential tails.
    \end{subproof}

    We now prove \Cref{lemma_wrong_coexploration_wrong_structure}.
    Let $u \in \braces{1, \ldots, c}$ be such that $\pairs_0^u$ is not closed.
    We have
    \begin{align*}
        \EE \brackets*{
            \sum_{t=1}^\infty
            \indicator{t \in \stimes^+_0 (\structure_0)}
        }
        & =
        \EE \brackets*{
            \sum_{j=1}^\infty
            \indicator{t_{k(j)} < \infty}
        }
        \\
        & \overset{(\dagger)}=
        \EE \brackets*{
            \sum_{j=1}^\infty
            \eqindicator{
                t_{k(j)} < \infty,
                \parens*{
                    \forall \pair \in {\textstyle \bigcup \structure_0},
                    \visits_{t_{k(j)}}(\pair) \ge \alpha \sqrt{j}
                }
            }
        } + \OH(1)
        \\
        & \overset{(\ddagger)}\le
        \EE \brackets*{
            \sum_{j=1}^\infty
            \eqindicator{
                T^u \ge \alpha \sqrt{j}
            }
        } + \OH(1)
        \\
        & \overset{(\S)}\le
        \sum_{j=1}^\infty
        \exp \parens*{
            - \frac{\alpha}{C} \sqrt{j} + \frac 1C
        }
        + \OH(1)
        < \infty
    \end{align*}
    where 
    $(\dagger)$ follows by \Cref{lemma_coexploration_structure_uniform};
    $(\ddagger)$ introduces $T^u := \sum_{i \in \mathcal{I}^u} (t_{k(i)} - t_{k(i)})$; and
    $(\S)$ follows from \STEP{1};
    This concludes the proof.
\end{proof}

\subsubsection{Correctness of empirical data on the skeleton}

In this paragraph, we bound the last term in \eqref{equation_coexploration_visits_decomposition}, accounting for wrong empirical data on the skeleton. 
The idea of the argument is that since a pair belongs to the skeleton if, and only if $\visits_t(\pair) \ge \log^2(t)$, the associated empirical data concentrates very fast.

\begin{lemma}
\label{lemma_wrong_coexploration_skeleton}
    Let $\event_t$ be the good event as given by \eqref{equation_coexploration_events}. 
    We have
    \begin{equation*}
        \EE \brackets*{
            \sum_{t=1}^{\infty} 
            \indicator{\event_t^c}
        }
        < 
        \infty
        .
    \end{equation*}
\end{lemma}

\begin{proof}
    Note that by \Cref{lemma_convert_snorm_to_norm}, $\snorm{\model'(\pair) - \model(\pair)} = \norm{\model'(\pair) - \model(\pair)}$ if $\norm{\model'(\pair) - \model(\pair)} < \dmin(\model)$. 
    Therefore, the event $\event_t$ holds if for all $\pair \in \skeleton_t$, we have
    \begin{equation}
    \label{equation_coexploration_skeleton_1}
        \norm{\hat{\model}_t(\pair) - \model(\pair)} < \frac{\epsilonflat(t)}{\kappa(\model)}
        \quad\text{and}\quad
        \epsilonflat(t) < \min \braces*{
            \frac{\gaingap(\model)}{3}, 
            \kappa(\model)
            \dmin(\model)
        }
        .
    \end{equation}
    By \Cref{assumption_regularization_hyperparameter} (\strong{A0}), we have $\epsilonflat(t) \to 0$ as $t \to \infty$, so the second requirement of \eqref{equation_coexploration_skeleton_1} is satisfied as soon as $t$ is large enough, say $t \ge \kappa_\EPSILON$ for some $\kappa_\EPSILON \in \NN$.
    So,
    \begin{align*}
        \EE \brackets*{
            \sum_{t=1}^\infty
            \indicator{\event_t^c}
        }
        & \overset{(\dagger)}\le
        \kappa_\EPSILON
        + \sum_{t=\kappa_\EPSILON}^\infty
        \Pr \parens*{
            \exists \pair \in \skeleton_t,
            \norm{\hat{\model}_t(\pair) - \model(\pair)}
            \le \frac{\epsilonflat(t)}{\kappa(\model)}
        }
        \\
        & \overset{(\ddagger)}=
        \kappa_\EPSILON
        + \sum_{t=\kappa_\EPSILON}^\infty
        \Pr \parens*{
            \exists \pair \in \pairs,
            \norm{\hat{\model}_t(\pair) - \model(\pair)}
            \le \frac{\epsilonflat(t)}{\kappa(\model)}
            \text{~and~}
            \visits_t(\pair) \ge \log^2(t)
        }
        \\
        & \overset{(\S)}\le
        \kappa_\EPSILON
        + 
        \abs{\pairs}
        \sum_{t = \kappa_\EPSILON}^\infty
        \exp \parens*{
            - \frac{2 \log^2(t) \epsilonflat(t)^2}{\kappa(\model)^2}
            + \frac 12 \log\parens*{1 + \log^2(t)}
            + \log\parens*{2^{\abs{\states}+1}}
        }
        \\
        & \overset{(\$)}\le
        \kappa_\EPSILON
        + 
        \abs{\pairs}
        \sum_{t = \kappa_\EPSILON}^\infty
        \exp \parens*{
            - \log^{\frac 32}(t)
            + \oh\parens*{
                \log^{\frac 32}(t)
            }
        }
        = \OH(1)
    \end{align*}
    where
    $(\dagger)$ follows by definition of $\kappa_\EPSILON$;
    $(\ddagger)$ follows by definition of the skeleton $\skeleton_t := \braces{\pair \in \pairs: \visits_t(\pair) \ge \log^2(t)}$;
    $(\S)$ follows from concentration results (\Cref{corollary_threshold_concentration}); and
    $(\$)$ follows from \Cref{assumption_regularization_hyperparameter} (\strong{A2}), that guarantees that $2 \log^2(t) \epsilonflat(t)^2 \kappa(\model)^{-2} \ge \log^{\frac 32}(t)$ for $t$ large enough. 
    This concludes the proof.
\end{proof}

\subsubsection{Proof of \Cref{lemma_wrong_coexploration}: Putting everything together}

We finally prove \Cref{lemma_wrong_coexploration}.
We have
\begin{align*}
    & \EE \brackets*{
        \sum_{t=1}^{T-1}
        \indicator{\Pair_t = \pair_0, t \in \stimes^+}
    }
    \\
    & \overset{(\dagger)}=
    \OH \parens*{
        \EE \brackets*{
            \sum_{t=1}^{T-1}
            \eqindicator{
                \pair_0 \in \flatme{\optpairs}{\iepsilonflat}(\hat{\model}_t)
                \atop
                t \in \stimes_0^+
            }
        }
    }
    \\
    & \overset{(\ddagger)}\le
    \OH \parens*{
        \EE \brackets*{
            \sum_{t=1}^{T-1}
            \eqindicator{
                \pair_0 \in \flatme{\optpairs}{\iepsilonflat}(\hat{\model}_t)
                \atop
                t \in \stimes_0^+,
                \event_t, \event'_t
            }
        }
        + \EE \brackets*{
            \sum_{t=1}^{T-1}
            \eqindicator{
                \pair_0 \in \flatme{\optpairs}{\iepsilonflat}(\hat{\model}_t)
                \atop
                t \in \stimes_0^+,
                (\event'_t)^c
            }
        }
        + \EE \brackets*{
            \sum_{t=1}^{T-1}
            \indicator{\event_t^c}
        }
    }
    \\
    & \overset{(\S)}=
    \OH \parens*{
        \sum_{t=1}^{T-1}
        \parens*{\frac 1t}^{1 + \frac 12 \epsilontest(t)}
    }
    + \OH \parens*{
        \sum_{\structure_0}
        \EE \brackets*{
            \sum_{t=1}^{T-1}
            \eqindicator{
                \pair_0 \in \flatme{\optpairs}{\iepsilonflat}(\hat{\model}_t)
                \atop
                t \in \stimes_0^+(\structure_0),
                (\event'_t)^c
            }
        }
    }
    + \OH(1)
    \\
    & \overset{(\$)}=
    \OH \parens*{
        \sum_{t=1}^{T-1}
        \parens*{\frac 1t}^{1 + \frac 12 \epsilontest(t)}
    }
    + \OH \parens*{
        \frac 1{\epsilonflat(T)^6}
    }
    + \OH(1)
\end{align*}
where 
$(\dagger)$ follows from \Cref{lemma_wrong_coexploration_initial};
$(\ddagger)$ invokes the decomposition from \eqref{equation_coexploration_visits_decomposition};
$(\S)$ bounds the first term using \Cref{lemma_wrong_coexploration_sanov}, the second is decomposed along the finitely many possible co-exploration structures $\structure_0$ and the third using \Cref{lemma_wrong_coexploration_skeleton}; and
$(\$)$ bounds the last term using \Cref{lemma_wrong_coexploration_correct_structure} and \Cref{lemma_wrong_coexploration_wrong_structure} depending on whether $\structure_0$ is closed or not. 
\qed

\subsection{Amount of exploration}
\label{appendix_exploration}

Let $\imeasure^*$ be the central optimal exploration probability measure of $\model$ (\Cref{definition_central_exploration_measure}), that is the limit point of optimal $\EPSILON$-regularized exploration measures $\imeasure^*_\EPSILON$ of $\model$ as $\EPSILON \to 0$ in the sense of \Cref{proposition_convergence_optimal_measure}.
Recall that $\ivalue(\imeasure^*, \model)$ denotes the information value of $\imeasure^*$ in $\model$ (\Cref{definition_information_value}).
The whole section is dedicated to the proof \Cref{lemma_exploration} below, that bounds the number of times a pair is being pulled during exploration. 

\begin{lemma}
\label{lemma_exploration}
    Let $\pair \in \pairs$ be an arbitrary state-action pair.
    As $T \to \infty$, we have
    \begin{equation}
        \EE \brackets*{
             \sum_{t=1}^{T-1} 
             \indicator{
                 \Pair_t = \pair,
                 t \in \stimes^-
             }
        }
        \le
        \frac{\imeasure^*(\pair) \log(T)}{\ivalue(\imeasure^*, \model)}
        + \oh(\log(T))
        .
    \end{equation}
\end{lemma}

\subsubsection{Idea of the proof}
\label{appendix_exploration_idea}

\begin{figure}[ht]
    \centering
    \resizebox{\linewidth}{!}{
    \begin{tikzpicture}
        \draw[line width=1] (0,0) to (7.65, 0); 
        \draw[line width=1, densely dotted] (0,0) to (11.5, 0); 
        \draw[line width=1, ->, >=stealth, dotted] (11.5, 0) to node[anchor=west, pos=1.0] {$\abs{\stimes^-}$} (14, 0); 

        \draw[line width=1] (0, -0.1) node[anchor=north, minimum height=0.8cm]{$0$} to (0, 0.1);
        \draw[line width=1] (2.25, -0.1) node[anchor=north, minimum height=1.2cm] {$\displaystyle \epsilon_0 \log(T) \atop \displaystyle \epsilon_0 \ll 1$} to (2.25, 0.1);
        \draw[line width=1] (7.25, -0.1) node[anchor=north, minimum height=1.1cm](optvalue){$\ivalue(\imeasure^*, \model)^{-1} \log(T)$}  to (7.25, 0.1);
        \draw[line width=1] (7.65, -0.1) to (7.65, 0.1);
        \draw[line width=1] (11.5, -0.1) node[anchor=north, minimum height=1.1cm]{$\log^3(T)$} to (11.5, 0.1);

        \fill[rounded corners, color=Crimson, opacity=0.33] (0.15, -.25) rectangle (2, 0.25);
        \node[color=Crimson,anchor=south] at (1.075, 0.25) {\sc Burn-In};

        \fill[rounded corners, color=DodgerBlue, opacity=0.33] (2.15, -.25) rectangle (7.75, 0.25);
        \node[color=DodgerBlue,anchor=south] at (4.75, 0.25) {\sc Near Optimal Exploration};
        \node[color=DodgerBlue,anchor=south] (cet) at (7.45, 0.75) {\sc Co-Exploration Travels};
        \draw[color=DodgerBlue, ->, >=stealth, line width=1] (cet) to (7.45, 0.1);
        \node[color=DodgerBlue] (optv) at (4.75, -1.25) {Most likely value};
        \draw[color=DodgerBlue,->,>=stealth,line width=1] (optv) to[in=180,out=0] (7.25, -1.25) to (7.25, -0.9);

        \fill[rounded corners, color=PineGreen,, opacity=0.33] (7.85, -.25) rectangle (13.75, 0.25);
        \node[color=PineGreen, anchor=south] at (9.575, 0.25) {\sc 1st failure};
        \node[color=PineGreen] (failp) at (9.575, -1.25) {$\Pr = \oh\parens*{e^{-\sqrt{\log(T)}}}$};
        \draw[color=PineGreen, ->,>=stealth, line width=1] (failp) to (9.575, -.1);

        \node[color=PineGreen, anchor=south] at (12.75, 0.25) {\sc 2nd failure};
        \node[color=PineGreen] (bigfailp) at (12.75, -1.25) {$\Pr = \oh\parens*{T^{-2}}$};
        \draw[color=PineGreen, ->,>=stealth, line width=1] (bigfailp) to (12.75, -.1);
    \end{tikzpicture}
}
    \caption{
    \label{figure_exploration}
        Global overview of the different exploration periods as $\abs{\stimes^-}$ increases.
    }
\end{figure}
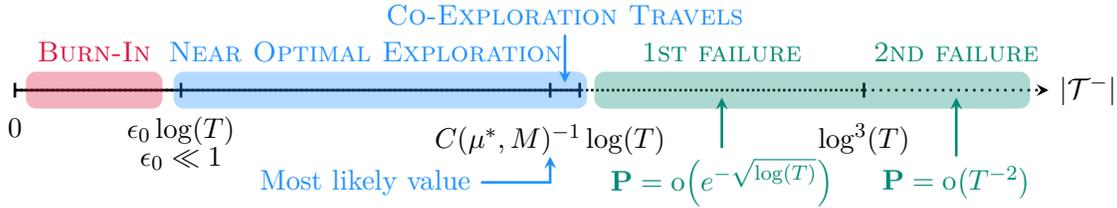

In \Cref{figure_exploration}, we provide a high level representation of how exploration is decomposed and analyzed. 
Overall, the set of exploration times $\braces{t \in \stimes^-: t \le T}$ is split into five periods: The ``burn-in'' period, the dominant ``near-optimal exploration'' period, the ``co-exploration travels'' period, a ``first order failure'' period and a ``second order failure'' period.

We fix $\epsilon_0 > 0$ a precision parameter.
Then, we begin by discarding what happens prior to time $\approx \epsilon_0 \log(T)$, during the \textcolor{Crimson}{\sc Burn-In} period. 
Past the burn-in period, we show that the empirical data is quite likely to concentrate well-enough so that the algorithm approximates the exploration measure and information values correctly. 
This phenomenon is called the \strong{trigger effect} (\Cref{appendix_trigger_effect}).
It is made possible by the uniform exploration guarantees of \texttt{ECoE*} allowed by the $\epsilonunif$-regularization of its exploration policies, that makes sure that all pairs are covered uniformly fast during exploration (\Cref{appendix_uniform_exploration}).
Following the burn-in period, \texttt{ECoE*} enters the \textcolor{DodgerBlue}{\sc Near Optimal Exploration} period during which it explores near-optimally (\Cref{appendix_near_optimal_exploration}).
During this period and because of the trigger effect, \texttt{ECoE*} further estimates the information values well-enough so that it does not overshot the duration of exploration.
After about $\ivalue(\imeasure^*, \model)^{-1} \log(T)$ exploration times, GLR exploration tests \eqref{equation_glr_exploration} do not provoke exploration phases anymore and \texttt{ECoE*} enters the \textcolor{DodgerBlue}{\sc Co-Exploration travels} period.
During this period, exploration phases can only be triggered by the co-exploration rule, for a total number of exploration phases that is negligible in front of $\log(T)$ (\Cref{appendix_coexploration_travels}).
Combining everything so far, we deduce that the number of exploration phases is quite likely to remain below $\ivalue(\imeasure^*, \model)^{-1} \log(T)$ (\Cref{appendix_weak_exploration_barrier,appendix_coexploration_travels}).

If $\abs{\stimes^-}$ exceeds $\ivalue(\imeasure^*, \model)^{-1} \log(T)$ significatively, it essentially means that the trigger effect failed to trigger after the burn-in period.
This is the \textcolor{PineGreen}{\sc First Order Failure} period, that \texttt{ECoE*} only enters with probability $\oh(\exp(-\sqrt{\log(T)}))$.
We show that at horizon $\OH(\log^3(T))$ and thanks to uniform exploration again, all pairs enter the skeleton $\braces{\pair \in \pairs: \visits_t (\pair) \ge \log^2(t)}$ and \texttt{ECoE*} naturally stops to explore by design. 
Because $\log^3(T) \exp(-\sqrt{\log(T)}) \to 0$ when $T \to \infty$, the number of exploration times induced by the first order failure period is negligible in expectation (\Cref{appendix_strong_exploration_barrier}).
This is the \strong{skeleton barrier}.
Beyond that barrier, the algorithm only continues to explore if pairs failed to entered the skeleton.
This is the \textcolor{PineGreen}{\sc Second Order Failure} period, that has so little probability to happen that its is completely negligible in expectation. 

Everything is compiled in \Cref{appendix_exploration_together} for a proof of \Cref{lemma_exploration}.

\subsubsection{Preliminaries: Exploration visit counts and truncated regularization}
\label{appendix_exploration_notations}

The \strong{exploration visit count} of $\pair$ at time $t \ge 1$ is given by
\begin{equation}
\label{equation_exploration_visit_count}
    \visits_t^- (\pair)
    :=
    \sum_{i=1}^{t-1}
    \indicator{
        \Pair_i = \pair,
        i \in \stimes^-
    }
    .
\end{equation}
We write $(\tau_\ell)$ the stopping-time enumeration of exploration times $\stimes^-$, i.e., $\tau_1 := \inf \stimes^-$ and $\tau_{\ell+1} := \inf \braces{t > \tau_\ell: t \in \stimes^-}$. 
Note that $\visits_{\tau_\ell}^- (\pair) = \sum_{i=1}^{\ell-1} \indicator{\Pair_{\tau_i} = \pair}$ for $\ell \ge 1$.
Given $T \ge 1$, we denote $\EPSILON''$ the \strong{$T$-truncated} regularization parameter, given by
\begin{equation}
\label{equation_truncated_regularizer}
\begin{gathered}
    \epsilonflat''(t) := \max \braces*{
        \epsilonflat'(t), \epsilonflat'(T)
    },
    \qquad\qquad
    \epsilontest''(t) := \epsilontest'(t),
    \\
    \epsilonunif''(m) := \epsilonunif'(m),
    \qquad\qquad
    \epsilonreg''(m) := \epsilonunif'(m).
\end{gathered}
\end{equation}
That is, $\EPSILON''$ and $\EPSILON'$ only differ at $\epsilonflat$. 

\subsubsection{Part I: Uniform exploration guarantees} 
\label{appendix_uniform_exploration}

We begin this adventure by establishing the uniform exploration guarantees of \texttt{ECoE*}, that are crucial to prove lower bound the likeliness of the trigger effect in \Cref{appendix_trigger_effect} and control the skeleton barrier in \Cref{appendix_strong_exploration_barrier}.
In \Cref{lemma_uniform_exploration} below, we provide a high probability bound of the uniform visit rates of \texttt{ECoE*} during exploration, as a function of the uniformization regularizer $\epsilonunif$. 

\begin{lemma}
\label{lemma_uniform_exploration}
    Let $(\tau_\ell)$ be the stopping-time enumeration of $\stimes^-$, i.e., $\tau_1 := \inf \stimes^-$ and $\tau_{\ell+1} := \inf\braces{t > \tau_\ell : t \in \stimes^-}$.
    If the regularizer $\EPSILON$ satisfies \Cref{assumption_regularization_hyperparameter}, there exist constants $\alpha, \kappa_\EPSILON > 0$ and a polynomial $\mathrm{poly}(-)$ of degree $4 \abs{\states}$ such that
    \begin{equation*}
        \Pr \parens*{
            \exists \pair \in \pairs,
            \exists \ell \ge m,
            ~\visits_{\tau_\ell} (\pair) < \alpha \epsilonunif(\ell)^{\abs{\states}} \ell 
        }
        \le
        \exp \parens*{
            - \mathrm{poly}(\epsilonunif(m)) m
            + \kappa_\EPSILON
        }
        .
    \end{equation*}
\end{lemma}
\begin{proof}
    Fix $\ell \ge m$ and denote $\eta \equiv \eta(\ell) := \epsilonunif(\ell)$ for short. 
    We consider the Markov decision process $\model_\eta := (\pairs, \rewardd, \kernel_\eta)$ with kernel given by
    \begin{equation*}
        \kernel_\eta (\state, \action)
        :=
        \parens*{
            1 - \eta \abs{\actions(\state)}
        } \kernel(\state, \action)
        + \eta \sum_{\action' \in \actions(\state)} \kernel(\state, \action')
        .
    \end{equation*}
    The model $\model_\eta$ is communicating, because the execution of the fully uniform policy $\policy_u$ on $\model_\eta$ is equivalent to the execution of the uniform policy on $\model$, in the sense that $\EE^{\model, \policy_u}[-] = \EE^{\model_\eta, \policy_u}[-]$. 
    By construction, the execution of any $\eta$-uniform policy $\policy$ on $\model$ can be seen as the execution of a randomized policy $\policy_\eta$ on $\model_\eta$, with $\EE^{\policy_\eta, \model}[-] = \EE^{\policy, \model_\eta}[-]$. 

    Fix $\pair_0 \in \pairs$ and introduce the reward function $f(\pair) := \indicator{\pair = \pair_0}$.
    Consider the model $\model_\eta^{-f} = (\pairs, - f, \kernel_\eta)$ with kernel $\kernel_\eta$ and deterministic reward function $-f$. 
    Introduce $\gainof{-f}_\eta$, $\biasof{-f}_\eta$ and $\gapsof{-f}_\eta$ the associated optimal gain, bias an gap functions of $\model_\eta^{-f}$.
    Because $\model_\eta^{-f}$ is communicating, we have $\vecspan{\gainof{-f}_\eta} = 0$.
    Let $\policy^*_\eta$ be a deterministic bias optimal policy of $\model_\eta^{-f}$.
    By construction, $\policy^*_\eta$ corresponds to an $\eta$-uniform policy of $\model$ under which all pairs are recurrent.
    By \Cref{lemma_uniform_measures_support_diameter}, we have $\min(\gainof{-f}_\eta) = \max(\gainof{-f}_\eta) \le - \alpha(\ell) := - \dmin(\kernel)^{\abs{\states}} \eta^{\abs{\states}}$. 
    Consider the gaps in $\model$, given by
    \begin{equation*}
        \gapsof{-f}(\state, \action)
        :=
        \gainof{-f}_{\eta}(\state) + \biasof{-f}_\eta(\state)
        + f(\state, \action) - \kernel(\state, \action) \biasof{-f}_\eta.
    \end{equation*}
    Following from the Bellman equations of $\model_\eta^{-f}$, we check that every $\eta$-uniform policy $\policy$ satisfies
    \begin{equation}
    \label{equation_uniform_exploration_1}
        \sum_{\action \in \actions(\state)}
        \policy_\eta (\action|\state)
        \gapsof{-f}_\eta(\state, \action)
        =
        \sum_{\action \in \actions(\state)}
        \policy(\action|\state) 
        \gapsof{-f}(\state, \action)
        \ge
        0
    \end{equation}
    where $\policy_\eta$ is the randomized policy induced by $\policy$ in $\model_\eta$.
    Introduce the function $\beta(\ell) := 1 + 2 \abs{\states} \dmin(\kernel)^{- \abs{\states}} \epsilonunif(\ell)^{-\abs{\states}}$.
    By \Cref{lemma_uniform_measures_support_diameter}, we have $\beta(\ell) \ge \max\braces{\vecspan{\biasof{-f}_\eta}, \norm{\gapsof{-f}}_\infty}$.
    We have
    \begin{align*}
        \visits_{\tau_\ell} (\pair_0)
        =
        \sum_{t=1}^{\tau_\ell-1}
        \indicator{\Pair_t = \pair_0}
        & \ge
        \sum_{i=1}^{\ell-1}
        \indicator{\Pair_{\tau_i} = \pair_0}
        \\ 
        & \overset{(\dagger)}=
        \sum_{i=1}^{\ell-1}
        \parens*{
            - \gainof{-f}_\eta(\State_{\tau_i})
            + \parens*{
                \kernel(\Pair_{\tau_i}) - \unit_{\State_{\tau_i}}
            } \biasof{-f}_\eta 
            + \gapsof{-f}(\Pair_{\tau_i})
        }
        \\
        & =
        (\ell - 1) \alpha(\ell) 
        + \underbrace{
            \sum_{i=1}^{\ell-1}
            \parens*{
                \kernel(\Pair_{\tau_i}) - \unit_{\State_{\tau_i}}
            } \biasof{-f}_\eta 
        }_{\mathrm{A}_1}
        + \underbrace{
            \sum_{i=1}^{\ell-1}
            \gapsof{-f}(\Pair_{\tau_i})
        }_{\mathrm{A}_2}
    \end{align*}
    where $(\dagger)$ unfolds the definition of $\gapsof{-f}$.
    We control the error terms as follows, starting with $\mathrm{A}_1$.
    We have
    \begin{align*}
        \mathrm{A}_1
        & =
        \sum_{i=1}^{\ell-1} \parens*{
            \biasof{-f}_\eta (\State_{\tau_i+1})
            - \biasof{-f}_\eta (\State_{\tau_i})
        }
        + \sum_{i=1}^{\ell-1} \parens*{
            \kernel(\Pair_{\tau_i}) - \unit_{\State_{\tau_i+1}}
        } \biasof{-f}_\eta
        \\
        & \overset{(\dagger)}\ge
        - \abs{\pairs} \abs{\states} \beta(\ell)
        + \sum_{i=1}^{\ell-1} \parens*{
            \kernel(\Pair_{\tau_i}) - \unit_{\State_{\tau_i+1}}
        } \biasof{-f}_\eta
    \end{align*}
    where 
    $(\dagger)$ follows from the observation that, between exploration times, the algorithm is exploiting.
    When it does, it does so until regeneration, hence coming back to the state where exploration was suspended; At the exception of phases killed by panicking, but there are at most $\abs{\pairs} \abs{\states}$ of them---one for each transition to be discovered, see \Cref{lemma_panic_times}.
    The remaining right-hand term is a martingale that we leave for later.

    We continue with $\mathrm{A}_2$.
    We have
    \begin{align*}
        \mathrm{A}_2
        & := 
        \sum_{i=1}^{\ell-1} 
        \gapsof{-f}(\Pair_{\tau_i})
        \\
        & =
        \sum_{i=1}^{\ell-1} 
        \sum_{\action \in \actions(\State_{\tau_i})}
        \policy^-_{\tau_i} (\action|\State_{\tau_i})
        \gapsof{-f}(\State_{\tau_i}, \action)
        +
        \sum_{i=1}^{\ell-1} 
        \sum_{\action \in \actions(\State_{\tau_i})}
        \parens*{
            \indicator{\Action_{\tau_i} = \action}
            - \policy^-_{\tau_i} (\action|\State_{\tau_i})
        }
        \gapsof{-f}(\State_{\tau_i}, \action)
        \\
        & \overset{(\dagger)}\ge
        \sum_{i=1}^{\ell-1} 
        \sum_{\action \in \actions(\State_{\tau_i})}
        \parens*{
            \indicator{\Action_{\tau_i} = \action}
            - \policy^-_{\tau_i} (\action|\State_{\tau_i})
        }
        \gapsof{-f}(\State_{\tau_i}, \action)
    \end{align*}
    where 
    $(\dagger)$ follows from the observation that $\policy_{\tau_i}^-$ is $\eta$-randomized (since $i \le \ell$), that $\epsilonunif$ is non-increasing by \Cref{assumption_regularization_hyperparameter} (\strong{A1}) combined with \eqref{equation_uniform_exploration_1}.
    We conclude that $\mathrm{A}_1$ and $\mathrm{A}_2$ both involve martingales obtained as the sum of $\ell - 1$ martingale differences whose terms have span at most $\beta(\ell)$.
    So, by a time-uniform Azuma-Hoeffing's inequality (\Cref{lemma_azuma_time_uniform}), we find that with probability $1 - \delta$ and uniformly for $\ell \ge m$, we have
    \begin{equation}
        \visits_{\tau_\ell} (\pair_0)
        \ge
        \alpha (\ell - 1)
        - \beta(\ell) \parens*{
            \abs{\pairs} \abs{\states}
            + 2 \sqrt{
                (\ell - 1) \log \parens*{\frac {1 + \ell}\delta}
            }
        }
        .
    \end{equation}
    We conclude by solving $m \alpha(m) > 4 \beta(m) \sqrt{m \log((1+m)/\delta)}$ in $\delta$. 
    We decouple the inequality by simplifying it to the pair of conditions (C1) $m \alpha(m) > 4 \beta(m) \sqrt{m \log(1+m)}$ and (C2) $m \alpha(m) > 4 \beta(m) \sqrt{m \log(1/\delta)}$.
    We know that $\epsilonunif(\ell) = \omega(\log(\ell)^{-1})$ by assumption, so that the condition (C1) is satisfied for $m$ large enough, say $m \ge C_\EPSILON$.
    As for the condition (C2), solving in $\delta$ provides the condition
    \begin{equation*}
        \delta \ge \exp \parens*{
            - \frac{\alpha(m)^2 m}{\beta(m)^2}
        }
        .
    \end{equation*}
    By assumption on $\epsilonunif$ again, see \Cref{assumption_regularization_hyperparameter} (\strong{A6}), the function $m \mapsto \alpha(m)^2 \beta(m)^{-2} m$ is increasing provided that $m$ is large enough, so provided that $m \ge C_\EPSILON$ up to increasing $C_\EPSILON$.
    In the end, we set 
    \begin{equation*}
        \delta := \exp \parens*{
            - \frac{\alpha(m)^2 m}{\beta(m)^2}
            + \indicator{m < C_\epsilon}
        }
        \le
        \exp \parens*{
            - \mathrm{poly}(\epsilonunif(m)) m
            + C_\EPSILON
        }
    \end{equation*}
    where $\mathrm{poly}(-)$ is of degree $4 \abs{\states}$.
\end{proof}

\subsubsection{Part II: Post burn-in properties \& the trigger effect}
\label{appendix_trigger_effect}

In this section, we show that after a few exploration phases called the \strong{burn-in} exploration phases, empirical data starts to concentrate in probability so that the algorithm explores near optimally: The exploration policy is near optimal and GLR exploration tests do not overshoot the required amount of information.

We refer this phenomenon to as the \strong{trigger effect}.

Consider the function $\beta(\ell) := 1 + 2 \abs{\states} \dmin(\kernel)^{-\abs{\states}} \epsilonunif'(\ell)^{- \abs{\states}}$ and let $\delta(\ell) := \alpha \epsilonunif(\ell)^{\abs{\states}}$ where $\alpha > 0$ is an alias for constant given by \Cref{lemma_uniform_exploration}.
Recall that $\probabilities_{\delta(\ell)}(\pairs)$ is the set of probability distributions $\imeasure$ over $\pairs$ such that $\min(\imeasure) \ge \delta(\ell)$.
Given a burn-in threshold $\epsilon_0 > 0$, we introduce the \strong{trigger event} as
\begin{equation}
\label{equation_definition_trigger_event}
    \event 
    \equiv \event_{\epsilon_0, T}
    := 
    \parens*{
        \forall \ell \ge \frac{\epsilon_0 \log(T)}{\beta(\epsilon_0 \log(T))}  
        :
        \hspace{-.5em}
        \begin{array}{c}
            \flatme{\optpairs}{\iepsilonflat''(\tau_\ell)}(\hat{\model}_{\tau_\ell}) = \optpairs(\model)
            \mathrm{~and~}
            \hat{\model}_{\tau_\ell} \sim \model
            \mathrm{~and~}
            \\
            \norm{\imeasure_{\EPSILON''}^*(\hat{\model}_{\tau_\ell}) - \imeasure^*_{\EPSILON''}(\model)}_\infty 
            < 
            \epsilon_0 \beta(\ell)^{-2}
            \mathrm{~and~}
            \\
            \sup_{\imeasure \in \probabilities_{\delta(\ell)}(\pairs)} \abs*{
                \ivalue_{\EPSILON''}(\imeasure, \hat{\model}_{\tau_\ell}) 
                - \ivalue_{\EPSILON''}(\imeasure, \model)
            } < \epsilon_0
        \end{array}
        \hspace{-.5em}
    }
\end{equation}
where $\EPSILON''$ is the $T$-truncated regularization hyperparameter, see \Cref{appendix_exploration_notations} and \eqref{equation_truncated_regularizer}.
It is to be noted that for $\epsilonunif(\ell) = \omega(\log^{-1}(\ell))$, we have $\beta(\ell) \ge 1$ and $\beta(\ell) = \oh(\log(\ell)^{\abs{\states}})$ when $\ell \to \infty$, so that $\beta(\ell)$ is negligible in front of any positive power of $\ell$. 

In \Cref{lemma_trigger_effect} below, we bound the probability of error of the trigger event $\event$ when the exploration threshold $\epsilon_0 \log(T)/\beta(\epsilon_0 \log(T))$ is replaced by a general $m / \beta(m)$ for $m \ge 1$, see \eqref{equation_trigger_effect_bis} below. 
For $m \ge 1$, let
\begin{equation}
\label{equation_trigger_effect_bis}
    \event_{m} \equiv \event_{\epsilon_0, T, m}
    := 
    \parens*{
        \forall \ell \ge \frac{m}{\beta(m)}
        :
        \hspace{-.5em}
        \begin{array}{c}
            \flatme{\optpairs}{\iepsilonflat''(\tau_\ell)}(\hat{\model}_{\tau_\ell}) = \optpairs(\model)
            \mathrm{~and~}
            \hat{\model}_{\tau_\ell} \sim \model
            \mathrm{~and~}
            \\
            \norm{\imeasure_{\EPSILON''}^*(\hat{\model}_{\tau_\ell}) - \imeasure^*_{\EPSILON''}(\model)}_\infty 
            < 
            \epsilon_0 \beta(\ell)^{-2}
            \mathrm{~and~}
            \\
            \sup_{\imeasure \in \probabilities_{\delta(\ell)}(\pairs)} \abs*{
                \ivalue_{\EPSILON''}(\imeasure, \hat{\model}_{\tau_\ell}) 
                - \ivalue_{\EPSILON''}(\imeasure, \model)
            } < \epsilon_0
        \end{array}
        \hspace{-.5em}
    }
    .
\end{equation}
In \Cref{corollary_trigger_effect}, we instantiate $m = \epsilon_0 \log(T)$ to obtain a final bound on $\Pr(\event^c)$.

\begin{lemma}
\label{lemma_trigger_effect}
    Let $(\tau_\ell)$ be the stopping-time enumeration of $\stimes^-$, i.e., $\tau_1 := \inf \stimes^-$ and $\tau_{\ell+1} := \inf \braces{t > \tau_\ell : t \in \stimes^-}$.
    For all error level $\epsilon_0 > 0$, there exists a constant $\kappa_\EPSILON > 0$ such that for all $T \ge \kappa_\EPSILON$, the event $\event_m$ as given by \eqref{equation_trigger_effect_bis} satisfies
    \begin{equation*}
        \Pr(\event_m^c) 
        \le 
        \exp \parens*{
            -\mathrm{poly}(\EPSILON'') 
            \exp\parens*{
                - \frac{\alpha}{\epsilonunif(m)^{\abs{\states}}}
            } m 
            + \kappa_\EPSILON
        }
    \end{equation*}
    for all $m \ge 1$, where $\mathrm{poly}(\EPSILON'') := \alpha' \epsilonflat(T)^2 \epsilonunif(m)^{11\abs{\states}} \epsilonreg(m)^2$ is polynomial and $\alpha, \alpha' \in \RR_+^*$ are constants that depend on $\model$. 
\end{lemma}
\begin{proof}
    Introduce the event $\event'_m := (\forall \pair \in \pairs, \forall \ell \ge m, \visits_{\tau_\ell}(\pair) \ge \alpha_1 \epsilonunif(\ell)^{\abs{\states}} \ell)$ where $\alpha_1 > 0$ is a constant given by \Cref{lemma_uniform_exploration}.
    We have
    \begin{align*}
        \Pr \parens*{
            \event_m^c 
        }
        \le
        \Pr \parens*{
            \event_m^c \cap \event_m'
        }
        + \Pr \parens*{
            (\event_m')^c
        }
        & \overset{(\dagger)}\le
        \Pr \parens*{
            \event_m^c \cap \event_m'
        }
        + \exp \parens*{
            - \mathrm{poly}(\epsilonunif(m)) m
            + \kappa_\EPSILON
        }
    \end{align*}
    where $(\dagger)$ follows by \Cref{lemma_uniform_exploration} again, and $\mathrm{poly}(-)$ is a polynomial of degree $4 \abs{\states}$ and $\kappa_\EPSILON < \infty$ is a constant.

    Now, we look at conditions on $\norm{\hat{\model}_{\tau_\ell} - \model}$ and $\EPSILON'$ under which the conditions of $\event_m$ are met, then we will conclude by combining $\event_m'$ with concentration bounds (\Cref{lemma_threshold_concentration}).
    First, by \Cref{lemma_convert_snorm_to_norm}, the absolute continuity condition ``$\hat{\model}_{\tau_\ell} \sim \model$'', which is equivalent to ``$\snorm{\hat{\model}_{\tau_\ell} - \model} = \norm{\hat{\model}_{\tau_\ell} - \model}$'', is met as soon as 
    \begin{equation}
    \label{equation_trigger_effect_1}
        \norm{\hat{\model}_{\tau_\ell} - \model} 
        < 
        \dmin(\model)
        .
    \end{equation}
    By \Cref{proposition_continuity_near_optimality}, the condition ``$\flatme{\optpairs}{\iepsilonflat'}(\hat{\model}_{\tau_\ell}) = \optpairs(\model)$'' is met as soon as
    \begin{equation}
    \label{equation_trigger_effect_2}
        \alpha_2 \norm{\hat{\model}_{\tau_\ell} - \model} 
        <
        \frac 13 \epsilonflat''(\tau_\ell)
        \quad \text{and} \quad
        \epsilonflat''(\tau_\ell)
        \le
        \frac 13 \epsilonflat''(\ell)
        < 
        \gaingap(\model)
    \end{equation}
    for some model dependent constant $\alpha_2 > 0$.
    Note that the second condition of \eqref{equation_trigger_effect_2} is satisfied as soon as $\ell, T$ are large enough, say $T, \ell \ge \kappa_\EPSILON$ up to increasing $\kappa_\EPSILON$.

    For the condition ``$\norm{\imeasure^*_{\EPSILON''}(\hat{\model}_{\tau_\ell}) - \imeasure^*_{\EPSILON''}(\model)}_\infty < \epsilon_0 \beta(\ell)^{-2}$,'' we rely on \Cref{proposition_continuity_regularized_lower_bound}.
    Because tht error function $\psi_\imeasure$ of \Cref{proposition_continuity_regularized_lower_bound} satisfies $\psi_\imeasure (x) = \OH(x)$ when $x \to 0$, there exists constants $\alpha_3, \alpha_4 \in \RR_+^*$ such that $\psi_\imeasure (x) \le \alpha_3 x$ for $x \le \alpha_4$. 
    Following \Cref{proposition_continuity_regularized_lower_bound}, we get the condition
    \begin{equation}
    \label{equation_trigger_effect_3}
        \sqrt{
            \frac{\norm{\hat{\model}_{\tau_\ell} - \model}}{\epsilonunif'(\ell)^{3 \abs{\states}} \epsilonreg'(\ell)}
        }
        + 
        \eqindicator{
            \norm{\hat{\model}_{\tau_\ell} - \model}
            < 
            \alpha_5 \min \braces*{
                \epsilonflat''(\ell),
                \exp \parens*{
                    - \frac{\alpha_6}{\epsilonunif'(\ell)^{\abs{\states}}}
                }   
            }
        }
        \le 
        \frac{\epsilon_0}{\alpha_3 \beta(\ell)^2}
    \end{equation}
    where $\alpha_5, \alpha_6$ are other model dependent constants. 
    Note that the indicator in \eqref{equation_trigger_effect_3} can be rewritten as the pair of conditions
    \begin{equation}
    \label{equation_trigger_effect_4}
        \norm{\hat{\model}_{\tau_\ell} - \model} 
        \le
        \alpha_5 \epsilonflat''(\tau_\ell) 
        \quad\text{and}\quad
        \norm{\hat{\model}_{\tau_\ell} - \model}
        \le
        \alpha_5 \exp \parens*{
            - \frac{\alpha_6}{\epsilonunif'(\ell)^{\abs{\states}}}
        }
        .
    \end{equation}
    Note that the first condition of \eqref{equation_trigger_effect_4} is essentially the same as the first condition of \eqref{equation_trigger_effect_2} up to tuning constants.

    To obtain the last condition ``$\sup_{\imeasure \in \probabilities_{\delta(\ell)}(\pairs)} \abs{\ivalue_{\EPSILON''}(\imeasure, \hat{\model}_{\tau_\ell}) - \ivalue_{\EPSILON''}(\imeasure, \model)} < \epsilon_0$'', we rely on \Cref{lemma_information_value_continuous_model}.
    Using that, for all $\alpha > 0$, $\epsilonunif(m)^{-\alpha} = \oh(\exp(1/\epsilonunif(m)))$ when $m \to \infty$ and that $\delta(\ell) = \Omega(\epsilonunif'(\ell)^{\abs{\states}})$, we obtain a pair of conditions where the first one is similar to \eqref{equation_trigger_effect_4} and the second is 
    \begin{equation}
    \label{equation_trigger_effect_5}
        \norm{\hat{\model}_{\tau_\ell} - \model} < \alpha_7 \epsilon_0.
    \end{equation}
    Since $\epsilonunif'(m) \to 0$ as $m \to \infty$ by \Cref{assumption_regularization_hyperparameter} (\strong{A0}), the condition \eqref{equation_trigger_effect_5} is already satisfied through \eqref{equation_trigger_effect_3} provided that $\ell$ is large enough, say $\ell > \kappa_\EPSILON$ up to increasing $\kappa_\EPSILON$. 
    Using that $\beta(\ell) = \OH(\epsilonunif'(\ell)^{-\abs{\states}})$ and that $\min \braces{a_1, \ldots, a_n} \ge \product_{i=1}^n a_i$ for $a_i \in [0, 1]$, we combine \eqref{equation_trigger_effect_1} to \eqref{equation_trigger_effect_5} into the single condition
    \begin{equation*}
        \norm{\hat{\model}_{\tau_\ell} - \model}
        \le 
        \alpha_8 
        \epsilonflat''(\tau_\ell)
        \epsilonunif'(\ell)^{7 \abs{\states}} 
        \epsilonreg'(\ell)
        \exp \parens*{
            - \frac{\alpha_6}{\epsilonunif'(\ell)^{\abs{\states}}}
        }
        \quad\text{and}\quad
        \ell \ge \beta_{\epsilon}
    \end{equation*}
    where $\alpha_8$ is some model dependent constant.
    Using that for all $\ell \ge 1$, we have $\epsilonflat''(\tau_\ell) \ge \epsilonflat(T)$, $\epsilonunif'(\ell) \ge \epsilonunif(\ell)$ and $\epsilonreg'(\ell) \ge \epsilonreg(\ell)$, we simplify the above into
    \begin{equation}
    \label{equation_trigger_effect_6}
        \norm{\hat{\model}_{\tau_\ell} - \model}
        \le 
        \underbrace{
            \alpha_8 
            \epsilonflat(T)
            \epsilonunif(\ell)^{7 \abs{\states}} 
            \epsilonreg(\ell)
        }_{\varphi(\EPSILON(\ell))}
        \exp \parens*{
            - \frac{\alpha_6}{\epsilonunif(\ell)^{\abs{\states}}}
        }
        \quad \text{and} \quad
        \ell \ge \beta_{\epsilon}
        .
    \end{equation}
    We conclude using concentration results (\Cref{lemma_threshold_concentration}).
    For $m \ge \kappa_\EPSILON$, we have
    \begin{align*}
        & \Pr(\event_m^c \cap \event_m')
        \\
        & \overset{(\dagger)}\le 
        \Pr \parens*{
            \parens*{
                \exists \ell \ge m,
                \norm{\hat{\model}_{\tau_\ell} - \model}
                \ge
                \varphi(\EPSILON(\ell)) \exp\parens*{
                    - \frac{\alpha_6}{\epsilonunif(\ell)^{\abs{\states}}}
                }
            }
            ,
            \event_m',
        }
        \\
        & =
        \Pr \parens*{
            \parens*{
                \exists \ell \ge m,
                \exists \pair \in \pairs,
                \norm{(\hat{\reward}_{\tau_\ell}(\pair), \hat{\kernel}_{\tau_\ell}(\pair)) - (\reward(\pair), \kernel(\pair))}_1
                \ge
                \varphi(\EPSILON(\ell)) \exp\parens*{
                    - \frac{\alpha_6}{\epsilonunif(\ell)^{\abs{\states}}}
                }
            },
            \event'_m
        }
        \\
        & \overset{(\ddagger)}\le
        \sup_{\ell \ge m}
        \exp \parens*{
            - 2 
            \varphi(\EPSILON(\ell))^2 
            \alpha_1 \epsilonunif(\ell)^{\abs{\states}}
            \exp \parens*{
                - \tfrac{2 \alpha_6}{\iepsilonunif(\ell)^{\abs{\states}}}
            } \ell
            + \tfrac 12 \log \parens*{
                1 + \alpha_1 \epsilonunif(\ell)^{2 \abs{\states}} \ell
            }
            + \alpha_8
        }
        \\
        & \overset{(\S)}=
        \exp \parens*{
            - 2 
            \varphi(\EPSILON(m))^2 
            \alpha_1 \epsilonunif(m)^{\abs{\states}}
            \exp \parens*{
                - \tfrac{2 \alpha_6}{\iepsilonunif(m)^{\abs{\states}}}
            } m
            + \tfrac 12 \log \parens*{
                1 + \alpha_1 \epsilonunif(m)^{2 \abs{\states}} m
            }
            + \alpha_8
        }
        \\
        & \overset{(\$)}\le
        \exp \parens*{
            - \varphi(\EPSILON(m))^2 
            \alpha_1 \epsilonunif(m)^{\abs{\states}}
            \exp \parens*{
                - \tfrac{2 \alpha_6}{\iepsilonunif(m)^{\abs{\states}}}
            } m
            + \alpha_8
        }
    \end{align*}
    where 
    $(\dagger)$ follows by \eqref{equation_trigger_effect_6};
    $(\ddagger)$ follows from concentration results (\Cref{lemma_threshold_concentration}); 
    $(\S)$ uses that the inner function in the exponential is decreasing for $\ell$ large enough by \Cref{assumption_regularization_hyperparameter} (\strong{A6}), so holds for $m \ge \kappa_\EPSILON$ up to increasing $\kappa_\EPSILON$; and
    $(\$)$ holds for $m$ large enough again, say $m \ge \kappa_\EPSILON$ up to increasing $\kappa_\EPSILON$ again.
    We conclude accordingly. 
\end{proof}

\begin{corollary}[Trigger effect]
\label{corollary_trigger_effect}
    Let $\event_{\epsilon_0, T}$ be the trigger event as given by \eqref{equation_definition_trigger_event}.
    For all $\alpha > 0$ and $\epsilon_0 > 0$, we have 
    \begin{equation*}
        \Pr\parens*{(\event_{\epsilon_0, T})^c}
        = 
        \oh\parens*{
            \exp\parens*{- \log^{1 - \alpha}(T)}
        }
        \quad\text{when}\quad
        T \to \infty
        .
    \end{equation*}
\end{corollary}
\begin{proof}
    Let $m_T := \epsilon_0 \log(T)$ and fix $\alpha > 0$.
    By (\strong{A2}-\strong{4}) from \Cref{assumption_regularization_hyperparameter}, we find $\mathrm{poly}(\EPSILON(m_T)) \ge \log^{-\alpha}(T)$ when $T \to \infty$ and by (\strong{A4}), it follows that $\exp(-\beta \epsilonunif(m_T)^{-\abs{\states}}) \ge \exp(-\alpha \beta \log \log(T))$ when $T \to \infty$.
    All together, we find that when $T \to \infty$, 
    \begin{equation*}
        \Pr\parens*{\event_{m_T}^c}
        \le
        \exp \parens*{
            - \Omega\parens*{
                \log(T)^{-\alpha}
                \log(T)^{-\alpha \beta}
                \cdot
                \epsilon_0 \log(T)
            }
        }
        = \exp \parens*{
            - \Omega\parens*{
                \log(T)^{1 - \alpha(1 + \beta)}
            }
        }
        .
    \end{equation*}
    As this holds for arbitrary $\alpha > 0$, we conclude accordingly.
\end{proof}

\subsubsection{Part III: Post burn-in quality of exploration}
\label{appendix_near_optimal_exploration}

In this section, we show that under the trigger event \eqref{equation_definition_trigger_event}, explorative visit counts concentrate around $\visits^-_{\tau_\ell}(\pair) \approx \ell \imeasure^*(\pair)$ in high probability after a burn-in time of $\sqrt{\epsilon_0} \log(T)$ exploration phases, see \Cref{lemma_exploration_efficient_visits} below. 
Roughly speaking, it means that once the trigger event holds, exploration is near optimal.

\begin{lemma}
\label{lemma_exploration_efficient_visits}
    Let $(\tau_\ell)$ be the stopping-time enumeration of $\stimes^-$, i.e., $\tau_1 := \inf \stimes^-$ and $\tau_{\ell+1} := \inf \braces{t > \tau_\ell : t \in \stimes^-}$. 
    Let $\event \equiv \event_{\epsilon_0, T}$ be the trigger event, see \eqref{equation_definition_trigger_event}.
    For all $\epsilon_0 \in (0, 1]$, there exists a constant $\alpha_\EPSILON(\epsilon_0) < \infty$ such that, for all $T \ge 1$, all $m \ge \sqrt{\epsilon_0} \log(T) + \alpha_\EPSILON (\epsilon_0)$ and all $\pair \in \pairs$, we have
    \begin{equation}
    \label{equation_exploration_step_2}
        \Pr \parens*{
            \exists \ell \ge m
            :
            \tau_\ell \le T 
            ~\mathrm{and}~
            \abs*{
                \frac 1{\ell} \visits_{\tau_\ell}^- (\pair) - \imeasure^*(\pair)
            }
            \ge
            8 \abs{\pairs}\sqrt{\epsilon_0}
            ,
            \event
        }
        \le \exp \parens*{
            - \frac{\abs{\pairs}^2 \epsilon_0 m}{\beta(m)^2}
        }
    \end{equation}
    where $\beta(m) := 1 + 2 \abs{\states} \dmin(\kernel)^{-\abs{\states}} \epsilonunif(m)^{- \abs{\states}}$. 
\end{lemma}

We start by showing the alternative statement of \Cref{lemma_exploration_efficient_visits} given by \Cref{lemma_exploration_efficient_visits_aux} below.
Then \Cref{lemma_exploration_efficient_visits_aux} is reworked into \Cref{lemma_exploration_efficient_visits} to better fit our needs. 

\begin{lemma}
\label{lemma_exploration_efficient_visits_aux}
    Let $(\tau_\ell)$ be the stopping-time enumeration of $\stimes^-$, i.e., $\tau_1 := \inf \stimes^-$ and $\tau_{\ell+1} := \inf \braces{t > \tau_\ell : t \in \stimes^-}$. 
    Let $\event \equiv \event_{\epsilon_0, T}$ be the trigger event, see \eqref{equation_definition_trigger_event}.
    There exists a function $\gamma_\EPSILON(m) = \OH\parens[\big]{\sum_{n=1}^m (\epsilonunif(n)/\epsilonreg(n) + \epsilonreg(n))^{1/2}} = \oh(m)$ when $m \to \infty$ such that, for all $\delta > 0$, we have
    \begin{equation}
    \label{equation_exploration_step_1}
        \Pr \parens*{
            \exists \ell \ge 1
            :
            \tau_\ell \le T 
            ~\mathrm{and}~
            \abs*{
                \visits_{\tau_\ell}^- (\pair) - \ell \imeasure^*(\pair)
            }
            > 
            \begin{pmatrix}
                \epsilon_0 \abs{\pairs} \parens*{
                    \log(T) + 4 \ell
                }
                \\
                + \beta(\ell) \sqrt{\ell \log \parens*{\frac{1 + \ell}{\delta}}}
                + \gamma_\EPSILON (\ell)
            \end{pmatrix}
            \hspace{-.33em}
            ,
            \event
        }
        \le 
        \delta
    \end{equation}
    where $\beta(m) := 1 + 2 \abs{\states} \dmin(\kernel)^{-\abs{\states}} \epsilonunif(m)^{- \abs{\states}}$. 
\end{lemma}

\paragraph{Discussion of \Cref{lemma_exploration_efficient_visits_aux}.}
Said roughly, \eqref{equation_exploration_step_1} states that on the good event $\event$, we have $\visits_{\tau_\ell}^- (\pair) \approx \ell \imeasure^*(\pair)$ up to a few error terms.
The first term $\epsilon_0 \abs{\pairs} \log(T)$ is due to the initial burn-in phase, when the number of exploration times is less than $\epsilon_0 \log(T)$ and that the algorithm explores very badly, due to potentially wrong empirical data.
The second term $4 \epsilon_0 \abs{\pairs} \ell$ comes from the algorithm navigating following a noisy version of the optimal exploration policy.
The third term $\beta(\ell) \sqrt{\ell \log((1+\ell)/\delta)}$ comes from the stochasticity of the navigation, namely, from the aggregate difference between $\unit_{\State_{t+1}}$ and $\kernel(\Pair_t)$ suffered during exploration.
The last term $\gamma_\EPSILON (\ell)$ encapsulates the cost of changing of exploration epochs and the cost on the quality of the exploration policy due to the regularization $\EPSILON$.

\medskip
\def\proofname{Proof of \Cref{lemma_exploration_efficient_visits_aux}}
\begin{proof}
    We begin by introducing a few notations. 
    Because the regularization is floored as in \eqref{equation_floored_regularization}, the uniformization $\epsilonunif'(\tau_\ell)$ and the convexification $\epsilonreg'(\tau_\ell)$ are constant over ranges of exploration times the form $\braces{\tau_{2^m}, \ldots, \tau_{2^{m+1}-1}}$.
    Such ranges are referred to as \strong{exploration epochs} and exploration will be decomposed along them. 
    Further note that for $\tau_\ell \le T$, the floored and $T$-truncated regularizers \eqref{equation_truncated_regularizer} are equal with $\EPSILON'(\tau_\ell) = \EPSILON''(\tau_\ell)$ and in particular, $\epsilonflat''(\tau_\ell) \ge \epsilonflat''(T)$. 
    For an exploration epoch $m \ge 0$, let 
    \begin{equation}
        \imeasure_m \in \imeasures(\model) \cap \probabilities(\pairs)
    \end{equation}
    the central optimal exploration measure (\Cref{definition_central_exploration_measure}) of $\model$ under regularization $\EPSILON''(\tau_{2^m})$.
    Let $\optpolicy_m$ the fully randomized policy induced by $\imeasure_m$ (see \Cref{proposition_measures_policies_correspondence}).
    For $\pair \in \pairs$, let $\bias_m^{\pair}$ and $\gaps_m^\pair$ the bias and gap functions of $\policy_m$ in $\model$ under the reward function $f(\pair') := \indicator{\pair' = \pair}$.

    Fix $\ell_0 \ge 1$ and assume that $\tau_{2^m} \le T$ for $m = \floor{\log_2 (\ell_0)}$. 
    We write
    \begin{align}
    \notag
        \visits_{\tau_{\ell_0}}(\pair)
        & = 
        \sum_{m=0}^\infty
        \sum_{\ell=2^m}^{2^{m+1}-1}
        \indicator{\tau_\ell < \tau_{\ell_0}}
        \indicator{\Pair_\ell = \pair}
        \\
        & =
    \label{equation_exploration_step_1_a}
        \sum_{m=0}^\infty
        \sum_{\ell=2^m}^{2^{m+1}-1}
        \indicator{\tau_\ell < \tau_{\ell_0}}
        \parens*{
            \imeasure_m (\pair) 
            + \parens*{\unit_{\State_{\tau_\ell}} - \kernel(\Pair_{\tau_\ell})} \bias_m^\pair
            - \gaps_m^\pair (\Pair_{\tau_\ell})
        }
        .
    \end{align}
    By \Cref{lemma_uniform_measures_support_diameter}, we have the two bounds $\beta(x) \ge \max_{m \le \log_2(x)} \max \braces{\vecspan{\bias_m^\pair}, \norm{\gaps_m^\pair}_\infty}$ and $\beta(x)^{-1} \le \min_{m \le \log_2(x)} \min(\imeasure_m)$.
    We continue by bounding the middle term in \eqref{equation_exploration_step_1_a}.
    We have
    \begin{align}
    \notag
        & 
        \sum_{m=0}^\infty
        \sum_{\ell=2^m}^{2^{m+1}-1}
        \indicator{\tau_\ell < \tau_{\ell_0}}
        \parens*{\unit_{\State_{\tau_\ell}} - \kernel(\Pair_{\tau_\ell})} \bias_m^\pair
        \\
    \notag
        & \overset{(\dagger)}=
        \pm \log_2(\ell_0) \beta(\ell_0)
        +
        \sum_{m=0}^\infty
        \sum_{\ell=2^m}^{2^{m+1}-1}
        \indicator{\tau_\ell < \tau_{\ell_0}}
        \parens*{
            \parens*{\unit_{\State_{\tau_{\ell+1}}} - \unit_{\State_{\tau_\ell+1}}}
            + \parens*{\unit_{\State_{\tau_{\ell}+1}} - \kernel(\Pair_{\tau_\ell})}
        } \bias_m^\pair
        \\
    \label{equation_exploration_step_1_b}
        & \overset{(\ddagger)}=
        \pm \beta(\ell_0) \parens*{
            \log_2(\ell_0) + \abs{\pairs}\abs{\states}
        }
        + 
        \sum_{m=0}^\infty
        \sum_{\ell=2^m}^{2^{m+1}-1}
        \indicator{\tau_\ell < \tau_{\ell_0}}
        \parens*{
            \unit_{\State_{\tau_{\ell}+1}} - \kernel(\Pair_{\tau_\ell})
        } \bias_m^\pair
    \end{align}
    where 
    $(\dagger)$ introduces the notation $x = \pm \alpha + y$ as a shorthand for $\abs{y - x} \le \alpha$; and
    $(\ddagger)$ follows from the observation that $\State_{\tau_{\ell+1}} \ne \State_{\tau_\ell+1}$ can only hold when a panic happens, and the algorithm panics at most $\abs{\pairs} \abs{\states}$ times, see \Cref{lemma_panic_times}. 
    The remaining term is a martingale that will be dealt with in time.

    Moving on to the bound of the last term in \eqref{equation_exploration_step_1_a}, we write
    \begin{align}
    \notag
        & 
        \sum_{m=0}^\infty
        \sum_{\ell=2^m}^{2^{m+1}-1}
        \indicator{\tau_\ell < \tau_{\ell_0}}
        \gaps_m^\pair (\Pair_{\tau_\ell})
        \\
    \notag
        & =
        \sum_{m=0}^\infty
        \sum_{\ell=2^m}^{2^{m+1}-1}
        \indicator{\tau_\ell < \tau_{\ell_0}}
        \parens*{
            \policy^*_m(\State_{\tau_\ell})
            + \parens*{\policy^*_m(\State_{\tau_\ell}) - \policy^-_{\tau_\ell}(\State_{\tau_\ell})}
            + \parens*{\policy^-_{\tau_\ell}(\State_{\tau_\ell}) - \unit_{\Action_{\tau_\ell}}}
        } \cdot \gaps_m^\pair
        \\
    \label{equation_exploration_step_1_c}
        & \overset{(\dagger)}=
        \sum_{m=0}^\infty
        \sum_{\ell=2^m}^{2^{m+1}-1}
        \indicator{\tau_\ell < \tau_{\ell_0}}
        \parens*{\policy^*_m(\State_{\tau_\ell}) - \policy^-_{\tau_\ell}(\State_{\tau_\ell})} 
        \cdot \gaps_m^\pair
        \quad \Big\} =: \mathrm{B_1}
        \\
    \notag
        & \phantom{{} \overset{(\dagger)}= {}}
        +
        \sum_{m=0}^\infty
        \sum_{\ell=2^m}^{2^{m+1}-1}
        \indicator{\tau_\ell < \tau_{\ell_0}}
        \parens*{\policy^-_{\tau_\ell}(\State_{\tau_\ell}) - \unit_{\Action_{\tau_\ell}}}
        \cdot \gaps_m^\pair
    \end{align}
    where $(\dagger)$ follows from the Poisson equation of $\policy_m^*$, stating that $\policy_m^* (\state) \cdot \gaps_m^\pair = 0$ for all $\state \in \states$.
    \Cref{equation_exploration_step_1_c} is the sum of a martingale and of an error term denoted $\mathrm{B}_1$.
    We focus on that error term. 
    On the trigger event $\event$ defined in \eqref{equation_definition_trigger_event} and combined with \Cref{lemma_measures_to_policies}, we see that for $\ell \in \braces{2^m, \ldots, 2^{m+1}-1}$, we have $\norm{\policy^-_{\tau_\ell} - \policy^*_m}_\infty \le 4 \epsilon_0 \beta(m)^{-2} \min(\imeasure_m)^{-1}$ provided that $m$ is large enough.
    Indeed, for $\ell \ge \epsilon_0 \log(T) / \beta(\epsilon_0 \log(T))$ and on the trigger event $\event$, we have 
    \begin{equation}
    \label{equation_exploration_step_1_c_bis}
        \norm{
            \imeasure^*_{\EPSILON''(\tau_\ell)}(\hat{\model}_{\tau_\ell})
            - \imeasure^*_{\EPSILON''(\tau_\ell)}(\model)
        }_\infty 
        \le \epsilon_0 \beta(\ell)^{-2}
    \end{equation}
    by definition. 
    Recall that $\EPSILON'' (\tau_\ell) = \EPSILON'(\tau_\ell)$ for $\tau_\ell \le T$.
    Accordingly, $\EPSILON''(\tau_\ell)$ can be changed to $\EPSILON'(\tau_\ell)$ for all $\ell \le \ell_0$ since $\tau_{\ell_0} \le T$. 
    By invoking \Cref{lemma_measures_to_policies}, we conclude that for all $2^m \ge \epsilon_0 \log(T) / \beta(\epsilon_0 \log(T))$ and $\ell \in \braces{2^m, \ldots, 2^{m+1}}$ with $\tau_\ell \le \tau_{\ell_0}$, we have
    \begin{equation}
    \label{equation_exploration_step_1_c_bbis}
        \norm{
            \policy^-_{\tau_\ell} 
            - \optpolicy_m
        }_\infty
        \le
        4 \epsilon_0 \beta(m)^{-2} \min(\imeasure_m)^{-1}
    \end{equation}
    So, on $\event$, we have
    \begin{align}
    \notag
        \abs{\mathrm{B}_1}
        & \overset{(\dagger)}\le
        \sum_{m=0}^\infty
        \sum_{\ell=2^m}^{2^{m+1}-1}
        \indicator{\tau_\ell < \tau_{\ell_0}}
        \norm{\policy^-_{\tau_\ell} - \optpolicy_m}_\infty
        \norm{\gaps_m^\pair}_1
        \\
        & \overset{(\ddagger)}\le
    \notag
        \frac{
            \epsilon_0 \log(T)
        \sup_{m \le \log_2(\epsilon_0 \log(T))}
        \norm{\gaps_m^\pair}_1
        }{
            \beta(\epsilon_0 \log(T))
        }
        +
        \sum_{m=0}^\infty
        \sum_{\ell=2^m}^{2^{m+1}-1}
        \indicator{\tau_\ell < \tau_{\ell_0}}
        \frac{
            4 \epsilon_0 \norm{\gaps_m^\pair}_1
        }{
            \beta(2^m)^2 \min(\imeasure_m)
        }
        \\
        & \overset{(\S)}\le
    \notag
        \abs{\pairs} \parens*{
            \frac{
                \epsilon_0 \log(T)
            \sup_{m \le \log_2(\epsilon_0 \log(T))}
            \norm{\gaps_m^\pair}_\infty
            }{
                \beta(\epsilon_0 \log(T))
            }
            +
            \sum_{m=0}^\infty
            \sum_{\ell=2^m}^{2^{m+1}-1}
            \indicator{\tau_\ell < \tau_{\ell_0}}
            \frac{
                4 \epsilon_0 \norm{\gaps_m^\pair}_\infty
            }{
                \beta(2^m)^2 \min(\imeasure_m)
            }
        }
        \\
    \label{equation_exploration_step_1_d}
        & \overset{(\$)}\le
        \epsilon_0 
        \abs{\pairs} 
        \parens*{
            \log(T) 
            + 
            4 \ell_0
        }
    \end{align}
    where 
    $(\dagger)$ follows from H\"older's inequality;
    $(\ddagger)$ bounds the first $\epsilon_0 \log(T) / \beta(\epsilon_0 \log(T))$ terms by $\norm{\gaps_m^\pair}_1$ and invoke \eqref{equation_exploration_step_1_c_bbis} for the others; 
    $(\S)$ uses that $\norm{\gaps_m^\pair}_1 \le \abs{\pairs} \norm{\gaps_m^\pair}_\infty$;
    $(\S)$ follows by bounding the first term using that $\beta(\ell) \le \beta(\epsilon_0 \log(T))$ for $\ell \le \epsilon_0 \log(T)$, then using \Cref{lemma_uniform_measures_support_diameter} to obtain that $\norm{\gaps_m^\pair}_\infty \le \beta(\epsilon_0 \log(T))$; 
    and $(\S)$ bounds the second term  with \Cref{lemma_uniform_measures_support_diameter} for the bound $\norm{\gaps_m^\pair}_\infty \le \min(\imeasure_m) \beta(2^m)^2$;
    and $(\$)$ follows from straight-forward algebra. 

    Lastly, we relate $\sum_{m=0}^\infty \sum_{\ell=2^m}^{2^{m+1}-1} \imeasure_m(\pair) \indicator{\tau_\ell < \tau_{\ell_0}}$ to $\ell_0 \imeasure^* (\pair)$. 
    Let $\psi$ be the error function given by \Cref{proposition_convergence_optimal_measure}, see \eqref{equation_convergence_optimal_measure}.
    For all $m \ge 0$, we have
    \begin{equation*}
        \norm{\imeasure_m - \imeasure^*}_\infty 
        \le 
        \abs{\pairs}^{\frac 32}
        \norm{\imeasure_m - \imeasure^*}_2
        \overset{(\dagger)}\le
        \abs{\pairs}^{\frac 32}
        \sqrt{\psi\parens*{
            \frac{\epsilonunif(2^m)}{\epsilonreg(2^m)}
            + \epsilonreg(2^m)
        }}
    \end{equation*}
    where $(\dagger)$ follows by definition of $\imeasure_m$ and \Cref{proposition_convergence_optimal_measure}. 
    Note that $\psi(x) = \OH(x)$ when $x \to 0$, see \Cref{proposition_convergence_optimal_measure}.
    Therefore,
    \begin{align}
    \notag
        & \sum_{m=0}^\infty
        \sum_{\ell=2^m}^{2^{m+1}-1}
        \indicator{\tau_\ell < \tau_{\ell_0}}
        \imeasure_m(\pair)
        \\
    \notag
        & \overset{(\dagger)}=
        \ell_0 \imeasure^*(\pair)
        \pm
        \abs{\pairs}^{\frac 32}
        \sum_{m=0}^\infty
        \sum_{\ell=2^m}^{2^{m+1}-1}
        \indicator{\tau_\ell < \tau_{\ell_0}}
        \sqrt{\psi\parens*{
            \frac{\epsilonunif(2^m)}{\epsilonreg(2^m)}
            + \epsilonreg(2^m)
        }}
        \\
    \label{equation_exploration_step_1_e}
        & =
        \ell_0 \imeasure^*(\pair)
        \pm
        \abs{\pairs}^{\frac 32}
        \sum_{m=0}^{\floor{\log_2(\ell_0)}}
        2^m
        \sqrt{\psi\parens*{
            \frac{\epsilonunif(2^m)}{\epsilonreg(2^m)}
            + \epsilonreg(2^m)
        }}
        =: \ell_0 \imeasure^*(\pair) \pm \eta_\EPSILON (\ell_0)
    \end{align}
    where $(\dagger)$ introduces the short-hand $x = y \pm \alpha$ for $\abs{x - y} \le \alpha$.

    We have $\psi(x) = \OH(x)$ when $x \to 0$ and $\epsilonunif(2^m)/\epsilonreg(2^m) + \epsilonreg(2^m) \to 0$ as $m \to 0$ by \Cref{assumption_regularization_hyperparameter} (\strong{A5}).
    Now, note that if a non-negative sequence $(u_n) \in \RR_+^\NN$ satisfies $u_n \to 0$, then we have $\sum_{m=0}^n 2^m u_{2^m} = \oh(2^n)$. 
    So, the quantity $\eta_\EPSILON (\ell_0)$ defined in \eqref{equation_exploration_step_1_e} satisfies
    \begin{equation}
        \eta_\EPSILON (\ell_0) = \OH \parens*{
            \sum_{n=1}^{\ell_0}
            \sqrt{
                \frac{\epsilonunif(n)}{\epsilonreg(n)} 
                + \epsilonreg(n)
            }
        } = \oh(\ell_0)
        .
    \end{equation}

    We finish the proof by merging everything.
    Combining \eqref{equation_exploration_step_1_a}, \eqref{equation_exploration_step_1_b}, \eqref{equation_exploration_step_1_c}, \eqref{equation_exploration_step_1_d} and \eqref{equation_exploration_step_1_e}, we find that on the trigger event $\event$ and for $\tau_{\ell_0} \le T$, we have
    \begin{align*}
        \abs*{
            \visits^-_{\tau_{\ell_0}}(\pair)
            -
            \ell_0 \imeasure^*(\pair)
        }
        \le
        \eta_\EPSILON(\ell_0)
        + \beta_{\ell_0} \parens*{\log_2(\ell_0) + \abs{\pairs} \abs{\states}}
        + \epsilon_0 \abs{\pairs} \parens*{\log(T) + 4 \ell_0}
        + \mathrm{MDS}
    \end{align*}
    where MDS is the martingale difference sequence given by
    \begin{equation*}
        \sum_{m=0}^\infty
        \sum_{\ell=2^m}^{2^{m+1}-1}
        \indicator{\tau_\ell < t}
        \parens*{
            \parens*{
                \unit_{\State_{\tau_{\ell}+1}} - \kernel(\Pair_{\tau_\ell})
            } \bias_m^\pair
            + \parens*{\policy^-_{\tau_\ell}(\State_{\tau_\ell}) - \unit_{\Action_{\tau_\ell}}}
            \cdot \gaps_m^\pair
        },
    \end{equation*}
    which is the sum of $\ell_0$ terms that are almost surely bounded by $2 \beta(\ell_0)$ because $\beta(-)$ is non-decreasing (see \Cref{assumption_regularization_hyperparameter} (\strong{A1})) and $\vecspan{\bias_m^\pair} \le \diameter(\policy_m^*; \model) \le \beta(2^m)$ by \Cref{lemma_bias_diameter,lemma_uniform_measures_support_diameter}.
    Therefore, using a time-uniform Azuma-Hoeffding inequality (\Cref{lemma_azuma_time_uniform}), it is bounded by $2 \beta(\ell_0) \sqrt{\ell_0 \log((1+\ell_0)/\delta)}$ with probability $1 - \delta$.
    To conclude the proof, we set $\gamma_\EPSILON(\ell) := \eta_\EPSILON (\ell) + \beta(\ell) (\log_2(\ell) + \abs{\pairs} \abs{\states})$.
    Note that the definition of $\gamma_\EPSILON$ does not depend on $\epsilon_0$.
    Because $\beta(\ell) = \mathrm{poly}(1/\epsilonunif(\ell))$ and $\epsilonunif(\ell)^{-\alpha} = \oh(\ell)$ for all $\alpha > 0$ by \Cref{assumption_regularization_hyperparameter} (\strong{A4}), we conclude that $\gamma_\EPSILON (\ell) = \oh(\ell)$.
\end{proof}
\def\proofname{Proof}

The proof of \Cref{lemma_exploration_efficient_visits_aux} used the result below (\Cref{lemma_measures_to_policies}), relating the distances between two induced policies $\policy(\imeasure)$ and $\policy(\imeasure')$ when $\imeasure, \imeasure'$ are two fully supported measures that are close to one another.

\begin{lemma}
\label{lemma_measures_to_policies}
    Let $\imeasure, \imeasure' \in (\RR_+^*)^\pairs$ be two fully supported measures on $\pairs$ and denote $\policy(\imeasure), \policy(\imeasure')$ their respective induced policies, see \Cref{proposition_measures_policies_correspondence}.
    If $\norm{\imeasure' - \imeasure}_\infty
    \le \epsilon \min(\mu)$ for $\epsilon < (2\abs{\pairs})^{-1}$, then
    \begin{equation*}
        \norm{\policy(\imeasure') - \policy(\imeasure)}_\infty
        :=
        \max_{(\state, \action) \in \pairs} 
        \abs*{
            \policy(\imeasure')(\action|\state) - \policy(\imeasure)(\action|\state)
        }
        \le
        4 \epsilon
        .
    \end{equation*}
\end{lemma}
\begin{proof}
    This is straight-forward algebra.
    For short, we write $\imeasure(\state) := \sum_{\action \in \actions(\state)} \imeasure(\state, \action)$.
    Recall that by definition, the policy induced by $\imeasure$ is $\policy(\imeasure)(\action|\state) := \imeasure(\state, \action) \imeasure(\state)^{-1}$.
    For $(\state, \action) \in \pairs$, we have
    \begin{align*}
        \abs*{
            \policy(\imeasure')(\action|\state) - \policy(\imeasure)(\action|\state)
        }
        & \le
        \abs*{
            \frac{
                \imeasure(\state, \action) - \imeasure'(\state, \action)
            }{\imeasure(\state)}
        }
        + 
        \abs*{
            \imeasure'(\state, \action) \parens*{
                \frac 1{\imeasure'(\state)} - \frac 1{\imeasure(\state)}
            }
        }
        \\
        & \le 
        \epsilon 
        + \frac{\imeasure'(\state, \action)}{\imeasure(\state)}
        \abs*{
            \frac 1{1 + \frac{\imeasure'(\state) - \imeasure(\state)}{\imeasure(\state)}}
            - 1
        }
        \\
        & \overset{(\dagger)}\le
        \epsilon 
        + \frac{2 \imeasure'(\state, \action) \epsilon}{\imeasure(\state)}
        \le 
        \epsilon \parens*{
            1 + \frac{2 (\imeasure(\state, \action) + \epsilon \min(\imeasure))}{\imeasure(\state)}
        }
        \le 4 \epsilon
    \end{align*}
    where $(\dagger)$ uses that $\abs{\frac 1{1+x} - 1} \le 2 \abs{x}$ for $x \le \frac 12$.
\end{proof}

We derive \Cref{lemma_exploration_efficient_visits} by reworking \eqref{equation_exploration_step_1} from \Cref{lemma_exploration_efficient_visits_aux} into a more convenient form.

\medskip
\def\proofname{Proof of \Cref{lemma_exploration_efficient_visits}}
\begin{proof}
    Borrowing the notations from \Cref{lemma_exploration_efficient_visits_aux}, it holds that with probability $1 - \delta$ and on the trigger event $\event$, that
    \begin{align*}
        \visits_{\tau_\ell}(\pair)
        & \overset{(\dagger)}\ge
        \ell \imeasure^*(\pair)
        - \beta(\ell) \sqrt{\ell \log\parens*{\frac{1+\ell}\delta}}
        - \epsilon_0 \abs{\pairs} \parens*{\log(T) + 4 \ell}
        - \gamma_\EPSILON (\ell)
        \\
        & \ge
        \ell \parens*{
            \imeasure^*(\pair)
            - {\beta(\ell)} \sqrt{\frac{\log((1+\ell)/\delta)}{\ell}}
            - {\epsilon_0 \abs{\pairs}} \parens*{\frac{\log(T)}\ell + 4}
            - \frac {\gamma_\EPSILON(\ell)}\ell
        }
        \\
        & \ge
        \ell \parens*{
            \imeasure^*(\pair)
            - \frac{\beta(\ell) \parens[\big]{\sqrt{\log(1+\ell)} + \sqrt{\log(1/\delta)}}}{\sqrt{\ell}} 
            - {\epsilon_0 \abs{\pairs}} \parens*{\frac{\log(T)}\ell + 4}
            - \frac{\gamma_\EPSILON(\ell)}{\ell}
        }
        \\
        & \overset{(\ddagger)}\ge
        \ell  \parens*{
            \imeasure^*(\pair) - {8 \sqrt{\epsilon_0} \abs{\pairs}}
        }
    \end{align*}
    where $(\dagger)$ readily follows from \Cref{lemma_exploration_efficient_visits_aux}, see \eqref{equation_exploration_step_1}; and
    $(\ddagger)$ holds provided that the following equations hold:
    \begin{equation}
    \notag
    \begin{gathered}
        {\sqrt{\epsilon_0}\abs{\pairs}}
        \overset{(\dagger)}\ge 
        {\beta(\ell)} \sqrt{\frac{\log(1+\ell)}\ell}
        ,
        \qquad
        {\sqrt{\epsilon_0}\abs{\pairs}}
        \overset{(\ddagger)}\ge 
        {\beta(\ell)} \sqrt{\frac{\log(1/\delta)}\ell}
        \\
        {\sqrt{\epsilon_0}\abs{\pairs}}
        \overset{(\S)}\ge 
        \frac{\epsilon_0 \abs{\pairs} \log(T)}{\ell}
        ,
        \qquad
        {\sqrt{\epsilon_0}\abs{\pairs}}
        \overset{(\$)}\ge 
        {\epsilon_0 \abs{\pairs}}
        \quad\text{and}\quad
        {\sqrt{\epsilon_0}\abs{\pairs}}
        \overset{(\#)}\ge 
        \frac{\gamma_\EPSILON(\ell)}{\ell}
        .
    \end{gathered}
    \end{equation}
    In the above,
    $(\dagger)$ is of the form $\sqrt{\epsilon_0} \abs{\pairs} \ge C\,\mathrm{poly}(\epsilonunif(\ell)) \sqrt{\log(1+\ell)/\ell}$, so using that $\epsilonunif(\ell) = \omega(\log^{-1}(\ell))$ by \Cref{assumption_regularization_hyperparameter} (\strong{A4}), so we see that $(\dagger)$ holds provided that $\ell$ is large enough with respect to $\epsilon_0$, say $\ell \ge \alpha^\dagger(\epsilon_0)$;
    $(\S)$~is equivalent to $\ell > \sqrt{\epsilon_0} \log(T)$; 
    $(\$)$~holds provided that $\epsilon_0 \le 1$; 
    $(\#)$~holds provided that $\gamma_\EPSILON(\ell) \le \ell \sqrt{\epsilon_0} \abs{\pairs}$ so since $\gamma_\EPSILON(\ell) = \oh(\ell)$, see \Cref{lemma_exploration_efficient_visits_aux}, so we see that $(\#)$ holds provided that $\ell$ is large enough with respect to $\epsilon_0$, say $\ell \ge \alpha^{\#}(\epsilon_0)$; and 
    $(\ddagger)$ can be solved in $\delta > 0$, to find the condition $\delta \ge \exp(-\abs{\pairs}^2 \epsilon_0 \ell \beta(\ell)^{-2})$. 
    Setting $\alpha(\epsilon_0) := \max\braces{\alpha^\dagger(\epsilon_0), \alpha^{\#}(\epsilon_0)}$, we finally find the conditions
    \begin{equation}
    \label{equation_exploration_step_2_b}.
        \ell \ge \sqrt{\epsilon_0} \log(T) + \alpha(\epsilon_0)
        \quad\text{and}\quad
        \delta \ge \exp \parens*{
            - \abs{\pairs}^2 \epsilon_0 
            \cdot \frac{\ell}{\beta(\ell)^2}
        }
        .
    \end{equation}
    Note that the condition on $\delta$ is monotone in $\ell$ when $\ell$ is large enough, see \Cref{assumption_regularization_hyperparameter} (\strong{A6}).
    So, up to increasing $\alpha(\epsilon_0)$, if \eqref{equation_exploration_step_2_b} holds for some $\ell \ge 1$, then it holds for all $\ell' \ge \ell$. 
    The upper-bound is obtained similarly. 
\end{proof}
\def\proofname{Proof}

\subsubsection{Part IV: The high probability exploration barrier}
\label{appendix_weak_exploration_barrier}

Following \eqref{equation_exploration_step_2} from \Cref{lemma_exploration_efficient_visits}, we have $\visits_{\tau_\ell}^- (\pair) \approx \ell \imeasure^*(\pair)$ with high probability on the trigger event $\event$.
In other words, exploration is near optimal.
Still under the trigger event, we know that the information value (\Cref{definition_information_value}) of exploration measures is well estimated by definition, see \eqref{equation_definition_trigger_event}. 
Following this, we show that the number of exploration phases is quite unlikely to exceed the \strong{exploration threshold} given by
\begin{equation}
\label{equation_exploration_step_3_a}
    m_0 \equiv m_{\epsilon_0, T}
    :=
    \frac {(1 + \epsilontest(T)) \log(T)}{
        \parens*{
            1 - \frac{8 \abs{\pairs} \sqrt{\epsilon_0}}{\dmin(\imeasure^*)}
        } \parens*{
            \ivalue(\imeasure^*, \model) - \epsilon_0
        }
    }
\end{equation}
where $\epsilon_0 > 0$ quantifies the desired precision of the trigger event, see \Cref{lemma_exploration_barrier_low}.

\begin{lemma}
\label{lemma_exploration_barrier_low}
    Let $T^- := \sup \braces{\ell \ge 1 : \tau_\ell \le T-1}$ be the (random) number of exploration times prior to $T \ge 1$.
    Let $m_{\epsilon_0, T}$ be as given by \eqref{equation_exploration_step_3_a}.
    There exists a constant $\kappa_\EPSILON \in \RR_+$ and a function $\psi \equiv \psi_{\epsilon_0, \EPSILON} : \NN \to \RR_+$ such that, for all $T \ge 1$, we have
    \begin{equation}
    \label{equation_exploration_step_4}
        \Pr \parens*{
            T^- \ge
            m_{\epsilon_0, T}
            + \kappa_\EPSILON \log^{\frac 34}(T)
            ,
            \event
        }
        \le
        \exp \parens*{
            - \psi(T)
        }
    \end{equation}
    with $\psi(T) = \omega(\log(T)^{1/2 - \alpha})$ as $T \to \infty$ for all $\alpha > 0$.
\end{lemma}

To establish \Cref{lemma_exploration_barrier_low}, we show first that after the exploration threshold $m_0$, exploration GLR tests \eqref{equation_glr_exploration} do not provoke exploration phases, see \Cref{lemma_exploration_glr_threshold} below.

\begin{lemma}
\label{lemma_exploration_glr_threshold}
    Let $T^- := \sup \braces{\ell \ge 1 : \tau_\ell \le T-1}$ be the (random) number of exploration times prior to $T \ge 1$.
    Let $m_0 \equiv m_{\epsilon_0, T}$ be as given by \eqref{equation_exploration_step_3_a}.
    There is a polynomial $\mathrm{poly}(-)$ of degree $4 \abs{\states}$ and some constant $\kappa_\EPSILON \in \RR_+$ such that, for $m_0 \equiv m_{\epsilon_0, T} \ge \sqrt{\epsilon_0} \log(T) + \alpha_\EPSILON (\epsilon_0)$, we have
    \begin{equation}
    \label{equation_exploration_step_3_b}
        \Pr \parens*{
            \sum_{\ell=m_0+1}^{T^-}
            \eqindicator{
                \State_{\tau_\ell} \in \states(\optpairs(\model))
                \atop
                \tau_\ell \notin \stimes^\pm
            }
            > 0,
            \event
        }
        \le \exp \parens*{
            - \mathrm{poly}(\epsilonunif(m_{\epsilon_0, T})) m_{\epsilon_0, T}
            \atop
            + \kappa_\EPSILON
        }
        .
    \end{equation}
\end{lemma}

\def\proofname{Proof of \Cref{lemma_exploration_glr_threshold}}
\begin{proof}
    Let $\beta(\ell) := 1 + 2 \abs{\states} \dmin(\kernel)^{-\abs{\states}} \epsilonunif(\ell)^{-\abs{\states}}$.
    Note that on $\event$ and for $\ell \ge \epsilon_0 \log(T) / \beta(\epsilon_0 \log(T))$, we have $\flatme{\optpairs}{\iepsilonflat''(\tau_\ell)}(\hat{\model}_{\tau_\ell}) = \optpairs(\model)$ where $\EPSILON''$ is the $T$-truncated regularizer, see \eqref{equation_truncated_regularizer}.
    We find condition on $\ell \ge 1$ such that the GLR exploration test \eqref{equation_glr_exploration} is guaranteed to fail at time $t = \tau_\ell$.
    Let $\model^\dagger \in \confusing_\EPSILON(\hat{\model}_{\tau_\ell})$.
    We have
    \begin{align*}
        \sum_{\pair \in \pairs}
        \visits_{\tau_\ell}(\pair)
        \KL(\hat{\model}_{\tau_\ell}(\pair)||\model^\dagger(\pair))
        & \overset{(\dagger)}\ge
        (1 - \lambda) \ell
        \sum_{\pair \in \pairs}
        \max \braces*{
            \imeasure^*(\pair),
             \alpha \epsilonunif(\ell)^{\abs{\states}}
        }
        \KL(\hat{\model}_{\tau_\ell}(\pair)||\model^\dagger(\pair))
        \\
        & \overset{(\ddagger)}\ge
        (1 - \lambda) \ell
        \sum_{\pair \in \pairs}
        \parens*{(1 - \delta_\ell) \imeasure^*(\pair) + \delta_\ell}
        \KL(\hat{\model}_{\tau_\ell}(\pair)||\model^\dagger(\pair))
        \\
        & \overset{(\S)}\ge
        (1 - \lambda) \ell ~
        \ivalue_\EPSILON \parens*{
            (1 - \delta)_\ell \imeasure^* + \delta_\ell \unit,
            \hat{\model}_{\tau_\ell}
        }
        \\
        & \overset{(\$)}\ge
        (1 - \lambda) \ell \cdot \parens[\Big]{
            \ivalue_\EPSILON \parens*{
                (1 - \delta_\ell) \imeasure^* + \delta_\ell \unit,
                \model
            }
            - \epsilon_0
        }
        \\
        & \ge
        (1 - \lambda) \ell \cdot \parens[\Big]{
            \ivalue_\EPSILON \parens*{
                (1 - \delta_\ell) \imeasure^*,
                \model
            }
            - \epsilon_0
        }
        \\
        & =
        (1 - \lambda) \ell \cdot \parens[\Big]{
            (1 - \delta_\ell) 
            \ivalue_\EPSILON \parens*{
                \imeasure^*,
                \model
            }
            - \epsilon_0
        }
        \\
        & \overset{(\#)}\ge
        (1 - \lambda) \ell \cdot \parens[\Big]{
            (1 - \delta_\ell) \ivalue_0 \parens*{
                \imeasure^*,
                \model
            }
            - \epsilon_0
        }
    \end{align*}
    where
    $(\dagger)$ introduces $\lambda := 8 \abs{\pairs} \sqrt{\epsilon_0} \dmin(\imeasure^*)^{-1}$ and holds, on $\event$, uniformly over $\ell$ with probability $1 - \exp(-\abs{\pairs}^2 \epsilon_0 \cdot \beta(\ell)^{-2} \ell) - \exp(-\mathrm{poly}(\epsilonunif(\ell))\ell + \kappa_\EPSILON)$, using a combination of \eqref{equation_exploration_step_2} from \Cref{lemma_exploration_efficient_visits} to claim that $\visits_{\tau_\ell}(\pair) \gtrsim \ell \imeasure^*(\pair)$ and of the uniform exploration result of \Cref{lemma_uniform_exploration} to claim that $\visits_{\tau_\ell}(\pair) \gtrsim \alpha \epsilonunif(\ell)^{\abs{\states}}$ for some $\alpha > 0$;
    $(\ddagger)$ introduces $\delta(\ell) := \alpha \epsilonunif(\ell)^{\abs{\states}}$;
    $(\S)$ holds by $\model^\dagger \in \confusing_\EPSILON(\hat{\model}_{\tau_\ell})$ and by definition of $\ivalue_\EPSILON$ (\Cref{definition_information_value});
    $(\$)$ holds because $(1 - \delta(\ell)) \imeasure^* + \delta(\ell) \unit$ is obviously $\delta(\ell)$-uniform and by definition of $\event$, see \eqref{equation_definition_trigger_event}; and 
    $(\#)$ holds when $\epsilonflat(\tau_\ell)$ is small enough so that $\confusing_{\iepsilonflat}(\model) = \confusing_0(\model)$, i.e., when $\epsilonflat(\tau_\ell) < \gaingap(\model)$, see \Cref{proposition_continuity_near_optimality} and \Cref{definition_near_confusing_set}.
    
    Solving $(\#) \ge (1 + \epsilontest(T)) \log(T)$, we derive the definition of $m_0 \equiv m_{\epsilon_0, T}$ in \eqref{equation_exploration_step_3_a}.  
    We conclude that
    \begin{align*}
        & \Pr \parens*{
            \sum_{\ell=m_0+1}^{T^-}
            \eqindicator{
                \State_{\tau_\ell} \in \states(\optpairs(\model))
                \atop
                \tau_\ell \notin \stimes^\pm
            }
            > 0,
            \event
        }
        \\
        & \le
        \exp \parens*{
            - \frac{\abs{\pairs}^2 \epsilon_0 m_0}{\beta(m_0)^2}
        }
        + \exp \parens*{
            - \mathrm{poly}(\epsilonunif(m_0)) m_0
            + \kappa_\EPSILON
        }
        \\
        & \le 
        \exp \parens*{
            - \min \braces*{
                \frac{\abs{\pairs}^2 \epsilon_0 m_0}{\beta(m_0)^2},
                \mathrm{poly}(\epsilonunif(m_0)) m_0
            }
            + \kappa_\EPSILON + \log(2)
        }
        \\
        & \overset{(\dagger)}\le 
        \exp \parens*{
            - \mathrm{poly}(\epsilonunif(m_0)) m_0
            + \kappa_\EPSILON + \beta' + \log(2)
        }
    \end{align*}
    where $(\dagger)$ follows from the observation that $\beta(\ell)^{-1} = \mathrm{poly}(\epsilonunif(\ell))$ is a polynomial of degree $\abs{\states}$ while $\mathrm{poly}(-)$ introduced by \Cref{lemma_uniform_exploration} is of degree $4 \abs{\states}$, so that $\mathrm{poly}(\epsilonunif(m_0)) m_0 \le \abs{\pairs}^2 \epsilon_0 \cdot \beta(m_0)^{-2} m_0$ for $m_0 > \kappa'$ where $\kappa' \in \RR_+$ is some constant.
    This proves the claim.
\end{proof}

Following \eqref{equation_exploration_step_3_b} from \Cref{lemma_exploration_glr_threshold}, we know that exploration GLR tests \eqref{equation_glr_exploration} do not provoke exploration phases after $m_{\epsilon_0, T}$ exploration phases---under the good event $\event$ given in \eqref{equation_definition_trigger_event} at least.
Beyond that threshold, exploration is mostly due to co-exploration and the number of induced phases can be controlled using \Cref{lemma_coexploration_travels}, concluding the proof of \Cref{lemma_exploration_barrier_low}.

\bigskip
\def\proofname{Proof of \Cref{lemma_exploration_barrier_low}}
\begin{proof}
    Let $m \ge m_0 \equiv m_{\epsilon_0, T}$ be as given in \eqref{equation_exploration_step_3_a}, and assume that $T \ge 1$ is large enough so that $m_{\epsilon_0, T} \ge \sqrt{\epsilon_0} \log(T) + \alpha_\EPSILON (\epsilon_0)$ where $\alpha_\EPSILON(\epsilon_0)$ is given by \Cref{lemma_exploration_efficient_visits}. 
    Let $\alpha > 0$ the constant given by \Cref{lemma_coexploration_travels}.
    We have
    \begin{align}
    \notag
        & \Pr \parens*{
            T^- 
            \ge 
            m 
            + \alpha \epsilonunif(T^-)^{-2 \abs{\states}} \log \log(T)
            ,
            \event
        }
        \\
    \notag
        & \overset{(\dagger)}\le 
        \Pr \parens*{
            T^- 
            \ge 
            m 
            + \alpha \epsilonunif(T^-)^{-2 \abs{\states}} \parens*{
                \log \log(T)
                +
                \sum_{\ell=m_0+1}^{T^-}
                \eqindicator{
                    \State_{\tau_\ell} \in \states(\optpairs(\model))
                    \atop
                    \tau_\ell \notin \stimes^\pm
                }
            }
            ,
            \event'
        }
        \\
    \notag
        & \phantom{{} \le {}}
        +
        \Pr \parens*{
            \sum_{\ell=m_0+1}^{T^-}
            \eqindicator{
                \State_{\tau_\ell} \in \states(\optpairs(\model))
                \atop
                \tau_\ell \notin \stimes^\pm
            }
            > 0,
            \event
        }
        \\
        & \overset{(\ddagger)}\le
    \label{equation_exploration_step_4_a}
        \exp \parens*{
            - \beta_1 \epsilonunif(m)^{4 \abs{\states}} (m - m_0)
            \atop
            + \kappa_\EPSILON
        }
        +
        \exp \parens*{
            - \mathrm{poly}(\epsilonunif(m_{\epsilon_0, T})) m_{\epsilon_0, T}
            \atop
            + \kappa'_\EPSILON
        }
    \end{align}
    where 
    $(\dagger)$ introduces $\event' := (\forall \ell \ge m_0, \flatme{\optpairs}{\iepsilonflat''}(\hat{\model}_{\tau_\ell}) = \optpairs(\model))$ where $\epsilonflat''$ is the $T$-truncated leveling regularizer \eqref{equation_truncated_regularizer}, and $(\dagger)$ uses that $\event \subseteq \event'$; and
    $(\ddagger)$ bounds the first term of $(\dagger)$ using the lemma on co-exploration travels (\Cref{lemma_coexploration_travels}) and the second term using \eqref{equation_exploration_step_3_b} from \Cref{lemma_exploration_glr_threshold}, where $\mathrm{poly}(-)$ is a polynomial of degree $4 \abs{\states}$.
    In the following, we we assume that $\mathrm{poly}(x) \le \beta_1 x^{4 \abs{\states}}$ up to decreasing $\mathrm{poly}(-)$.

    We set $m_T := (1 + \log(T)^{-1/2}) m_{\epsilon_0, T}$, so that $m_T - m_{\epsilon_0, T} = \log(T)^{-1/2} m_{\epsilon_0, T}$.

    Setting $m \equiv m_T$ in \eqref{equation_exploration_step_4_a}, we find
    \begin{align*}
        & \Pr \parens*{
            T^- 
            \ge 
            m_{\epsilon_0, T} 
            + \frac{m_{\epsilon_0, T}}{\sqrt{\log(T)}}
            + \epsilonunif(T^-)^{-2 \abs{\states}} \log \log(T)
            ,
            \event
        }
        \\
        & \le
        \exp \parens*{
            - \mathrm{poly}(\epsilonunif(m_T)) (m - m_{\epsilon_0,T})
            \atop
            + \kappa_\EPSILON
        }
        +
        \exp \parens*{
            - \mathrm{poly}(\epsilonunif(m_{\epsilon_0, T})) m_{\epsilon_0, T}
            \atop
            + \kappa'_\EPSILON
        }
        \\
        & \overset{(\dagger)}\le
        \exp \parens*{
            - \omega \parens*{
                \log \log(T)^{- 4 \abs{\states}} \sqrt{\log(T)}
            }
        }
        + \exp \parens*{
            - \omega \parens*{
                \log \log(T)^{- 4 \abs{\states}} \log(T)
            }
        }
        \\
        & \overset{(\ddagger)}\le
        \exp \parens*{
            - \omega\parens*{
                \log(T)^{\frac 12 - \alpha}
            }
        }
    \end{align*}
    where 
    $(\dagger)$ uses that $m_{\epsilon_0, T} = \Theta(\log(T))$, that $m_T - m_{\epsilon_0, T} = \Theta(\sqrt{\log(T)})$ and that $\epsilonunif(m) = \omega(\log(m)^{-1})$, see \Cref{assumption_regularization_hyperparameter} (\strong{A4}); and
    $(\ddagger)$ holds when $T \to \infty$ for all (fixed) $\alpha > 0$.

    We conclude by simplifying the bound on $T^-$.
    Assume that $T^- \le (1 + \log(T)^{-1/2}) m_{\epsilon_0, T} + \epsilonunif(T^-)^{-2\abs{\states}} \log \log(T)$.
    Because $\epsilonunif(m)^{-2\abs{\states}} = \oh(\log(m))$ by \Cref{assumption_regularization_hyperparameter} (\strong{A4}), we have $\epsilonunif(T^-)^{-2 \abs{\states}} \le \log(T^-)$ provided that $T^-$ is large enough, say $T^- \ge \kappa''_{\EPSILON}$. 
    It follows that
    \begin{align*}
        T^- 
        & \le
        \parens*{1 + \frac 1{\sqrt{\log(T)}}} m_{\epsilon_0, T} 
        + \log(T^-) \log\log(T)
        + \kappa''_\EPSILON
        \\
        & \overset{(\dagger)}\le
        \inf_{\delta \in (0, 1)}
        \braces*{
            \frac 1{1 - \delta}
            \parens*{
                \parens*{1 + \frac 1{\sqrt{\log(T)}}} m_{\epsilon_0, T} 
                + \kappa''_\EPSILON
            }
            + \parens*{
                \frac{\log\log(T)}{\delta}
            }^2
        }
        \\
        & \overset{(\ddagger)}\le
        \inf_{\delta \in (0, \frac 12)}
        \braces*{
            (1 + 2 \delta)
            \parens*{
                \parens*{1 + \frac 1{\sqrt{\log(T)}}} m_{\epsilon_0, T} 
                + \kappa''_\EPSILON
            }
            + \parens*{
                \frac{\log\log(T)}{\delta}
            }^2
        }
        \\
        & \overset{(\S)}\le
        \parens*{1 + \frac 2{\log(T)^{1/4}}}
        \parens*{1 + \frac 1{\sqrt{\log(T)}}} 
        m_{\epsilon_0, T} 
        + \log\log(T)^2 \sqrt{\log(T)}
        + 2 \kappa''_\EPSILON
        \\
        & \le
        \parens*{1 + \frac 5{\log(T)^{1/4}}}
        m_{\epsilon_0, T} 
        + \log\log(T)^2 \sqrt{\log(T)}
        + 2 \kappa''_\EPSILON
    \end{align*}
    where
    $(\dagger)$ uses that $\log(T^-) \log\log(T)/T^- \le \delta$ for $T^- \ge (\log\log(T)/\delta)^2$;
    $(\ddagger)$ follows from $\frac 1{1 - \delta} \le 1 + 2 \delta$ for $\delta < \frac 12$; and 
    $(\S)$ sets $\delta = \log(T)^{-1/4}$.
    This proves the claim.
\end{proof}
\def\proofname{Proof}

\subsubsection{Part V: First order failure and the skeleton exploration barrier} 
\label{appendix_strong_exploration_barrier}

If either the trigger event $\event$ fails, or if $\visits^-_{\tau_\ell}(\pair)$ drastically deviates from its expected value $\ell \imeasure^*(\pair)$, the number of exploration phases can be as big as $T$ in the worst scenario. 
Combining \Cref{lemma_trigger_effect,lemma_exploration_efficient_visits,lemma_exploration_barrier_low}, this can happen with probability of order $\exp(-\log(T)^{1 - \alpha})$ for some $\alpha > 0$. 
In expectation, the associated error becomes $T \cdot \exp(-\log(T)^\alpha)$, that is $\Omega(T^{1 - \beta})$ for all $\beta > 0$ .
One cannot afford an error term that large---bounding the number of exploration phases by $T$ when our good events fail is too rough. 

That is why, in this paragraph, we show that with \strong{very} high probability, exploration cannot go beyond a threshold of order $\log(T)^3$ (\Cref{lemma_exploration_barrier} and \Cref{corollary_exploration_barrier}): This is the \strong{heavy} exploration barrier motivated earlier. 
We start with a complete (but heavy) statement in \Cref{lemma_exploration_barrier} that we simplify in \Cref{corollary_exploration_barrier} under assumptions on the hyperparameter $\EPSILON$.
The idea behind the exploration barrier (\Cref{lemma_exploration_barrier}) is that after enough exploration times, all pairs must enter the skeleton because exploration is slightly uniformized. 
Once that all pairs are in the skeleton, the GLR exploration test \eqref{equation_glr_exploration} never triggers exploration phases, so that all further exploration times are exclusively produced by the co-exploration mechanism. 
Because all the empirical data concentrates quickly on the skeleton, all the conditions are realized so that co-exploration induces very few exploration times, see \Cref{lemma_coexploration_travels}.

Therefore, the exploration barrier consists in showing that exploration naturally stops for good once all pairs enter in the skeleton, and quantifying how quickly this happens. 

\begin{lemma}
\label{lemma_exploration_barrier}
    Let $T^- := \sup \braces{\ell \ge 1: \tau_\ell \le T-1}$ the (random) number of exploration times prior to $T \ge 1$.
    There exist constants $\alpha, \beta > 0$ and a function $\psi \equiv \psi_\EPSILON : \NN \to \RR_+$ such that
    \begin{equation*}
        \Pr \parens*{
            T^-
            \ge
            2 \log^3(T) 
            + \alpha \epsilonunif(T^-)^{-2 \abs{\states}} \log \log(T)
        }
        \le
        \exp \parens*{
            - \psi(T)
        }
    \end{equation*}
    where $\psi(T) \sim
        - \beta \min \braces*{
            \epsilonunif(4 \log^3(T))^{4\abs{\states}} \log^3(T),
            \epsilonflat(T)^{2} \log^2(T)
        }
    $ when $T \to \infty$.
\end{lemma}

Under \Cref{assumption_regularization_hyperparameter}, the dominant term in the condition on $T^-$ is $T^- \ge 2 \log^3(T)$ and the dominant term in the error probability is $\log^2(T) \epsilonflat(T)^{-2}$. 
This provides the alternative version below (\Cref{corollary_exploration_barrier}) that is easier to carry out.

\begin{corollary}[Skeleton barrier]
\label{corollary_exploration_barrier}
    Under \Cref{assumption_regularization_hyperparameter}, there exists a constant $\alpha > 0$ such that for all $\delta > 0$, we have
    \begin{equation}
    \label{equation_exploration_barrier}
        \Pr \parens*{
            \sum_{t=1}^{T-1} \indicator{t \in \stimes^-}
            \ge
            \alpha \log^3 (T)
        }
        \le
        \oh \parens*{ \exp \parens*{
            - \log^{2 - \delta}(T)
        } }
        = \oh(T^{-2})
        \text{\quad when $T \to \infty$}.
    \end{equation}
\end{corollary}

\def\proofname{Proof of \Cref{lemma_exploration_barrier}}
\begin{proof}
    Let $(\tau_\ell)$ be the stopping-time enumeration of $\stimes^-$, i.e., $\tau_1 := \inf \stimes^-$ and $\tau_{\ell + 1} := \inf \braces{t > \tau_\ell : t \in \stimes^-}$. 
    Recall that $\EPSILON'$ denotes the floored regularizer \eqref{equation_floored_regularization} and that $\EPSILON''$ denotes the $T$-truncated regularizer \eqref{equation_truncated_regularizer}. 
    Recall that $\epsilonunif''(\ell) = \epsilonunif'(\ell) \ge \epsilonunif(\ell)$, $\epsilonreg''(\ell) = \epsilonreg'(\ell) \ge \epsilonreg(\ell)$ and $\epsilontest''(t) = \epsilontest'(t) = \epsilontest(t)$. 
    Let $m_0 \ge 1$ to be tuned later and introduce the events
    \begin{equation*}
    \begin{aligned}
        \event_{m_0} 
        & := \parens*{
            \forall \ell \ge m_0,
            \flatme{\optpairs}{\iepsilonflat''(\tau_\ell)}(\hat{\model}_{\tau_\ell}) \ne \optpairs(\model)
        }
        ,
        \\
        \event'_{m_0}
        & := \parens*{
            \forall \ell \ge m_0,
            \forall \pair \in \pairs,
            \visits_{\tau_\ell} (\pair) \ge \log^2 (T)
        }
    \end{aligned}
    \end{equation*}
    and introduce the short-hand random variables
    \begin{align*}
        T^- 
        & := 
        \sum_{t=1}^{T-1} 
        \indicator{t \in \stimes^-} 
        = 
        \sup\braces{\ell \ge 1 : \tau_\ell \le T - 1}
        ,
        \\
        \psi_1(m_0, T) 
        & := 
        2 m_0 
        + \alpha_1 \epsilonunif(T^-)^{-2\abs{\states}} \log \log(T)
        ,
        \\
        \psi_2(m_0, T)
        & :=
        2 m_0 
        + \alpha_1 \epsilonunif(T^-)^{-2\abs{\states}} \parens*{
            \log \log(T)
            + \sum_{\ell=2m_0+1}^{T^-}
            \eqindicator{
                \State_{\tau_\ell} \in \states(\optpairs(\model))
                \atop
                \tau_\ell \notin \stimes^\pm
            }
        }
    \end{align*}
    where $\alpha_1$ is the alias for the constant $\alpha$ that appears in the co-exploration travel lemma (\Cref{lemma_coexploration_travels}).
    We have
    \begin{align}
    \notag
        \Pr \parens*{
            T^-
            \ge
            2m_0 + \psi_1 (m_0, T)
        }
    \notag
        & \le
        \Pr \parens*{
            T^-
            \ge
            2m_0 + \psi_1 (m_0, T)
            ,
            \atop
            \event_{m_0}, \event'_{m_0}
        }
        + 
        \Pr \parens*{
            \parens*{\event_{m_0}}^c, \event'_{m_0}
        }
        + 
        \Pr \parens*{
            \parens*{\event'_{m_0}}^c
        }
        \\
    \notag
        & \overset{(\dagger)}\le
        \Pr \parens*{
            T^-
            \ge
            2m_0 + \psi_1 (m_0, T)
            ,
            \atop
            \event_{m_0}, \event'_{m_0}
        }
        + 
        \Pr \parens*{
            \parens*{\event_{m_0}}^c, \event'_{m_0}
        }
        + 
        \Pr \parens*{
            \parens*{\event'_{m_0}}^c
        }
        \\
    \label{equation_exploration_barrier_1}
        & \overset{(\ddagger)}\le
        \exp\parens*{
            - \beta_1 \epsilonunif(2m_0)^{4 \abs{\states}} m_0
            \atop
            + 2 \log(m_0) + \kappa_\EPSILON
        }
        + 
        \Pr \parens*{
            \parens*{\event_{m_0}}^c, \event'_{m_0}
        }
        + 
        \Pr \parens*{
            \parens*{\event'_{m_0}}^c
        }
    \end{align}
    where
    $(\dagger)$ follows from the observation that under $\event'_{m_0}$ and for $t \le T-1$, all pairs are within the skeleton hence no GLR exploration test \eqref{equation_glr_exploration} may trigger an exploration phase, so that $\psi_1 (m_0, T) = \psi_2 (m_0, T)$; and
    $(\ddagger)$ follows by the co-exploration lemma (\Cref{lemma_coexploration_travels}) and $\beta_1, \kappa_\EPSILON$ are aliases for the constants $\beta, \kappa_\EPSILON$ that appear in that result.

    This leaves to bound $\Pr \parens{ \parens{\event_{m_0}}^c, \event'_{m_0} }$ and $\Pr \parens{ \parens{\event'_{m_0}}^c }$. 

    Starting with $\Pr \parens{ \parens{\event_{m_0}}^c, \event'_{m_0} }$, we could use the trigger effect (\Cref{lemma_trigger_effect}) but this is a bit of an overshoot---instead, we specialize the argument of the proof of \Cref{lemma_trigger_effect} on the fly. 
    By \Cref{proposition_continuity_near_optimality}, the condition ``$\flatme{\optpairs}{\iepsilonflat''}(\hat{\model}_{\tau_\ell}) = \optpairs(\model)$'' is met as soon as 
    \begin{equation}
    \label{equation_exploration_barrier_2}
        \alpha_2 \snorm{\hat{\model}_{\tau_\ell} - \model}
        < 
        \frac 13 \epsilonflat''(\tau_\ell)
        \le
        \frac 13 \epsilonflat''(\ell)
        < \gaingap(\model)
    \end{equation}
    where $\alpha_2$ is an alias for the constant $C(\model)$ that appears in \Cref{proposition_continuity_near_optimality}.
    To convert $\snorm{\hat{\model}_{\tau_\ell} - \model}$ into $\norm{\hat{\model}_{\tau_\ell} - \model}$, we rely on \Cref{lemma_convert_snorm_to_norm}, that provides the additional condition ``$\hat{\model}_{\tau_\ell} \sim \model$'' that is met as soon as
    \begin{equation}
    \label{equation_exploration_barrier_3}
        \norm{\hat{\model}_{\tau_\ell} - \model} 
        < 
        \dmin(\model)
        =: \alpha_3
        .
    \end{equation}
    Now, the equations \eqref{equation_exploration_barrier_2} and \eqref{equation_exploration_barrier_3} can be combined into a single one, of the form $\alpha_4 \norm{\hat{\model}_{\tau_\ell} - \model} < \frac 13 \epsilonflat''(\tau_\ell)$ together with $\epsilonflat'(\ell) < \gaingap(\model)$.
    The second condition is about $\ell$ and $T$ being large enough because $\epsilonflat(\ell) \to 0$ when $\ell \to \infty$. 
    All together, we have
    \begin{align}
    \notag
        \Pr \parens*{
            \parens*{\event_{m_0}}^c, \event'_{m_0}
        }
    \notag
        & \overset{(\dagger)}\le
        \Pr \parens*{
            \exists \ell \ge m_0
            :
            \norm{\hat{\model}_{\tau_\ell} - \model}
            \ge
            \frac 1{3 \alpha_4} \epsilonflat''(\tau_\ell)
            \text{~and~}
            \forall \pair \in \pairs,
            \visits_{\tau_\ell}(\pair) \ge \log^2(T)
        }
        \\
    \notag
        & \overset{(\ddagger)}\le
        \Pr \parens*{
            \exists \ell \ge m_0
            :
            \norm{\hat{\model}_{\tau_\ell} - \model}
            \ge
            \frac 1{3 \alpha_4} \epsilonflat(T)
            \text{~and~}
            \forall \pair \in \pairs,
            \visits_{\tau_\ell}(\pair) \ge \log^2(T)
        }
        \\
    \notag
        & \overset{(\S)}\le
        2 \abs{\pairs} \exp \parens*{
            - {18 \alpha_4^2 {\epsilonflat(T)^{2}} \log^2(T)}
            + \frac 12 \log\parens*{1 + \log^2(T)}
            + \abs{\states} \log(2)
        }
        \\
    \label{equation_exploration_barrier_4}
        & =
        \exp \parens*{
            - {18 \alpha_4^2 {\epsilonflat(T)^{2}} \log^2(T)}
            + \OH (\log \log(T))
        }
    \end{align}
    where
    $(\dagger)$ holds when $\epsilonflat(m_0) < \gaingap(\model)$ and $T \ge m_0$, i.e., when $T$ is large enough by \Cref{assumption_regularization_hyperparameter} (\strong{A0});
    $(\ddagger)$ holds because $\epsilonflat$ is decreasing; and
    $(\S)$ follows from concentration results (\Cref{corollary_threshold_concentration}).

    We continue with $\Pr \parens{ \parens{\event'_{m_0}}^c }$.
    Assume that $m_0$ is large enough so that, for all $\ell \ge m_0$, we have $\alpha_5 \epsilonunif(\ell)^{\abs{\states}} \ell \ge \log^2(T)$ where $\alpha_5$ is an alias for the constant $\alpha$ appearing in \Cref{lemma_uniform_exploration}.
    Because $\epsilonunif(\ell)^{\abs{\states}} \ell = \omega(\ell^{1 - \eta})$ for all $\eta > 0$ when $\ell \to \infty$ by \Cref{assumption_regularization_hyperparameter} (\strong{A4}), we see that $m_0 \asymp \alpha_4^{-1}\log^{2}(T)$ is asymptotically the right threshold. 
    In the sequel, set 
    \begin{equation*}
        m_0 := \log^3 (T)
    \end{equation*}
    and consider that $T \to \infty$.
    For $T$ large, we have $\alpha_5 \epsilonunif(\ell)^{\abs{\states}} \ell \ge \log^2(T)$ for all $\ell \ge m_0 \equiv \log^3(T)$. 
    So, by \Cref{lemma_uniform_exploration}, we find
    \begin{equation}
    \label{equation_exploration_barrier_5}
        \Pr \parens{\parens{\event'_{m_0}}^c}
        \le 
        \exp \parens*{
            - \mathrm{poly}(\epsilonunif(m_0)) m_0
            \atop
            + \kappa_\EPSILON
        }
        =
        \exp \parens*{
            - \mathrm{poly}(\epsilonunif(\log^3(T))) \log^3(T)
            \atop
            + \kappa_\EPSILON
        }
    \end{equation}
    where $\kappa_\EPSILON$ is the constant $\kappa_\EPSILON$ appearing in \Cref{lemma_uniform_exploration} and $\mathrm{poly}(-)$ is a polynomial of degree $4 \abs{\states}$.
    So, setting $m_0 = \log^3(T)$, we conclude by combining \eqref{equation_exploration_barrier_1}, \eqref{equation_exploration_barrier_4} and \eqref{equation_exploration_barrier_5}.
\end{proof}
\def\proofname{Proof}

\subsubsection{Part VI: Putting everything together and proving \Cref{lemma_exploration}}
\label{appendix_exploration_together}

We can finally conclude the proof.
Let $m_{\epsilon_0, T}$ the exploration threshold as given by \eqref{equation_exploration_step_3_a}.
Introduce
\begin{equation}
    \beta_1 (T) 
    := m_{\epsilon_0, T} + \kappa_\EPSILON \log^{\frac 34}(T)
    \quad\text{and}\quad
    \beta_2 (T)
    := \alpha \log^3(T)
\end{equation}
where 
$\kappa_\EPSILON$ is the constant that appears in the high probability exploration barrier (\Cref{lemma_exploration_barrier_low}) and 
$\alpha$ is the constant that appears in the skeleton exploration barrier (first order failure, see \Cref{corollary_exploration_barrier}).
Introduce
\begin{equation}
\begin{gathered}
    \event_0
    := \parens*{
        \forall \ell \ge \beta_1 (T),
        \forall \pair \in \pairs,
        \visits_{\tau_\ell}^-(\pair)
        \le \parens*{
            \imeasure^*(\pair) + 8 \abs{\pairs} \sqrt{\epsilon_0}
        } \ell
    }, 
    \\
    \event_1 := \parens*{ T^- < \beta_1(T) }
    \quad\text{and}\quad
    \event_2 := \parens*{ T^- < \beta_2 (T) }
    .
\end{gathered}
\end{equation}
The event $\event_1^c$ corresponds to the first order failure and $\event_2^c$ corresponds to the second order failure in \Cref{figure_exploration}. 
Recall that $\event \equiv \event_{\epsilon_0, T}$ is the trigger event, see \eqref{equation_definition_trigger_event}.
We have
\begin{align*}
    & \EE \brackets*{
        \visits_T^- (\pair)
    }
    \\
    & \overset{(\dagger)}\le
    \EE \brackets*{
        \parens*{\imeasure^*(\pair) + 8 \abs{\pairs} \sqrt{\epsilon_0}}
        \beta_1 (T) 
        \cdot 
        \indicator{\event_0 \cap \event_1 \cap \event}
        + 
        \beta_2 (T) 
        \cdot 
        \indicator{\event_0^c \cup \event_1^c \cup \event^c}
        \indicator{\event_2}
        +
        T 
        \cdot 
        \indicator{\event_2^c}
    }
    \\
    & \le
    \parens*{\imeasure^*(\pair) + 8 \abs{\pairs} \sqrt{\epsilon_0}}
    \beta_1 (T)
    + \beta_2(T) \parens*{
        \Pr \parens*{\event_0^c, \event}
        + \Pr \parens*{\event_1^c, \event}
        + \Pr \parens*{\event^c}
    }
    + T \Pr \parens*{
        \event_2^c
    }
    \\
    & \overset{(\ddagger)}=
    \parens*{\imeasure^*(\pair) + 8 \abs{\pairs} \sqrt{\epsilon_0}}
    \beta_1 (T)
    + \OH \parens*{ \log^3(T) e^{- \Omega\parens*{\sqrt{\log(T)}}} }
    + \OH \parens*{ \log^3(T) e^{- \Omega\parens*{\log(T)^{1/4}}} }
    \\
    & 
    \phantom{ 
        {}\overset{(\ddagger)}=
        \parens*{\imeasure^*(\pair) + 8 \abs{\pairs} \sqrt{\epsilon_0}}
        \beta_1 (T)
    }
    + \OH \parens*{ \log^3(T) e^{- \Omega\parens*{\sqrt{\log(T)}}} }
    + \OH \parens*{ T e^{- \Omega\parens*{\log(T)^{3/2}}} }
    \\
    & =
    \parens*{\imeasure^*(\pair) + 8 \abs{\pairs} \sqrt{\epsilon_0}}
    \beta_1 (T)
    + \oh(1)
\end{align*}
where 
$(\dagger)$ follows from the observation that $\visits_T^-(\pair) = \visits^-_{\tau_{T^-}}(\pair)$ and $\visits^-_T(\pair) \le T^-$ a.s.; and
$(\dagger)$ bounds the first error using \eqref{equation_exploration_step_2} from \Cref{lemma_exploration_efficient_visits}, the second error using \eqref{equation_exploration_step_4} from \Cref{lemma_exploration_barrier_low} (exploration barrier), the third error using \Cref{corollary_trigger_effect} (trigger effect), and the fourth error using \Cref{corollary_exploration_barrier} (skeleton exploration barrier).
Unfolding the definition of $m_{\epsilon_0, T}$, see \eqref{equation_exploration_step_3_a}, we obtain
\begin{equation*}
    \EE \brackets*{
        \visits_T^- (\pair)
    }
    \le
    \frac{
        \parens*{\imeasure^*(\pair) + 8 \abs{\pairs} \sqrt{\epsilon_0}}
        (1 + \epsilontest(T)) \log(T)
    }{
        \parens*{
            1 - \frac{8 \abs{\pairs} \sqrt{\epsilon_0}}{\dmin(\imeasure^*)}
        } \parens*{
            \ivalue(\imeasure^*, \model)
            - \epsilon_0
        }
    }
    + \OH\parens*{\log^{\frac 34}(T)}
    \quad \text{when} \quad
    T \to \infty,
\end{equation*}
where $\imeasure^*$ is the central optimal probability exploration measure of $\model$ (\Cref{definition_central_exploration_measure}).
Letting $T \to \infty$ then taking $\epsilon_0$ vanishing, we conclude that 
\begin{equation*}
     \limsup_{T \to \infty}
     \frac{\EE\brackets*{\visits_T^- (\pair)}}{\log(T)}
     \le
     \frac{\imeasure^*(\pair)}{\ivalue(\imeasure^*, \model)}
     .
\end{equation*}
This concludes the proof of \Cref{lemma_exploration}.

\subsubsection{Auxiliary: Co-exploration travels}
\label{appendix_coexploration_travels}

In this section, we provide an instrumentary result (\Cref{lemma_coexploration_travels}) to bound the number of exploration phases triggered by co-exploration.
In spirit, \Cref{lemma_coexploration_travels} states that the number of exploration phases triggered by co-exploration is of order $\log \log(T)$. 

Recall that $\EPSILON''$ denotes the $T$-truncated regularizer, see \eqref{equation_truncated_regularizer}.

\begin{lemma}
\label{lemma_coexploration_travels}
    Let $(\tau_\ell)$ be the stopping-time enumeration of $\stimes^-$, i.e., $\tau_1 := \inf \stimes^-$ and $\tau_{\ell+1} := \inf \braces{t > \tau_\ell : t \in \stimes^-}$.
    Let $T^- := \sup \braces{\ell \ge 1 : \tau_\ell \le T-1}$ be the number of exploration times prior to $T \ge 1$.
    There exist constants $\alpha, \beta, \kappa_\EPSILON > 0$ such that, for all $m_0$ and $T \ge 1$, introducing the event $\event \equiv \event_{m_0} := (\forall \ell \ge m_0, \flatme{\optpairs}{\iepsilonflat''}(\hat{\model}_{\tau_\ell}) = \optpairs(\model))$, for all $m \ge m_0$, we have
    \begin{equation}
    \label{equation_coexploration_travels}
    \begin{gathered}
        \Pr \parens*{
            T^- 
            \ge
            m 
            + \alpha \epsilonunif(T^-)^{- 2 \abs{\states}} \parens*{
                \log \log(T)
                + \sum_{\ell=m_0+1}^{T^-} \eqindicator{
                    \State_{\tau_\ell} \in \states(\optpairs(\model))
                    \atop
                    \tau_\ell \notin \stimes^\pm
                }
            },
            \event
        }
        \\
        \le \exp \parens*{
            - \beta \epsilonunif(m)^{4 \abs{\states}} (m - m_0)
            + \kappa_\EPSILON
        }
        \quad
        \text{when $m \to \infty$.}
    \end{gathered}
    \end{equation}
\end{lemma}

\paragraph{Interpretation of \Cref{lemma_coexploration_travels}.}
The statement of \Cref{lemma_coexploration_travels} reads as follows: $\event$ is a good event, stating that beyond the exploration threshold $m_0$, the co-exploration structure is correct, with $\flatme{\optpairs}{\iepsilonflat''}(\hat{\model}) = \optpairs(\model)$. 
Then, \Cref{equation_coexploration_travels} states that beyond the threshold $m_0$ and under $\event$, exploration times are due to 
(I)~co-exploration times for a total of $\OH(\log\log(T))$ induced exploration times; and
(II)~exploration GLR tests \eqref{equation_glr_exploration} that explicitely cause exploration phases (exploration times for which $\tau_\ell \notin \stimes^\pm$ while $\State_{\tau_\ell} \in \states(\optpairs(\model))$).
\Cref{lemma_coexploration_travels} is to be applied when (II) is unlikely to happen beyond $m_0$, i.e., when the exploration GLR test is very likely to claim that enough information has been gathered beyond the exploration threshold $m_0$.
In such circumstances, \Cref{equation_coexploration_travels} essentially says that there are $\OH(\log\log(T))$ exploration times beyond $m_0$ under the good event $\event$.

\medskip
\begin{proof}
    Fix $m_0 \ge 1$.
    Let $\pairs_1, \ldots, \pairs_{c_0}$ be the communicating components of $\optpairs(\model)$, that the algorithm guesses correctly on $\event$ by assumption.
    Let $c^-(t) := \min \braces{c : \min\braces{\visits_t(\pair) : \pair \in \pairs_c} = \min\braces{\visits_t (\pair) : \pair \in \optpairs(\model)}}$ be the least visited component at time $t \ge 1$.
    We decompose $\braces{\tau_{m_0}, \tau_{m_0+1}, \ldots}$ into segments alternating between least visited components.
    Specifically, let $(j(i))_{i \ge 1}$ be the sequence given by
    \begin{equation}
    \label{equation_coexploration_travels_0_a}
    \begin{aligned}
        j(1) & := m_0,
        \\
        j(i+1) & := \inf \braces*{
            \ell > j(i)
            :
            \State_{\tau_\ell} \in \states(\pairs_{c^-(\tau_\ell)})
        }.
    \end{aligned}
    \end{equation}
    We relate the number of exploration times in $\braces{\tau_{m_0}, \tau_{m_0+1}, \ldots}$ to $j(-)$ as follows.

    \par
    \medskip
    \noindent
    \STEP{1}
    \textit{
        \allowdisplaybreaks[1]
        Denote $I(m) := \inf \braces{i : j(i) < m}$ the pseudo-inverse of $j(-)$.
        For $m \ge 1$, let $\alpha_m := \dmin(\kernel)^{\abs{\states}} \epsilonunif(m)^{\abs{\states}}$ and $\beta_m := 1 + 2 \abs{\states} \dmin(\kernel)^{-\abs{\states}} \epsilonunif(m)^{-\abs{\states}}$.
        For all $\delta \ge 1$, we have
        \begin{gather*}
            \Pr \parens*{
                \exists m \ge m_0,
                (1 + \beta_m) I(m)
                \ge
                \alpha_m (m - m_0)
                - \beta_m \parens*{
                    \abs{\pairs} \abs{\states}
                    + 
                    2 \sqrt{
                        \textstyle
                        (m - m_0)
                        \log \parens*{
                            \frac{1 + m - m_0}\delta
                        }
                    }
                },
                \event
            }
            \\
            \le \delta
            .
        \end{gather*}
        \vspace{-1em}
    }
    \begin{subproof}
        We use a technique that is similar to the proof of the uniform exploration lemma (\Cref{lemma_uniform_exploration}).
        Given $\eta > 0$, we consider the Markov decision process $\model_\eta := (\pairs, \reward, \kernel_\eta)$ with kernel given by
        \begin{equation*}
            \kernel_\eta (\state, \action)
            :=
            \parens*{
                1 - \eta \abs{\actions(\state)}
            } \kernel(\state, \action)
            + \eta \sum_{\action' \in \actions(\state)} \kernel(\state, \action')
            .
        \end{equation*}
        The model $\model_\eta$ is communicating and by construction, the execution of any $\eta$-uniform policy $\policy$ on $\model$ can be seen as the execution of a randomized policy $\policy_\eta$ on $\model_\eta$, in the sense that $\EE^{\model_\eta, \policy}[-] = \EE^{\model, \policy_\eta}[-]$.

        Given $\pairs^i \equiv \pairs_{c^-(\tau_{j(i)})}$, introduce the reward function $f_i(\pair) := \indicator{\pair \in \pairs^i}$.
        Consider the model $\model_\eta^{-f_i} = (\pairs, - f_i, \kernel_\eta)$ with kernel $\kernel_\eta$ and deterministic reward function $-f_i$ and introduce $\gainof{-f_i}_\eta$, $\biasof{-f_i}_\eta$ and $\gapsof{-f_i}_\eta$ the associated optimal gain, bias an gap functions of $\model_\eta^{-f_i}$.
        Because $\model_\eta^{-f_i}$ is communicating, we have $\vecspan{\gainof{-f_i}_\eta} = 0$.
        Let $\policy^*_\eta$ be a deterministic bias optimal policy of $\model_\eta^{-f_i}$.
        By construction, $\policy^*_\eta$ corresponds to an $\eta$-uniform policy of $\model$ for which all pairs are recurrent.
        By \Cref{lemma_uniform_measures_support_diameter}, we have $\min(\gainof{-f_n}_\eta) = \max(\gainof{-f_i}_\eta) \le - \dmin(\kernel)^{\abs{\states}} \eta^{\abs{\states}}$. 
        Consider the gaps in $\model$, given by
        \begin{equation}
        \label{equation_coexploration_travels_1_a}
            \gapsof{-f_i}(\state, \action)
            :=
            \gainof{-f_i}_{\eta}(\state) + \biasof{-f_i}_\eta(\state)
            + f(\state, \action) - \kernel(\state, \action) \biasof{-f_i}_\eta.
        \end{equation}
        Note that we also have $\max\braces{\vecspan{\biasof{-f_i}_\eta}, \norm{\gapsof{-f_i}_\eta}_\infty} \le \beta_\eta := 1 + 2\,\dmin(\kernel)^{-\abs{\states}} \eta^{-\abs{\states}}$ by \Cref{lemma_uniform_measures_support_diameter} again.
        Following from the Bellman equations of $\model_\eta^{-f_i}$, check that every $\eta$-uniform policy $\policy$ satisfies
        \begin{equation}
        \label{equation_coexploration_travels_1_b}
            \sum_{\action \in \actions(\state)}
            \policy_\eta (\action|\state)
            \gapsof{-f_i}_\eta(\state, \action)
            =
            \sum_{\action \in \actions(\state)}
            \policy(\action|\state) 
            \gapsof{-f_i}(\state, \action)
            \ge
            0
        \end{equation}
        where $\policy_\eta$ is the randomized policy induced by $\policy$ (i.e., bisimulating $\policy$) in $\model_\eta$.

        Let $m \ge m_0$ and set $\eta := \epsilonunif(m)$.
        For conciseness, let $j'(i) := \min \braces{j(i), m}$. 
        We have
        \begin{align*}
            I(m)
            =
            \sum_{i=1}^{I(m)}
            1
            & \overset{(\dagger)}=
            \sum_{i=1}^{I(m)}
            \sum_{\ell=j(i)}^{j'(i+1)-1}
            \indicator{
                \State_{\tau_\ell} \in \states(\pairs_{c^-(\tau_{j(i)})})
            }
            \\
            & \overset{(\ddagger)}=
            \sum_{i=1}^{I(m)}
            \sum_{\ell=j(i)}^{j'(i+1)-1}
            \parens*{
                - \gainof{-f_i}_\eta(\State_{\tau_\ell})
                + \parens*{
                    \kernel(\Pair_{\tau_\ell}) - \unit_{\State_{\tau_\ell}} 
                } \biasof{-f_i}_\eta
                + \gapsof{-f_i}(\Pair_{\tau_\ell})
            }
            \\
            & \ge
            \alpha_m (m - m_0)
            + 
            \underbrace{
                \sum_{i=1}^{I(m)}
                \sum_{\ell=j(i)}^{j'(i+1)-1}
                \parens*{
                    \kernel(\Pair_{\tau_\ell}) - \unit_{\State_{\tau_\ell}} 
                } \biasof{-f_i}_\eta
            }_{\mathrm{A}_1}
            +
            \underbrace{
                \sum_{i=1}^{I(m)}
                \sum_{\ell=j(i)}^{j'(i+1)-1}
                \gapsof{-f_i}(\Pair_{\tau_\ell})
            }_{\mathrm{A}_2}
        \end{align*}
        where 
        $(\dagger)$ follows by definition of $j(i+1)$, see \eqref{equation_coexploration_travels_0_a}; and
        $(\ddagger)$ follows by definition of $\gapsof{-f_i}$, see \eqref{equation_coexploration_travels_1_a}.
        We control the error terms as follows, starting with $\mathrm{A}_1$.
        We have
        \begin{align*}
            \mathrm{A}_1
            & =
            \sum_{i=1}^{I(m)}
            \sum_{\ell=j(i)}^{j'(i+1)-1}
            \parens*{
                \unit_{\State_{\tau_\ell+1}}  - \unit_{\State_{\tau_\ell}} 
            } \biasof{-f_i}_\eta
            +
            \sum_{i=1}^{I(m)}
            \sum_{\ell=j(i)}^{j'(i+1)-1}
            \parens*{
                \kernel(\Pair_{\tau_\ell+1}) - \unit_{\State_{\tau_\ell+1}} 
            } \biasof{-f_i}_\eta
            \\
            & \overset{(\dagger)}\ge
            - \beta_m \parens*{
                I(m) + \abs{\pairs}\abs{\states}  
            }
            +
            \sum_{i=1}^{I(m)}
            \sum_{\ell=j(i)}^{j'(i+1)-1}
            \parens*{
                \kernel(\Pair_{\tau_\ell+1}) - \unit_{\State_{\tau_\ell+1}} 
            } \biasof{-f_i}_\eta
        \end{align*}
        where 
        $(\dagger)$ follows from the observation that for $\ell \in \braces{j(i), \ldots, j'(i+1)-1}$, we have $\tau_{\ell+1} = \tau_{\ell}+1$ unless $\ell \in \stimes^!$ is a panic time---and by bounding the number of panic times by $\abs{\pairs} \abs{\states}$, the maximum number of times a new transition is discovered, see \Cref{lemma_panic_times}.
        The remaining right-hand term is a martingale that we leave for later.

        We continue with $\mathrm{A}_2$.
        We have
        \begin{align*}
            \mathrm{A}_2 
            & := 
            \sum_{i=1}^{I(m)}
            \sum_{\ell=j(i)}^{j'(i+1)-1}
            \gapsof{-f_i}(\Pair_{\tau_\ell})
            \\
            & =
            \sum_{i=1}^{I(m)}
            \sum_{\ell=j(i)}^{j'(i+1)-1}
            \sum_{\action \in \actions(\State_{\tau_\ell})}
            \parens*{
                \policy_{\tau_\ell}^- (\action|\State_{\tau_\ell})
                \gapsof{-f_i}(\State_{\tau_\ell}, \action)
                + 
                \parens*{
                    \indicator{\Action_{\tau_\ell} = \action} - \policy_{\tau_\ell}^- (\action|\State_{\tau_\ell})
                } \gapsof{-f_i}(\State_{\tau_\ell}, \action)
            }
            \\
            & \overset{(\dagger)}\ge
            \sum_{i=1}^{I(m)}
            \sum_{\ell=j(i)}^{j'(i+1)-1}
            \sum_{\action \in \actions(\State_{\tau_\ell})}
            \parens*{
                \indicator{\Action_{\tau_\ell} = \action} - \policy_{\tau_\ell}^- (\action|\State_{\tau_\ell})
            } \gapsof{-f_i}(\State_{\tau_\ell}, \action)
        \end{align*}
        where
        $(\dagger)$ follows from the observation that $\policy_{\tau_\ell}^-$ is $\eta$-randomized (since $\ell \le m$) and from \eqref{equation_coexploration_travels_1_b}.
        We conclude that $\mathrm{A}_1$ and $\mathrm{A}_2$ both involve martingales obtained as the sum of $m - m_0$ martingale differences whose terms hav e span at most $\beta_m$. 
        So, by a time-uniform Azuma-Hoeffding's inequality (\Cref{lemma_azuma_time_uniform}), we find that with probability $1 - \delta$ and uniformly for $m \ge m_0$, we have
        \begin{equation*}
            (1 + \beta_m) I(m)
            \ge
            \alpha_m (m - m_0)
            - \beta_m \parens*{
                \abs{\pairs} \abs{\states}
                + 
                2 \sqrt{
                    (m - m_0)
                    \log \parens*{
                        \frac{1 + m - m_0}\delta
                    }
                }
            }
            .
        \end{equation*}
        This proves the claim.
    \end{subproof}

    \par
    \medskip
    \noindent
    \STEP{2}
    \textit{
        For short, denote $I(m) := \inf \braces{i : j(i) < m}$ the pseudo-inverse of $j(-)$.
        There exist constants $\alpha, \alpha', \kappa_\EPSILON > 0$ such that, for all $m_0$, we have
        \begin{equation}
        \label{equation_coexploration_travels_1}
        \begin{aligned}
            & \Pr \parens*{
                \exists m' \ge m,
                I(m') > \alpha \epsilonunif(m')^{2 \abs{\states}} (m' - m_0),
                \event
            }
            \\
            & \qquad \le
            \exp \parens*{
                - \alpha' \epsilonunif(m)^{4 \abs{\states}} (m - m_0)
                + 2 \log(m_0) + \kappa_\EPSILON
            }
            .
        \end{aligned}
        \end{equation}
        \vspace{-1em}
    }
    \begin{subproof}
        We continue from the result of \STEP{1}.
        Recall that $\alpha_m := \dmin(\kernel)^{\abs{\states}} \epsilonunif(m)^{\abs{\states}}$ and $\beta_m := 1 + 2 \abs{\states} \dmin(\kernel)^{- \abs{\states}} \epsilonunif(m)^{- \abs{\states}}$.
        We are to solve the equation
        \begin{equation}
        \label{equation_coexploration_travels_1_b_bis}
            \alpha_m \cdot (m - m_0) 
            > 
            4 \beta_m \sqrt{
                (m - m_0) \log\parens*{
                    \frac{1 + m - m_0}\delta
                }
            }
        \end{equation}
        in $\delta$. 
        We first decouple \eqref{equation_coexploration_travels_1_b_bis} as the pair of sufficient conditions 
        \begin{align*}
            \text{({\it i})}
            \quad
            & \alpha_m \cdot (m - m_0) > 8 \beta_m \sqrt{(m - m_0) \log(1 + m - m_0)}, \text{and}
            \\
            \text{({\it ii})}
            \quad
            & \alpha_m \cdot (m - m_0) > 8 \beta_m \sqrt{(m - m_0) \log\parens*{\frac 1\delta}}
            .
        \end{align*}
        The first condition (\textit{i}) is equivalent to
        \begin{equation}
        \label{equation_coexploration_travels_1_c}
            \frac{m - m_0}{\log(1 + m - m_0)}
            \ge
            64 
            \dmin(\kernel)^{-2\abs{\states}}
            \epsilonunif(m)^{-2 \abs{\states}}
            \parens*{1 + 2 \abs{\states} \dmin(\kernel)^{-\abs{\states}} \epsilonunif(m)^{-\abs{\states}}}^2
        \end{equation}
        and using $\log(1 + x) \le \sqrt{x}$, we deduce that there is some $c > 0$ such that \eqref{equation_coexploration_travels_1_c} holds if $m - m_0 \ge c ~ \epsilonunif(m)^{- 8 \abs{\states}}$. 
        As $\epsilonunif(m)^{-8\abs{\states}} = \oh(\log(m))$ when $m \to \infty$ by \Cref{assumption_regularization_hyperparameter} (\strong{A4}), we have $\epsilonunif(m)^{-8\abs{\states}} \le \log(m)/c$ provided that $m$ is large enough, say $m \ge \kappa_\EPSILON$. 
        We obtain the sufficient condition $m - m_0 \ge \log(m)$ that is again satisfied provided that
        \begin{equation*}
            m - m_0 \ge 2 \log(m_0)
            \quad\text{and}\quad
            m - m_0 \ge 2 \log \parens*{1 + \frac{m - m_0}{m_0}}
            .
        \end{equation*}
        The second is always satisfied provided that $m$ is large enough. 
        We conclude that there is some $\kappa_\EPSILON \in \RR_+$ such that \eqref{equation_coexploration_travels_1_c} is satisfied as soon as $m \ge 2 \log(m_0) + \kappa_\EPSILON$.

        The second condition (\textit{ii}) solves as $\delta \ge \exp(-c (m - m_0) \epsilonunif(m)^{4 \abs{\states}})$ for some constant $c > 0$. 
        We conclude by using that $\psi_{m}(x) = (x - m_0) \epsilonunif(x)^{4\abs{\states}}$ is asymptotically increasing when $x \to \infty$, see \Cref{assumption_regularization_hyperparameter} (\strong{A6}):
        There exist constants $\alpha' \in (0, 1]$ and $\kappa_\EPSILON \in \RR_+^*$ such that 
        \begin{align*}
            & \Pr \parens*{
                \exists m' \ge m,
                I(m') \le \alpha \epsilonunif(m')^{2 \abs{\states}} (m' - m_0)
            }
            \\
            & \le
            \exp \parens*{
                - 
                \indicator{m \ge 2 \log(m_0) + \kappa_\EPSILON}
                \cdot 
                \alpha'~\epsilonunif(m)^{4 \abs{\states}} (m - m_0)
            }
            \\
            & \overset{(\dagger)}\le
            \exp \parens*{
                - 
                \alpha'~\epsilonunif(m)^{4 \abs{\states}} (m - m_0)
                + 2 \log(m_0) + \kappa_\EPSILON
            }
        \end{align*}
        where $(\dagger)$ follows from $\alpha'~\epsilonunif(m)^{4 \abs{\states}} \le 1$.
        This proves the claim.
    \end{subproof}

    \par
    \noindent
    \STEP{3}
    \textit{
        Denote $I^\pm(m) := \abs{\braces{i \le I(m): \tau_{j(i)} \in \stimes^\pm}}$.
        On $\event$, we have
        \begin{equation}
        \label{equation_coexploration_travels_2}
            I^\pm (m) 
            \le
            c_0 + \frac{\log\log(T)}{\log(2)} 
            \le
            \abs{\states} + 2 \log \log(T)
            .
        \end{equation}
        \vspace{-1em}
    }
    \begin{subproof}
        Let $c_t$ be the component of $\State_t$ at time $t$, if any (if it exists, it has to be unique).
        For $c = 1, \ldots, c_0$, further introduce
        \begin{equation*}
            L (t; c) 
            := 
            \log \min \braces*{
                \visits_t (\pair)
                :
                \pair \in \pairs_c
            }
        \end{equation*}
        the minimal visit count of component $c$ at time $t$.
        Note that under the event $\event := (\forall \ell \ge m, \flatme{\optpairs}{\iepsilonflat''}(\hat{\model}_{\tau_\ell}) = \optpairs(\model))$ and by definition of $j(-)$, see \eqref{equation_coexploration_travels_0_a}, $c_{\tau_{j(i)}}$ is well-defined for $i \ge 2$ and $c \equiv c_{\tau_{j(i+1)}}$ is a least visited component of time $\tau_{j(i)}$. 
        Under $\event \cap (\tau_{j(i+1)} \in \stimes^\pm)$, all the time instants $t \in \braces{\tau_{j(i+1)-1}+1, \tau_{j(i+1)-1}+2, \ldots, \tau_{j(i+1)-1}}$ are exploitation times that play the uniform policy on $\pairs_c$, and $\tau_{j(i+1)} \in \stimes^{\pm}$ means that $\pairs_c$ is over-visited compared to other components of $\optpairs(\model)$.
        Moreover, under $\event$, the algorithm cannot panic by running such a policy on $\pairs_c$ because no transition discovery may ever lead to a state outside of $\states(\pairs_c)$.
        In other words, under $\event \cap (\tau_{j(i+1)} \in \stimes^\pm)$, we have
        \begin{equation}
        \label{equation_coexploration_travels_2_a}
            L(\tau_{j(i+1)}; c)
            >
            2 \min_{c' \in \braces{1, \ldots, c_0}}
            L(\tau_{j(i+1)}; c')
            \ge
            2 \min_{c' \in \braces{1, \ldots, c_0}}
            L(\tau_{j(i)}; c')
            = 2 L(\tau_{j(i)}; c).
        \end{equation}
        From \eqref{equation_coexploration_travels_2_a}, we conclude that under $\event$, every time $\tau_{j(i+1)} \in \stimes^\pm$, the least visited component of time $\tau_{j(i)}$ has doubled its visited count by time $\tau_{j(i+1)}$. 
        On $\event$, we have $\min_c L(\tau_m; c) \ge 2^{\floor{I^\pm(m)/c_0} - 1}$.
        Moreover, for $\tau_m \le T$, we have $L(\tau_m; c) \le \log(T)$ for all $c = 1, \ldots, c_0$.
        Solving in $I^\pm(m)$, we obtain the claim.
    \end{subproof}

    Now, we conclude the proof of \Cref{lemma_coexploration_travels} by combining \eqref{equation_coexploration_travels_2} from \STEP{3} together with \eqref{equation_coexploration_travels_1} from \STEP{2}.

    For $m \ge m_0$ and as $m \to \infty$, we have
    \begin{align*}
        & \exp \parens*{
            - \beta \epsilonunif(m)^{4 \abs{\states}} (m - m_0)
            + 2 \log(m_0) + \kappa_\EPSILON
        }
        \\
        & \overset{(\dagger)}\ge
        \Pr \parens*{
            \exists m' \ge m,
            m'
            \ge 
            m_0 + 
            \frac 1\alpha 
            \epsilonunif(m')^{-2 \abs{\states}}
            I(m') 
        }
        \\
        & \overset{(\ddagger)}\ge
        \Pr \parens*{
            T^- 
            \ge 
            m + \frac {\epsilonunif(T^-)^{-2 \abs{\states}}}\alpha  
            I(T^-) 
        }
        \\
        & \ge
        \Pr \parens*{
            T^- 
            \ge 
            m + \frac {\epsilonunif(T^-)^{-2 \abs{\states}}}\alpha  
            \sum_{i=1}^{I(T^-)}
            \parens*{
                \eqindicator{
                    \tau_{j(i)} \in \stimes^\pm
                } + \eqindicator{
                    \tau_{j(i)} \notin \stimes^\pm
                }
            }
            , \event
        }
        \\
        & \overset{(\S)}\ge
        \Pr \parens*{
            T^- 
            \ge 
            m + 
            \frac {\epsilonunif(T^-)^{-2 \abs{\states}}}\alpha  
            \parens*{
                \abs{\states} + 2 \log \log(T) 
                +
                \sum_{i=1}^{I(T^-)}
                \eqindicator{
                    \tau_{j(i)} \notin \stimes^\pm
                }
            }
            , \event
        }
        \\
        & \overset{(\$)}\ge
        \Pr \parens*{
            T^- 
            \ge 
            m + 
            \frac {\epsilonunif(T^-)^{-2 \abs{\states}}}{\alpha/2 \cdot \abs{\states}^{-1}}
            \parens*{
                \log \log(T) 
                +
                \sum_{\ell=1}^{T^-}
                \eqindicator{
                    \State_{\tau_\ell} \in \states(\optpairs(\model))
                    \atop
                    \tau_\ell \notin \stimes^\pm
                }
            }
            , \event
        }
    \end{align*}
    where
    $(\dagger)$ follows from \STEP{2};
    $(\ddagger)$ instantiates $m' = T'$ and simplifies the expression with straight-forward algebra;
    $(\S)$ unfolds $I^\pm (m) = \sum_{j=1}^{I(m)} \indicator{\tau_{j(i)} \in \stimes^\pm}$ then invokes \eqref{equation_coexploration_travels_2} from \STEP{3}; and
    $(\$)$ uses that by definition of $j(-)$, see \eqref{equation_coexploration_travels_0_a}, we have $\State_{\tau_{j(i)}} \in \states(\optpairs(\model))$ almost surely.
    This proves \Cref{lemma_coexploration_travels}.
\end{proof}

\subsection{Amount of panic times}
\label{appendix_panic_times}

Bounding the amount of panic times is much easier.

\begin{lemma}
\label{lemma_panic_times}
    We have $\sum_{t=1}^{T-1} \indicator{t \in \stimes^!} \le \abs{\pairs} \abs{\states}$ almost surely.
\end{lemma}
\begin{proof}
    By construction of the \texttt{ECoE*}, \Cref{algorithm_ecoe_annotated} can only panic if it has discovered a transition $(\State_t, \Action_t, \State_{t+1})$ that it has never seen before. 
    As $(\State_t, \Action_t, \State_{t+1})$ lives in the space $\pairs \times \states$, we deduce that there is a injection $\stimes^! \to \pairs \times \states$.
    So $\abs{\stimes^!} \le \abs{\pairs} \abs{\states}$. 
\end{proof}

\subsection{Combining everything together and proving \Cref{theorem_visit_suboptimal_precise}}

Now, \Cref{theorem_visit_suboptimal_precise} is a direct consequence of \Cref{lemma_wrong_coexploration,lemma_exploration,lemma_panic_times}.
Indeed, for $\pair \notin \optpairs(\model)$ and applying \Cref{lemma_wrong_coexploration,lemma_exploration,lemma_panic_times}, we obtain
\begin{align*}
    & \EE \brackets*{
        \visits_T (\pair)
    }
    =
    \EE \brackets*{
        \sum_{t=1}^{T-1}
        \eqindicator{
            \Pair_t = \pair
            \atop
            t \in \stimes^+
        }
    }
    + \EE \brackets*{
        \sum_{t=1}^{T-1}
        \eqindicator{
            \Pair_t = \pair
            \atop
            t \in \stimes^-
        }
    }
    + \EE \brackets*{
        \sum_{t=1}^{T-1}
        \eqindicator{
            \Pair_t = \pair
            \atop
            t \in \stimes^!
        }
    }
    \\
    & \le
    \frac{\imeasure^*(\pair) \log(T)}{\ivalue(\imeasure^*, \model)}
    + \OH \parens*{
        \frac 1{\epsilonflat(T)^6}
        + \sum_{t=1}^{T-1} \exp\parens*{
            - \parens*{1 + \frac 12 \epsilontest(t)} \log(t)
        }
    }
    + \abs{\pairs} \abs{\states}
    + \oh(\log(T))
    .
\end{align*}
This proves \Cref{theorem_visit_suboptimal_precise}.

\subsection{Auxiliary: Adaptations of standard concentration results}

In this paragraph, we adapt a few classical probability results (see \Cref{appendix_probability}) into the versions that are used all along the proof of \Cref{theorem_visit_suboptimal_precise}.

\begin{lemma}
\label{lemma_threshold_concentration}
    Let $m \ge 1$ and $d \ge 2$.
    Fix $q$ a distribution on $\braces{1, \ldots, d}$ and denote $q^n$ the empirical distribution on $\braces{1, \ldots, d}$ obtained after $n$ i.i.d.~samples of $q$. 
    For every sequence $(\epsilon_n) \in (\RR_+^*)^\NN$, we have
    \begin{equation*}
        \Pr \parens*{
            \exists n \ge m
            :
            \norm{q^n - q}_1 > \epsilon_n
        }
        \le 
        \sup_{n \ge m}
        \exp\parens*{
            - 2 n \epsilon_n^2 
            + \frac 12 \log(1 + n)
            + d \log(2)
        }
        .
    \end{equation*}
\end{lemma}
\begin{proof}
    By \Cref{lemma_weissman_time_uniform}, we have
    \begin{align*}
        \delta 
        & \ge 
        \Pr \parens*{
            \exists n \ge m,
            \norm*{
                n(q^n - q)
            }_1^2
            \ge
            \tfrac 14 n \parens*{1 + \tfrac 1n} 
            \log \parens*{
                \tfrac{2^d \sqrt{1 + n}} \delta
            }
        }
        \\
        & \ge
        \Pr \parens*{
            \exists n \ge m,
            \norm*{
                q^n - q
            }_1^2
            \ge
            \tfrac 1{2n} 
            \log \parens*{
                \tfrac{2^d \sqrt{1 + n}} \delta
            }
        }
        \\
        & \overset{(\dagger)}\ge
        \Pr \parens*{
            \exists n \ge m,
            \norm*{q^n - q}_1^2
            \ge
            \epsilon_n^2
        }
    \end{align*}
    where $(\dagger)$ holds if $\frac 1{2n} \log(2^d\sqrt{1+n}/\delta) \le \epsilon^2_n$ for all $n \ge m$.
    We conclude by solving in $\delta$. 
\end{proof}

\begin{corollary}[Threshold concentration]
\label{corollary_threshold_concentration}
    Let $m \ge 1$ and $d \ge 2$.
    Fix $q$ a distribution on $\braces{1, \ldots, d}$ and denote $q^n$ the empirical distribution on $\braces{1, \ldots, d}$ obtained after $n$ i.i.d.~samples of $q$. 
    For all $\epsilon > 0$, we have
    \begin{equation*}
        \Pr \parens*{
            \exists n \ge m
            :
            \norm{q^n - q}_1 > \epsilon
        }
        \le 
        \exp\parens*{
            - 2 m \epsilon^2 
            + \frac 12 \log(1 + m)
            + d \log(2)
        }
        .
    \end{equation*}
\end{corollary}
\begin{proof}
    Starting from \Cref{lemma_threshold_concentration}, we introduce the function
    \begin{equation*}
        \varphi(n)
        :=
        \exp \parens*{
            - 2 n \epsilon^2 + \frac 12 \log(1 + n) + d \log(2)
        }
        .
    \end{equation*}
    The function $\varphi$ is decreasing in $[\frac 1{4\epsilon^2} - 1, \infty)$ and satisfies $\varphi \ge 4$ on $[0, \frac 1{4\epsilon^2}-1]$, so that we may set $\inf_{n \ge m} \varphi(n) = \varphi(m) \equiv \exp(-2m\epsilon^2 + \frac 12 \log(1 + m) + d \log(2))$. 
\end{proof}

\begin{lemma}[Converting $\snorm{-}$ to $\norm{-}$]
\label{lemma_convert_snorm_to_norm}
    Let $\model \equiv (\pairs, \kernel, \reward)$ and $\model' \equiv (\pairs, \kernel', \reward')$ be two Markov decision processes with $\model' \ll \model$. 
    For all $\pair \in \pairs$, if $\snorm{\model'(\pair) - \model(\pair)} < \dmin(\model)$, then $\snorm{\model'(\pair) - \model(\pair)} = \norm{\model'(\pair) - \model(\pair)}$.
\end{lemma}
\begin{proof}
    By definition, we have 
    \begin{align*}
        \snorm{\model'(\pair) - \model(\pair)} 
        & = 
        \snorm{\reward'(\pair) - \reward(\pair)}_\infty 
        + \snorm{\kernel'(\pair) - \kernel(\pair)}_1,
        \text{~and}
        \\
        \norm{\model'(\pair) - \model(\pair)} 
        & = 
        \norm{\reward'(\pair) - \reward(\pair)}_\infty 
        + \norm{\kernel'(\pair) - \kernel(\pair)}_1
        .
    \end{align*}
    By assumption, $\model'(\pair) \ll \model(\pair)$ so we just have to show that $\model(\pair) \ll \model'(\pair)$. 
    Assume that $\norm{\model'(\pair) - \model(\pair)} < \dmin(\model)$.
    For kernels, we have 
    \begin{align*}
        \max_{\state \in \states}
        \abs{\kernel'(\state|\pair) - \kernel(\state|\pair)}
        & \le 
        \norm{\kernel'(\pair) - \kernel(\pair)}_1
        < \dmin(\model) \le \dmin(\kernel(\pair))
    \end{align*}
    so $\kernel'(\state|\pair) > 0$ for all $\state \in \support(\kernel(\pair))$.
    So $\kernel'(\pair) \gg \kernel(\pair)$.
    With the same rationale, we show that $\reward'(\pair) \gg \reward(\pair)$ and conclude that $\model'(\pair) \gg \model(\pair)$.
\end{proof}

\subsection{Auxiliary: Standard probability results}
\label{appendix_probability}

In this paragraph, we recall a few classical probability results. 

\begin{lemma}[Time-uniform Azuma-Hoeffding, \cite{bourel_tightening_2020}]
\label{lemma_azuma_time_uniform}
    Let $(U_t)$ be a martingale difference sequence sequence with $\sigma$-subgaussian increments, i.e., for all $\lambda \in \R$, we have $\EE[\exp(\lambda U_n)|U_1, \ldots, U_{n-1}] \le \exp(\frac{\lambda^2 \sigma^2}2)$.
    Then, for all $\delta > 0$, we have
    \begin{equation*}
        \Pr \parens*{
            \exists n \ge 1,
            \quad
            \parens*{
                \sum_{k=1}^n U_k
            }^2
            \ge
            n \sigma^2 \parens*{
                1 + \frac 1n
            } \log \parens*{
                \frac{\sqrt{1 + n}}\delta
            }
        }
        \le
        \delta
        .
    \end{equation*}
\end{lemma}

\begin{lemma}[Time-uniform Weissman, \cite{boone_logarithmic_2025}]
\label{lemma_weissman_time_uniform}
    Let $d \ge 2$ and $q$ be a distribution over $\braces{1, \ldots, d}$.
    Let $(U_t)$ be a sequence of i.i.d.~random variables of distribution $q$.
    Then, for all $\delta > 0$, we have
    \begin{equation*}
        \Pr \parens*{
            \exists n \ge 1,
            \quad
            \norm*{
                \sum_{k=1}^n 
                \parens*{
                    \unit_{U_k} - q
                }
            }_1^2
            \ge
            \frac 14 n \parens*{1 + \frac 1n} 
            \log \parens*{
                \frac{2^d \sqrt{1 + n}} \delta
            }
        }
        \le
        \delta
        .
    \end{equation*}
\end{lemma}
\begin{proof}
    Note that 
    \begin{equation*}
        \norm*{
            \sum_{k=1}^n (\unit_{U_k} - q)
        }_1 
        = 
        \max_{v \in \braces{-1, 1}^d} 
        \braces*{
            \sum_{k=1}^n (\unit_{U_k} - q) \cdot v
        }
        .
    \end{equation*}
    Let $W_k^v := (\unit_{U_k} - q) \cdot v$.
    Then, for each $v \in \braces{-1, 1}^d$, $(W_k^v)$ is a family of i.i.d.~random variables with $\EE[W_k^v] = 0$ and $- qv \le W_k^v \le 1 - qv$ a.s., so $\EE[\exp(\lambda W_k^v)] \le \exp(\lambda^2/8)$ by Hoeffding's Lemma. 
    So, by a time-uniform Azuma-Hoeffding inequality (\Cref{lemma_azuma_time_uniform}), we obtain
    \begin{align*}
        & 
        \Pr \parens*{
            \exists n \ge 1,
            \quad
            \norm*{
                \sum_{k=1}^n 
                \parens*{
                    \unit_{U_k} - q
                }
            }_1^2
            \ge
            \frac 14 n \parens*{1 + \frac 1n} 
            \log \parens*{
                \frac{2^d \sqrt{1 + n}} \delta
            }
        }
        \\
        & = 
        \Pr \parens*{
            \exists v \in \braces{-1, 1}^d,
            \exists n \ge 1,
            \quad
            \parens*{
                \sum_{k=1}^n W_k^v
            }^2
            \ge 
            \frac 14 n \parens*{1 + \frac 1n} 
            \log \parens*{
                \frac{2^d \sqrt{1 + n}} \delta
            }
        }
        \\
        & \le
        \sum_{v \in \braces{-1, 1}^d}
        \Pr \parens*{
            \exists n \ge 1,
            \quad
            \parens*{
                \sum_{k=1}^n W_k^v
            }^2
            \ge 
            \frac 14 n \parens*{1 + \frac 1n} 
            \log \parens*{
                \frac{2^d \sqrt{1 + n}} \delta
            }
        }
        \\
        & \le
        \sum_{v \in \braces{-1, 1}^d}
        \Pr \parens*{
            \exists n \ge 1,
            \quad
            \parens*{
                \sum_{k=1}^n W_k^v
            }^2
            \ge 
            \frac 14 n \parens*{1 + \frac 1n} 
            \log \parens*{
                \frac{\sqrt{1 + n}}{2^{-d} \delta}
            }
        }
        \le 2^d \cdot 2^{-d} \delta = \delta.
    \end{align*}
    This concludes the proof.
\end{proof}

\begin{lemma}[All-time Sanov]
\label{lemma_sanov_all_time}
    Let $\model \equiv (\pairs, \rewardd, \kernel)$ be a Markov decision process with Bernoulli rewards.
    Fix $n \in (\N^*)^{\pairs}$ and denote $\hat{\model}_{(n)}$ the empirical model obtained obtained by sampling independently $n(\pair)$ times from $\rewardd(\pair) \otimes \kernel(\pair)$.
    Let $\models^{(n)}$ be the discrete space of all possible values of $\hat{\model}_{(n)}$.
    For all $\model' \in \models^{(n)}$, we have
    \begin{equation*}
        \Pr \parens*{
            \hat{\model}_{(n)} = \model'
        }
        \le 
        \exp \parens*{
            - \sum_{\pair \in \pairs} 
            \KL(\model'(\pair)||\model(\pair))
        }
    \end{equation*}
    where $\KL(\model'(\pair)||\model(\pair)) := \KL(\rewardd'(\pair)||\rewardd(\pair)) + \KL(\kernel'(\pair)||\kernel(\pair))$. 
\end{lemma}
\begin{proof}
    This result is conveniently obtained with the method of types, see \cite{csiszar_simple_2006} for example.
    For $m \ge 1$, write $\reward_{(m)}(\pair)$ and $\kernel_{(m)}(\pair)$ the empirically observed reward and kernel after $m$ independent samples of $\rewardd(\pair) \otimes \kernel(\pair)$. 
    We have
    \begin{equation}
    \label{equation_sanov_all_time_1}
        \Pr \parens*{
            \hat{\model}_{(n)} = \model'
        }
        =
        \product_{\pair \in \pairs}
        \Pr \parens*{
            \hat{\reward}_{(n(\pair))}(\pair) = \reward'(\pair)
        }
        \Pr \parens*{
            \hat{\kernel}_{(n(\pair))}(\pair) = \kernel'(\pair)
        }
        .
    \end{equation}
    Let $\pair \in \pairs$.
    By definition, $\reward'(\pair)$ is of the form $\mathrm{Ber}(k(\pair)/n(\pair))$ for $k \in \braces{0, \ldots, n(\pair)}$.
    So,
    \begin{align}
    \notag
        \Pr \parens*{
            \hat{\reward}_{(n(\pair))}(\pair) = \reward'(\pair)
        }
        & =
        \binom{n(\pair)}{k(\pair)}
        \reward(\pair)^{k(\pair)}
        (1 - \reward(\pair))^{n(\pair) - k(\pair)}
        \\
    \notag
        & = 
        \exp \parens*{
            - n \KL(\rewardd'(\pair)||\rewardd(\pair))
        }
        \cdot 
        \binom{n(\pair)}{k(\pair)}
        \exp \parens*{
            - n(\pair) \entropy(\reward'(\pair))
        }
        \\
    \label{equation_sanov_all_time_2}
        & \overset{(\dagger)}\le
        \exp\parens*{
            - n(\pair) \KL(\rewardd'(\pair)||\rewardd(\pair))
        }
    \end{align}
    where $(\dagger)$ follows from the classical inequality $\binom{a}{b} \le \exp(a \entropy(\frac{b}{a}))$ where $\entropy(-)$ is the entropy in base $e$. 
    With the same computation for kernels, we find that
    \begin{equation}
    \label{equation_sanov_all_time_3}
        \Pr \parens*{
            \hat{\kernel}_{(n(\pair))}(\pair) = \kernel'(\pair)
        }
        \le
        \exp \parens*{
            - n(\pair) \KL(\kernel'(\pair)||\kernel(\pair))
        }
    \end{equation}
    for all $\pair \in \pairs$.
    Combining \eqref{equation_sanov_all_time_1}, \eqref{equation_sanov_all_time_2} and \eqref{equation_sanov_all_time_3}, we conclude accordingly.
\end{proof}

\begin{lemma}[Combinatorics, folklore]
\label{lemma_combinatorial_bounds}
    Let $k \ge 1$ and denote $\probabilities_n[k]$ the set of probability distributions over $\braces{1, \ldots, k}$ of the form $\parens{\frac{n_1}{n}, \ldots, \frac{n_k}{n}}$ for $n_i \in \N$. 
    We have $\abs{\probabilities_n[k]} \le (n + 1)^k$.
\end{lemma}

    \clearpage
    \section{Deviation bounds on MDP specific quantities}
    \label{appendix_mdpart}

The goal of this section is to provide a complete machinery to bound the variations of various policy related quantities, such as the gain, the bias, the diameter, the invariant measure and the reaching probabilities to components of recurrent sets. 
One important consequence of the developed results is \Cref{proposition_continuity_near_optimality}, consisting in that if $\model, \model'$ are two Markov decision processes, then
\begin{equation}
    \flatme{\optpairs}{\epsilon}(\model') = \optpairs(\model)
    \quad\text{when}\quad
    C(\model)
    \norm{\model' - \model}^\opt
    \ll 
    \epsilon
\end{equation}
where $C(\model)$ is directly related to $\wdiameter(\model)$, that is a notion of \textbf{worst diameter} (see \Cref{section_leveling} and \Cref{proposition_continuity_near_optimality}), and $\epsilon > 0$ is smaller than the gain gap of~$\model$.

All bounds are developed using a martingale technique coupled with a Poisson equation, sometimes under a transform of the initial kernel and reward function.
To provide a general flavor of the technique, consider two Markov reward processes $(\reward_1, \kernel_1)$ and $(\reward_2, \kernel_2)$, and denote $\gain_j, \bias_j$ their respective gain and bias functions.
We have the Poisson equation $\gain_j(\state) + \bias_j(\state) = \reward_j(\state) + \kernel_j(\state) \bias_j$.
We write $\State_t$ the random state at time $t$, and $\Pr_{\state}^{\kernel_j}(-), \E_{\state}^{\kernel_j}[-]$ the probability and expectation of the trajectory governed by $\kernel_j$ initialized at $\State_1 = \state$.
Consider a stopping time $\tau$ such that $\E_{\state}^{\kernel_2}[\tau] < \infty$.
The heart of the technique lies in the following computation:
\begin{equation}
\nonumber
\begin{split}
    & \E_{\state}^{\kernel_2} \brackets*{
        \sum_{t=1}^{\tau-1} \reward_2(\State_t)
    }
    \\
    & =
    \E_{\state}^{\kernel_2} \brackets*{
        \sum_{t=1}^{\tau-1} \parens*{\reward_2(\State_t) - \reward_1(\State_t)}
    }
    +
    \E_{\state}^{\kernel_2} \brackets*{
        \sum_{t=1}^{\tau-1} \reward_1(\State_t)
    }
    \\
    & \overset{(\dagger)}=
    \E_{\state}^{\kernel_2} \brackets*{
        \sum_{t=1}^{\tau-1} \parens*{\reward_2(\State_t) - \reward_1(\State_t)}
    }
    +
    \E_{\state}^{\kernel_2} \brackets*{
        \sum_{t=1}^{\tau-1} 
        \parens*{
            \gain_1(\State_t)
            + \parens*{
                \mathbf{e}_{\State_t} - \kernel_1(\State_t)
            } \bias_1
        }
    }
    \\
    & =
    \E_{\state}^{\kernel_2} \brackets*{
        \sum_{t=1}^{\tau-1} \gain_1(\State_t)
    }
    +
    \E_{\state}^{\kernel_2} \brackets*{
        \sum_{t=1}^{\tau-1} \parens*{\reward_2(\State_t) - \reward_1(\State_t)}
    }
    +
    \E_{\state}^{\kernel_2} \brackets*{
        \sum_{t=1}^{\tau-1} \parens*{\kernel_2(\State_t) - \kernel_1(\State_t)} \bias_1
    }
    \\
    & 
    \phantom{
        {} =
        \E_{\state}^{\kernel_2} \brackets*{
        \sum_{t=1}^{\tau-1} \gain_1(\State_t)
        }
    }
    + \E_{\state}^{\kernel_2} \brackets*{
        \bias_1(\state) - \bias_1(\State_{\tau})
    }
    \\
    & \overset{(\ddagger)}\le 
    \E_{\state}^{\kernel_2} \brackets*{
        \sum_{t=1}^{\tau-1} \gain_1(\State_t)
    }
    + \E_{\state}^{\kernel_2} \brackets*{
        \bias_1(\state) - \bias_1(\State_{\tau})
    }
    + \E_{\state}^{\kernel_2} \brackets*{\tau} \parens*{
        \norm{\reward_2 - \reward_1}_\infty
        + 
        \tfrac 12 \vecspan{\bias_1} \norm{\kernel_2 - \kernel_1}_\infty
    }
\end{split}
\end{equation}
where $(\dagger)$ follows from the Poisson equation of $(\reward_1, \kernel_1)$ and $(\ddagger)$ from standard norm bounds.
Coupled with model transformation and careful choices of stopping times, this technique provides tight bounds for the variations of the gain, bias, diameter, reaching time and invariant measures.
The obtained bounds on the variations of gain for instance (see \cref{lemma_unichain_gain_variations} and \cref{lemma_multichain_gain_variations}) are much better than those obtained using algebraic approaches involving the Drazin inverse with very little structural assumptions, going way beyond the scope of ergodic Markov reward processes.

\subsection{A multichain-friendly diameter notion for Markov chains}

The diameter provided in the definition is below generalizes the well-known notion of diameter in the communicating/recurrent setting.
It is tuned to provide a simple bound on the span of the bias function.

\begin{definition}[Diameter of a policy/Markov chain]
\label{definition_policy_diameter}
    Let $\kernel$ be the kernel of a Markov chain with recurrent components $\states_1, \ldots, \states_k$.
    The \textbf{(policy) diameter} of $\kernel$ is given by
    \begin{equation}
    \label{equation_definition_policy_diameter}
        \diameter(\kernel)
        :=
        \sup_{(\state_i) \in \prod_i \states_i}
        \sup_{\state \in \states} \E_{\state}^{\kernel}\brackets{\tau_{\set{\state_1, \ldots, \state_k}}}
        < \infty.
    \end{equation}
\end{definition}

For Markov decision processes, the policy diameter of $\policy \in \policies$ in $\model$ is also denoted $\diameter(\policy; \model) \equiv \diameter(\kernel_\policy)$, where $\kernel_\policy(\state'|\state) := \sum_{\action \in \actions(\state)} \kernel(\state'|\state,\action) \policy(\action|\state)$ is the transition kernel of $\policy$. 

When the underlying chain is irreducible (i.e., recurrent), this diameter is related to return times hence providing lower bounds on invariant measures.

\begin{lemma}[Diameter and invariant measures]
\label{lemma_diameter_invariant_measures}
    The unique probability invariant measure $\imeasure$ of a irreducible Markov chain with kernel $\kernel$ satisfies $\min(\imeasure) \ge \diameter(\kernel)^{-1}$.
\end{lemma}
\begin{proof}
    Let $\tau_\state^+ := \inf \set{t \ge 2: \State_t = \state}$ the first return time to $\state \in \states$.
    It is known, see for e.g.~\cite[Proposition~1.19]{levin_markov_2017}, that $\imeasure(\state) = (\EE_{\state}^{\kernel}[\tau_\state^+] - 1)^{-1}$.
    Now, note that
    \begin{equation*}
        \EE_{\state}^{\kernel}[\tau_{\state}^+]
        \le
        1 
        + \max_{\state ' \ne \state} \EE_{\state'}^{\kernel}[\inf\set{t \ge 1 : \State_t = \state}]
        \le
        1 + \diameter(\kernel)
    \end{equation*}
    and conclude accordingly. 
\end{proof}

\subsubsection{Policy diameter and diameters of Markov decision processes}
\label{appendix_diameter_and_policy_diameter}

The policy diameter is related to the standard notion of diameter in Markov decision processes.
Recall that the \strong{diameter} of a Markov decision process $\model$ is given by $\diameter(\model) := \max_{\state \ne \state'} \min_{\policy \in \policies} \E_{\state}^{\policy, \model}[\inf\braces{t \ge 1 : \State_t = \state'}]$ and is finite if, and only if $\model$ is communicating. 
Note that, for fixed $\state \ne \state'$, the inner minimum can be taken over the set of unichain policies, i.e., policies with a unique recurrent component on $\model$; Indeed, by taking a policy $\policy$ achieving the minimum, we can modify $\policy$ into $\policy'$ by setting $\policy'(\state'') = \mathrm{Uniform}(\actions(\state''))$ for every $\state'' \in \states$ from which $\state'$ is not reachable, i.e., $\Pr_{\state''}^{\policy, \model}(\inf\braces{t \ge 2: \State_t = \state'}) = 0$, and  check that the reaching time to $\state'$ from $\state$ has to be the same under $\policy$ and $\policy'$. 
Then, note that if $\state'$ is recurrent under a unichain policy $\policy \in \policies$, then $\E_{\state}^{\policy, \model}[\inf\braces{t \ge 1: \State_t = \state'}] \le \diameter(\policy; \model)$. 
This shows that the diameter of $\model$ is upper-bounded relatively to the policy diameters of its unichain policies as given by
\begin{equation}
\label{equation_diameter_and_policy_diameter}
    \diameter(\model)
    \le
    \max_{\state' \in \states}
    \min \braces[\Big]{
        \diameter(\policy; \model)
        :
        \policy
        \text{ unichain and $\state'$ recurrent under $\policy$}
    }
    .
\end{equation}
In particular, we have the following result.

\begin{proposition}[Diameter and worst diameter]
\label{proposition_diameter_and_worst_diameter}
    In communicating Markov decision processes, the diameter is bounded by the worst diameter, i.e., $\diameter(\model) \le \worstdiameter(\model)$.
\end{proposition}

The inequality in \eqref{equation_diameter_and_policy_diameter} is strict in general; Take for instance the deterministic transition Markov decision process induced by the complete graph with 3 vertices (the undirected $3$-cycle).
However, if we denote $\model_\state$ the copy of $\model$ obtained by making $\state$ absorbing, then it can be shown that
\begin{equation}
\label{equation_diameter_and_absorbing_policy_diameter}
    \diameter(\model)
    =
    \max_{\state' \in \states}
    \min \braces[\Big]{
        \diameter(\policy; \model_{\state'})
        :
        \policy
        \text{ unichain and $\state'$ recurrent under $\policy$ in $\model$}
    }
    .
\end{equation}
Indeed, regarding \eqref{equation_definition_policy_diameter}, the set of recurrent states is reduced to $\braces{\state'}$ because if $\state'$ is recurrent under some unichain policy $\policy$, then it is the only recurrent state of $\policy$ in $\model_{\state'}$.

\subsubsection{Policy diameter and bias function}

Following these observations, we can generalize the links between the bias function and the diameter to the policy diameter as follows.

\begin{lemma}[Policy bias and diameter]
\label{lemma_policy_bias_diameter}
    Let $(\reward, \kernel)$ be a Markov reward process and denote $\gain, \bias$ its gain and bias functions.
    Then $\vecspan{\bias} \le 2 ~ \vecspan{r}\diameter(\kernel)$.
\end{lemma}

\begin{proof}
    For $(\state_i) \in \prod_{i=1}^k \states_i$ a covering of the recurrent components, we denote $\tau \equiv \tau_{\set{\state_1, \ldots, \state_k}}$ for short.
    We have $\E_{\state}^{\kernel}[\tau] \le \diameter(\kernel) < \infty$ for all $\state \in \states$ by construction.
    Because $\imeasure \bias = 0$ for every invariant measure of $\kernel$, we see that for all $i = 1, \ldots, k$, there must be $\state_i \in \states_i$ such that $\bias(\state_i) \ge 0$, and another $\state_i \in \states_i$ such that $\bias(\state_i) \le 0$.
    Assume that $\bias(\state_i) \le 0$ for all $i = 1, \ldots, k$.
    Then
    \begin{equation}
    \label{equation_policy_bias_diameter_1}
    \begin{split}
        \max(\reward - \gain) \diameter(\kernel)
        & \ge
        \E^{\kernel}_{\state} \brackets*{
            \sum_{t=1}^{\tau-1}
            (\reward(\State_t) - \gain(\State_t))
        }
        \\
        & \overset{(\dagger)} =
        \E^{\kernel}_{\state} \brackets*{
            \sum_{t=1}^{\tau-1}
            \parens*{\mathbf{e}_{\State_t} - \kernel(\State_t)} \bias
        }
        \\
        & = 
        \bias(\state) - \E_{\state}^{\kernel}\brackets*{\bias(\State_{\tau})}
        =
        \bias(\state)
        -
        \sum_{i=1}^k \Pr^{\kernel}_{\state}\parens*{\tau_{\states_i} < \infty} \bias(\state_i)
        \overset{(\ddagger)} \ge 
        \bias(\state).
    \end{split}
    \end{equation}
    where $(\dagger)$ follows from the Poisson equation $\gain(\state) + \bias(\state) = \reward(\state) + \kernel(\state) \bias$ and $(\ddagger)$ by $\bias(\state_i) \le 0$.
    With similar computations, and picking $\state_i$ such that $\bias(\state_i) \ge 0$ for $i = 1,\ldots,k$, we have
    \begin{equation}
    \label{equation_policy_bias_diameter_2}
        \min(\reward - \gain) \diameter(\kernel)
        \le 
        \bias(\state)
        -
        \sum_{i=1}^k \Pr^{\kernel}_{\state}\parens*{\tau_{\states_i} < \infty} \bias(\state_i)
        \le 
        \bias(\state).
    \end{equation}
    Combining \eqref{equation_policy_bias_diameter_1} and \eqref{equation_policy_bias_diameter_2}, we obtain
    \begin{equation}
        \max(\bias) - \min(\bias) 
        \le 
        \diameter(\kernel) \parens*{
            \max(\reward - \gain) 
            - 
            \min(\reward - \gain)
        }
        \le 2 ~ \vecspan{r} \diameter(\kernel)
    \end{equation}
    where the second inequality uses that $\min(\reward) \le \gain(\state) \le \max(\reward)$ for all $\state$.
\end{proof}

\subsubsection{Sensitivity of the policy diameter to perturbations}

In this paragraph, we quantify the sensitivity of hitting times and of the policy diameter to perturbations of the transition kernel (\Cref{lemma_policy_diameter_variations}).
For the general setting, we make a absolute continuity assumption.
This assumption can be dropped completely under an ergodic assumption (\Cref{lemma_policy_recurrent_diameter_variations}), and partially under a unichain assumption (\Cref{corollary_unichain_diameter_variations}).

\begin{lemma}[Variations of policy diameter]
\label{lemma_policy_diameter_variations}
    Let $\kernel_1 \sim \kernel_2$ be two equivalent Markov chains with (common) recurrent components $\states_1, \ldots, \states_k$.
    For all $(\state_i) \in \prod_{i=1}^k \states_i$ covering of the recurrent components, the variations of the hitting time $\tau \equiv \tau_{\set{\state_1, \ldots, \state_k}}$ are bounded as:
    \begin{equation}
        \abs*{
            \sup_{\state} \E_{\state}^{\kernel_2}[\tau]
            - 
            \sup_{\state} \E_{\state}^{\kernel_1}[\tau]
        }
        \le
        \frac 12 
        \sup_{\state} \E_{\state}^{\kernel_2}[\tau]
        \sup_{\state} \E_{\state}^{\kernel_1}[\tau]
        \norm{\kernel_2 - \kernel_1}_\infty.
    \end{equation}
    In particular $\diameter(\kernel_2) \le \diameter(\kernel_1) + \frac 12 \diameter(\kernel_1) \diameter(\kernel_2) \norm{\kernel_2 - \kernel_1}_\infty$.
\end{lemma}
\begin{proof}
    Let $(\state_i) \in \prod_{i=1}^k \states_i$ be a covering of the recurrent components and denote $\tau \equiv \tau_{\set{\state_1, \ldots, \state_k}}$ for short.
    Consider the kernel $\kernel'_j$ obtained by making every $\state_i$ absorbing and consider the reward function $\reward'(\state) := \indicator{\state \notin \set{\state_1, \ldots, \state_k}}$.
    The gain and bias functions of the Markov reward process $(\reward', \kernel'_j)$ are denoted $\gain'_j$ and $\bias'_j$ for $j = 1,2$.
    Remark that $\gain'_j = 0$ and that the bias satisfies
    \begin{equation}
    \label{equation_policy_diameter_variations_1}
        \bias'_j(\state) 
        = 
        \Clim_{T \to \infty} \E_{\state}^{\kernel'_j} \brackets*{
            \sum_{t=1}^{\tau \wedge T - 1}
            \parens*{\reward'(\State_t) - \gain'_j(\State_t)}
        }
        = \E_{\state}^{\kernel_j}[\tau]
    \end{equation}
    hence $\bias'_j$ are the reaching times of which we want to bound the variations.
    We have
    \begin{equation}
    \nonumber
    \begin{split}
        \bias'_2(\state)
        =
        \E_{\state}^{\kernel_2} 
        \brackets*{
            \sum_{t=1}^{\tau - 1} \reward'(\State_t)
        }
        & =
        \bias'_1(\state)
        +
        \E_{\state}^{\kernel_2}
        \brackets*{
            \sum_{t=1}^{\tau-1}
            \parens*{\kernel_2(\State_t) - \kernel_1(\State_t)}
            \bias'_1
        }
        .
    \end{split}
    \end{equation}
    The error term is bounded by $\abs{\E_{\state}^{\kernel_2} \brackets{ \sum_{t=1}^{\tau-1} \parens{\kernel_2(\State_t) - \kernel_1(\State_t)} \bias'_1}} \le \tfrac 12 \vecspan{\bias_1'} \E_{\state}^{\kernel_2}[\tau] \norm{\kernel_2 - \kernel_1}_\infty$.
    By \eqref{equation_policy_diameter_variations_1}, we clearly have $\vecspan{\bias_1'} = \max(\bias'_1) = \sup_{\state} \E^{\kernel_1}_{\state}[\tau]$.
    Accordingly, we have obtained the self-bound:
    \begin{equation}
        \abs*{
            \sup_{\state} \E_{\state}^{\kernel_2}[\tau]
            - 
            \sup_{\state} \E_{\state}^{\kernel_1}[\tau]
        }
        \le
        \tfrac 12 
        \sup_{\state} \E_{\state}^{\kernel_2}[\tau]
        \sup_{\state} \E_{\state}^{\kernel_1}[\tau]
        \norm{\kernel_2 - \kernel_1}_\infty.
    \end{equation}
    This concludes the proof.
\end{proof}

The assumption $\kernel_1 \sim \kernel_2$ is not always necessary and can be dropped under a recurrent assumption.
In turn, this allows to control the deviations of reaching times to recurrent states of unichain Markov chains. 

\begin{lemma}[Variations of policy diameter, recurrent case]
\label{lemma_policy_recurrent_diameter_variations}
    Let $\kernel_1, \kernel_2$ be two recurrent Markov chains.
    We have
    \begin{equation}
        \abs*{
            \diameter(\kernel_1)
            - 
            \diameter(\kernel_2)
        }
        \le
        \frac 12 
        \diameter(\kernel_1) \diameter(\kernel_2)
        \norm{\kernel_2 - \kernel_1}_\infty
        .
    \end{equation}
\end{lemma}
\begin{proof}
    Same proof as \cref{lemma_policy_diameter_variations}.
\end{proof}

\begin{corollary}[Variations of reaching times, unichain case]
\label{corollary_unichain_diameter_variations}
    Let $\kernel_1, \kernel_2$ be two {unichain} Markov chains that share one recurrent state $\state' \in \states$.
    Then 
    \begin{equation*}
        \abs*{
            \sup_{\state \in \states}
            \E_{\state}^{\kernel_2}[\tau_{\state'}]
            -
            \sup_{\state \in \states}
            \E_{\state}^{\kernel_1}[\tau_{\state'}]
        }
        \le
        \frac 12
        \sup_{\state \in \states}
        \E_{\state}^{\kernel_1}[\tau_{\state'}]
        \sup_{\state \in \states}
        \E_{\state}^{\kernel_2}[\tau_{\state'}]
        \norm{\kernel_2 - \kernel_1}_\infty
        .
    \end{equation*}
\end{corollary}
\begin{proof}
    For $i = 1, 2$, introduce $\kernel'_i$ the copy of $\kernel_i$ where $\state'$ is made absorbing, and apply \Cref{lemma_policy_diameter_variations} to $\kernel'_1, \kernel'_2$.
\end{proof}

In \Cref{lemma_policy_recurrent_diameter_variations}, the inequality $\diameter(\kernel_2) \le \diameter(\kernel_1) + \frac 12 \diameter(\kernel_1) \diameter(\kernel_2) \norm{\kernel_2 - \kernel_1}_\infty$ is a self-bound for $\diameter(\kernel_2)$. 
It can be decoupled when $\norm{\kernel_2 - \kernel_1}_\infty$ is small.
If $\norm{\kernel_2 - \kernel_1}_\infty < \tfrac 2{\diameter(\kernel_1)}$, we obtain
\begin{equation}
\label{equation_decoupled_diameter_bound}
    \diameter(\kernel_2) 
    \le
    \frac{1}{1 - \tfrac 12 \diameter(\kernel_1) \norm{\kernel_2 - \kernel_1}_\infty}
    \diameter(\kernel_1)
    .
\end{equation}
For instance, if $\norm{\kernel_2 - \kernel_1}_\infty \le {\diameter(\kernel_1)}^{-1}$, then $\diameter(\kernel_2) \le 2 \diameter(\kernel_1)$.
The same technique can be used to untangle the inequalities of \Cref{lemma_policy_diameter_variations,corollary_unichain_diameter_variations}.

\subsubsection{Sensitivity of the diameter of MDPs to perturbations}

Using the links between the diameter of Markov decision processes and the diameter of unichain policies, see \eqref{equation_diameter_and_policy_diameter} and \eqref{equation_diameter_and_absorbing_policy_diameter}, \Cref{lemma_policy_diameter_variations} allows to bound the variations of the diameter in a neighborhood of $\model$.

\begin{lemma}[Variations of diameter]
\label{lemma_variations_diameter}
    Let $\model \equiv (\pairs, \kernel, \reward)$ be a communicating Markov decision process. 
    For $\model' \equiv (\pairs, \kernel', \reward')$ such that $\kernel' \sim \kernel$, we have
    \begin{equation}
        \abs{\diameter(\model') - \diameter(\model)}
        \le
        \frac 12 \diameter(\model') \diameter(\model) \norm{\kernel' - \kernel}_\infty
        .
    \end{equation}
    In particular, if $\norm{\kernel' - \kernel}_\infty \le 2 \diameter(\model)^{-1}$, then $\diameter(\model') \le (1 - \frac 12 \diameter(\model) \norm{\kernel' - \kernel}_\infty)^{-1} \diameter(\model)$. 
\end{lemma}
\begin{proof}
    Following \eqref{equation_diameter_and_absorbing_policy_diameter}, we introduce $\policy_{\state'}$ the policy minimizing $\diameter(\policy; \model_{\state'})$ over the set of unichain policies under which $\state' \in \states$ is recurrent in $\model$, where $\model_{\state'} \equiv (\pairs, \kernel_{\state'}, \reward)$ is the copy of $\model$ obtained by making $\state'$ absorbing. 
    By \Cref{lemma_policy_diameter_variations}, we have
    \begin{equation*}
        \abs{
            \diameter(\policy_{\state'}; \model'_{\state'})
            - 
            \diameter(\policy_{\state'}; \model_{\state'})
        }
        \le
        \frac 12 
        \diameter(\policy_{\state'}; \model'_{\state'})
        \diameter(\policy_{\state'}; \model_{\state'})
        \norm{\kernel'_{\state'} - \kernel_{\state'}}_\infty
    \end{equation*}
    for $\state' \in \states$. 
    By construction, $\norm{\kernel'_{\state'} - \kernel_{\state'}}_\infty \le \norm{\kernel' - \kernel}_\infty$.
    Conclude by \eqref{equation_diameter_and_absorbing_policy_diameter}.
\end{proof}

In particular, if $\snorm{\model' - \model} \le {\diameter(\model)}^{-1}$, then $\diameter(\model') \le 2 \diameter(\model)$. 

\subsection{Results for unichain Markov reward processes}

The first lemma is the conclusion of the tutorial computations that start \cref{appendix_mdpart}.

\begin{lemma}[Unichain gain variations]
\label{lemma_unichain_gain_variations}
    Let $(\reward_1, \kernel_1)$ and $(\reward_2, \kernel_2)$ be two Markov reward processes.
    Denote $\gain_j, \bias_j$ the gain and bias functions of $(\reward_j, \kernel_j)$ for $j = 1, 2$.
    Assume that $\vecspan{\gain_1} = 0$.
    Then
    \begin{equation*}
        \norm*{\gain_2 - \gain_1}_\infty
        \le 
        \norm{\reward_2 - \reward_1}_\infty + \tfrac 12 \vecspan{\bias_1} \norm{\kernel_2 - \kernel_1}_\infty.
    \end{equation*}
    If in addition, $\kernel_1 \sim \kernel_2$, then $\vecspan{\bias_1}$ can be changed to $\vecspan{\bias_1|_{\states_1}}$ where $\states_1$ is the collection of recurrent states under $\kernel_1$.
\end{lemma}
\begin{proof}
    This is direct adaptation of tutorial computations.
\end{proof}

\begin{lemma}[Unichain invariant measure variations]
\label{lemma_unichain_invariant_measure_variations}
    Let $\kernel_1 \sim \kernel_2$ be two Markov chains and assume that $\kernel_1$ is unichain.\footnote{Hence $\kernel_2$ is equally unichain with the same recurrent class.}
    Denote $\imeasure_j$ the (unique) probability invariant measure under $\kernel_j$.
    We have
    \begin{equation*}
        \norm{\imeasure_2 - \imeasure_1}_\infty
        \le
        \min_{j \in \set{1,2}} \diameter(\kernel_j) 
        ~
        \norm*{\kernel_2 - \kernel_1}_\infty.
    \end{equation*}
\end{lemma}
\begin{proof}
    Fix $\state_0 \in \states$ a recurrent state.
    Consider the reward function $\reward(\state) = \indicator{\state = \state_0}$, and denote $\gain_j, \bias_j$ the gain and bias functions of the Markov reward process $(\reward, \bias_j)$.
    Remark that $\gain_j(\state) = \imeasure_j(\state_0)$.
    By \cref{lemma_policy_bias_diameter}, we have $\vecspan{\bias_j} \le 2 \vecspan{\reward} \diameter(\kernel_j) \le 2 \diameter(\kernel_j)$.
    Continuing with \cref{lemma_unichain_gain_variations}, we have
    \begin{equation*}
        \abs*{\imeasure_2(\state) - \imeasure_1(\state)}
        \le
        \diameter(\kernel_j) \norm*{\kernel_2 - \kernel_1}_\infty
    \end{equation*}
    for both $j = 1,2$. 
    This concludes the proof.
\end{proof}

\begin{lemma}[Unichain bias variations]
\label{lemma_unichain_bias_variations}
    Let $(\reward_1, \kernel_1)$ and $(\reward_2, \kernel_2)$ be two Markov reward processes, assume that $\kernel_1$ is unichain and that $\kernel_1 \sim \kernel_2$.
    Denote $\gain_j$ and $\bias_j$ the gain and bias function of $(\reward_j, \kernel_j)$ for $j = 1, 2$.
    We have
    \begin{equation*}
        \norm{\bias_2 - \bias_1}_\infty 
        \le
        4 \diameter(\kernel_2) \norm{\reward_2 - \reward_1}_\infty
        +
        \parens*{
            2 \diameter(\kernel_2) \vecspan{\bias_1} 
            + \tfrac 12 \vecspan{\bias_1^1}
        }\norm*{\kernel_2 - \kernel_1}_\infty
        .
    \end{equation*}
\end{lemma}

\begin{proof}
    We start by bounding $\vecspan{\bias_2 - \bias_1}$.
    Fix $\state_0$ a recurrent state and consider the transform $\kernel'_j$ that makes $\state_0$ absorbing as well as the reward function $\reward'_j(\state) := \indicator{\state \ne \state_0} \parens{\reward_j(\state) - \gain_j(\state)}$.
    Denote $\gain'_j$ and $\bias'_j$ the associated gain and bias functions.
    From direct computations, we see that $\gain'_j = 0$ and that $\bias'_j(\state) = \bias_j(\state) - \bias_j(\state_0)$, so in particular $\bias'_j - \bias_j \in \R \mathbf{e}$ so $\vecspan{\bias'_j} = \vecspan{\bias_j}$ and $\vecspan{\bias'_2 - \bias'_1} = \vecspan{\bias_2 - \bias_1}$.
    Moreover, $\bias'_2$ and $\bias'_1$ are related as follows:
    \begin{equation*}
    \begin{split}
        \bias'_2(\state) 
        & = 
        \E_{\state}^{\kernel'_2} 
        \brackets*{
            \sum_{t=1}^{\tau_{\state_0}-1}
            \reward'_2(\State_t)
        }
        \\
        & \le
        \E_{\state}^{\kernel_2} \brackets*{\tau_{\state_0}} 
        \norm*{\reward'_2 - \reward'_1}_\infty
        +
        \E_{\state}^{\kernel_2} 
        \brackets*{
            \sum_{t=1}^{\tau_{\state_0}-1}
            \parens*{
                \mathbf{e}_{\State_t}
                -
                \kernel_1(\State_t)
            } \bias'_1
        }
        \\
        & \le
        \bias_1'(\state)
        +
        \E_{\state}^{\kernel_2} \brackets*{\tau_{\state_0}} 
        \parens*{
            \norm*{\reward_2 - \reward_1}_\infty
            + 
            \norm*{\gain_2 - \gain_1}_\infty
            +
            \tfrac 12 \vecspan{\bias'_1} \norm*{\kernel'_2 - \kernel'_1}_\infty
        }
        \\
        & \le 
        \bias_1'(\state)
        +
        \E_{\state}^{\kernel_2} \brackets*{\tau_{\state_0}} 
        \parens*{
            2\norm*{\reward_2 - \reward_1}_\infty
            +
            \vecspan{\bias_1} \norm*{\kernel'_2 - \kernel'_1}_\infty
        }
    \end{split}
    \end{equation*}
    where the last inequality is a consequence of \cref{lemma_unichain_gain_variations} and $\vecspan{\bias'_1} \le \vecspan{\bias_1}$.
    With the same technique, we lower bound $\bias'_2(\state)$ by $\bias'_1(\state)$ with the same error term. 
    Remark that $\E_{\state}^{\kernel_j}[\tau_{\state_0}] \le \diameter(\kernel_j)$.
    So:
    \begin{equation}
    \label{equation_unichain_bias_variations_1}
    \begin{split}
        \vecspan{\bias_2 - \bias_1} = \vecspan{\bias'_2 - \bias'_1}
        & = 
        (\bias'_2 - \bias'_1)(\state_\mathrm{max}) - (\bias'_2 - \bias'_1)(\state_\mathrm{\min})
        \\
        & \le 
        2 ~
        \diameter(\kernel_2)
        \parens*{
            2\norm*{\reward_2 - \reward_1}_\infty
            +
            \vecspan{\bias_1} \norm*{\kernel'_2 - \kernel'_1}_\infty
        }
        .
    \end{split}
    \end{equation}
    To transform the bound on the span to a bound on the norm, remark if $\mathbf{u}, \mathbf{v}$ are two vectors and $\mathbf{q}$ is a probability distribution, then $\norm{\mathbf{u} - \mathbf{v}}_\infty \le \vecspan{\mathbf{u}-\mathbf{v}} + \abs{\mathbf{q} (\mathbf{u} - \mathbf{v})}$.
    Let $\mu_j$ be the (unique) invariant probability measure under $\kernel_j$. 
    We have
    \begin{equation}
    \label{equation_unichain_bias_variations_2}
    \begin{split}
        \norm{\bias_2 - \bias_1}_\infty
        & \le
        \vecspan{\bias_2 - \bias_1}
        + 
        \abs{
            \imeasure_2 (\bias_2 - \bias_1)
        }
        .
    \end{split}
    \end{equation}
    The left term of \eqref{equation_unichain_bias_variations_2} is taken care of with \eqref{equation_unichain_bias_variations_1} and we are left to bound $\abs{\imeasure_2(\bias_2 - \bias_1)}$.
    Remark that $\bias_j^1$, the $1$-th higher order bias, is the bias of the Markov reward process $(-\bias_j, \kernel_j)$.
    We have
    \begin{align}
    \notag
        \abs{\imeasure_2(\bias_2 - \bias_1)}
        = \abs{\imeasure_2 \bias_1}
        & = 
        \lim_{T \to \infty}
        \frac 1T
        \abs*{
            \E_{\state}^{\kernel_2} \brackets*{
                \sum_{t=1}^{T-1}
                \bias_1(S_t) 
            }
        }
        \\
    \label{equation_unichain_bias_variations_3}
        & = 
        \lim_{T \to \infty}
        \frac 1T
        \abs*{
            \E_{\state}^{\kernel_2} \brackets*{
                \sum_{t=1}^{T-1}
                \parens*{\mathbf{e}_{\State_t} - \kernel_1(\State_t)} \bias_1^1
            }
        }
        \\
    \notag
        & = 
        \lim_{T \to \infty}
        \frac 1T
        \abs*{
            \E_{\state}^{\kernel_2} \brackets*{
                \sum_{t=1}^{T-1}
                \parens*{\kernel_2(\State_t) - \kernel_1(\State_t)} \bias_1^1
            }
        }
        \le 
        \tfrac 12 \vecspan{\bias_1^1} \norm{\kernel_2 - \kernel_1}_\infty.
    \end{align}
    Combining \eqref{equation_unichain_bias_variations_1}, \eqref{equation_unichain_bias_variations_2} and \eqref{equation_unichain_bias_variations_3}, we obtain the desired bound.
\end{proof}

\subsection{Results for multichain Markov reward processes}

We generalize the inequalities of the unichain setting to general multichain Markov processes.
The new difficulty is that in the multichain setting, a perturbation of the kernel changes the probability of reaching a given recurrent component.
The first result below is key.
It is perhaps surprising, because it shows that the reaching probabilities vary additively and not multiplicatively with respect to kernel modifications.

\begin{definition}[Equivalent chains]
    We say that two Markov chains are \textbf{equivalent} and write $\kernel_1 \sim \kernel_2$ if $\kernel_1 \ll \kernel_2 \ll \kernel_1$.
\end{definition}

\begin{lemma}[Reaching probabilities variations]
\label{lemma_reaching_probabilities}
    Let $\kernel_1 \sim \kernel_2$ be two equivalent Markov chains and $\states^1_1, \ldots, \states^k_1$ be the recurrent classes of $\kernel_1$.
    Let $\tau_\infty := \inf \set{t \ge 1 : \State_t \in \states^1_1 \cup \ldots \cup \states^k_1}$ be the reaching time to one of the recurrent classes and $\tau_i := \inf\set{t \ge 1: \State_t \in \states^i_1}$ be the reaching time to $\states_1^i$.
    We have
    \begin{equation*}
        \abs*{
            \Pr^{\kernel_1}_\state(\tau_i < \infty)
            -
            \Pr^{\kernel_2}_\state(\tau_i < \infty)
        }
        \le 
        \min_{j \in \set{1, 2}}
        \tfrac 12 \E_s^{\kernel_j}[\tau_\infty]
        ~
        \norm*{\kernel_1 - \kernel_2}_\infty
        .
    \end{equation*}
    In case, the bound can be simplified using $\E_{\state}^{\kernel_j}[\tau_\infty] \le \diameter(\kernel_j)$.
\end{lemma}
\begin{proof}
    Consider the reward function $\reward(\State_t, \State_{t+1}) = \indicator{\State_t \notin \states_1^i \text{~and~} \State_t \in \states_1^i}$, providing reward when the walk crosses the frontier between $\states_1^i$ and its $\states \setminus \states_1^i$.
    For $j \in \set{1, 2}$, we denote $\gain_j, \bias_j$ the gain and bias functions under the Markov reward process $(\reward, \kernel_j)$.
    Since $\kernel_1 \sim \kernel_2$, they have the same recurrent classes, we have $\E^{\kernel_j}[\tau_i] < \infty$ for $j = 1, 2$ and $\gain_j(\state) = 0$. 
    Therefore, for $\state \notin \states_1^i$, we have
    \begin{align*}
        \bias_j(\state) 
        & := 
        \lim_{T \to \infty}
        \E^{\kernel_j}_\state \brackets*{
            \reward(S_1, S_2) + \ldots + \reward(S_{T-1}, S_T)
        }
        \\
        & = 
        \lim_{T \to \infty}
        \E^{\kernel_j}_\state \brackets*{
            \sum_{t=1}^{T-1}
            \indicator{\State_t \notin \states_1^i \text{~and~} \State_{t+1} \in \states_1^i}
        }
        \\
        & = 
        \lim_{T \to \infty}
        \E^{\kernel_j}_\state \brackets*{
            \sum_{t=1}^{T-1}
            \indicator{\tau_i = t+1}
        }
        \\
        & =
        \lim_{T \to \infty}
        \sum_{t=1}^{T-1}
        \Pr^{\kernel_j}_\state \parens*{
            \tau_i= t+1
        }
        = \Pr^{\kernel_j}_\state \parens*{2 \le \tau_i < \infty} = \Pr^{\kernel_j}_\state(\tau_i < \infty)
    \end{align*}
    where the last line use that $s \notin \states_1^i$.
    For $\state \in \states_1^i$, we have $\kernel_j(\state) = 0$.
    Our goal is therefore to bound $\bias_2(\state)$ with respect to $\bias_1(\state)$.
    For $\state \notin \states_1^i$, we have
    \begin{align*}
        \bias_2(\state)
        & =
        \lim_{T \to \infty}
        \E^{\kernel_2}_\state \brackets*{
            \sum_{t=1}^{T-1} \reward(\State_t, \State_{t+1})
        }
        \\
        & \overset{(\dagger)}= 
        \lim_{T \to \infty}
        \E^{\kernel_2}_\state \brackets*{
            \sum_{t=1}^{T-1} 
            \parens*{\mathbf{e}_{\State_t} - \kernel_1(\State_t)} 
            \bias_1
        }
        =
        \bias_1(\state) 
        + 
        \lim_{T \to \infty}
        \E^{\kernel_2}_\state \brackets*{
            \sum_{t=1}^{T-1} 
            \parens*{\kernel_2(\State_t) - \kernel_1(\State_t)} 
            \bias_1
        }
    \end{align*}
    where $(\dagger)$ follows from Poisson's equation.
    We are left to bound the RHS. 
    Observe that once $\State_t$ belongs to one of the recurrent components of $\kernel_1$, every state $\state'$ that is reachable under $\kernel_1$ satisfies $\bias_1(\state') = 0$.
    Because $\kernel_1 \sim \kernel_2$, this is also true for any state that is reachable under $\kernel_2$.
    Therefore
    \begin{equation*}
        \abs*{
            \lim_{T \to \infty}
            \E^{\kernel_2}_\state \brackets*{
                \sum_{t=1}^{T-1} 
                \parens*{\kernel_2(\State_t) - \kernel_1(\State_t)} 
                \bias_1
            }
        }
        \le 
        \tfrac 12 \E^{\kernel_2}_\state[\tau_\infty] \vecspan{\bias_1} 
        \norm{\kernel_2 - \kernel_1}_\infty
        \le 
        \tfrac 12 \E^{\kernel_2}_\state[\tau_\infty]
        ~ \norm{\kernel_2 - \kernel_1}_\infty
        .
    \end{equation*}
    In other words, $\abs{\kernel_1(\state) - \kernel_2(\state)} \le \tfrac 12 \E_{\state}^{\kernel_2}[\tau_\infty]$.
    Because $j =1$ and $j=2$ play a symmetric role, $\E_{\state}^{\kernel_2}[\tau_\infty]$ can be changed to $\E_{\state}^{\kernel_1}[\tau_\infty]$.
\end{proof}

\begin{lemma}[Multichain gain variations]
\label{lemma_multichain_gain_variations}
    Let $\kernel_1 \sim \kernel_2$ be two equivalent Markov chains and $\states_1, \ldots, \states_k$ be the (common) recurrent classes.
    Let $\tau_\infty := \inf \set{t \ge 1 : \State_t \in \states_1 \cup \ldots \cup \states_k}$ be the reaching time to one of the recurrent classes.
    We have
    \begin{equation*}
        \norm{\gain_2 - \gain_1}_\infty
        \le
        \norm{\reward_2 - \reward_1}_\infty
        + 
        \min_{j \in \set{1,2}}
        \tfrac 12 \parens*{
            \max_i \vecspan{\bias_j|_{\states_i}}
            + 
            \tfrac 12 k \E_{\state}^{\kernel_j}[\tau_\infty]
        } 
        \norm{\kernel_2 - \kernel_1}_\infty
    \end{equation*}
    In case, the bound can be simplified using $\tfrac 12 \parens{\max_i \vecspan{\bias_j|_{\states_i}} + \tfrac 12 k \E_{\state}^{\kernel_j}[\tau_\infty]} \le \parens{1 + \tfrac 14 k} \diameter(\kernel_j)$.
\end{lemma}
\begin{proof}
    Let $\state \in \states$. 
    We have
    \begin{align*}
        \abs{
            \gain_2(\state) - \gain_1(\state)
        }
        & \le
        \abs*{
            \sum_{i=1}^k 
            \Pr_{\state}^{\kernel_2}(\tau_i < \infty)
            \gain_2(\states_i)
            -
            \sum_{i=1}^k 
            \Pr_{\state}^{\kernel_1}(\tau_i < \infty)
            \gain_1(\states_i)
        }
        \\
        & \le 
        \sum_{i=1}^j 
        \Pr_{\state}^{\kernel_2} \parens{\tau_i < \infty}
        \abs*{\gain_2(\states_i) - \gain_1(\states_i)}
        +
        \abs*{
            \sum_{i=1}^j 
            \parens*{\Pr_{\state}^{\kernel_2}(\tau_i < \infty) - \Pr_{\state}^{\kernel_1}(\tau_i<\infty)}
            \gain_1(\states_i)
        }
        \\
        & \overset{(\dagger)}\le 
        \sum_{i=1}^j 
        \Pr_{\state}^{\kernel_2} \parens{\tau_i < \infty}
        \parens*{
            \norm{\reward_2 - \reward_1}_\infty
            + 
            \tfrac 12 \vecspan{\bias_1|_{\states_i}} \norm{\kernel_2 - \kernel_1}_\infty
        }
        \\
        & \phantom{{} \overset{(\dagger)}\le {}}
        +
        \tfrac 12 \norm*{
            \parens*{\Pr_{\state}^{\kernel_2}(\tau_i < \infty) - \Pr_{\state}^{\kernel_1}(\tau_i<\infty)}_{i=1}^k
        }_1
        \\
        & \overset{(\ddagger)}\le 
        \norm{\reward_2 - \reward_1}_\infty
        + 
        \tfrac 12 \parens*{
            \max_i \vecspan{\bias_1|_{\states_i}}
            + 
            \tfrac 12 k \E_{\state}^{\kernel_1}[\tau_\infty]
        } 
        \norm{\kernel_2 - \kernel_1}_\infty
    \end{align*}
    where $(\dagger)$ follows by applying \cref{lemma_unichain_gain_variations} to the LHS and $(\ddagger)$ by applying \cref{lemma_reaching_probabilities} to the RHS.
\end{proof}

\begin{lemma}[Multichain invariant measure variations]
\label{lemma_multichain_invariant_measure_variations}
    Let $\kernel_1 \sim \kernel_2$ be two equivalent Markov chains and $\states_1, \ldots, \states_k$ be the (common) recurrent classes.
    Let $\tau_\infty := \inf \set{t \ge 1 : \State_t \in \states_1 \cup \ldots \cup \states_k}$ be the reaching time to one of the recurrent classes.
    Denote $\imeasure_j(-|\state)$ the asymptotic empirical distribution of visits starting from $\state$ under $\kernel_j$.
    We have
    \begin{equation*}
        \forall \state \in \states,
        \quad
        \norm*{\imeasure_2(-|\state) - \imeasure_1(-|\state)}_\infty
        \le 
        \min_{j \in \set{1,2}} 
        \tfrac 12 \parens*{
            \max_i \diameter(\kernel_j|_{\states_i})
            +
            \tfrac 12 k \E_{\state}^{\kernel_j}[\tau_\infty]
        }
        \norm{\kernel_2 - \kernel_1}_\infty
        .
    \end{equation*}
    In case, the bound can be simplified using $\tfrac 12 \parens{\max_i \diameter(\kernel_j|_{\states_i}) + \tfrac 12 k \E_{\state}^{\kernel_j}[\tau_\infty]} \le \parens{1 + \tfrac 14 k} \diameter(\kernel_j)$.
\end{lemma}
\begin{proof}
    This is a consequence of \cref{lemma_multichain_gain_variations} by considering the reward function $\reward(\state) = \indicator{\state = \state_\infty}$ for $\state_\infty \in \states$ a fixed state, similarly to \cref{lemma_unichain_invariant_measure_variations}.
    Denote $\gain_j, \bias_j$ denote the gain and bias function of the Markov reward process $(\reward, \kernel_j)$.
    Remark that $\gain_j(\state_0) = \imeasure_j(\state_\infty|\state_0)$.
    By \cref{lemma_multichain_gain_variations}, we have
    \begin{equation*}
        \abs{
            \imeasure_2(\state_\infty|\state_0) - \imeasure_1(\state_\infty|\state_0)
        }
        \le
        \min_{j \in \set{1,2}}
        \tfrac 12 \parens*{
            \max_i \vecspan{\bias_j|_{\states_i}}
            +
            \tfrac 12 k \E_{\state_0}^{\kernel_j}[\tau_\infty]
        } \norm{\kernel_2 - \kernel_1}_\infty
        .
    \end{equation*}
    Because $\states_i$ is a recurrent class under $\kernel_j$, it corresponds to the optimal bias of a communicating model, so by \cref{lemma_policy_bias_diameter}, we have $\vecspan{\bias_j|_{\states_i}} \le \vecspan{\reward} \diameter(\kernel_j|_{\states_i}) \le \diameter(\kernel_j|_{\states_i})$.
    This provides the desired bound.       
\end{proof}

\begin{lemma}[Multichain bias variations]
\label{lemma_multichain_bias_variations}
    Let $(\reward_1, \kernel_1)$ and $(\reward_2, \kernel_2)$ be two Markov reward processes with $\kernel_1 \sim \kernel_2$ and let $\states_1, \ldots, \states_k$ be the (common) recurrent classes.
    Denote $\gain_j, \bias_j$ the gain and bias functions of $(\reward_j, \kernel_j)$ for $j = 1, 2$.
    Let $\tau_\infty := \inf \set{t \ge 1 : \State_t \in \states_1 \cup \ldots \cup \states_k}$ be the reaching time to one of the recurrent classes.
    We have
    \begin{equation*}
        \norm*{\bias_2 - \bias_1}_\infty
        \le
        6 \diameter(\kernel_2)
        \norm{\reward_2 - \reward_1}_\infty
        +
        \parens*{
            \parens*{7 + \tfrac 12k} \diameter(\kernel_2) \diameter(\kernel_1)
            + 2 \diameter(\kernel_1)^2
        }
        \norm{\kernel_2 - \kernel}_\infty
        .
    \end{equation*}
    Following \cref{lemma_policy_diameter_variations}, when $\norm{\kernel_2 - \kernel_1}_\infty \le {\diameter(\kernel_1)}^{-1}$, the quantity $\diameter(\kernel_2)$ can be changed to $2 \diameter(\kernel_1)$.
\end{lemma}

The result can be improved by more carefully tracking the Lipschitz constants throughout.
To lighten the typography---and because our application do not require an optimal result, we overshoot the error term.

\begin{proof}
    For $i = 1, \ldots, k$, pick $\state^i_0 \in \states_i$ and denote $\states_0 = \set{\state^1_0, \ldots, \state^k_0}$.
    For $j = 1,2$, let $\kernel'_j$ be the kernel obtained by copying $\kernel_j$ while making every $\state^i_0$ absorbing, and set $\reward'_j(\state) := \indicator{\state\notin\states_0} \parens{\reward_j(\state) - \gain_j(\state)}$.
    Denote $\gain'_j$ and $\bias'_j$ the gain and bias functions of the Markov reward process $(\reward'_j, \kernel'_j)$.
    Observe that $\gain'_j = 0$ and $\bias'_j(s_0^i) = 0$ for all $i = 1,\ldots, k$.
    Moreover, the biases $\bias_j$ and $\bias'_j$ are linked as follows
    \begin{equation}
    \label{equation_multichain_bias_variations_1}
    \begin{split}
        \bias'_j(\state)
        = 
        \Clim_{T \to \infty}
        \E_{\state}^{\kernel'_j}\brackets*{
            \sum_{t=1}^{T-1}
            \reward'_j(\State_t)
        }
        & = 
        \Clim_{T \to \infty}
        \E_{\state}^{\kernel'_j}\brackets*{
            \sum_{t=1}^{\tau_{\states_0} \wedge T-1}
            \parens*{\reward_j(\State_t) - \gain_j(\State_t)}
        }
        \\
        & =
        \E_{\state}^{\kernel_j}\brackets*{
            \sum_{t=1}^{\tau_{\states_0} -1}
            \parens*{\reward_j(\State_t) - \gain_j(\State_t)}
        }
        \\
        & =
        \bias_j(\state) 
        -
        \E_{\state}^{\kernel_j}\brackets*{\bias(\State_{\tau_{\states_0}})}
        = 
        \bias_j(\state) 
        -
        \sum_{i=1}^k 
        \Pr^{\kernel_j}_{\state}\parens*{\tau_{\states_i} < \infty} \bias_j(\state_0^i)
        .
    \end{split}
    \end{equation}
    In particular, $\vecspan{\bias'_j} \le 2~\vecspan{\bias_j} \le 4 \diameter(\kernel_j)$ by \cref{lemma_policy_bias_diameter}.
    We have
    \begin{equation*}
    \begin{split}
        \bias'_2(\state)
        = 
        \E_{\state}^{\kernel_2}\brackets*{
            \sum_{t=1}^{\tau_{\states_0} -1}
            \reward'_2(\State_t)
        }
        & \le
        \E_{\state}^{\kernel_2}\brackets*{\tau_{\states_0}} \norm*{\reward'_2 - \reward'_1}_\infty
        +
        \E_{\state}^{\kernel_2}\brackets*{
            \sum_{t=1}^{\tau_{\states_0} -1}
            \parens*{\mathbf{e}_{\State_t} - \kernel_1(\State_t)} \bias'_1
        }
        \\
        & \overset{(\dagger)} \le
        \bias'_1(\state)
        +
        \E_{\state}^{\kernel_2}\brackets*{\tau_{\states_0}}
        \parens*{
            \norm*{\reward'_2 - \reward'_1}_\infty
            +
            \tfrac 12 \vecspan{\bias'_1}
            \norm*{\kernel'_2 - \kernel'_1}_\infty
        }
        \\
        & \overset{(\ddagger)}\le 
        \bias'_1(\state)
        +
        \diameter(\kernel_2)
        \parens*{
            \norm*{\reward_2 - \reward_1}_\infty
            +
            \norm*{\gain_2 - \gain_1}_\infty
            +
            2 \diameter(\kernel_1)
            \norm*{\kernel_2 - \kernel_1}_\infty
        }
        \\
        & \overset{(\S)}\le 
        \bias'_1(\state)
        +
        \diameter(\kernel_2)
        \parens*{
            2 \norm*{\reward_2 - \reward_1}_\infty
            +
            \parens*{3 + \tfrac 14 k}
            \diameter(\kernel_1)
            \norm*{\kernel_2 - \kernel_1}_\infty
        }
        .
    \end{split}
    \end{equation*}
    where $(\dagger)$ follows from Poisson's equation, $(\ddagger)$ bounds bias span and hitting times by diameter and expands the definition of $\reward'_j$, and $(\S)$ invokes the (simplified version of) \cref{lemma_multichain_gain_variations} to bound $\norm{\gain_2 - \gain_1}_\infty$.
    With the same computations, we upper-bound $\bias'_2(\state)$ relatively to $\bias'_1(\state)$, leading to the equation:
    \begin{equation}
    \label{equation_multichain_bias_variations_2}
        \abs*{\bias_2'(\state) - \bias_1'(\state)}
        \le
        \diameter(\kernel_2)
        \parens*{
            2 \norm*{\reward_2 - \reward_1}_\infty
            +
            \parens*{3 + \tfrac 14 k}
            \diameter(\kernel_1)
            \norm*{\kernel_2 - \kernel_1}_\infty
        }
        .
    \end{equation}
    Combining \eqref{equation_multichain_bias_variations_1} and \eqref{equation_multichain_bias_variations_2}, we obtain
    \begin{align*}
        &
        \abs*{\bias_2(\state) - \bias_1(\state)}
        \\
        & \le 
        \abs*{\bias'_2(\state) - \bias'_1(\state)}
        +
        \abs*{
            \sum_{i=1}^k \parens*{
                \Pr^{\kernel_2}_{\state}\parens*{\tau_{\states_i} < \infty} \bias_2(\state_0^i)
                -
                \Pr^{\kernel_1}_{\state}\parens*{\tau_{\states_i} < \infty} \bias_1(\state_0^i)
            }
        }
        \\
        & \le 
        \abs*{\bias'_2(\state) - \bias'_1(\state)}
        +
        \abs*{
            \sum_{i=1}^k 
                \Pr^{\kernel_2}_{\state}\parens*{\tau_{\states_i} < \infty} 
                \parens*{\bias_2(\state_0^i) - \bias_1(\state_0^i)}
        }
        \\
        & 
        \phantom{
            {} \le
            \abs*{\bias'_2(\state) - \bias'_1(\state)}
        }
        + 
        \abs*{
            \sum_{i=1}^k \parens*{
                \Pr^{\kernel_2}_{\state}\parens*{\tau_{\states_i} < \infty} 
                -
                \Pr^{\kernel_1}_{\state}\parens*{\tau_{\states_i} < \infty} 
            } \bias_1(\state_0^i)
        }
        \\
        & \overset{(\dagger)}\le 
        6 \diameter(\kernel_2)
        \norm{\reward_2 - \reward_1}_\infty
        +
        \parens*{
            \parens*{7 + \tfrac 12k} \diameter(\kernel_2) \diameter(\kernel_1)
            + \tfrac 12 \vecspan{\bias_1^1}
        }
        \norm{\kernel_2 - \kernel}_\infty
        \\
        & \overset{(\ddagger)}\le 
        6 \diameter(\kernel_2)
        \norm{\reward_2 - \reward_1}_\infty
        +
        \parens*{
            \parens*{7 + \tfrac 12k} \diameter(\kernel_2) \diameter(\kernel_1)
            + 2 \diameter(\kernel_1)^2
        }
        \norm{\kernel_2 - \kernel}_\infty
    \end{align*}
    where $(\dagger)$ is obtained by bounding the first term with \eqref{equation_multichain_bias_variations_2}, the second term with \cref{lemma_unichain_bias_variations} and the third term with \cref{lemma_reaching_probabilities}; and $(\ddagger)$ follows from $\vecspan{\bias_1^1} \le 4 \diameter(\kernel_1)^2$ by applying \cref{lemma_policy_bias_diameter} twice.
\end{proof}

    \clearpage
    \bibliography{biblio}

    \clearpage
    \section*{Index of notations}
    \label{appendix_notations}

\small
\begin{multicols}{2}
    \begin{description}[itemsep=-.25em]

    \item[$\indicator{-}$]
        indicator function
    \item[$:=$]
        defining equality, \emph{defined as}
    \item[$\equiv$]
        syntactical equality, \emph{equal by definition}
    \item[$=$]
        propositional equality, \emph{shown equal}
    \item[$\snorm{-}$]
        support aware norm, \Cref{definition_stared_distance}
    
    \item[$\action$]
        action
    \item[$\actions$]
        action space
    \item[$\actions(\state)$]
        playable actions from $\state$
    \item[$\Action_t$]
        random action at time $t \ge 1$

    \item[$\biasof{\policy}(\model)$]
        bias function of $\policy$, \Cref{section_policies}
    \item[$\optbias(\model)$]
        optimal bias, \Cref{section_optimal_policies}

    \item[$\ivalue(\imeasure, \model)$]
        information value, \Cref{definition_information_value}
    \item[$\ivalue_\epsilon (\imeasure, \model)$]
        $\epsilon$-leveled information value, \Cref{definition_leveled_information_value}
    \item[$\confusing(\model)$]
        confusing set, \Cref{definition_confusing_set}
    \item[$\confusing_\epsilon (\model)$]
        $\epsilon$-leveled confusing set, \Cref{definition_near_confusing_set}

    \item[$\diameter(\model)$]
        diameter, \Cref{assumption_communicating}
    \item[$\worstdiameter(\model)$]
        worst diameter, \Cref{definition_worst_diameter}
    \item[$\diameter(\policy; \model)$]
        policy diameter, \Cref{definition_policy_diameter}
    \item[$\diameter(\kernel)$]
        kernel diameter, \Cref{definition_policy_diameter}
    \item[$\dmin(-)$]
        definite minimum
    \item[$\dmin(\model)$]
        definite minimum of $\model$, \Cref{definition_definite_minimum}
    \item[$\gapsof{\policy}(\model)$]
        gap function of $\policy$, \Cref{section_policies}
    \item[$\ogaps(\model)$]
        Bellman gaps, \Cref{section_optimal_policies}
    \item[$\gaingap(\model)$]
        gain gap, \Cref{definition_gain_gap}

    \item[$\unit$]
        vector full of ones, Euler constant
    \item[$\unit_\state$]
        canonical basis of $\RR^\states$
    \item[$\unit_\pair$]
        canonical basis of $\RR^\pairs$
    \item[$\event$]
        event
    \item[$\entropy(-)$]
        entropy in base $e$
    \item[$\EPSILON$]
        regularization hyperparameter, \Cref{appendix_choice_regularization_hyperparameter}
    \item[$\EPSILON'$]
        floored regularizer, \Cref{appendix_choice_regularization_hyperparameter}
    \item[$\epsilonreg$]
        convexification regularizer, \Cref{section_definition_regularized} and \Cref{appendix_choice_regularization_hyperparameter}
    \item[$\epsilonflat$]
        leveling regularizer, \Cref{section_definition_regularized} and \Cref{appendix_choice_regularization_hyperparameter}
    \item[$\epsilontest$]
        test power, \Cref{appendix_choice_regularization_hyperparameter}
    \item[$\epsilonunif$]
        uniformization regularizer, \Cref{section_definition_regularized} and \Cref{appendix_choice_regularization_hyperparameter}

    \item[$\gainof{\policy}(\model)$]
        gain of $\policy$, \Cref{section_policies}
    \item[$\optgain(\model)$]
        optimal gain, \Cref{section_optimal_policies}

    \item[$\History_t$]
        random history at time $t \ge 1$
    \item[$\histories$]
        history space

    \item[$\imeasures(\model)$]
        invariant measures, \Cref{definition_invariant_measure}
    \item[$\imeasures(\epsilon; \model)$]
        $\epsilon$-uniform invariant measures, \Cref{definition_uniform_invariant_measure}
    \item[$\imeasures(\policy; \model)$]
        invariant measures of $\policy$, \Cref{definition_invariant_measure_policy}
    \item[$\pimeasures(\model)$]
        probability invariant measures $\imeasures(\model) \cap \probabilities(\pairs)$, \Cref{equation_definition_probability_invariant_measures}

    \item[$\regretlb(\model)$]
        regret lower bound, \Cref{theorem_lower_bound}
    \item[$\regretlb_\EPSILON (\model)$]
        regularized regret lower bound, \Cref{definition_regularized_regret_lower_bound}
    \item[$\kl(x, y)$]
        Kullback-Leibler divergence from $\mathrm{Ber}(y)$ to $\mathrm{Ber}(x)$
    \item[$\KL(-||-)$]
        Kullback-Leibler divergence

    \item[$\learner$]
        learning algorithm, \Cref{section_algorithms}

    \item[$\model$]
        Markov decision process, model
    \item[$\model^\dagger$]
        confusing model, \Cref{definition_confusing_set}
    \item[$\models$]
        ambient space of models
    \item[$\imeasure$]
        invariant measure, \Cref{definition_invariant_measure}
    \item[$\imeasureof{\policy}$]
        probability invariant measure of $\policy$, \Cref{proposition_measures_policies_correspondence}
    \item[$\imeasure^*$]
        optimal exploration measure, \Cref{section_lower_bound}
    \item[$\imeasure^*_\EPSILON$]
        $\EPSILON$-optimal exploration measure, \Cref{section_definition_regularized}

    \item[$\visits_t (\pair)$]
        visit count of $\pair$, \Cref{section_algorithms}
    \item[$\rewardd$]
        function of reward distributions


    \item[$\kernel$]
        transition kernel
    \item[$\probabilities(\mathcal{X})$]
        probability vectors over $\mathcal{X}$
    \item[$\policy$]
        policy
    \item[$\policy^+$] 
        exploitation policy, \Cref{section_coexploration} and \eqref{equation_exploration_policy}
    \item[$\policy^-$]
        exploration policy, \Cref{section_exploration} and \eqref{equation_exploration_policy}
    \item[$\policies$]
        deterministic policies
    \item[$\randomizedpolicies$]
        randomized policies
    \item[$\policies_u$]
        unichain deterministic policies
    \item[$\optpolicies(\model)$]
        gain optimal policies, \Cref{section_optimal_policies}

    \item[$\reward$]
        mean reward function
    \item[$\Reward_t$]
        random reward at time $t \ge 1$
    \item[$\Reg(T; \model, \learner)$]
        expected regret, \Cref{definition_regret}

    \item[$\state$]
        state
    \item[$\states$]
        state space
    \item[$\State_t$]
        random state at time $t \ge 1$
    \item[$\sp{-}$]
        span semi-norm

    \item[$t$]
        time, number of learning step
    \item[$T$]
        learning horizon
    \item[$\stimes^-$]
        exploration times, \Cref{section_final_ecoe}
    \item[$\stimes^+$]
        exploitation times, \Cref{section_final_ecoe}
    \item[$\stimes^\pm$] 
        co-exploration times, \Cref{section_final_ecoe}
    \item[$\stimes^!$]
        panic times, \Cref{section_final_ecoe}
    \item[$\stimes_0^-$]
        initial exploration times, \Cref{section_final_ecoe}
    \item[$\stimes_0^+$]
        initial exploitation times, \Cref{section_final_ecoe}

    \item[$\pair$]
        state-action pair
    \item[$\pairs$]
        pair space
    \item[$\Pair_t$]
        random pair at time $t \ge 1$
    \item[$\wkoptpairs(\model)$]
        weakly optimal pairs, \Cref{definition_classification_pairs}
    \item[$\optpairs(\model)$]
        optimal pairs, \Cref{definition_classification_pairs}
    \item[$\flatme{\optpairs}{\epsilon}(\model)$]
        $\epsilon$-leveled optimal pairs, \Cref{definition_near_optimal_pairs}
    \item[$\structure, \structure^+$]
        co-exploration structure, \Cref{appendix_wrong_coexploration}

    \end{description}
\end{multicols}

    \clearpage
    \tableofcontents
\end{document}